# Corpus-Based Approaches to Igbo Diacritic Restoration

## An IGBONLP Project

**Ignatius Majesty Ezeani**

**Supervisors:** Dr. Mark Hepple
Dr. Mark Stevenson

Department of Computer Science
The University of Sheffield

This dissertation is submitted for the degree of
*Doctor of Philosophy*

Faculty of Engineering September 2019

To Babiscum and the four musketeers: Popolistic, Chumeemee, Rịkọ-Rịkọ and Bụbụlistic

# Declaration

I hereby declare that except where specific reference is made to the work of others, the contents of this dissertation are original and have not been submitted in whole or in part for consideration for any other degree or qualification in this, or any other university. This dissertation is my own work and contains nothing which is the outcome of work done in collaboration with others, except as specified in the text and Acknowledgements.

Ignatius Majesty Ezeani
September 2019

# Acknowledgements

The PhD journey is never made alone and mine is not an exception. To get to this stage, I received all kinds of support from numerous people to whom I pour out my thankful heart. This little space may not be enough to mention all the names let alone expressing how deeply grateful I am for each person's contribution. So if I left anyone out, then he or she has the greatest thanks.

My project supervisor, Dr Mark Hepple, is one of a kind. Clearly, without his support, it would have been extremely difficult if not impossible. He was gentle but firm; kind but thorough. He doused my fears, worries and confusions and gave me numerous opportunities to prove myself. I am not sure I convinced him but fortunately his belief in me was enough to sustain the journey and we eventually made it. Thank you, Mark! I learnt a lot.

I wish to also thankfully extend my gratitude to other members of my supervisory team: Dr Mark Stevenson, my second supervisor, whose insightful contributions helped to focus this work and Prof Jon Barker, my panel chair, for the constructive suggestions you gave at the panel meetings. I also owe a debt of gratitude to my examiners, Professor Eric Atwell (the external examiner) and Dr. Diana Maynard (the internal examiner) for the well thought-out questions and suggestions that greatly improved the final work. Worthy of mention is Diana's patience and painstaking approach to correcting this work – it was, to say the least, remarkable. I thank the post-graduate admin team of the Computer Science Department for all their amazing support and guidance. They are simply the best!

I am immensely indebted to Dr. Ikechukwu Onyenwe for introducing me to NLP and literally hand-holding me through each step at the beginning of this journey. I call him *the father of IgboNLP* (and watch him blush of course!) because he did most of the ground work that made it easy for us to get in. I also thank Chioma Enemuo whose contributions as a team member opened us to other possibilities. I am also grateful to Mark Tice for his technical support and enlightening discussions which he often volunteered at very huge personal costs. Mark's support was not only timely and handy but also most invaluable.

I also deeply appreciate all my friends and colleagues in the NLP Group especially those I shared G28 Regent Court within the last 4 years (or a fraction of that). To many other members of the university community with whom I interfaced on daily basis: the porters, cleaners, IT support guys etc, thank you all.

To my dear wife Babiscum, the pillar of my strength and to our lovely children: Mmesoma (*Popolistic*), Chukwuemeka (*Chumeemee*), Ifunanya (*Rịkọ-Rịkọ nwa*) and Ebubechukwu (*Bubulistic nwa*), I know it's been a journey and you all sacrificed a lot. But I promise that someday, hopefully in my life time, I will make it up to you guys. *Doonu shinne!*

I am also immensely grateful to my parents, Onyeagwalukwe and Ogolinaeriaku, my siblings especially Dr. Ernest Ezeani and their families who are here with me, and our family friends especially the Nnolịs, Warmates, Ezurikes, Nwagbos, Nnanwas, Ezeribes, Okos for all the prayers and various kinds of supports we got from you during this journey. Thank you all so much and God bless you all!

And finally, to *Him Who Sits on the Throne*, be all the Glory!

# Abstract

With natural language processing (NLP), researchers aim to get the computer to identify and understand the patterns in human languages. This is often difficult because a language embeds many dynamic and varied properties in its syntaxes, pragmatics and phonology, which needs to be captured and processed. The capacity of computers to process natural languages is increasing because NLP researchers are pushing its boundaries. But these research works focus more on well resourced languages such as English, Japanese, German, French, Russian, Mandarin Chinese etc. Over 95% of the world's 7000 languages are low-resourced for NLP i.e. they have little or no data, tools and techniques for NLP work.

In this thesis, we present an overview of diacritic ambiguity and a review of previous diacritic disambiguation approaches on other languages. Focusing on Igbo language, we report the steps taken to develop a flexible framework for generating datasets for diacritic restoration. Three main approaches, the standard *n-gram model*, the *classification models* and the *embedding models* were proposed.

The standard *n*-gram models use a sequence of previous words to the target stripped word as key predictors of the correct variants. For the classification models, a window of words on both sides of the target stripped word were use. The embedding models compare the similarity scores of the combined context word embeddings and the embeddings of each of the candidate variant vectors.

The processes and techniques involved in projecting embeddings from a model trained with English texts to an Igbo embedding space and the creation of intrinsic evaluation tasks to validate the models were also discussed. A comparative analysis of the results indicate that all the approaches significantly improved on the baseline performance which uses the unigram model. The details of the processed involved in building the models as well as the possible directions for future work are discussed in this work.

# Table of contents

# List of figures

# List of tables

# Chapter 1

# Introduction

This chapter presents a background to this thesis report by introducing the motivation, the research questions and a set of aims and objectives we started with. It then goes further to summarize our main contributions to this body of research with references to our conference publications. The structure of the rest of the thesis is given at the end of the chapter.

## 1.1 Background

The period from the early 1940s to just after the World War II is regarded as "intellectually fertile" and will take the credit of actually giving birth to the computer itself. It was about the same time (1950s) that Alan Turing, through his work, *Computing Machinery and Intelligence* [100] proposed what is now called the *Turing Test* as a criterion of intelligence. This is generally believed to be the birth of AI research and consequently the beginning of research efforts towards developing man-machine language understanding techniques.

Natural language processing (NLP) is an area of research that aims at developing computer systems with the capacity to understand and manipulate human language text and speech in performing useful tasks. It is an aspect of a vibrant interdisciplinary field that connects such disciplines as computer and information sciences, linguistics, mathematics, statistics, electrical and electronic engineering, artificial intelligence and robotics, psychology and even the arts and social sciences. In NLP, computational techniques are used to *learn*, *understand* and *produce* human language contents for such real-world applications as spoken dialogue systems, speech-to-speech translation engines, mining social media for information about health or finance, and identifying sentiment and emotion toward products and services [48].

The basic goal of this field of research is to get the computer to perform useful tasks with human language such as enabling man-machine communication, improving human-human communication, or simply doing useful processing of text or speech [53]. NLP researchers aim to model human understanding and use of language so that appropriate tools and techniques can be developed to make computer systems mimic, understand and manipulate natural languages in performing the desired tasks. Traditionally, natural language processing often involves classifying the language analysis tasks to reflect the theoretical linguistic distinctions between syntax, semantics and pragmatics [22].

The problem, however, is that NLP research often focuses on major languages such as English, Japanese, German, French, Russian, Mandarin Chinese and others. This non-integration of minority languages can potentially marginalise their speakers especially with the changing trends in global information access and technology applications.

## 1.2 Motivation

With the proliferation of smart speech and text processing technologies for richly resourced languages, the chance of integration of low-resourced languages keeps diminishing. This is especially the case for Africa, which is home to speakers of over 2000 different minority languages. About two decades ago Church and Lisa [15] observed that:

> "Monolingual speakers of English (like the authors) sometimes forget that their language is not the only language in the world. The major word processing applications are being sold throughout Europe and much of the rest of the world. The overseas markets are large and growing. To be successful in these important markets, products have to be *localized* so that they conform to the language and cultural norms of the target customers."

This assertion may not be entirely correct now as English, though still at the top, is not the only strong player among the languages of technologies. Other languages such as Japanese, German, French, Russian and Mandarin Chinese also have a substantial share of the market [97]. But this was re-echoed in a more recent review by Hirschberg and Manning [48]:

> A major limitation of NLP today is the fact that most NLP resources and systems are available only for high-resource languages (HRLs), such as English, French, Spanish, German, and Chinese. In contrast, many

> low-resource languages (LRLs)—such as Bengali, Indonesian, Punjabi, Cebuano, and Swahili—spoken and written by millions of people have no such resources or systems available. A future challenge for the language community is how to develop resources and tools for hundreds or thousands of languages, not just a few.

So today, it is still a huge problem for the NLP community that more than 95% of the world's 7000+ languages cannot be fully processed by today's technologies. New technologies are often built with these established languages in mind mainly because there are lots of resources to test them on. Speakers of under-resourced languages are therefore by this, not only marginalised economically, politically and socially but are also under the threat of losing their identities. With the commercial interest around well resourced languages, the advancements in technologies that support them are likely to continue in upward trend, considering the enormous amount of resources committed to developing them[1]. This will definitely widen the gap between the people on both sides of the divide and put more pressure on low resource language speakers to "upgrade" to well resourced languages.

As Ruder [87] recently noted, our globalised society has national borders that increasingly blur and the Internet gives everyone equal access to information. This therefore calls for the need to not only seek to eliminate gender and racial bias noted in the work of Bolukbasi *et al.* [13], but also aim for more inclusion of languages with limited resources. This project therefore is motivated by the need to redefine the success of modern NLP tools in terms of their capacity to support minority languages. This includes the ability to formulate strategies, build frameworks and develop tools for low resource languages that can efficiently leverage existing models in building language processing resources.

Our focus in the project is Igbo, a south-eastern Nigerian language. As may have been observed even from the quotes above, languages like Igbo are so marginalised that they have not caught the attention of the NLP research community and, as such, are not even being mentioned among the low resourced languages. Igbo is low-resourced and has orthographic and tonal diacritics. These diacritics capture distinctions between words that are important for both meaning and pronunciation, and hence are of potential value for a range of language processing tasks. A more detailed description of the language in presented in Chapter 2.

[1]Common examples are Amazon's *Echo*, Apple's *Siri*, Microsoft's *Cortana* and Google's *Google Now*.

## 1.3 Research Questions

An efficient approach to developing NLP resources for Igbo may involve adapting existing methods and models. This will surely require Igbo corpora with proper diacritic marks. While the web provides a good source of data to bootstrap the process of building resources for Igbo, web texts are often lacking in proper diacritic marks making it difficult to process or assemble them into corpora.

Diacritics play a huge role in defining the pronunciation and meaning of words in Igbo. The absence of proper diacritics in Igbo words causes ambiguities that can affect NLP systems (e.g machine translation). Diacritic restoration is the process of replacing missing diacritics in texts. So there is a need for effective methods for automatic diacritic restoration for Igbo. Thus our main research questions are:

1. *Can we construct a standard dataset for the Igbo diacritic restoration task?* This will be a useful resource for other researchers who may be interested in working on this task and will provide a benchmark for evaluating the performance of Igbo diacritic restoration systems.

2. *Can we build a robust automatic diacritic restoration system for the Igbo language?* Ideally, this system should be able to attach the right diacritics to a *wordkey*[2] given the context e.g. **akwa**: *àkwá* (egg), *ákwà* (cloth), *ákwá* (cry/wail), *àkwà* (bed/bridge).

3. *Can we take advantage of existing high resource language models in diacritic restoration?* The variants of an ambiguous wordkey in Igbo map to unique words in other languages, say English. We will consider taking advantage of the semantic properties captured by an English language model for diacritic restoration.

## 1.4 Aims and Objectives

This project aims at developing methods for automatic restoration of diacritics. This will hopefully be a step to enhancing the quality of existing corpora and contribute to the broader plan of developing fully functional language processing modules.

Our work will follow the guidelines suggested in the BLARK project [57] for defining, adopting and implementing a standard resource toolkit for *all* languages,

[2] A wordkey is a word stripped of its diacritics if it has any. Wordkeys could have multiple diacritic variants, one of which could be the same as the wordkey itself

*despite their size or importance*, and thereby creating better starting conditions for research, education and development in language and speech technology.

BLARK is defined in principle to be language independent. It recognizes the specificity of the requirements for different languages and therefore provides room for varied instantiations for languages. Krauwer [57] recommends that a typical BLARK specification for a language should contain:

- a multigenre text corpus size of 10 million words, from contemporary materials, with standard annotations
- a similar size and structure for the speech corpus
- a collection of basic tools for manipulating and analyzing the corpora
- a sizeable number of experts and researchers with various skills in language processing
- an open source policy that guarantees free accessibility and usability to ensure maximum impact

Given the amount of time and resources available for this work, it would not be possible to achieve all the goals pointed above. In this work therefore, we aim at contributing to this vision by building resources for Igbo diacritic restoration. Our aim is to introduce a language-independent approach to this task that is efficient, generalisable and involves very little human effort. The core objectives set out at the beginning of our work are:

1. reviewing the state-of-the-art approaches in the literature
2. acquiring a fully diacritically marked Igbo corpus for model building
3. developing and evaluating a baseline system for Igbo diacritic restoration
4. applying state-of-the-art techniques to improve the performance of the tool
5. extending the system for Igbo word sense disambiguation task
6. building a framework for adaptability to other languages

## 1.5 Contributions of this project

At the end of this project, we aim at making the following contributions to the body of knowledge and the NLP research community which will serve as reference for other researchers and support future work in this area.

1. A good review of reviewing what has been reported on this task and, where applicable, on Igbo language.

2. A standard dataset for Igbo diacritic restoration experiments will be created to enable other researchers to train their models, evaluate their and, possibly, suggest improvements on the dataset.

3. Build a benchmark system that will serve a baseline future explorations and experimentation on the diacritic restoration task.

4. Building a robust restoration model through extensive experimental design and implementation involving other language models (e.g. embedding models) and techniques such as embedding models.

5. Providing a comprehensive report and analysis of the results of the experiments and the evaluation of the models built.

## 1.6 Thesis Structure

The rest of this thesis is structured as follows:

- **Chapter 2** presents an overview of Igbo people, their language and culture as well as their origin and history. It also presents the language characteristics: script and orthography, phonology and tonology, vocabulary and morphology. It ends by highlighting recent efforts in the IgboNLP research.

- **Chapter 3** discusses diacritics in languages and reviewed a broad range of diacritic restoration techniques for a cross-section of languages with diacritics in literature. It then introduced the ambiguities and complexities caused by the presence of both tonal and orthographic diacritics in Igbo and explained why it is an important problem in Igbo.

- In **Chapter 4**, we lay the foundation for the rest of the project creating a standards dataset for this task. We will present the process and considerations involved in building an extensible framework for automatically creating such datasets from any plain text data of any language with diacritics. The core evaluation methods and metrics used throughout this work will also be defined in this chapter.

- **Chapter 5** introduces the first set of experiments on diacritic restoration with the *unigram* as the baseline model. The design, implementation and execution of experiments with other higher n-gram models as well as the analysis of the results will also be presented.

- **Chapter 6** presents the application of classification models built with different machine learning algorithms to the diacritic restoration task. It discusses the high-level process flow as well as the feature extraction and vectorization techniques used and then ends with the analysis of results.

- In **Chapter 7**, we delve into a whole new world of word embedding models and the possibility of applying them to diacritic restoration. This chapter is particularly interesting because it provides a fairly good background to embedding models and in addition to the main project task introduced other intrinsic evaluation tasks.

  Also, given that trained Igbo embedding models are not readily available and there is not enough data for training one, we had to expand the scope of this chapter to cover some of the transfer learning techniques used to adapt models trained in one domain to be used in another domain.

  The process of applying embedding models to our core task of diacritic restoration is also defined along with some of the enhancement schemes that improve the performance of the models. Experiments to compare the performances of trained and projected models on both the intrinsic evaluation and the diacritic restoration tasks are conducted.

- **Chapter 8** presents a concise summary of the work done, the resources created and results obtained in this thesis with a discussion on the challenges met and the direction for future work.

## 1.7 Chapter Summary

This introductory chapter started with a broad overview of NLP and the challenges of building resources for low resource languages. The motivation, research questions as well as the aims and objectives of the research on Igbo language were then presented. The chapter ended with our main contributions, including research presentations made at conferences and workshops

# Chapter 2

# Igbo Language and IgboNLP

This chapter presents an overview of Igbo, the language of our study. It highlights the characteristics of the people and their culture, historical developments as well as other language characteristics e.g script, orthography, phonology and tonology which are very essential for our research as well as other features like vocabulary, morphology.

## 2.1 Overview

At the turn of the decade, there were fears that Igbo language will go extinct by the year 2050 [6]. Africa has 2,138 living languages out of which 376 are either dying or in trouble[1]. Nigeria, popularly referred to as the *Giant of Africa*, is the most populous country in Africa and the 7th in the world with UN projected population estimate of 183,523,000 by July 1, 2015[2]. Also as of 2015, the economy of Nigeria was the largest in Africa and 20th in the world. The Igbo people constitute 18% of Nigeria's population and their language is one of the three major Nigerian languages. So the possibility of the extinction of Igbo language posed a big threat not only to Nigeria but also to Africa.

Fortunately, more recent studies suggest that Igbo language is no longer considered endangered. In fact current statistics indicate that it is quite institutionalized[3]. However, as observed by [8] Igbo still faces endangerment largely due to the "*colonial mentality*" that imposes the "*so-called receptivity to change*" which then gives rise to "*loss of identity in every new situation of culture contact*". Igbo speakers are constantly under immense pressure to "upgrade" themselves to this superior language as both a status symbol and a meal ticket.

[1] https://www.ethnologue.com/region/Africa

[2] https://en.wikipedia.org/wiki/List_of_countries_and_dependencies_by_population

[3] See http://www.ethnologue.com/cloud/ibo The Expanded Graded Intergenerational Disruption Scale (EGIDS) level for Igbo in Nigeria is 2 (Provincial) i.e. it is used in education, work, mass media, and government within major administrative subdivisions.

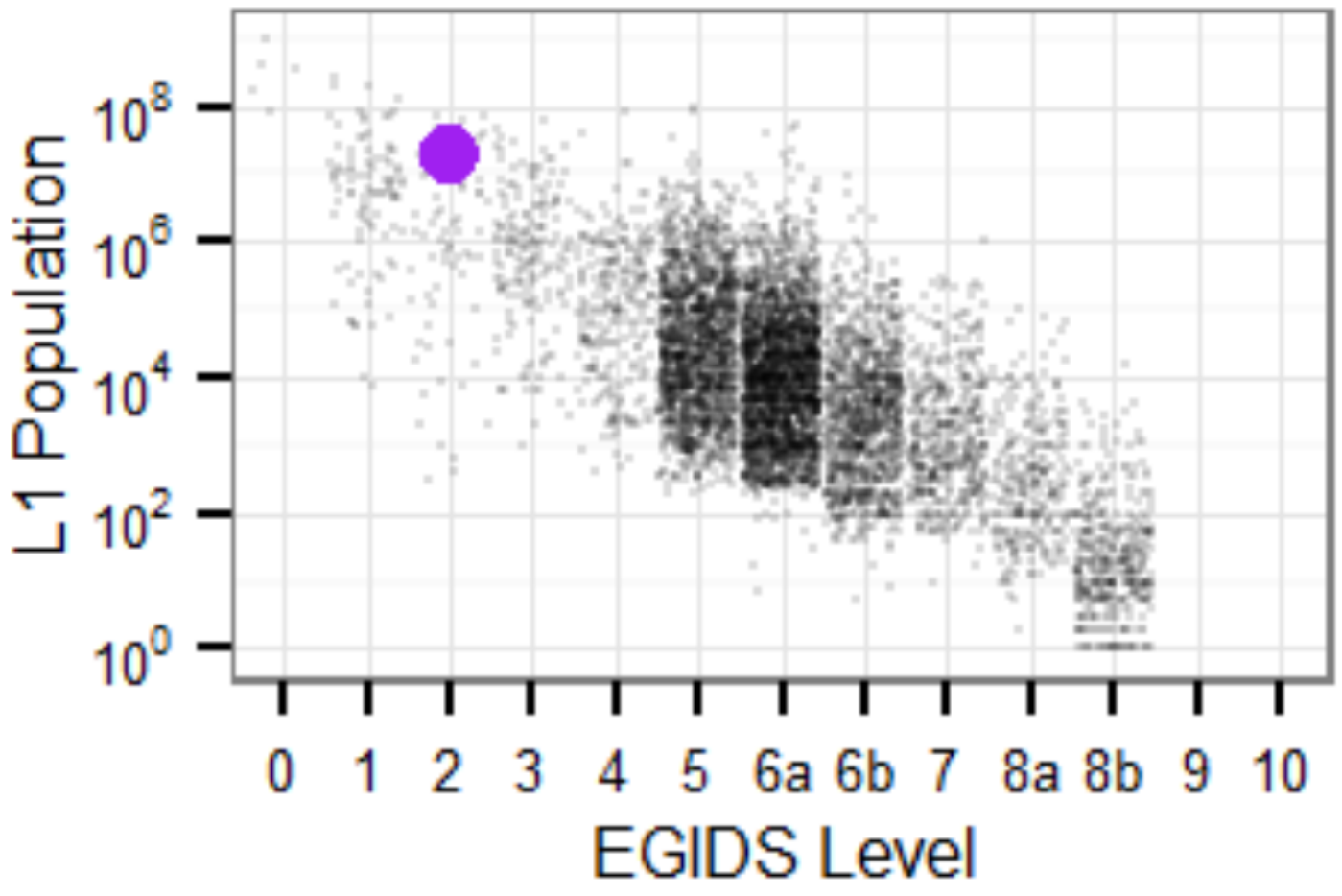

Fig. 2.1 Graph showing Igbo in the Language Cloud (*Source*: Ethnologue: Languages of the World, 21st edition (2018) [96])

## 2.2 Igbo People and Culture

Igbo is the primary native language of the Igbo people of southeastern Nigeria. There are over 30 million speakers of Igbo language who are mostly resident in Nigeria and are of predominantly Igbo descent. Igbo is written with the Latin scripts and has about 20 different dialects. Some Igboid languages such as *Ekpeye* and *Ukwuani*, though more divergent from Igbo, are often considered Igbo dialects. Equatorial Guinea officially recognises Igbo as a minority language.

### 2.2.1 Origin and History

Igbo is one of the 1538 languages of the Niger-Congo family. Its trace on the language family tree is: *Niger-Congo, Atlantic-Congo, Volta-Congo, Benue-Congo, Igboid, Igbo.* In the pre-1500 era, Igbo was written with the *Nsibidi* [23] writing system. This system used formalized pictograms but later died out.

At the height of the trans-Atlantic slave trade (1500-1700), which seriouly affected the Igbos, there was no recorded development of the language. However, in the second half of the 18th century studies of Igbo language made strides with the Isuama

Igbo[4] [72] and the works of G. C. A. Oldendorp, a German missionary. Oldendorp's works, *History of the Evangelistic Mission of the Brothers in the Caribbean*, which was published in 1777 contained a few Igbo words and sentences. By 1789, the work of Olaudah Equiano, a former slave *The Interesting Narrative of the Life of Olaudah Equiano* featuring 79 Igbo words was published.

The rest of the 18th century was dominated by further development of the Isuama studies and the works of different missionary workers notable among whom were, Hannah Kilham, McGregor Laird, Jacob Friedric Schon, Edwin Norris, Samuel Ajayi Crowther and Julius Spencer. Crowther, an ex-slave who later became the first African Anglican bishop, made remarkable contributions to the Isuama Igbo with his works *Isoama Igbo Primer (1857)* and *Vocabulary of the Ibo Language, Part I & II (1883)*[5]. Other notable publications of this era include *Oku Ibo: Grammatical Elements of the Ibo Language (1861)* by J F Schon, *An Ibo Primer (1870)* by F. W. Smart, *An Elementary Grammar of the Igbo Language (1892)* by J Spencer.

Between 1900 and 1929, Church Missionary Society (CMS) began the Union Igbo Studies with the aim of binding all Igbo dialects with the intent to create a "compromise" or "central" Igbo. The key dialects were Onitsha, Owerri, Unwana, Arochukwu and Bonny and the central Igbo was to be a standard literary medium. Major translations of materials on Igbo culture including proverbs, folktales, riddles and customs happened in this period and the most prominent publication was Bible Nso (Holy Bible).

Other notable publications include the trilingual dictionary, *English, Ibo and French Dictionary (1904)* by A Gabot, a French missionary and the 6-volume *Anthropological Report on the Igbo Speaking People of Nigeria (1914)* by N W Thomas which covered the Onitsha, Awka and Asaba dialects. Most of the publications at this point used the *Lepsius orthography*[6]. However, a new orthography[7] by the International Institute of African Languages and Culture (IIALC) was introduced in 1927 [51].

Between 1930 and 1970, Igbo language scholars were enmeshed in the orthography controversies. The African orthography by IIALC was adopted by the government and the Roman Catholic mission but was not well received by the Anglican missions. This gave rise to what eventually became the *CMS orthography* and the *Roman Catholic orthography.* There were also controversies surrounding the dialects. While the Anglicans adopted Union Igbo, the Catholics and the Methodists took the Onitsha and Central dialects respectively.

[4] *Isuama* is the name given to the south-central part of Igboland, which was a major source of slaves during the period of the trans-Atlantic slave trade

[5] Part I was the first comprehensive Igbo dictionary while Part II extended to be an English-Igbo dictionary

[6] https://en.wikipedia.org/wiki/Standard_Alphabet_by_Lepsius

[7] IIALC released a pamphlet called *The Practical Orthography of African Languages*

Some of the key literary works of this era include *Omenuko (1933)* by Peter Nwana which was the first Igbo novel, *Ala Bingo (1937)* (Bingo's Island) by D N Achara and *Ije Odumodu Jere (1963)* (Odumodu's Adventures) by Léopold Bell-Gam. Also the Society for Promoting Igbo Language and Culture (SPILC) was formed in 1949 by F C Ogbalu, a mission tutor at DMGS, Onitsha[8] The main aim of SPILC was to oppose the new orthography which was seen at that point as the imposition of the '*whiteman's will*'. There are many failed efforts to resolve the differences in the major opposing forces in the orthography quagmire.

Eventually in 1961, a government commitee headed by Dr S E Onwu[9] produced *The Official Igbo Orthography* which was the most widely accepted. Despite persistent subtle attacks on this new orthography, most people believe that it is designed to be flexible and more easily understood by all.

### 2.2.2 Script and Orthography

Igbo language has many dialects but most written works are based on the standard Igbo that uses the official orthography produced by the Ọnwụ Committee [10]. This orthography contains 8 vowels (*a, i, o, u, ị, ọ, ụ*) and 28 consonants (*b, gb, ch, d, f, g, gw, gh, h, j, k, kw, kp, l, m, n, nw, ny, ṅ, p, r, s, sh, t, v, w, y, z*).

The significant features of this orthograpy includes the presence of 3 dot-below vowels (*ị, ọ, ụ*), 9 digraph consonants (*gb, ch, gw, gh, kw, kp, nw, ny, sh*) and a tilde n (ñ). Also with this orthography, long vowels are often marked by doubling and there is no emphasis on tone marking despite its phonemic characteristics.

It is important to note that many researchers of Igbo dialects consider the Ọnwụ orthography inadequate because of the inability to represent many dialectal sounds with it. A Composite Synchronic Alphabet of Igbo Dialects (CSAID) by the Igbo Archival Dictionary Project (IADP) group [1] is an attempt to produce a more comprehensive orthography that spans a broad range of Igbo dialects. CSAID has, at the moment, 96 sound segments comprising of 10 vowels and 86 consonants.

[8] Dennis Memorial Grammar School (DMGS), Onitsha, a prominent mission secondary school, founded in 1927, is one of the CMS's strongest legacy of the missionaries through which the SPILC commanded a lot of influence in the south-east.

[9] This committee is popularly referred to as *The Ọnwụ Committee*.

[10] http://www.columbia.edu/itc/mealac/pritchett/00fwp/igbo/txt_onwu_1961.pdf

### 2.2.3 Phonology and Tonology

The Igbo vowels are divided into two mutually incompatible harmony groups based on the advanced tongue root (ATR): The ATR+ vowels are *e,i,o,u*, which are usually pronounced by moving the base of the tongue forward thereby expanding the pharyngeal cavity. However in pronouncing the ATR- vowels, *a,ị,ọ,ụ*, the tongue usually remains in neutral position. In the Igbo syllable structure, syllables end with a vowel or *m* with the most common structures as *V*-vowel, *CV*-consonant-vowel and *CVm*-consonant-vowel-*m*

Igbo language is tonal i.e. one can use the voice pitch to infer the meaning of a word or utterance [108, 31]. It belongs to the 'terraced-level' tone languages in which the high- or the mid-tones step downward (also known as *downstep*) in pitch after certain tones. Welmers [108] observed that Igbo language has two distinct tones: high(´)[**H**], and low(`)[**L**], with an additional mid tone(¯)[**S**], also referred to as *phoneme downstep*.

Emenanjo [31] observed that the downstep tone does not occur as commonly as the other two. For example, all words begin with either high tone or low tone but not the downstep (mid) tone. Another distinction in the behaviour of the the downstep tone is that it does not follow a low-tone sound in a word. For example, we can have words like *òké* (rat) or *òkè* (share, part, portion) but there is no word like *òkē* (rat). But all tones can follow the high tone e.g *áká* (hand)[HH], *ázụ̀* (fish)[HL] or *írī*(to eat)[HS]. So basically, there are two possible phonemic contrast: a 2-way contrast after **L** (i.e. L: LL and LH ) and 3-way contrast after **H** (i.e. H: HH, HL and HS).

| | **akwa** | **isi** | **igwe** | **ọcha** |
|---|---|---|---|---|
| **HH** | cry | head | – | white |
| **HL** | cloth | smell | iron/metal | – |
| **LH** | egg | – | – | whiteness |
| **LL** | bed;bridge | blindness | crowd | – |
| **HS** | – | – | sky | – |
| | **oke** | **mma** | **ide** | |
| **HH** | male | – | – | |
| **HL** | boundary | knife | flood;earthworm | |
| **LH** | rat | – | pillar, basket | |
| **LL** | share | blindness | raffia palm wine | |
| **HS** | – | beauty | – | |

Table 2.1 Lexical Classification with Tones [31, 1]

Tone plays two distinct roles in Igbo language:

**Lexical:** Tones can function as lexical classifiers. They differentiate between lexical items or words which could otherwise be written in the same way. For example, it is difficult to say what the word *isi* means without a context. But with tone marks, the disambiguation becomes easier even without additional information e.g. ***isi*** could be *ísí* (head), *ísì* (smell), *ísī* (to cook), *ìsì* (blindness). Tonal distinction of words can separate words into different meanings and classes or parts of speech e.g. the word *ọ̀chá* (whiteness) is a *nominal* while *ọ́chá* (whiteness) is a *nominal modifier.* Other examples are given in Table 2.1.

**Grammatical:** Grammatically, declarative sentences are easily distinguished from interrogative ones with similar structure using tone marks (e.g. *Ọ́ ga-abịa* "He will come." and *Ọ̀ ga-abịa?* "Will he come?"). Also, the numerical expressions in the statements *ụ́lọ̀ ạ̀tọ́* (three houses) and *ụ́lọ̄ ạ̄tọ́* (the third house) are differentiated by the tones they both bear. While the first is cardinal, the second is ordinal.

Although the letters of the Igbo alphabet are not pre-tone-marked, the importance of tone in Igbo is not in doubt. In fact, as Achebe *et al.* [1] asserted:

> *... tone is a very important feature of [Igbo] and ... should be marked if the intended meaning of words and grammatical structures should be understood.*

There are different tone-marking conventions but a common convention adopts a form of tone economy in which the high tone (often marked with the acute accent (´) in the examples above) is left unmarked, low tone is marked with the grave accent (`) and the downstep (or mid) tone is marked with macron high(¯).

### 2.2.4 Vocabulary

Table 2.2 shows a list of some common Igbo words or phrases and their meanings. As can be observed, any Igbo nouns are actually fusions of older original words and phrases. For example, one Igbo word for vegetable leaves is *akwụkwọ nri*, which literally translates to 'leaves for eating' or 'edible leaves'. Green leaves are called *akwụkwọ ndụ*, because ndụ means 'life'. Another example is train (ụgbọ igwe), which comes from the words ụgbọ (vehicle, craft) and igwe (iron, metal); thus a locomotive train is "iron (rail) vehicle"; a car, *ụgbọ ala* (land vehicle); and an aeroplane, *ụgbọ elu* (air vehicle).

Words may also take on multiple meanings depending on the context of use. Take for example the word akwụkwọ which originally means "leaf" (as on a tree), but since the colonization period, akwụkwọ also came to be linked to paper, book, school, and education, to become akwụkwọ édémédé, akwụkwọ ọgụgụ, ụlọ akwụkwọ, and mmụta akwụkwọ, respectively. This is because paper comes in sheets, or leaves; books contain 'leaves' of paper, a school is a 'book building', and so on. Combined with other words, akwụkwọ can take on many forms; for example, *akwụkwọ ego* (money paper) means printed money or bank notes, and *akwụkwọ ejị èjé njem* (journey paper) means passport.

| *Word* | *Meaning* | *Word* | *Meaning* | *Word* | *Meaning* |
|---|---|---|---|---|---|
| *otu* | one | *nna* | father | *isi* | head |
| *abụọ* | two | *nne* | mother | *ihu* | face |
| *atọ* | three | *nwanne nwoke* | brother | *anya* | eye |
| *anọ* | four | *nwanne nwanyi* | sister | *ntị* | ears |
| *ise* | five | *nwa nwoke* | son | *imi* | nose |
| *isii* | six | *nwa nwanyi* | daughter | *ọnụ* | mouth |
| *asaa* | seven | *okpara* | first son | *onu* | neck |
| *asatọ* | eight | *ada* | first daughter | *aka* | hand |
| *itolu* | nine | *nwa* | child | *ụkwụ* | leg |
| *iri* | ten | *ụmụ* | children | *ire* | tongue |
| *narị* | hundred | *ụmụnna* | kinsmen | *afọ* | stomach |
| *puku* | thousand | *ụmụnne* | siblings | *obo aka* | palm |
| *nde* | million | *di* | husband | *mkpịsị aka* | fingers |
| *ijeri* | billion | *nwunye* | wife | *mkpịsị ukwu* | toes |

Table 2.2 Some common Igbo words and meanings

### 2.2.5 Morphology

Morphologically, Igbo language is considered agglutinative[11] i.e. words usually contain different but easily deducible morphemes that determine their meanings. The lexical categories (nouns and verbs) derive their forms through affixations. Though, at nominal levels, there is little inflection, the verb forms can be heavily influenced by the affixes. Consider the conjugation of the stem ***ri*** (to eat) as presented in Table 2.3 and the impact on tense, aspect and mood.

## 2.3 IGBONLP: The Vision

NLP research for Igbo language, IGBONLP, is a project that aims to galvanize researchers to articulate, build and document processes, models, data and output geared towards developing automatic linguistic analysis tools for Igbo language in particular and for other low resource languages in general. The idea originated in 2013 at the University of Sheffield, UK, from the work of Onyenwe [74] but it is still an on-going project which our work directly contributes to.

[11] https://en.wikipedia.org/wiki/Agglutinative_language

| Stems: **ri**(to eat), **bịa**(to come) | | |
|---|---|---|
| *Tense* | **ri** | **bịa** |
| present: | *e***ri** | *a***bịa** |
| past: | **ri***ri* | **bịa***ra* |
| present continuous: | *na-e***ri** | *na-a***bịa** |
| future: | *ga-e***ri** | *ga-a***bịa** |
| imperative: | **rie** | **bịa** |
| present perfect: | *e***ri***ela* | *a***bịa***la* |
| negative past: | *e***ri***ghị* | *a***bịa***ghị* |
| negative present cont: | *naghị e***ri** | *naghị a***bịa** |
| negative future: | *gaghị e***ri** | *gaghị a***bịa** |
| negative imperative: | *e***ri***la* | *a***bịa***la* |
| infinitive: | *i***ri** | *ị***bịa** |

Table 2.3 Example of Igbo Morphological Inflection

Simply put, the IGBONLP project aims at providing a Basic Language Resource Kit (BLARK) as proposed by Krauwer [57] which not only provides a substantial collection of Igbo corpora and datasets across different genres but also enables the development of basic NLP tools such as tokenizers, diacritic restorers, PoS-taggers, parsers, chunkers, stemmers, morphological analysers, etc., that are tuned to the nuances of the language. The development of applications from simple spell checkers, hyphenation tools, keyword-in-context (KWIC) to grammar checkers, information retrieval systems, electronic dictionaries, thesauri and named-entity recognisers and even machine translation systems will hopefully be supported by these tools.

The IGBONLP project is, expectedly, hindered by similar challenges faced by such projects viz data sparsity, non-suitable models and no standard evaluation mechanisms as highlighted by Palmer & Regneri [78], isolated development occasioned by inadequate funding. Also for Igbo, even with a well-developed orthography and writing system, the lack of adequate keyboard and input system makes the creation of electronic text a major challenge [102].

Streiter *et al.* [97] also observed that lack of funding and infrastructure, enabling academic environment and culture, suitably qualified personnel and ineffective national policies and strategies on education and development make NLP research in Nigeria very unattractive to researchers. However, in spite of these challenges, it is believed that IGBONLP could benefit from the generalization of models and frameworks that target low resource languages to avoid reinventing the wheel. These models are likely going to address the problem of data sparsity.

### 2.3.1 Current Research Efforts

The work of Onyenwe [75] set the tone for the IgboNLP research. As shown in § 2.3.1, they made a lot of progress in the development of a sizeable Igbo corpus, a tagset and tagged corpus for IGBONLP research as well as a number of academic publications [77], [73], [75], [74] [76]. Also a considerable amount of effort has been put into the Igbo diacritic restoration task some of which have been reported in [32],[33],[34]. These will be briefly highlighted in § 2.3.1 and discussed throughout this thesis report.

#### Igbo POS Tagger

An Igbo tagging scheme, tagset and tagged corpus for IGBONLP research was developed [77] and extended [73] by creating a novel approach to refining the tagset and improving the quality of the manually tagged corpus using an inter-annotation agreement and a semi-automated process driven by transformation-based learning (TBL). Handling previously unseen words with "morphological reconstruction" improved the performance of the tagger on morphologically-complex words [75].

This work, which is still in progress, produced an automatic tagger for a standard version of Igbo language. Efforts are on-going to improve the robustness of the tagger across Igbo dialects and, hopefully in future, standardizing the entire process for other low-resource languages. Their key contributions, as presented in their final thesis report [74], are summarised below:

1. One million token Igbo corpus comprising two genres: the Bible and other literary text
2. Igbo tagsets in three categories: fine-grain (85 tags), medium-grain (70 tags) and coarse-grain (15 tags).
3. An Igbo sentence and word tokenizer.
4. A POS-tagged Igbo corpus: 263856 tokens from the New Testament and 39960 words from the literary text
5. A morphological parser based on a morphological reconstruction method.
6. An automatic part-of-speech (POS) tagger trained on the manually tagged Igbo New Testament Bible corpus.

### Igbo Diacritic Restoration

Igbo has orthographic and tonal diacritics, which capture distinctions in both meaning and pronunciation of words that are spelt the same way with the Latin characters. These diacritics are, however, often largely absent from the electronic texts we might want to process, or assemble into corpora. Therefore, there is a need for effective methods for automatic diacritic restoration for Igbo.

We have experimented with different approaches to the task and have also shared some of our findings at conferences. Brief descriptions of the contents of our conference publications are listed below:

1. *Automatic Restoration of Diacritics for Igbo Language* TSD2016 Brno, Czech Republic: we investigate a number of word-level diacritic restoration methods, based on n-grams, under a closed-world assumption, achieving a very high accuracy.

2. *Lexical Disambiguation of Igbo through Diacritic Restoration* EACL2017 Sense Workshop, Valencia, Spain: we modelled the problem as a classification task and applied machine learning methods. A number of machine learning algorithms were introduced and their individual performances on the task were compared.

3. *Learning Diacritic Embedding* WiNLP2018 Workshop, co-located with NAACL2018, New Orleans, Louisiana, US: word embedding models are commonly applied to NLP tasks but not for diacritic restoration. We introduced a new approach to diacritic restoration using embedding models which basically updates the original vectors of diacritic variants with a composition of the vectors of their co-occurring words.

4. *Igbo Diacritic Restoration using Embedding Models* NAACL-SRW2018, New Orleans, Louisiana, US: we applied word embedding models diacritic restoration by learning diacritic embeddings. We used two classes of word embeddings: *trained* and *projected* from the English embedding space and they performed well on the task

5. *Multi-task Projected Embedding for Igbo* TSD2018 Brno, Czech Republic: we created Igbo intrinsic evaluation datasets: *odd-word*, *word similarity* and *analogy*, for the embedding models we used to ensure the generalisability of the models. These were also tested on both the trained and projected embeddings.

6. *Transferred Embedding for Igbo Similarity, Analogy and Diacritic Restoration Tasks* SemDeep-3 Workshop COLING2018, Santa Fe, New-Mexico, USA: on the transferred learning, we expanded the dataset beyond the Igbo bible and standardized our n-gram models for the diacritic restoration task.

## 2.4 Chapter Summary

In this chapter, we presented an overview of Igbo people and culture as well as the historical development of Igbo language script and orthography. Other characteristic features of Igbo language that were presented include: phonology, tonology, vocabulary and morphology. Diacritics occur in the orthography in the form of dots on certain letters (i.e. **ị, ọ, ụ, ṅ**) and also in the tonology, we can have high (´), low (`) and downstep or mid (¯) tones on all vowels (**a, e, i, o, u, ị, ọ, ụ**) and Igbo nasal consonants (**m, n**).

We introduced IGBONLP as a platform for developing NLP resources for Igbo language and also presented our research efforts up to now. IGBONLP originated with the works of Onyenwe *et al.* which was mostly articulated in the thesis report [74] and a journal publication [76]. On the diacritic restoration task, we also presented our efforts in developing and effective strategies for the task as presented at conferences and workshops. In the next chapter we shall discuss in detail the procedures adopted for building our n-gram models for diacritic restoration.

# Chapter 3

# Review of Diacritic Restoration Methods

Diacritic restoration is not a very popular task within the NLP community. This is partly because there are little or no diacritics in English and most of the high resource languages and also because there is not much NLP research on languages with diacritics. In this chapter, we present a review of key literature on automatic diacritic restoration systems (ADRS) with particular attention on the methods and data used.

## 3.1 What are diacritics?

Diacritics[1] are simply defined as marks placed over, under, or through a letter in some languages to indicate a different sound value from the same letter without the diacritics. The word *diacritics* was derived from the Greek word *diakritikós*, meaning *distinguishing*[2]. Typical diacritic marks such as the acute (´) and grave (`) are commonly referred to as accents. However, there are other diacritics, such as dots (ọ, ṅ), cedilla(ç), slashed (ø), macron(m̄), tilde (õ), and so on. Diacritics may appear above or below a letter, or in some cases, even between letters (e.g. o͡o).

The seeming low interest in this area of language processing research may be attributed to the lack of diacritics in English language which is arguably the most dominant language on the web [39] and therefore, the most researched language in NLP. It therefore appears that the clamour for ADRS is often from the researchers working on relatively less popular languages. Although English language does not have much diacritics (apart from a few borrowed words), many of the world's language groups (Germanic, Celtic, Romance, Slavic, Baltic, Finno-Ugric, Turkic etc), as well as many African languages, use a wide range of diacritized letters in their orthography. As shown in the figure above, Milhalcea & Nastase [62] listed the diacritics in 32 European languages with latin-based scripts.

[1]http://www.merriam-webster.com/dictionary/diacritic
[2]http://dictionary.reference.com/browse/diacritic

| Language | Diacritics | Language | Diacritics |
|---|---|---|---|
| Albanian | ç ë | Italian | à é è í ì ï ó ò ú ù |
| Basque | ñ ü | Lower Sorbian | ć č ě ł ń ŕ ś š ź ž |
| Breton | â ê ñ ù ü | Maltese | ċ ġ ħ ż |
| Catalan | à ç è é í ï l· ò ó ú ü | Norwegian | å æ ø |
| Czech | á č d' é ě í ň ó ř š t' ů ú ý ž | Polish | a̧ ć ȩ ł ń ó ś ź ż |
| Danish | å æ ø | Portuguese | â ã ç ê ó ô õ ü |
| Dutch | á à â ä é è ê ë i í ì î ï ó ò ô ö ú ù û ü | Romanian | â ă î ş ţ |
| English | none | Sami | á ï č d- ń ņ š t- ž |
| Estonian | ä č õ ö š ü ž | Serbo-Croatian | ć č d- š ž |
| Faroese | á æ d- í ó øú ý | Slovak | á ä č d' é í ĺ ñ ó ô ŕ š t' ú ý ž |
| Finnish | ä å ö š ž | Slovene | č š ž |
| French | à â æ ç è é ê ë î ï ô œ ù û ÿ | Spanish | á é í ó ú ü ñ |
| Gaelic | á é í ó ú | Swedish | ä å ö |
| German | ä ö ü ß | Turkish | ç ğ i ı ö ş ü |
| Hungarian | á é í ó ö ő ú ü ű | Upper Sorbian | ć č ě ł ń ó ř š ž |
| Icelandic | á æ∂ é í ó ö ú ý ♭ | Welsh | â ê î ô û ŵ ŷ |

Fig. 3.1 Diacritics in European languages with Latin based alphabets. (*Source*: Mihalcea [62])

# 3.2 Why restore diacritics?

## 3.2.1 Resolving ambiguities

According to Yarowsky [112]

> *"...accent restoration is merely an instance of a closely-related class of problems including word-sense disambiguation, word choice selection in machine translation, homograph and homophone disambiguation, and capitalization restoration".*

In processing diacritic languages, the need to restore diacritics often arises given that most of the available texts for development miss a substantial amount of diacritics. Since diacritics could determine the meaning or the pronunciation of certain words, it follows that the correctness of grammar may be questionable without them in proper places.

It is also argued that besides deteriorating the language in its own right, the lack of diacritics causes a serious impediment to language processing tasks [104]. Most of the works in diacritic restoration, [62, 63, 111–113, 20] were inspired by the need to resolve lexical (syntactic and semantic) ambiguities caused by lack of, or wrong use of, diacritics especially in low-resource languages.

### 3.2.2 Improving diacritic text creation

Simard [94, 95], Tufiş & Ceauşu, [98, 99] and Šantić et al [104] through their works set out to develop tools such as REACC, AAI and DIAC+ that would standardise as well as speed up the creation of diacritic texts as both stand-alone application and plug-in tools to existing text editors. Simard [94] asserts that

> *"This type of feature could significantly facilitate the inputting of ... texts, ...in light of the lack of uniformity and ergonomic soundness in producing accents on computer keyboards."*

Also Scannell [91] stated their intention to:

> *...integrate our unicodification software into free text editors like Vim and OpenOffice.org, allowing users to enter text in plain ASCII and have the correct orthography appear on the screen "magically".*

### 3.2.3 Corpus construction from legacy texts

In many diacritic languages, existing texts were generally created and stored in non-standard formats. This non-standardization of texts persisted and has been tolerated because the absence of diacritics will very rarely render the text incomprehensible to the human reader [95]. Luu & Yamamoto [60] noted that such texts are often easy to read by human speakers but difficult to process using language tools due to a high level of ambiguity caused by lack of diacritics.

Though these texts are of immense value to native speakers, they are not useful in furthering NLP research and development due to their poor standards. Diacritic restoration, as suggested by Wagacha *et al* [105], can be used for effective recovery and standardization of such legacy texts to construct useful corpora.

### 3.2.4 Quicker integration of low resource languages

De Pauw *et al.* [24] explained in their work that, diacritic restoration process will help to shorten the development time for resource-scarce languages. This assertion agrees with the suggestion that most of the available pieces of non-diacritic texts on the web, can be used to make a substantial pool of training data for many language processing tasks [63, 60, 20].

Scannell [91] also argues that even in cases where the performance of a diacritic restorer is not perfect, it tends to reduce the amount of manual correction needed to create a high-quality corpus.

### 3.2.5 Printing and publishing

Turčić *et al.* [101] argues that “printing text implies printing letters as well...”. Editors of international publications, especially those that may contain diacritics, may require a tool that can restore the original intents and meanings of words. Such languages span Africa, South America, Asia and, of course, Europe. To omit diacritics in such publications amounts to underestimating their importance.

### 3.2.6 Internationalization and localization of tools

*Internationalization* and *localization* refer to software product design considerations that enable easy adaptation of tools for target audiences that vary in culture, region or language. It is believed that in spite of the popularity of the concepts, very little has been achieved in the area of language support especially for low resource languages. In fact, web publishers choose in many cases to avoid diacritics for reasons of “simplicity, uniformity or lack of means” [63]. Also Brezina [14] argues in his work *On Diacritics* that:

> “*...despite the proliferation of new multilingual typefaces, many still do not support some European languages, let alone cater for African and Asian languages. In fact, contrary to the claims of advertisements, the offering is, in respect to language support, quite limited*”.

### 3.2.7 Search engines

Using diacritic text in a search query may likely not return the results from ASCII based files and *vice verse*. An example mentioned in [91] is that of the Irish language. It was observed that in the 1990s, an acute accent (*síneadh fada*) in Irish was often typed as a forward slash following the vowel (*si/neadh fada*, for example). Because of this, some of the largest repositories of Irish language material on the web are essentially invisible to the standard search engines

## 3.3 Diacritic Restoration and Sense Disambiguation

Word sense disambiguation (WSD) is a popular task in NLP which involves "*the computational identification of meaning for words in context*" [68]. Historically, WSD emerged as a subtask of the machine translation (MT) in the 1950s [106]. It was then said to be *AI-complete* [61] given that it presents some obvious challenges. There were different formalizations to address issues of sense representations, sense granularity, domain-based versus multi-genre, target words for disambiguation.

Although diacritic restoration does not feature as frequently as other tasks in NLP research, it shares similar properties with such tasks as word sense disambiguation with regards to resolving both syntactic and semantic ambiguities [111]. Indeed it was referred to as an instance of a closely related class of problems which includes word choice selection in machine translation, homograph and homophone disambiguation and capitalisation restoration [112].

Diacritic restoration is similar to WSD in the sense that it is not an end in itself but an intermediate task which supports better understanding and representation of meanings in human-machine interactions. In most non-diacritic languages, WSD systems can directly support such tasks as machine translation, information retrieval, text processing, speech processing etc. [50]. Also, like WSD, diacritic restoration also applies computational techniques to disambiguating a non-diacritic word that could have diacritic variants and, by so doing, have multiple meanings or pronunciations.

However, WSD relies heavily on human annotated resources like dictionaries and sense inventories which are not only too expensive and time-consuming to create [69, 30] but are also, in themselves, products of subjective human judgements. On the other hand, the diacritic restoration task can be designed to be mostly data-driven and completely automated. The diacritics, which are the major distinguishing elements, are largely part of the characters. This makes the formalization of the task a bit more defined than in the case of WSD.

So for languages with diacritics, diacritic restoration could be a very essential pre-processing task that could guarantee better results with other NLP systems including WSD. However, we note that diacritic restoration does not eliminate the need for WSD. For example, if the wordkey *akwa* is successfully restored to *àkwà*, it could still be referring to either *bed* or *bridge.* Another good example is the behaviour of *Google Translate* as the context around the word *àkwà* changes.

| **Statement** | ***Google Translate*** | **Comment** |
|---|---|---|
| *Akwa* ya di n'elu *akwa* | It was on the high | confused |
| *Akwa* ya di n'elu *akwa* ya | It was on the bed in his room | fair |
| **Ákwà** ya di n'elu **àkwà** | His clothing was on the bridge | okay |
| **Ákwà** ya di n'elu **àkwà** ya | His clothing on his bed | good |

Table 3.1 Disambiguation challenge for *Google Translate*

The last two statements, with proper diacritics on the ambiguous wordkey *akwa* seem both correct. Some disambiguation system in Google Translate must have been used to select the right form. However, it highlights the fact that such a disambiguation system may perform better when diacritics are restored.

## 3.4 Diacritic Restoration Systems

Automatic Diacritic Restoration Systems (ADRS) are tools that enable the restoration of missing diacritics in texts. Many forms of such tools have been proposed, designed and developed. While some ADRS work on existing texts, others insert appropriate diacritics "on-the-fly" during text creation [113]. In this review, we shall mainly be looking at the two common approaches to automatic diacritic restoration for the Latin based text described in literature: *word based* and *character based.* Also, we will be highlighting the a few works on the restoration of texts written with the Arabic script.

### 3.4.1 Techniques for Word Level Approaches

Word based approach, as reported by Cocks & Keegan [16], Simard [94, 95], Yarowsky [111–113], Crandall [20], Tufiş & Chiţu [99], Tufiş & Ceauşu [98], often relies on an extensive collection of lexical resources and language models as well as already existing processing tools for such tasks as tokenisation, tagging and and other lexical analysis to produce good results. It also requires a comprehensive and robust dictionary of properly marked words that can be used to replace unmarked ones.

#### Decision List

One application of decision list to ADRS is presented by Yarowsky[112] who views accent restoration as an instance of a class of disambiguation problems which requires the resolution of both semantic and syntactic ambiguity and offers an objective ground truth for automatic evaluation.

| Stripped word | Patterns | % | Number |
|---|---|---|---|
| cesse | cesse | 53% | 669 |
| | cessé | 47% | 593 |
| cout | coût | 100% | 330 |
| couta | coûta | 100% | 41 |
| coute | coûté | 53% | 107 |
| | coûte | 47% | 96 |
| cote | côté | 69% | 2645 |
| | côte | 28% | 1040 |
| | cote | 3% | 99 |
| | coté | <1% | 15 |

Table 3.2 Corpus Analysis for Pattern Distribution

| Pattern | Context | | |
|---|---|---|---|
| (1)côté | du laisser de | **cote** | faute de temps |
| (1)côté | appeler l'autre | **cote** | de l'atlantique |
| (1)côté | passe de notre | **cote** | de la frontiere |
| (2)côte | vivre sur notre | **cote** | ouest toujours verte |
| (2)côte | creer sur la | **cote** | du labdrador des |
| (2)côte | travaillaient cote a | **cote** | , ils avaient |

Table 3.3 Sample instances of the training data for the ambiguous words

The corpora used in this experiment were the Spanish AP Newswire (1991-1993, 49 million words), the French Canadian Hansards (1986-1988, 19 million words) and a collection from *Le Monde* (1 million words). He proposed a 7-step approach to diacritic restoration which are described below:

- **Step 1: Identifying ambiguities** – Analyse the corpus and generate and a table of accent pattern distribution (see Table 3.2).
  Unambiguous words were replaced with their usual accent patterns but the ambiguous words were further processed as described in Steps 2-5.
- **Step 2: Building training context** – This is done by: 1. extracting $\pm k$ words from both sides of each ambiguous word; 2. labelling the instance with the observed accent pattern; 3. stripping the accents from all the instance words.
  Sample instances of the training data for the ambiguous words *cote* are as shown in Table 3.3 below:
- **Step 3: Measuring collocation distribution** – Distributions of collocations of the ambiguous tokens are computed for the instances:

| Position | Collocations | côte | côté |
|---|---|---|---|
| -1 w | du *cote* | 0 | 536 |
| | la *cote* | 766 | 1 |
| | un *cote* | 0 | 216 |
| | notre *cote* | 10 | 70 |
| +1 w | *cote* ouest | 288 | 1 |
| | *cote* est | 174 | 3 |
| | *cote* du | 55 | 156 |
| +1w,+2w | *cote* du gouvernement | 0 | 62 |
| -2w,-1w | cote a *cote* | 23 | 0 |
| $\pm k$ w | poisson (in $\pm k$ words) | 20 | 0 |
| $\pm k$ w | ports (in $\pm k$ words) | 22 | 0 |
| $\pm k$ w | opposition (in $\pm k$ words) | 0 | 39 |

Table 3.4 Sample collocations and their distributions

- **-1W:** Immediate preceding word
- **+1W:** Immediate following word
- $\pm k$ **W:** word found in $k$ words window[3]
- **-2,-1 W:** Offsets -2 and -1 words
- **+2,+1 W:** Offsets +2 and +1 words

Table 3.4 shows a sample of these collocations and their distributions: Yarowsky noted that these core sets of evidence assume no additional language specific knowledge but where that exists, it may likely improve the information base.

- **Step 4: Log-Likelihood Ranking** – The collocation evidences generated above are ranked as decision lists using the log-likelihood ratio given as:

$$abs\left(log\left(\frac{P(accent_1|collocation_i)}{P(accent_2|collocation_i)}\right)\right)$$

This strategy lists the strongest and most reliable evidence first as seen in 3.5: Lines 2,3 and 7 in Table 3.5 will be pruned in the next step.

- **Step 5: Pruning and Interpolation** – Pruning the list increases efficiency. Lines 2 and 3 are satisfied by line 1 while line 5 deals with line 7. So word classes preceded by their members are pruned. Also using cross-validation, lines that contribute more incorrect classifications than correct ones can be removed.

[3]Optimal value of $k$ depends on the type of ambiguity, for semantic and topic based ambiguities, $k \approx 20 - 50$; for local syntactic ambiguities $k \approx 3$ *or* 4

| # | LogL | Evidence | | Classification |
|---|---|---|---|---|
| 1. | 8.28 | prep + "que" *terminara* | ⇒ | terminara |
| 2. | 7.24 | de que *terminara* | ⇒ | terminara |
| 3. | 7.14 | para que *terminara* | ⇒ | terminara |
| 4. | 6.87 | y *terminara* | ⇒ | terminará |
| 5. | 6.64 | weekday (within $\pm k$ words) | ⇒ | terminará |
| 6. | 5.82 | noun+"que" *terminara* | ⇒ | terminará |
| 7. | 5.45 | domingo (within $\pm k$ words) | ⇒ | terminará |

Table 3.5 Sample decision list *Source* [112]

- **Step 6: Training the decision list** – On the assumption that for similar types of ambiguities, there are similar basic classification evidences, a general class decision list for similar types of ambiguities is trained. The comparative accuracies determine the cases that should apply the class decision list and those that should apply their individual list.
- **Step 7: Using the decision list** – The decision list created in Steps 1–6 is used to determine accent patterns in new texts. Unambiguous words are simply replaced by their valid patterns while ambiguous words have a table of possible patterns and a pointer to a decision list for either the particular case or its ambiguity class.

Statistically, the highest ranking evidence will most likely disambiguate the target word. Other evidences are inefficient and rarely improve accuracy and so are often ignored. Test data are realised by stripping accents from "correctly" accented texts. The original corpus may contain some errors which may affect the performance of the algorithm. The evaluation technique is a simple comparison of the restored patterns against the corresponding original actual patterns in the test data.

The baseline algorithm chooses the most common pattern. The last column shows the number of occurrences of each of the ambiguity instances in the test corpus. 80% of the corpus was used for training while 20% was held-out for testing on a 5-fold cross validation process. The results for the most problematic cases in both languages are shown in Table 3.6.

Yarowsky argues that the key advantage of this approach to other works is its ability to combine multiple, non-independent evidence types e.g. root form, parts of speech, thesaurus category or application-specific clusters. It is also more cost effective to create the training corpus for this technique than other techniques such as part-of-speech tagger because the test data is produced by stripping diacritics from the original properly marked text.

| Pattern1 | Pattern2 | Result | Baseline | # |
|---|---|---|---|---|
| **Spanish** | | | | |
| anuncio | anunció | 98.40% | 57% | 9456 |
| registro | registró | 98.40% | 60% | 2596 |
| marco | marcó | 98.20% | 52% | 2069 |
| completo | completó | 98.10% | 54% | 1701 |
| retiro | retiró | 97.50% | 56% | 3713 |
| duro | duró | 96.80% | 52% | 1466 |
| paso | pasó | 96.40% | 50% | 6383 |
| regalo | regaló | 90.70% | 56% | 280 |
| terminara | terminará | 82.90% | 59% | 218 |
| Ilegara | Ilegaraá | 78.40% | 64% | 860 |
| deje | dejé | 89.10% | 68% | 313 |
| gane | gané | 80.70% | 60% | 279 |
| secretaria | secretariá | 84.50% | 52% | 1065 |
| seria | sería | 97.70% | 93% | 1065 |
| hacia | haciá | 97.30% | 91% | 2483 |
| esta | está | 97.10% | 61% | 14140 |
| mi | mí | 93.70% | 80% | 1221 |
| **French** | | | | |
| cessé | cesse | 97.70% | 53% | 1262 |
| décidé | décide | 96.50% | 64% | 3667 |
| laisse | laissé | 95.50% | 50% | 2624 |
| commence | commencé | 95.20% | 54% | 2105 |
| côté | côte | 98.10% | 69% | 3893 |
| traité | traite | 95.60% | 71% | 2865 |

Table 3.6 Sample results for the most difficult cases in both languages, *Source* [112]

### Hidden Markov Model

Simard [94] applied the Hidden Markov model (HMM) in their accent restoration technique for French. 85% of French words are normally unaccented and about half of the accented words are unambiguous. So with a good dictionary and a baseline strategy of using the most frequent words, an accuracy of up to 97.5% can be achieved. To improve on this, Simard implemented two basic steps: *hypothesis generation* and *candidate selection*.

Hypothesis generation produces valid alternatives for the target word using a list of over 250k valid words from a French morpho-syntactic electronic dictionary. If an unknown word is encountered, reproduce it verbatim. Candidate selection chooses the candidate that maximises the likelihood of the output sentence. The language model used for this task is a 2-tag HMM with a sequence of tags from 350 tags from the French morpho-syntactic electronic dictionary. The model is defined by the parameters:

- $P(t_i|h_{i-1})$: the probability of tag $t_i$ given the previous tag history.
- $P(w_i|t_i)$: the probability of word $w_i$ given tag $t_i$. Given the above parameters, tag set $T$ and $T^n$ then:

$$P(w) = \sum_{t \in T^n} \prod_{i=1}^{n} P(t_i|h_{i-1})P(w_i|t_i)$$

An illustration of how the above process works as presented by Simard is demonstrated with Figure 3.2 below.

Sentence segments and sub-segments are syntactically independent. Sub-segments are regions within the sentence separated by semi-colons, commas etc. or regions of "low ambiguities". Restoring accents on segment words fixes the entire text. The negative impact of "sub-optimal" segmentation is minimised by prepending the last few words of the previous segment to the current segment.

Over 250,000 valid words extracted from the French morpho-syntactic electronic dictionary were used. The HMM probabilities were first estimated by direct frequency counts on a 60k word, hand-tagged extract of the Canadian Hansard and were then refined on a 3 million word extract from documents of Hansards, Canadian National Defence documents and French press revues.

Evaluation involves counting the number of words, from the "re-accented" stripped version, that differ from their corresponding words in the original text. The test corpus is a multi-genre collection distinct from the training data totalling 57 966 words. A key factor in the process is the maximum number of combinations per sub-segment used in the segmentation heuristic (called $S$). Table 3.7 indicates that $S$ values of 16 or 32 gave optimal results.

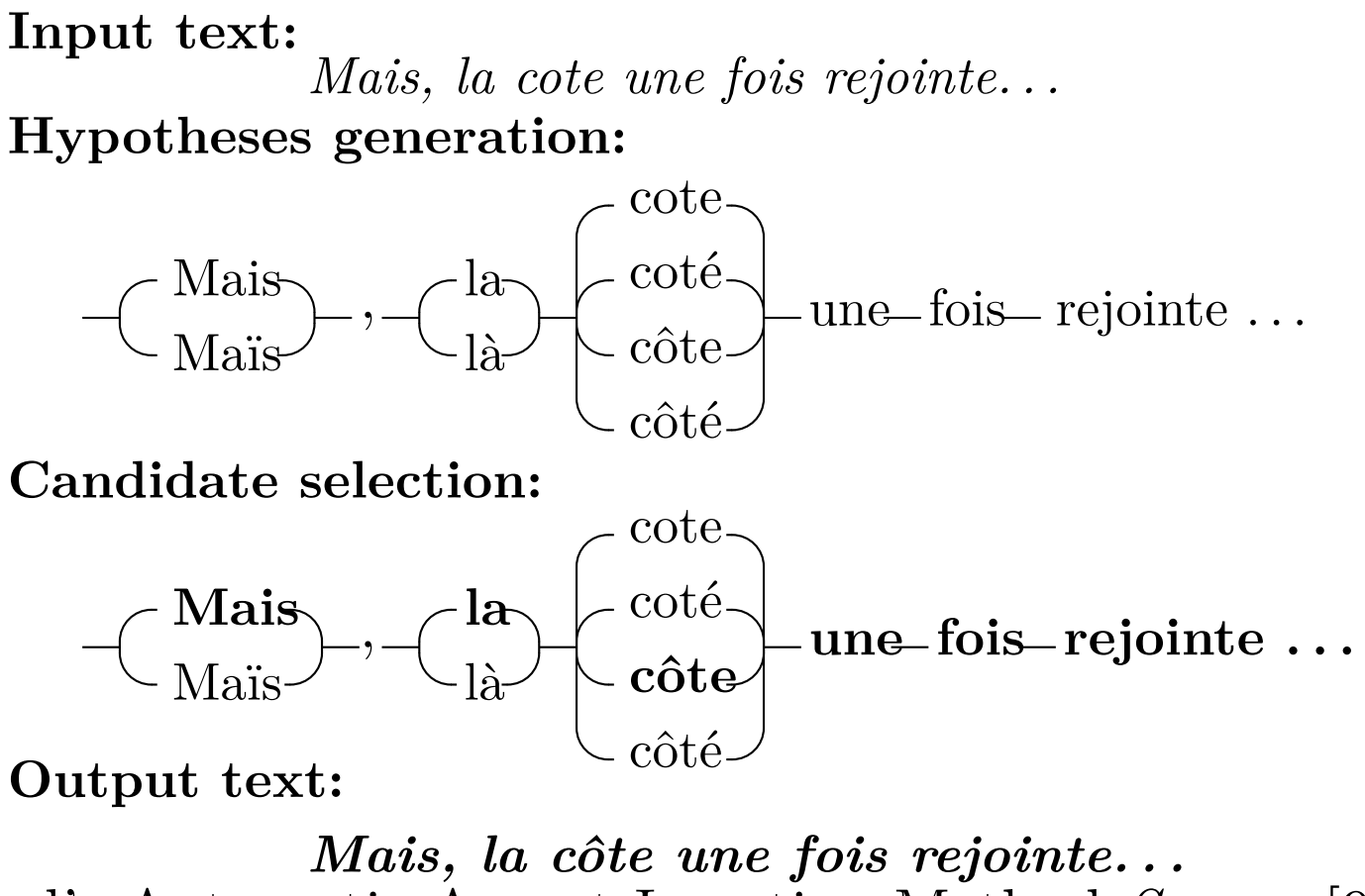

Fig. 3.2 Simard's Automatic Accent Insertion Method *Source* [94]

| Max comb | RunTime | Errors | AvgErrDist |
|---|---|---|---|
| 2 | 68 | 821 | 70 |
| 4 | 85 | 560 | 103 |
| 8 | 132 | 466 | 124 |
| **16** | **169** | **441** | **130** |
| **32** | **277** | **429** | **134** |
| 64 | 429 | 425 | 136 |
| 128 | 731 | 420 | 137 |

Table 3.7 Results of AAI Experiments on 58K-word Test Corpus *Source* [95]

Table 3.8 provides a rough classification of errors with $S = 16$ with the highest as the *e* vs *é* and also the inability of the generator to provide the correct word forms.

The two major previous works that were compared with this are:

1. **El-Beze *et al.***[26] This work also adopted the concept of *hypotheses generation* and *candidate selection.* However, instead of word segments, it processes texts from left to right using a fixed width sliding window as follows:

   - For each word $w_i$, generate a list of possible *word/tag* alternatives i.e.

$$(w_{i1}, t_{i1}), \ldots, (w_{ik}, t_{ik})$$

| Error type | Occurrences | Percentage |
|---|---|---|
| endswith -*e* vs -*é* | 171 | 38.8% |
| unknown words | 111 | 25.2% |
| *a* vs *à* | 69 | 15.7% |
| others | 90 | 20.4% |

Table 3.8 Classification of Accent Restoration Errors ($S = 16$) *Source:* [95]

- Find the sequence of *word/tag* pairs that maximises the probability:

$$\prod_{i=1}^{n} P(w_{ij_i}|t_{ij_i})P(t_{ij_i}|t_{i-1j_{i-1}}, t_{i-2j_{i-2}})$$

- Approximate with the first three terms (i.e. only the product of $P_i, P_{i+1}$ and $P_{i+2}$) to avoid combinatorial explosions.
- The process is from left to right i.e. optimal tag for position $i$ is used to compute for positions $i + 1$ and

El-Beze *et al.*[26] reported success levels slightly superior to the work done by Simard [95]. This may be due to their 3-tag HMM (Simard [95] used 2 tag2) and a relatively small test-corpus (a little over 8000 words). Also the performances varied wildly from text to text, with *average distances* between errors between 100 and 600 words.

2. **Yarowsky**[112] Yarowsky's method applied decision lists to exploit different local and global context information around the ambiguous word (i.e., words within a $k$ window around the current word), part-of-speech of surrounding words, etc. Their system chooses a candidate with the single most reliable piece of evidence in the list.

   However, this work focuses more on Spanish than French. Also, the evaluation focuses on specific ambiguities, from which it is impossible to get a global performance measure. Again, this approach performed poorly on "syntactically interchangeable" candidates[4].

### Bayesian Framework and Hidden Markov Model

Crandall[20] described two statistical algorithms for diacritic restoration of Spanish – Bayesian framework and Hidden Markov Model – as well as a hybrid of both techniques.

[4]These refer to candidates with similar morpho-syntactic features. Simard also claims that the use of *ad hoc* tags would weaken the performance of the Yarowsky's model.

Their work also measures the effect of the quality and size of the training corpus on the performance of the proposed algorithms. Previous works cited and their shortcomings include:

- Yarowsky [113] applied Bayesian classifiers, HMMs and decision lists but the task was for only a few set of words;
- Simard & Deslaurier [94] used HMM and POS-tags but creating the training data is a costly process and the method could not resolve ambiguities from same parts of speech.
- Galicia-Haro *et al* [38] adopted a rule-based approach that separates nouns and verbs by looking at the context information and Spanish noun-adjective agreement rules but it could not deal with words of the same part-of-speech.
- Mihalcea & Natase [63] introduced a letter level approach for languages without lexicons, but this method handles characters and doesn't reflect word level accuracies.

Crandall's [20] work focused only on the acute accents and identified the following categories:

- **Unambiguous words**: have only one correct accentuation pattern which is restored by looking up the correct form in a lexicon.
- **Ambiguous words**: have multiple patterns and are further classified as:
    1. ***Conjugations of the same verb***: Candidate patterns are all verbs of the same sense but with differences in tenses, moods and persons e.g. *canto* (I sing) and *cantó* (he/she sang).
    2. ***Different parts of speech***: Candidates are of different parts of speech e.g. *de* (preposition, $of$) and *dé* (imperative verb, *give*)
    3. ***Same part of speech***: Candidates are of the same part-of-speech but not conjugated verbs e.g. *papa* (noun, *potato*) and *papá* (noun, $father$)

In Crandall's work [20], three basic techniques were reported: a Bayesian Framework, an HMM framework and a hybrid of both techniques:

**Bayesian Framework:** The motivation for this method is that, according to Yarowsky [111], words co-occurring with an ambiguous stripped word may give clues to its correct diacritic form. The model assumes that if a stripped word, $w_0$, has multiple diacritic forms $a_0, a_1, \ldots, a_n$, the most likely form can be chosen by looking at other surrounding words within a specified window size.

Then the probability of a given diacritic form, $a_i$ within a window size of $s$ is defined as $P(a_i|w_{-s}, \ldots, w_{-1}, w_1, \ldots, w_s)$ and the task is to select the form with the maximum probability i.e. chose $a_i$ if:

$$i* = argmax_{0 \leq i \leq n} P(a_i|w_{-s}, \ldots, w_{-1}, w_1, \ldots, w_s)$$

Using Bayes rule on an unordered bag-of-words model, we have:

$$\begin{aligned}
i* &= argmax_{0 \leq i \leq n} P(a_i|w_{-s}, \ldots, w_{-1}, w_1, \ldots, w_s) && (3.1)\\
&= argmax_{0 \leq i \leq n} \frac{P(w_{-s}, \ldots, w_{-1}, w_1, \ldots, w_s|a_i)P(a_i)}{P(w_{-s}, \ldots, w_{-1}, w_1, \ldots, w_s)} && (3.2)\\
&= argmax_{0 \leq i \leq n} P(a_i) \prod_{-s \leq j \leq s} \frac{P(w_j|a_i)}{P(w_j)} && (3.3)\\
&= argmax_{0 \leq i \leq n} P(a_i) \prod_{-s \leq j \leq s} P(w_j|a_i) && (3.4)
\end{aligned}$$

where:

- $P(w_j|a_i)$ is the probability that a word $w_j$ appeared within the context of $w_0$ with accent pattern $a_i$.
- $P(w_j|a_i)$ and $P(a_i)$ can be directly estimated from the training data.
- The denominator is dropped in equation (4) since it is independent of the accent pattern being considered.
- A simple smoothing technique sets zero probabilities to a small non-zero constant.
- The window size parameter $s$ is specified at the beginning with varying values applied.

**HMM Method:** This method is motivated by its successful application to POS-tagging and the effect of POS-tags on diacritic restoration. Crandall [20] applied the morphological information from the words as suggested by Yarowsky [112]. The tag of a word can often be inferred from the word suffixes alone e.g. words ending in *mente* are almost always adverbs while those ending in *ar*, *er* and *ir* are mostly verb infinitives.

Although this assumption is not perfect, simple pattern-matching rules can be used to tag most words and generate a labelled training corpus for HMM training procedure. Viterbi decoding is then performed on a per-sentence basis to extract the set of possible labels for each stripped word. The HMM model is then trained and applied to assign the most likely tag to the ambiguous word while the corresponding accent pattern is chosen.

For example, in the Spanish sentence *Ella canto* with the ambiguous word *canto*, the morphological analyser suggests that *Ella* is a third person pronoun and that *canto* can either be a first-person present verb (*canto*) or a third-person past verb (*cantó*). The HMM model therefore concludes that it is more likely to transit from a third-person pronoun to a third-person verb and so *cantó* is selected.

**Hybrid approach:** This deals with the problems of the two methods described above. The Bayesian approach ignores word order and also performs poorly for infrequent words. The performance of the HMM model depends on the accuracy of the morphological analysis and does not deal with words from the same part-of-speech.

The hybrid approach takes advantage of the strengths of the two techniques by choosing the method that works best for each ambiguity set. Both techniques were trained and also tested on the same training data. For each stripped word, the performance of each technique is noted and the better model is used to disambiguate the word on the test data. The test process involves alternating between the two models depending on the stripped word in view.

**Experimental data:** Web-crawled data was used for training and testing from professional websites such as newspapers, government offices and religious institutions. Preprocessing operations –removing documents with below 3% diacritic content and deleting parts of a document written in other languages – were performed on the resulting corpus to improve its orthographic qualities.

A total of 35,318,775 words (with 350,592 unique words), excluding punctuation, was collected. These comprise articles, legal and scientific documents, essays and encyclopaedic articles, works of literature and religious articles.

**Evaluation and Results:** Accuracy is the fraction of words in the output of a model that match corresponding words in the original version. Two baseline models were used:

- - *no_accents*: returns the stripped words as received, 89.22% accuracy while
- - *most_likely* chooses the most common pattern gave 98.82%.

The Bayesian model used different window sizes: $\pm 1, \ldots, \pm 10$, $\pm 12$, $\pm 15$ and $\pm 30$ as well as asymmetric window sizes: $-3$, $-2$, $-1$, $+1$, $+2$ and $+3$. The best accuracy, 99.118%, was achieved with a window size of $\pm 2$. However, an optimal performance of 99.211% was achieved by using the optimal window size for each ambiguity set.

With the HMM method, words were classified into 98 classes - 52 function words, 45 suffix groups and a default class - using a regular expression pattern matching scheme. An accuracy of 99.05% was recorded. The hybrid approach alternates between the two on a per-word basis using the better technique and parameter for each word. It achieved an accuracy of 99.24% on the test data.

Crandall made the following observations from the results analysis:

- *Corpus Genre:* The techniques are genre sensitive. The literary categories were the most difficult with HMM performing best while the least challenging were the medical encyclopedic articles with the hybrid doing better than the others.
- *Corpus Size:* Small training data (<500 tokens) affected the performance of HMM and hybrid negatively. Good performance began from data size of about 10 million tokens and kept the upward trend even with upto 35 million tokens and beyond.
- *Corpus Pollution:* The orthographic quality of the training data may also cause sub-optimal learning for algorithms and decrease their performance.

The closest comparison to Crandall's work [20] is Yarowsky's [112] but there are differences in methods. Although Yarowsky's work appears to have performed better in the few words examined, their methods were significantly simpler.

- *Context:* Yarowsky's context around an ambiguous word are well accentuated words while Crandall's work used fully stripped context words. This is more realistic and when Yarowsky's method was applied on the best algorithm (HMM+Bayesian), an overall 28% error reduction was recorded.
- *Complexity:* Yarowsky considered ambiguity pairs while this work uses ambiguity sets.
- *Genre:* Yarowsky's corpus comprises only one genre i.e. newspaper articles while this work used a diverse collection of different genres.

According to Crandall [20], future works were expected to include:

1. Replacing the pattern matching algorithm with a full POS tagger
2. Replacing the bigram model with a trigram model in the tagging process
3. Using a hybrid model that allows the two models to vote on a candidate i.e instead of using one or the other.
4. Creating a real-time version for word processing or email applications.

.

| Character | Possible substitutes |
|---|---|
| **č, ć** | c, cc, ch, cx, cy |
| **š** | s, ss, sh, sx, sy |
| **ž** | z, zz, zh, zx, zy |
| **đ** | d, dj, dy |

Table 3.9 Common diacritic substitutions in Croatia. *Source:*[71]

### Dictionary and Language Models

Nikola *et al.* [71] presented an automatic diacritic restoration system for Croatian texts that combines dictionary look-up and statistical language modelling. Croatian has five diacritic characters: č, ć, š, ž and đas well as a diacritic in the digraph dž. Diacritic characters are commonly represented with a combination of ASCII characters to resolve ambiguity thereby producing less readable texts e.g. writing *čišći* (cleaner) as *cxisxcxi*. Common substitution schemes are shown in Table 3.9 below:

The experimental data were collected from 100k words of newspaper articles and 30k words of forum posts which were then grouped into:

1. *valid:* newspaper articles assumed to have correct diacritics
2. *mixed:* discussion forum posts with both diacritics and substitutions
3. *removed:* both newspaper articles and discussion forum with diacritics stripped

This study then classified Croatian words into these two overlapping categories:

- **C-words:** words that are *potentially* diacritic i.e. contain at least one of *c*,*s*,*z* or *d*. They are to be considered for restoration.
- **D-words:** words that are *actually* diacritic i.e. contain at least one of *č*, *ć*, *š*, *ž*, *đ*.

Note that a word can be a *C-word* as well as a *D-word* and all *D-words* collapse to *C-words* when stripped of diacritics. Also the restoration task involves the replacement of substitute patterns (e.g. using another ASCII character to indicate diacritic presence) where they exist. Table 3.10 shows the analysis of diacritic distribution data used.

- Valid text(Newspaper articles):
  - The *C-word* ratio indicates that over half (45.7%+16.3%) of the words are correct, needing no further processing.
  - The *D-word* ratio shows that only 1/6 (16.30%) of the words contain a diacritic character, the rest (5/6) are already correct without diacritics.

| Class | Valid | Mixed | Removed |
|---|---|---|---|
| **C-Word content** | 45.70% | 48.50% | 53.80% |
| **D-Word content** | 16.30% | 10.10% | – |
| **Substitutes** | 10.50% | 12.20% | 14.10% |
| **Diacritics** | 3.20% | 1.30% | – |

Table 3.10 Statistical analysis of diacritics on the data classes. *Source:*[71]

  – There were generally fewer substitute characters than in other texts and the diacritic content is highest

- Mixed text (Forum posts):
  – The *C-word* ratio expectedly increased due to missing diacritics
  – The *D-word* ratio decreased for the same reason
  – Substitute characters increased and the diacritic content reduced by more than half
- Removed (stripped valid text):
  – Has 53.8% *C-word* content, no *D-word*s, 14.1% substitute characters

Also, on valid *C-word* variants: 88.6% are unambiguous, 7.4% 2-variants and 0.1% 3-variants. The rest, 3.9% are not in the dictionary, either mispelt or uncommon. Given a wide coverage dictionary, most *C-word*s are easily restored. An illustrative description of their proposed restoration model is shown below:

- Tokenization:
  - Input lines (could be more than a sentence) split into word tokens; left and right context with varying window size
- Candidate generation:
  - Non *C-word*s are assumed correct; variants of a *C-word* are generated using all diacritic characters; a left to right process;
  - *C-word* variants of the right contexts required, increases exponentially with the context size; variants are validated using a 750k-entries dictionary.
- Selection of correct word form:
  - Baseline = dictionary restoration (up to 88% accuracy);

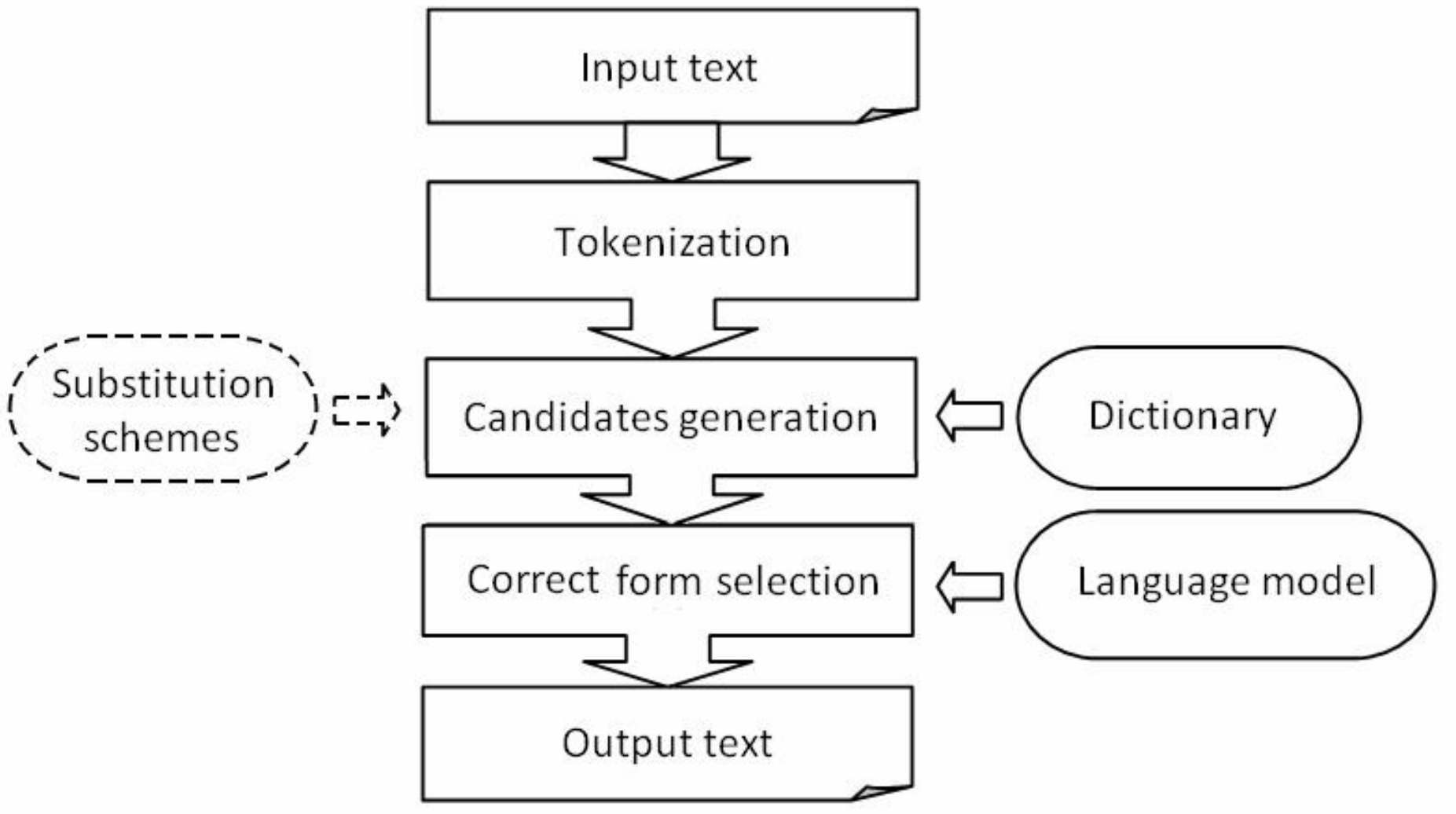

Fig. 3.3 Architecture of the Croatian Diacritics Restoration System. *Source:*[71]

| | Newspapers | Forum posts |
|---|---|---|
| *Unrestored* | 80.72% | 92.00% |
| *baseline(dict only)* | 97.07% | 97.07% |
| *Dict+LM* | 97.65% (+0.6%) | 98.38% (+0.4%) |
| *Dict+LM+Smoothing* | 98.81% (+1.8%) | 99.35% (+1.4%) |

Table 3.11 Results of the restoration on different corpora

- Ambiguity resolved using a bigram language model with a WB smoothing technique i.e.

$$\begin{aligned} P(w_1^n) &= P(w_1)P(w_2|w_1)P(w_3|w_1^2)\ldots P(w_n|w_1^{n-1}) \\ &= \prod_{k=1}^{n} P(w_k|w_1^{k-1}) \\ &\approx \prod_{k=1}^{n} P(w_k|w_{k-1}) \end{aligned}$$

The results (see Table 3.11) show that the best context window size is one word on each side. Also, incorrect restorations occur mostly on named entities in newspaper corpus and mostly as spelling errors in forum posts. Again, the use of language models also reduced the error frequencies on the most frequent mistaken C-Word variants.

Nikola *et al.* [71] work presented a system for diacritic restoration for Croatian that relies on a dictionary and a bigram language model, and is computationally inexpensive.

From the results, the dictionary look-up shows good results, but language models improved accuracy to almost 99%.

### Naïve Bayes Classifier

Cocks and Keegan [16] proposed a naïve Bayes classifier in developing a word-based diacritic restorer for Māori. Māori alphabet consists of 15 characters: 10 consonants and 5 vowels. The vowels are categorised into *short* (**a e i o u**) and *long* (**ā ē ī ō ū**).

Omitting the macron on a long vowel automatically defaults to its short version creating ambiguity e.g. *wāhine* (women) and *wahine* (woman). Since there are no electronic lexicons for Māori and the existing tools for resource rich languages cannot be easily adapted to Māori, a machine learning method that creates training instances from local word context is proposed.

Two baseline models were defined: the first assumes no diacritic markings exist while the second chooses the most frequent pattern from candidates. If the two competing candidates are observed equally, the model randomly chooses one of them.

Naïve Bayes classifiers apply Bayes theorem with the naive assumption of independence between features. A simple definition of a classification function $cf$ for a class variable $c$ and a dependent feature vector $x_1, \ldots, x_n$, using Bayes Theorem is given as:

$$cf = argmax P(c) \prod_{i=1}^{n} P(x_i|c)$$

Below is the equivalent form that avoids floating point underflow

$$cf = argmax \left( log P(c) + \sum_{i=1}^{n} log P(x_i|c) \right)$$

The *prior* probabilities for the parameters $P(c)$ and $P(x_i|c)$ were estimated on the training data. The relevant context as presented in the study may be explained thus:

Let $T = (t_1, \ldots, t_n)$ be a sequence of all tokens in the training data and $S = (s_1, \ldots, s_n)$ be the stripped version of T. Also let $D_T = dt_1, \ldots, dt_n$ be a set of distinct tokens in T and $D_S = ds_1, \ldots, ds_n$ be a set of distinct stripped tokens. If $f : ds_i \rightarrow cands \subset D_T$ then the goal is to maximise the probability for all words in $fds_i$. Therefore in relation to the Bayes formula above, the probability of a candidate *cand* in the set $cand_1, \ldots, cand_n$ is defined as:

$$P(cand) = \frac{C(cand)}{\sum_{i=1}^{n} C(cand_i)}$$

where $C(cand)$ is the number of occurrences of *cand* and $\sum_{i=1}^{n} C(cand_i)$ is the summation of the number of occurrences of each of the other candidates. Extending it further to include the features as well as smoothing, we have:

$$P(cand) = \frac{C(cand + feat_1) + 1}{\sum_{i=1}^{n} C(cand_i + feat_1)}$$

$C(cand+feat_1)$ is the number of occurrences of *cand* in T with some feature, $feature1$, and $\sum_{i=1}^{n} C(cand_i + feat_1)$ is the summation of occurrences of other candidates with $feature_1$. Estimates of zero were avoided with Laplace smoothing.

The feature sets for the machine learning algorithms were modelled after those used for the grapheme based approach Scannell [91]. Word-based n-grams relative to the target word are extracted as outlined below:

- **FSW1-Features(-1,1):** That is the monogram[5] preceding the target word.
- **FSW2-Features(-2,2):** That is the bigram preceding the target word.
- **FSW3-Features(-3,3):** That is the trigram preceding the target word.
- **FSW4-Features(1,1):** That is the monogram following the target word.
- **FSW5-Features(1,2):** That is the bigram following the target word.
- **FSW6-Features(1,3):** That is the trigram following the target word.
- **FSW7-Features(-1,1),(-2,2):** That is the monogram and bigram preceding the target word.
- **FSW8-Features(1,1),(1,2):** That is the monogram and bigram following the target word.
- **FSW9-Features(-1,1),(1,1):** That is the monogram on either side of the target word.
- **FSW10-Features(-2, 2),(-1,1),(1,1),(1,2):**That is the monogram and bigram on either side of the target word.

[5]Cocks *et al.* refer to *unigram* as *monogram*

| Item | Quantity | Percentage |
|---|---|---|
| Words with Diacritics | 859,083 | 20.06% |
| Unambiguous Words | 1,656,051 | 38.68% |
| Words+1 ambiguity | 2,345,874 | 54.81% |
| Words+2 ambiguities | 98,995 | 2.31% |
| Words+3 ambiguities | 179,788 | 4.20% |
| Total words | 4,281,708 | 100% |

Table 3.12 Statistical Distribution of Diacritic Ambiguities in the Corpus

- **FSW11-Features(-1,3),(-2,2),(1, 2),(-1,4),(-2,4):** A combination of features around the target word.

This study used a fully diacritically marked multi-genre corpus of about 4.2 million words. Table 3.12 shows the statistical distribution[6] of diacritic ambiguities in the data used for this study.

From the data, almost 80% of the training corpus have no diacritics. 39% of the total words have no ambiguities and can simply by replaced by looking them up in some lexicon. The task is to properly classify the remaining 61% into their appropriate forms as used in the text corpus.

Evaluation is done with a 10-fold cross validation method that partitioned that data into 10 subsets. Each round of the experiment used a different subset for testing and the average score is taken. The comparison of the results with the grapheme models indicate that the word based approach performed much better. FSW11 gave the best accuracy with a score of 99.01% and a 1.9% of the $Baseline2$ accuracy score. The scores are shown in the Table 3.13 below.

In this paper, Cocks and Keegan [16] attempted to contrast the version originally proposed by Scannell [91] and therefore reproduced their work using a larger, better quality corpus. Scannell, however, proposed a grapheme-based model while this is word-based models but the results suggest that the performances of the models proposed in this study are better than those of Scannell's under similar circumstances. For future work, several languages will be used to evaluate these methods with properly marked training data.

In their work, Tufiş and Chiţu [99] reported a word based approach based on POS tagging to restore diacritics in Romanian texts with an accuracy of 97.4%. Also in extending the original work on automatic diacritic insertion for Romanian texts,[99],

[6]This statistics do not seem to add up though...

| Model | Accuracy | Model | Accuracy |
|---|---|---|---|
| Baseline1 | 79.94% | Baseline2 | 97.11% |
| FSG1 | 79.94% | FSG2 | 79.94% |
| FSG3 | 84.45% | FSG4 | 87.02% |
| FSG5 | 95.07% | | |
| FSW1 | 98.50% | FSW2 | 98.33% |
| FSW3 | 97.94% | FSW4 | 98.28% |
| FSW5 | 98.34% | FSW6 | 98.01% |
| FSW7 | 98.65% | FSW8 | 98.54% |
| FSW9 | 98.65% | FSW10 | 98.85% |
| FSW11 | 99.01% | | |

Table 3.13 Accuracy scores for different models

Tufiş & Ceauşu built the DIAC+ [98] both as an MS-Word based DLL and a web service which deals with "unknown words" using character based back-off.

Atserias *et al* [7] argued, however, that despite the number of works done on diacritic restoration of Spanish, on an orthographically rich language such as Spanish, no spellchecker exists that tackles effectively the existence of word forms whose diacritic and non-diacritic versions are valid dictionary entries e.g. such as continuo 'continuous' and continuó 'he/she/it continued'. They went on to implement a simple bigram-model based solution and also an evaluation mechanism that prevents the biasing of high frequency.

### 3.4.2 Techniques for Grapheme Level Approaches

Grapheme level methods may not be as reliable as the word-based method in languages where diacritics have grammatical or semantic role but they appear to present better alternatives for low resourced languages. Mihalcea [62] observed that the word-based methods may not be very useful for languages with limited or no electronic dictionaries, no tools for morphological and/or syntactic analysis, no sizable diacritically marked corpora for training algorithms.

Various forms of the grapheme-based (or letter- or character-based) approach as reported by Mihalcea *et al* [62, 63], De Pauw *et al* [24], Wagacha *et al* [105], Nguyen [70] and Zweigenbaum [114] explore simple techniques to get around these deficiencies and speed up the process of developing resources for such languages. Its major appeal includes language independence, ease of implementation and non-requirement of expensive wide-coverage lexicons and processing tools.

### Instance-based Learning Algorithm

Mihalcea *et al* [62], in their works, which focused initially on Romanian but later extended to Czech, Hungarian and Polish,[63] presented not just a strong argument in favour of letter based learning for low resource languages but also a clear approach to data gathering and experimental setup. This work is motivated by the lack of diacritics on the large 75k-entries electronic dictionary and the Romanian corpora available for other NLP research. The core experimental data used a 3M-word corpus of Romanian texts with diacritics assembled from 2,780 articles of *România Literară*, a weekly Romanian publication of literary works which were downloaded, preprocessed and standardised.

Instance-based (also called *memory-based*) learning algorithm[21] was used for the experiments because it learns efficiently and considers every single training example when making a classification decision. The target attributes of learning are the ambiguous letters: s - ş , t - ţ , a - ă and i - î. Also a decision tree classifier, C4.5 [83] was used in this experiment and although similar results were obtained, it was slower in learning. At the letter level, an accuracy of 99% on the experiment with only Romanian [62] was obtained while an average accuracy of 98% was reported for four languages (i.e. Romanian, Czech, Hungarian and Polish) in a later experiment [63]. The key considerations and structure for this approach are as highlighted below:

- Core language tools – part of speech taggers, morphological or syntactic analysers – are not available.
- Surrounding words cannot be relied on due to data sparsity and many cases of unknown words.
- Surrounding letters are used with special notations assigned to white spaces, commas, dots and colons[7].
- The attributes in the training data are formed by $N$ letters to the left and right of the ambiguous letter, while the target attribute is the ambiguous letter itself[8].
- Training instances are based only on the ambiguous pair in question i.e.:
  - a – ă: 2,161,556
  - i – î: 2,055,147
  - t – ţ: 1,257,458
  - s – ş: 866,964

[7]These are generally considered special characters by C4.5 & Timbl.
[8]N=5 is the best accuracy in Mihalcea's experiments.

| | Ambiguous pair | | | | |
|---|---|---|---|---|---|
| | a -ă | a -ă(2) | i -î | s -ş | t-ţ |
| Data set size | 2,161,556 | 1,369,517 | 2,055,147 | 866,964 | 1,157,458 |
| Baseline | 74.70% | 85.90% | 88.20% | 76.53% | 85.81% |
| Training size | Precision for a test set of 50,000 examples | | | | |
| 2,000,000 | 95.56% | - | 99.69% | - | - |
| 1,000,000 | 95.10% | 99.14% | 99.58% | - | 98.75% |
| 750,000 | 94.83% | 98.97% | 99.53% | 99.07% | 98.63% |
| 500,000 | 94.57% | 98.79% | 99.46% | 98.86% | 98.40% |
| 250,000 | 94.00% | 98.37% | 99.28% | 98.87% | 98.26% |
| 100,000 | 93.03% | 97.56% | 98.96% | 98.54% | 97.81% |
| 50,000 | 92.10% | 96.86% | 98.57% | 98.13% | 97.40% |
| 25,000 | 90.99% | 95.75% | 98.11% | 97.58% | 96.92% |
| 10,000 | 88.99% | 93.75% | 97.31% | 96.53% | 96.20% |
| 5,000 | 87.56% | 92.76% | 96.65% | 95.61% | 95.10% |
| 4,000 | 86.91% | 91.86% | 96.49% | 94.99% | 94.53% |
| 3,000 | 86.39% | 90.99% | 96.19% | 94.18% | 94.30% |
| 2,000 | 85.81% | 89.93% | 95.49% | 93.47% | 93.56% |
| 1,000 | 83.49% | **88.36%** | 93.78% | 92.31% | 91.85% |
| 500 | 80.61% | 85.66% | 93.07% | 90.75% | 89.74% |
| 250 | 77.89% | 83.17% | 92.75% | 87.41% | **87.23%** |
| 100 | **74.80%** | 84.04% | **91.41%** | 82.13% | 84.46% |
| 50 | 72.79% | 82.73% | 88.05% | 86.53% | 77.54% |
| 25 | 72.45% | 81.34% | 88.15% | **78.26%** | 78.52% |
| 10 | 73.38% | 85.90% | 88.20% | 75.88% | 85.81% |

Table 3.14 Mihalcea's Results on Diacritic Restoration for Romanian with N=5. *Source:* [62]

The results of the experiments are as presented in Table 3.14. For the results shown on Table 3.14, the training sets range from 2,000,000 examples to as few as 10 examples, to optimize learning rate and minimum size of a corpus required for a satisfactory result. The size of the test set is 50,000 examples for all experiments. A 10-fold cross validation scheme was used and the baseline was the most frequent letter of each pair.

Most crucial learning achieved with the first 10,000 examples. 100,000-250,000 running characters (approx. 25-60 pages of text) generates a small corpus of about 10,000 examples with diacritics. Beyond that, a significant number of examples is required for little improvement in accuracy. First precision figures to exceed the baseline are shown in bold. The best performing pair was *i* - *î* with almost 100% accuracy. For precision, the worst pair was *a* – *ă* because many Romanian nouns have base forms ending in ă, and their articulated forms *a* e.g. *masă* and *masa.* Also, some verb tenses end with an *a* or *ă.*

This is solved by avoiding those examples that contain an *a* or *ă* letter at the end of a word, as reported in Table 2 under the heading *a –ă(2)* with over 4% gained in precision (87% error reduction). C4.5 was applied without improvements and was slower in learning than the Timbl implementation.

### Extended Trigram Based Technique

One major drawback in Mihalcea's works [62, 63] is the evaluation method. The authors admitted that their works cannot be fairly compared to similar research works reported due to differences in evaluation methods. For example, while Tufiş & Ceauşu [98] reported a word level accuracy of 97.4%, Mihalcea *et al.* [63] got an average character level accuracy of 98.30% on the four languages used which may imply a much worse word level accuracy.

In their work, Wagacha *et al.* [105] highlighted the same problem while proposing a letter based approach with an extended trigram based technique. This method was applied to Gĩkũyũ, an East African language, and contrasted with Dutch, German and French. Gĩkũyũ has seven vowels with two extra diacritically marked graphemes: the cardinal vowels *ĩ* and *ũ* which are different from the unmarked *i* and *u* graphemes.

Due to lack of extensive text corpora or electronic dictionary, a 14,000 word fully diacritic text was manually created and 4,500 unique word tokens were extracted for the experiment. These were extracted from scanned and corrected short stories, letters, poems, proverbs and riddles, songs, bible verses and other religious materials. This is contrasted with a 23m word French corpus with a 45k word lexicon and 330k word lexicon each for German and Dutch. 10 Folds cross-validation was also applied.

There were two baseline models: the first uses the most frequent candidate while the other uses the training set (lexicon and plain text) to translate the unmarked words in the test set to their corresponding accented words. Expectedly the lexicon method failed completely during cross validation because the test data contains only "unknown" words. There were also two memory based learning approaches: *unigram* and *trigram.*

The memory-based classifier TiMBL[21] was used. Training instances are generated using a windowing approach (see Table 3.15) with disambiguated left context and ambiguous right context. Each grapheme belongs to a class explicitly for diacritic candidates i/ĩ and u/ũ.

In Table 3.16 the context is made of grapheme trigrams using a similar windowing method as in the unigram experiment. This captures a larger graphemic context and effectively provides three separate classification decisions for each grapheme (see Table 3.17 below). In the example, the classifier predicted 7 trigram classes for the word *mbũri* and selected the most likely grapheme by majority voting.

| L | L | L | F | R | R | R | C |
|---|---|---|---|---|---|---|---|
| - | - | - | **m** | b | u | r | - |
| - | - | m | **b** | u | r | i | - |
| - | m | b | **u** | r | i | - | ũ |
| m | b | ũ | **r** | i | - | - | - |
| b | ũ | r | **i** | - | - | - | i |

Table 3.15 Training instances for the unigram approach. *Source:* [105]

| L | L | L | F | R | R | R | C |
|---|---|---|---|---|---|---|---|
| - - - | - - - | - - - | **- -m** | -mb | mbu | bur | - -m |
| - - - | - - - | - -m | **-mb** | mbu | bur | uri | -mb |
| - - - | - -m | -m b | **mbu** | bur | uri | ri- | mbũ |
| - -m | -mb | mbũ | **bur** | uri | ri- | i- - | bũr |
| -mb | mbũ | bũr | **uri** | ri- | i- - | - - - | ũri |
| mbũ | bũr | ũri | **ri-** | i- - | - - - | - - - | ri- |
| bũr | ũri | ri- | **i–** | - - - | - - - | - - - | i- - |

Table 3.16 Training instances for the trigram approach. *Source:* [105]

From the results shown in Table 3.18, the trigram approach substantially improved the accuracy for the Gĩkũyũ accent restoration task and significantly outperforms the unigram approach scoring up to 90% range for the individual graphemes. It also achieved a word level accuracy increase of about 14% on plain text but did not do as well on unknown words. Wagacha recommended this system as a tool for Gĩkũyũ corpus construction due to its 90% accuracy.

| | | | | | | | | | |
|---|---|---|---|---|---|---|---|---|---|
| Predicted Class 1 | - | - | m | | | | | | |
| Predicted Class 2 | | - | m | b | | | | | |
| Predicted Class 3 | | | m | b | u | | | | |
| Predicted Class 4 | | | | b | ũ | r | | | |
| Predicted Class 5 | | | | | ũ | r | i | | |
| Predicted Class 6 | | | | | | r | i | - | |
| Predicted Class 7 | | | | | | | ĩ | - | - |
| Majority Vote: | - | - | m | b | ũ | r | i | - | - |

Table 3.17 Predicted Classes for the Restoration of the word *mbũri*. *Source:* [105]

| All Words | Unknown Words | | | | Plain Text | | | |
|---|---|---|---|---|---|---|---|---|
| | Dutch | French | German | Gĩkũyũ | Dutch | French | German | Gĩkũyũ |
| | Dut | Fre | Ger | Gik | Dut | Fre | Ger | Gik |
| **Baseline 1 (Most Frequent)** | 98.9 | 71.5 | 43.9 | 46.6 | 99.7 | 79.7 | 71.1 | 57.1 |
| **Baseline 2 (Lexicon)** | 0 | 0 | 0 | 0 | 99.9 | 86.5 | 96.2 | 74.9 |
| **MBL - Grapheme (unigram)** | 99.5 | 82.2 | 91.6 | 68.0 | 99.8 | 88.3 | 95.3 | 77.5 |
| **MBL - Grapheme (trigram)** | 99.7 | 82.8 | 89.5 | 72.2 | 99.5 | 89.0 | 94.3 | 91.4 |

Table 3.18 Word level accuracy scores for all the methods and languages. *Source:* [105]

Word level ambiguity was not dealt with by this method. In plain text Dutch and German, the unigram did better than the trigram because the trigram has a "stronger tendency to place diacritics", consequently making mistakes on "words that don't need them". The grapheme-based approach improves on the lexicon lookup approach for Gĩkũyũ and even compares with the lexicon lookup scores achieved on German and Dutch. Wagacha, however, admits that there is no silver bullet to grapheme-based diacritic restoration due to noticeable differences even among related languages. Also, this approach can be used to process related Bantu languages like Kĩembu, Kĩmerũ and Kĩkamba, and help to digitally preserve these resource-scarce, endangered languages in their full orthographic forms.

### Lexicon Lookup and Instance based learning

De Pauw *et al* [24] extended the work of Wagacha *et al* [105] by applying a combination of lexicon lookup and instance based learning methods to African languages (Cilubà, Gĩkũyũ, Kĩkamba, Maa, Sesotho sa Leboa, Tshivenda and Yoruba) while contrasting with experiments on some European languages (Czech, Dutch, French German and Romanian) as well as Vietnamese and Chinese Pinyin.

The experimental data for all the languages used in this work were collected from different sources: manually built, web crawled and available. They are also of various sizes ranging from 14.8k tokens for Gĩkũyũ to 23.2M tokens for French. Metrics like **T(d)** and **LexDif** were used for quantitative analysis of each corpus. T(d) is percentage of words with at least one diacritic, while the **LexDif** (Lexical diffusion) is a measure of the difficulty of the restoration task for a language which is calculated by dividing the number of distinct word types by the number of latinized (stripped) word forms.

A "rigid" pre-processing of the data improved the noise-robustness of the output thereby making the use of trigram models, as in the work of Wagacha *et al.*[105], almost unnecessary. A memory-based classifier was trained with instances extracted from the training data. Unlike the work of Mihalcea, the evaluation is done primarily on the word level i.e. the percentage of words in the text that have been predicted completely correctly. The baseline model chooses the most frequent candidate observed in the training set. The word level and grapheme level accuracy scores are presented in Table 3.20 below.

| Language | Tokens | Types | n | T(n) | LexDif |
|---|---|---|---|---|---|
| Cilubà | 144.7k | 20.0k | 17 | 71.8% | 1.17 |
| Gĩkũyũ | 14.8k | 9.1k | 2 | 64.9% | 1.03 |
| Kĩkamba | 38.3k | 9.7k | 2 | 65.7% | 1.07 |
| Maa | 22.2k | 22.2k | 11 | 46.9% | 1.05 |
| Sesotho sa Leboa | 6.9M | 157.8k | 1 | 23.3% | 1.04 |
| Tshivenda | 249.0k | 9.6k | 5 | 18.2% | 1.03 |
| Yoruba | 65.6k | 4.2k | 21 | 61.3% | 1.26 |
| Czech | 123.9k | 105.8k | 15 | 66.3% | 1.05 |
| Romanian | 3.3M | 146.9k | 5 | 39.9% | 1.05 |
| French | 23.2M | 258.6k | 19 | 21 % | 1.04 |
| Dutch | 301.9k | 301.9k | 18 | 1.5 % | 1 |
| German | 365.6k | 365.6k | 4 | 23.9% | 1.03 |
| Vietnamese | 2.6M | 50.9k | 26 | 61.3% | 1.21 |
| Chinese Pinyin | 73.5k | 12.0k | 25 | 97.1% | 1.12 |

Table 3.19 Corpus Statistics. *Source:* [105]

| Word | Ci | Gĩ | Kĩ | Ma | Se | Ts | Yo | Cz | Ro | Fr | Du | Ge | Vi | Ch |
|---|---|---|---|---|---|---|---|---|---|---|---|---|---|---|
| Baseline | 28.2 | 48.7 | 58.4 | 53.1 | 76.2 | 81.8 | 35.4 | 33.7 | 60.6 | 75.2 | 98.5 | 78.3 | 29.4 | 6.7 |
| MBL | 36.6 | 74.9 | 73.5 | 58.6 | 90.1 | 89.3 | 40.6 | 74.4 | 83.2 | 88.2 | 99.6 | 92.7 | 63.1 | 31.5 |
| Grapheme | Ci | Gĩ | Kĩ | Ma | Se | Ts | Yo | Cz | Ro | Fr | Du | Ge | Vi | Ch |
| Baseline | 69.8 | 58.9 | 66.7 | 76.8 | 50.6 | 87.2 | 54.0 | 83.2 | 92.5 | 93.8 | 99.7 | 83.1 | 65.8 | 40.4 |
| MBL | 77.4 | 83.1 | 80.4 | 85.4 | 80.9 | 92.9 | 68.2 | 95.2 | 97.3 | 97.2 | 99.9 | 94.3 | 82.7 | 69.0 |

Table 3.20 Results on Word- and Grapheme-Level Evaluations. *Source:* [105]

The results from this experiments as shown in Table 3.20 indicate that the grapheme-based MBL approach improved both word level and grapheme level accuracy especially for Gĩkũyũ, Kĩkamba, Sesotho sa Leboa, Czech, Romanian and Vietnamese. For Czech and Romanian a modest increase of accuracy on the grapheme level has a major impact on the accuracy on the word level although their accuracy scores are below those reported by Mihalcea[62] using similar approach and data. This is likely to be due to their experimental design which, unlike what was reported in [62, 63], split the training and testing instances at the word-level before being extracted from the different pools to be able to assess the performance on unseen words.

| Word | Ci | Gĩ | Kĩ | Se | Ts | Yo | Cz | Ro | Fr | Du | Ge | Vi | Ch |
|---|---|---|---|---|---|---|---|---|---|---|---|---|---|
| LLU | 77 | 77.3 | 79.4 | 97.6 | 97.7 | 67.8 | 61.8 | 94 | 89.1 | 99.9 | 96.2 | 74.5 | 78.5 |
| MBL | 85.3 | 92.4 | 91.6 | 99.2 | 99.4 | 76.8 | 89.2 | 96.5 | 88.3 | 99.8 | 95.3 | 73.5 | 83.9 |
| LLU+MBL | 79.6 | 91.5 | 90.4 | 99.4 | 99.2 | 68.5 | 90.1 | 96.6 | 89.3 | 99.9 | 96.8 | 75.5 | 80.3 |

Table 3.21 Results on LLU and MBL. *Source:* [105]

Though Cilubà and Yoruba improved significantly, they still gave comparatively poor results due to tonal diacritics[9]. A language data could be used to bootstrap the process of restoring diacritics on a language with similar orthography (e.g. Gĩkũyũ and Kĩkamba. Experiments were conducted to confirm this and could be extended to such pairs as Kĩembu or Kĩmerũ. Another round of experiments was conducted to compare the three approaches: lexicon look up, MBL and a combination of both (see Table 3.21).

Table 3.21 shows that Dutch is almost a solved problem with only LLU and German also performs quite well. Using MBL decreased accuracy for Dutch, German and French. Generally, MBL outperformed the LLU as expected especially for the resource scarce languages and, in some cases, did better than the combined method[10].

The plain text LLU approach gave more accuracy because it used small domain-specific corpora, with a typically restricted lexicon. Language data could be used to bootstrap the process of restoring diacritics on a language with similar orthography (e.g. Gĩkũyũ and Kĩkamba. Experiments were conducted to confirm this and could be extended to such pairs as Kĩembu or Kĩmerũ.

### C4.5 and AdaBoost

Nguyen & Ock [70] worked on the restoration of diacritics in Vietnamese which is heavily marked with *phonetic diacritics* (e.g. a, ă, and â) and *tonic accents* (e.g. a, á, à, ã using C4.5 and AdaBoost. Table 3.4 shows the ambiguous letters in Vietnamese. In their work, the neighbouring words to the ambiguous pattern are used as features. A sliding window is scanned through training corpus to build data instances.

From popular experiments in the literature of lexical disambiguation [62, 24], a window size of 5 characters to both sides of the ambiguous pattern was chosen. The ambiguous pattern is centered on the sliding window. No additional feature selection mechanism or parameter tuning was applied. Default parameters of C.45 implemented in Weka are used.

[9] I do not understand the assertion that *Tonal diacritics can simply not be solved on the level of the grapheme (pg 7).*

[10] This could be because these languages do not yet contain enough lexical information to deal with accurate lexicon look up.

| Letters | Classes |
|---|---|
| a | [a, à, á, ả, ã, ạ, ă, ằ, ắ, ẳ, ẵ, ặ, â, ầ, ấ, ẩ, ẫ, ậ] |
| e | [e, è, é, ẻ, ẽ, ẹ, ê, ề, ế, ể, ễ, ệ] |
| o | [o, ò, ó, ỏ, õ, ọ, ô, ồ, ố, ổ, ỗ, ộ, ơ, ờ, ớ, ở, ỡ, ợ] |
| u | [u, ù, ú, ủ, ũ, ụ, ư, ừ, ứ, ử, ữ, ự] |
| i | [i, ì, í, ỉ, ĩ, ị] |
| y | [y, ỳ, ý, ỷ, ỹ, ỵ] |
| d | [d, đ] |

Fig. 3.4 Ambiguous Letters in Vietnamese. *Source:*[70]

The experimental data contains 3.7K articles (2.2M tokens, 20K unique tokens) in the education category. 4.5K syllables in the Vietnamese dictionary appeared in the corpus as tokens. The remaining 15.5K tokens are out-of-dictionary words, each of which rarely appears in the corpus and are mostly English named entities which do not contain diacritics.

Some tokens with diacritics are acronyms, noisy or misspelt words. To eliminate the effect of noisy data and to reduce feature space in decision tree learning, all out-of-vocabulary tokens are tagged "UNKNOWN". There were five experimental strategies applied in this work:

- **Learning from letters**: Ambiguous patterns are letters that may have different diacritics (Table 3.4). Attribute values are case sensitive. Delimiters (space, comma, dot, question marks, colons, date, and number) are tagged as SPACE, COMMA, DOT, QUESTION, COLON, DATE, and NUMBER, respectively.
- **Learning from syllables**: In 20K unique tokens in the corpus, 15.5K tokens are out-of-vocabulary tokens, tagged "UNKNOWN". 4.5K tokens which are syllables used in Vietnamese dictionary, have equivalent 1.3K diacritic-free tokens or word keys after removing diacritics.
- **Learning from semi-syllables**: An approach based on construction rules of syllables in Vietnamese was used. In learning from syllables, each attribute has 1.3K values. Semi-syllables are extracted by omitting head consonants from syllables. As the result, each attribute has about 100 values.
- **Learning from words**: To prepare data for learning from words, training text is preprocessed by a word segmenter that was reported to have achieved a 90% accuracy [49].
- **Learning from bi-grams**: To differentiate between learning from syllables (unigrams) and learning from words, learning from $n$-grams is considered as an "intermediate approach". Bigram based learning was used because in the Vietnamese dictionary, the majority of words are composed of 2 syllables as shown in Table 3.22.

| Sylables/word | # words | Percentage |
|---|---|---|
| 1 | 5208 | 17.27% |
| 2 | 22,866 | 75.81% |
| 3 | 1,362 | 4.52% |
| 4 | 653 | 2.16% |
| $\geq 5$ | 75 | 0.25 % |

Table 3.22 Syllables/Word in Vietnamese 30k-Entries Dictionary *Source:*[70]

| Learning Method | Accuracy |
|---|---|
| Baseline (most freq) | 45.15% |
| C4.5 + Letters | 93.00% |
| C4.5 + Semi-syllable | 88.20% |
| C4.5 + Words | 91.90% |
| C4.5 + Bigrams | 88.80% |
| AdaBoost+C4.5+Letters | 94.70% |

Table 3.23 Syllables/Word in Vietnamese 30k-Entries Dictionary *Source:*[70]

These approaches were compared using only the C4.5 classifier while performance of the best approach was enhanced by a combination of AdaBoost and C4.5. In the experiment, 2M data instances of all ambiguous patterns were created from training corpus for each learning approach. The 10-fold cross validation scheme was applied. Best accuracy with C4.5 and AdaBoost for boosting gave an improvement of 1.4% against individual C4.5 on letters level approach. The results of the experiments are presented in Table 3.23.

With simple features set, learning from letters performed best. It is however not clear if this is a word or letter level accuracy. Learning from semi-syllables was the worst due to loss of information when all head consonants are omitted but it lost only 1.6% accuracy over learning from syllables while reducing the feature space from 1000 to 100 candidates for each feature.

The accuracy of word segmentation is yet to be improved on. However, learning from words performs better than learning from syllables and learning from bigrams. It is assumed that a more accurate word-segmenter may improve the results of learning from words.

### Naïve Bayes with Layered Lexicon

Scannell [91], in their *unicodification* experiments with about 115 languages, introduced a *layered lexicon* Naïve bayes implementation for both word- and character-level models. It took a lot of ideas from the approaches presented above but assumed a possible existence of a layered lexicon for each language.

*Layer1* is the list of correct words in the language; *Layer2* contains words with non-standard but commonly used spellings while *Layer3* is a collection of words seen in the raw text that do not appear in the first two layers. *Layer3* becomes the default lexicon where *Layer1* and *Layer2* do not exist.

The training data was assembled using the Crúbadán web crawler [90]. Correct documents for each language were selected using manual inspection with the help of native speakers. These were then segmented into sentences eliminating sentences that contain English texts. Open source spell-checkers were used to provide the first layer of the lexicon for languages for which they are available.

Ten-fold cross validation was used while word-level accuracy was reported contrary to Mihalcea's letter-level approach. A modified version of lexical diffusion[11], LD1, which computes the percentage of words in the training corpus that are incorrectly resolved by always choosing the most frequent candidate, was applied as a measure of the difficulty of the task. The algorithms proposed for the restoration task are described below:

- **BL:** Baseline - leaves all characters as ASCII
- **LL:** Lexicon-Lookup - assumes a 3-layer lexicon and looks for the diacritic variants of a word in *layer1* and chooses the most frequent, if not in *layer1*, then *layer2* is used else *layer3*.
- **LL2:** Lexicon-Lookup2 - similar to LL but uses a word level bigram model to choose a candidate instead of selecting the most frequent.
- **FS1:** Feature Set1 – character level approach with the feature set consisting of 3 single characters on either side of the target character i.e.: (-3, 1), (-2, 1), (-1, 1), (+1, 1), (+2, 1), (+3, 1) as used by Mihalcea [62]
- **FS2:** Feature Set2 – (-5, 1), (-4, 1), (-3, 1), (-2, 1), (-1, 1), (+1, 1), (+2, 1), (+3, 1), (+4, 1), (+5, 1), i.e. five single characters on each side as used by De Pauw *et al.* [24]

[11]Lexical Diffusion was introduced by De Pauw *et al.* [24]

- **FS3:** Feature Set3 – (-4, 3), (-3, 3), (-2, 3), (-1, 3), (0, 3), (+1, 3), (+2, 3) i.e. the character trigrams on each side of the target character similar to Wagacha's[105] approach but different in application.
- **FS4:** Feature Set4 – (-3, 3), (-1, 3), (+1, 3) i.e. the character trigrams immediately preceding the target, centred on it, and immediately following it
- **CMB:** Combined techniques – Uses LL2 for lexicon words and the best performing algorithm (FS1–FS4) for the language in question

The results for some of the 115 languages used for the experiment are as shown in Table 3.24.

| 639 | Train | Lex | **LD1** | **BL** | **LL** | **LL2** | **FS1** | **FS2** | **FS3** | **FS4** | **CMB** |
|---|---|---|---|---|---|---|---|---|---|---|---|
| ada | 14k | 1.2k | 4.44 | 62.8 | 93.8 | 93.8 | 87.8 | 87.0 | *92.6* | 92.5 | **94.0** |
| aka | 177k | 16k | 4.18 | 70.6 | 94.1 | 95.8 | 84.3 | 84.9 | *90.3* | 90.1 | **95.9** |
| bam | 342k | 17k | 2.60 | 69.8 | 95.2 | 95.4 | 83.7 | 83.2 | *89.2* | 89.2 | **95.6** |
| bas | 13k | 1.7k | 1.39 | 72.0 | 96.0 | 96.0 | 80.2 | 80.9 | 88.2 | *88.3* | **96.1** |
| bci | 15k | 1.5k | 4.90 | 59.7 | 92.3 | 92.4 | 75.5 | 74.2 | *83.3* | 82.8 | **93.1** |
| bfa | 12k | 1.8k | 0.31 | 76.5 | 97.4 | 97.4 | 84.1 | 84.3 | *93.4* | 92.4 | **97.9** |
| bin | 11k | 1.5k | 1.98 | 66.5 | 94.7 | 94.7 | 80.5 | 80.7 | *92.6* | 92.3 | **95.9** |
| bum | 39k | 4.1k | 3.65 | 69.6 | 92.4 | 92.4 | 79.5 | 79.1 | *85.4* | 85.2 | **92.8** |
| byv | 8k | 1.0k | 6.86 | 59.4 | 89.0 | 89.0 | 68.5 | 67.8 | *79.7* | 79.0 | **89.4** |
| dua | 36k | 4.5k | 7.82 | 74.5 | 88.4 | **88.8** | 76.0 | 75.4 | *81.4* | 80.2 | 88.5 |
| dyo | 12k | 3.5k | 1.40 | 78.0 | 92.9 | 92.9 | 78.0 | 79.2 | *87.3* | 85.0 | **93.1** |
| dyu | 10k | 1.1k | 0.52 | 72.7 | 97.2 | 97.2 | 84.4 | 84.6 | *91.8* | 91.4 | **98.2** |
| efi | 20k | 2.9k | 5.08 | 71.4 | 90.8 | 90.8 | 76.3 | 74.7 | 87.5 | *88.2* | **91.5** |
| ewe | 19k | 3.2k | 5.24 | 59.8 | 89.1 | 89.2 | 75.9 | 76.7 | *82.7* | 81.7 | **90.5** |
| fon | 36k | 3.4k | 29.81 | 32.3 | 66.1 | 66.1 | 55.0 | 54.8 | *59.3* | 59.2 | **69.1** |
| fub | 873k | 49k | 1.07 | 77.4 | 98.1 | 98.1 | 84.1 | 84.4 | 90.1 | *90.7* | **98.3** |
| gaa | 11k | 2.0k | 2.30 | 44.3 | 91.1 | 91.2 | 78.9 | 77.2 | 90.8 | *90.9* | **94.6** |
| gba | 9k | 0.7k | 1.58 | 89.6 | **97.8** | **97.8** | 92.2 | 92.1 | *95.3* | 95.0 | 97.7 |
| guw | 21k | 2.3k | 3.88 | 45.4 | 93.2 | 93.4 | 72.6 | 72.2 | *86.9* | 85.7 | **94.2** |
| hau | 472k | 42k | 0.83 | 93.5 | 97.5 | **97.7** | 95.0 | 94.4 | *96.9* | 96.6 | 97.6 |
| her | 9k | 2.5k | 0.06 | 95.5 | 98.7 | 98.7 | 95.5 | 95.5 | *97.2* | 96.9 | **98.8** |
| ibo | 31k | 4.3k | 7.48 | 54.7 | 88.6 | 89.5 | 75.0 | 75.8 | *81.7* | 81.3 | **89.5** |

Table 3.24 Sample Results from the Diacritic Restoration Experiment [Igbo (ibo) on the last row] *Source:* [91].

The closest work to this is the work of De Pauw *et al*[24]. However due to the differences in data set and learning techniques (naïve Bayes vs memory-based learning), the results are not directly comparable. But column FS2 uses the same features as were used by De Pauw but reported generally poorer results which Scannell attributed to noise in the web corpora.

The trigram models perform consistently better than the ones reported in literature. However, comparing the trigram models, FS3 is superior on small data while FS4 performs better with a larger amount of data. Also, LL2 records high performance improvement with languages with high LD1 values and large corpora for accurate bigram model. Surprisingly, LL2 often outperforms CMB i.e. words not in the lexicon are better left as pure ASCII than trying to restore them.

### 3.4.3 Techniques for Non-Latin (Arabic) Scripts

Although Igbo is written with a Latin-based orthography, it is important to mention that other writing systems, e.g. Arabic scripts, also deal with the challenge of diacritic restoration. By the number of countries using it, the Arabic script is the second most-widely-used[12] writing system in the world after the Latin script and the third by the population of users, after the Latin and Chinese characters scripts.

In Arabic language, diacritics often carry phonological information. They occur in the form of short vowels, *Shaddah*, *tanween*, *Maddah* as well as *hamzah* [2]. These are often omitted in text and their absence leads to high word-level ambiguity. On average, there are 11.5 possible variants per word in Arabic [25]. Alosaimy & Atwell [2] observed that although diacritic could be '*full*' (where the diacritics for all letters are specified), '*partial*' (where some of diacritics are restored) or even '*minimal*' (only enough restoration to remove ambiguity is done), the depth of the restoration is relative to the task at hand.

In their work on the Arabic diacritic restoration, Alosaimy and Atwell [2] applied a 'semi-automatic' method in restoring the diacritics on the *Sunnah Arabic Corpus* [3] which is only partially marked with diacritics. However, their method extracts a highly reproduced part of the *Sunnah Arabic Corpus*, the *Riyāḏu Aṣṣāliḥīn* and aligns it with the texts from the other sources that cited it. Since these other sources have different levels of the diacritic markings, they could compare the aligned texts and select the most appropriate variants of the words given the context.

A morphological analyser was also introduced to enhance the system. To evaluate their work, they used different metrics: *diacritic error rate (DER)* (basically a reverse accuracy), *coverage* (the percentage of letters with at least one diacritic) and *ambiguity* (the measure of of ambiguity left in the restored text). Their baseline score for *DER* was not reported but *coverage* and *ambiguity* were 48.66% and 17.42% respectively. The best performing system, a trigram model enhanced with the morphological analyser achieved a *coverage* of 81.26%, a *DER* of 0.007 and reduced *ambiguity* to 1.56%.

## 3.5 Evaluation Criteria and Standards

Atserias *et al* [7] made this general comment on evaluation mechanisms used for the experiments reported:

[12] Source: https://en.wikipedia.org/wiki/List_of_writing_systems

> *...there is still no consensus on how to evaluate the results of the experiments. Different evaluation methodologies on different datasets render incommensurable evaluation figures... there are many factors that can artificially increase precision, such as counting the number of errors per total number of words in running text, because many words in the text do not contain any ambiguity and therefore would always count as correct. In other cases, such ambiguity can be negligible and thus picking the most frequent option already produces positive figures.*

Many of the ADRS systems view evaluation methods as a language independent process. There has been a debate on what the basic unit of language should be despite the approach used. For instance, Mihalcea [62] reported a letter-level accuracy of 99% on Romanian language over the 97.4% earlier reported by Tufiş & Chiţu [99] on word-level. There will be no basis to make direct comparisons on such works.

Systems that handle multiple languages tend to ignore the nuances of the individual languages. For example, Scannell [91] was quick to admit that the evaluations performed in his research work should be taken with a "grain of salt". The ideal approach, according to him, would have been training and evaluating models based on "writing systems". He gave an example with Hausa language with the following distinguishing features for the training sets: "no length or tone marks", "with tone but no length", "with tone and long vowels doubled", "with tone and long vowels with macrons", "with tone, long vowels unmarked, short vowels marked with cedilla", as well as considering the "hooked y" used in Niger.

Dealing with unknown (or out-of-lexicon) words is also a key issue in diacritic restoration especially when there is no existing dictionary [24]. While some works did not explicitly deal with this issue, some recognise that "guessing" the diacritic forms of words not found in the dictionary could be a challenging task [94].

Luu & Yamoto [60] argued that approaches that examine neighboring labels such as HMMs and conditional random fields (CRFs) are not effective when unknown words are encountered. Interestingly, even for works on well resourced languages that used a word-based approach [98], character-based back-off models are recommended as being more effective in dealing with unknown words [62].

Generally, models can be evaluated *extrinsically* or *intrinsically.* Extrinsic evaluation requires embedding the model to an application and assess its performance with the model as compared to without it. In practice, this is often difficult as it is expensive to keep running some NLP systems end-to-end while developing a language model. With a intrinsic evaluation method, the model's performance is measured independent of any running system. The model is developed on some *training set* or *training corpus* and its performance is measured on some unseen data called the *test set* or *test corpus*.

It is important to clearly separate the training set and the test set, i.e. there should be no intersection between the two sets in order to avoid "training on the test set" and thereby introducing some bias. Also in practice, the aim of having a separate test set is often defeated if, by using it so much during development, the model's behaviour is over-tuned to it. So, to avoid this scenario, a *development set* is often used to store unseen instances during model training. The recommended percentage of the entire corpus for training, development and test sets generally 80%,10% and 10% respectively [53].

## 3.6 Review of works on Igbo ADRS

As referenced in §3.4.2, Scannell [91] included Igbo among the other 115 mostly low-resourced languages in their diacritic restoration project. They implemented a variety of restoration techniques including word- and character-level as well as naïve Bayes approaches (§3.4.2). For the word-level approach, they used two lexicon lookup methods, *LL* which replaces ambiguous words with the most frequent word and *LL2* that uses a bigram model to determine the most probable candidate. They reported accuracies of 88.6% and 89.5% for Igbo language on the *LL* and *LL2* models respectively.

In one of our publications [32], we extended Scannell's work on the word-level diacritic restoration for Igbo language. We note, however, that although we based our work on Scannell's *unicodification* work, there are some key distinctions which made it difficult to directly compare both works and they are highlighted below:

**Data Size:** Their experiment used a data size of 31k tokens with 4.3k word types. In our case, for these experiments, we used only the Igbo Bible data with $902,150$ word tokens and a vocabulary size of 16,061 unique tokens as shown in Table 4.3.

**Data Quality:** They trained their model with web-crawled data which are often heterogeneous and also riddled with words with wrong diacritic forms. We used the Igbo Bible which, in addition to being homogeneous, was produced by human translators and so should be far better in terms of proper diacritic contents.

**Pre-Processing:** Our tokenizer considered certain language information such as the ones described in §4.2.1 which their method may not have considered.

**Baseline Model:** The baseline accuracy reported in their work was a measure of the percentage correctness of a stripped version of the text against the *gold standard* i.e. the original diacritically marked corpus. In our experiment, the baseline is the performance of the *unigram* model which simply selects the most-frequent variant as a replacement for the wordkey.

***n*-gram Models:** We extended the bigram models with different smoothing techniques, *words vs keys* approach, and *backward replacement* features. We also implemented different trigram models with similar structures as the bigram models which were not included in their work.

## 3.7 Challenges for Igbo ADRS

One of the major challenges for low resourced languages is the unavailability of sizeable, well annotated corpora for development experiments. Like other NLP tasks, this adversely affects the ability to develop well trained and robust systems for such languages. For the diacritic restoration task for instance, Table 3.25 shows the corpus sizes of some of the works reviewed.

| Language | Source | Tokens | UnqTokens | Genre | Citation |
|---|---|---|---|---|---|
| Vietnamese | articles | 2.2M | 20k | education | [70] |
| Spanish | articles | 35.3M | 351K | mixed | [20] |
| Romanian | newspaper | 3M | - | news | [62] |
| Croatian | newspaper | 2.5M | - | news | [71] |
| French | web | 1.5M | 655k | mixed | [91] |
| Igbo | web | 31k | 4.3k | mixed | [91] |
| Mongo-Nkundu | web | 1k | 0.5k | mixed | [91] |

Table 3.25 A simple survey of data sizes used by researchers

As we see in Table 3.25, under-resourced languages like Igbo and Mongo-Nkundu are trailing behind others in terms of corpus availability. Even at that, a manual scrutiny of the data on Igbo, for example, shows a lot of impurities such as foreign words, web addresses, tweet handles email addresses and so on. So clearly having a large pool of normative corpus gives a good head start for a task like this.

## 3.8 Chapter Summary

In this chapter, we introduced diacritic restoration as an NLP task in more depth with examples of the challenges they pose to meanings and pronunciations of words in languages that have them. A review of some of the popular literature on the task of diacritic restoration was presented which shows the two broad categories of approaches to solving the diacritic restoration problem: *word*- and *character*- (or *grapheme*-) based approaches.

### 3.8.1 Summary of techniques

We summarise the key techniques found in the literature from which we derived the intuition for our work as follows:

**Decision List:** This technique was applied by Yarowsky [112] for the Spanish diacritic restoration. Their experiment used a large amount of data (≈ 70 million tokens): Spanish AP Newswire (1991-1993, 49 million words), the French Canadian Hansards (1986-1988, 19 million words) and a collection from *Le Monde* (1 million words).

Our training instance are generated in a similar to theirs but we did not completely adopt this technique. The key challenge with the approach is its dependence on an existing Spanish lemmatizer and a set of hand-crafted rules for likelihood ranking. Besides, compared to our corpus size and distribution, they have a very large amount of well curated data to work with.

**Bayesian Framework and Hidden Markov Model:** Simard [94] applied the HMM on French text based on the original work by El-Beze *et al.* [26]. However, their approach required a large ≈ 250k word French morpho-syntactic electronic dictionary. Also, accent restoration, as they termed it, is not as challenging in French as in Igbo (85% of French words are normally unaccented and about half of the accented words are unambiguous.).

Also Crandall [20], in his work on Spanish, applied both the Bayesian framework and Hidden Markov Model separately and as a hybrid. His work expanding the works of Yarowsky [113], which focused on only a few sets of ambiguous words. This work is not well suited for our task because it focused only on *accentuation* i.e. replacing only the acute accents. Besides, it requires a pre-built pattern-matching module in Spanish.

**Bayesian Framework and Hidden Markov Model:** Nikola *et al.* [71] presented a method that combines dictionary look-up and statistical language modelling in the restoration of Croatian text. They used the data built from a combination of 100k words of newspaper articles and 30k words of forum posts. Their look-up approach relies on a 750k-entry electronic dictionary to validate their variants. Our method of stripping diacritics to generate test data is similar to theirs but we could not directly apply their method because there is no available electronic dictionary for Igbo.

**Classification Models:** Perhaps the method that most aligned with our work is the one proposed by Cocks and Keegan [16] for their word-based diacritic restorer for Māori. They used a naïve Bayes classifier with the features similar to the ones used for the character-based approach by Scannell [91]. These were produced by extracting word-based n-grams relative to the target word which is similar to what we did in the classification models. However, we extended their method beyond the trigrams to enrich the context information. We also vectorised our input to improve the efficiency of our system.

**Part-of-speech tagging:** Tufiş & Chiţu [99] used part-of-speech tagging in their approach to the restoration of diacritics in Romanian texts which was later extended to an MS-Word based DLL and a web service [98]. The method uses character-based backoff to deal with "unknown words".

**Language Models:** One thing we observed is that a basic language model such as the n-gram model is fundamentally useful in most of the approaches presented in this review. A form of n-gram model is either used on its own in the restoration system or it is applied in the extraction of features for training other kinds of restoration models. For example, for their diacritic restoration system that was integrated into the Spanish spellchecker, Atserias *et al* [7] basically applied a simple bigram-model with an evaluation mechanism that prevents the biasing of high frequency. In this work, we shall be comparing the n-gram model approach to other approaches.

**Charater-based Approaches:** Charater-based approaches were also used in some of the works we have reviewed [62, 63, 105, 24, 114, 70] and presented as better alternatives for low resourced languages. These methods are generally similar to the word-based methods in the way they use the character models of languages, e.g character n-grams, as features to a learning algorithm or directly into the restoration models. In some cases, the character-based approach augments the word level approach especially in dealing with 'unknown words' [91]. Although we did not implement any character-based approach in this work, it will be considered in our future work.

### 3.8.2 Conclusion

We highlighted a number of popular techniques that have been applied to the task in various forms which includes: dictionary and language models, decision list, HMM, Bayesian and different types of classification methods.There is a wide variation in evaluation methods. In some cases, character based evaluation methods are applied as opposed to word based methods. In others, a different set of ambiguous words is used by different works on the same language.

Generally, there is a lack of standard data sources, pre-processing methods or experimental datasets, and frameworks for generating them, in these languages. We also observed that only a few languages have been thoroughly studied. We noted that, although Scannell [91] covered a large number of languages, their work did not cover enough depth with the majority of them e.g. Igbo. This is largely due to lack of adequate data and unavailability of language speakers. In the subsequent chapters, we will introduce the problem of diacritic restoration in Igbo and our approaches to solving it.

# Chapter 4

# Igbo Diacritic Restoration (IDR)

This chapter lays out the task of diacritic restoration with regards to the Igbo language. As with most low resource languages, the absence of well-structured datasets and evaluation methods to test our NLP models posed a challenge. So we shall also describe the processes involved in creating the various datasets used for the IDR system and the evaluation methods applied.

In section 4.1 of this chapter we defined the problem of diacritic restoration. Section 4.2 introduces the data and the methods used to build the dataset as well as the generic framework for similar datasets from any text with diacritics irrespective of the language. The next section defines the evaluation methods and the metrics that are used for assessing the performance of models throughout this work.

## 4.1 Problem Definition

Recall that in §3.1 we defined diacritics as "marks placed over, under, or through a letter in some languages to indicate a different sound value from the same letter without diacritics". Igbo is one of such languages. As highlighted in §1.3, Igbo has orthographic and tonal diacritics which distinguish the meaning and pronunciation of words and therefore are essential for language processing tasks.

Diacritics are often absent from the electronic texts available on the web which we can process and/or build into corpora. This creates the need for automatic diacritic restoration methods. As shown in Figures 4.1 and 4.2 as well as Table 4.1, the absence of diacritics can lead to semantic ambiguity in written sentences. More often than not, a human reader can understand the intended meaning from context but the machine may not. Before we proceed, let us define some of the key terms commonly referred to in this report.

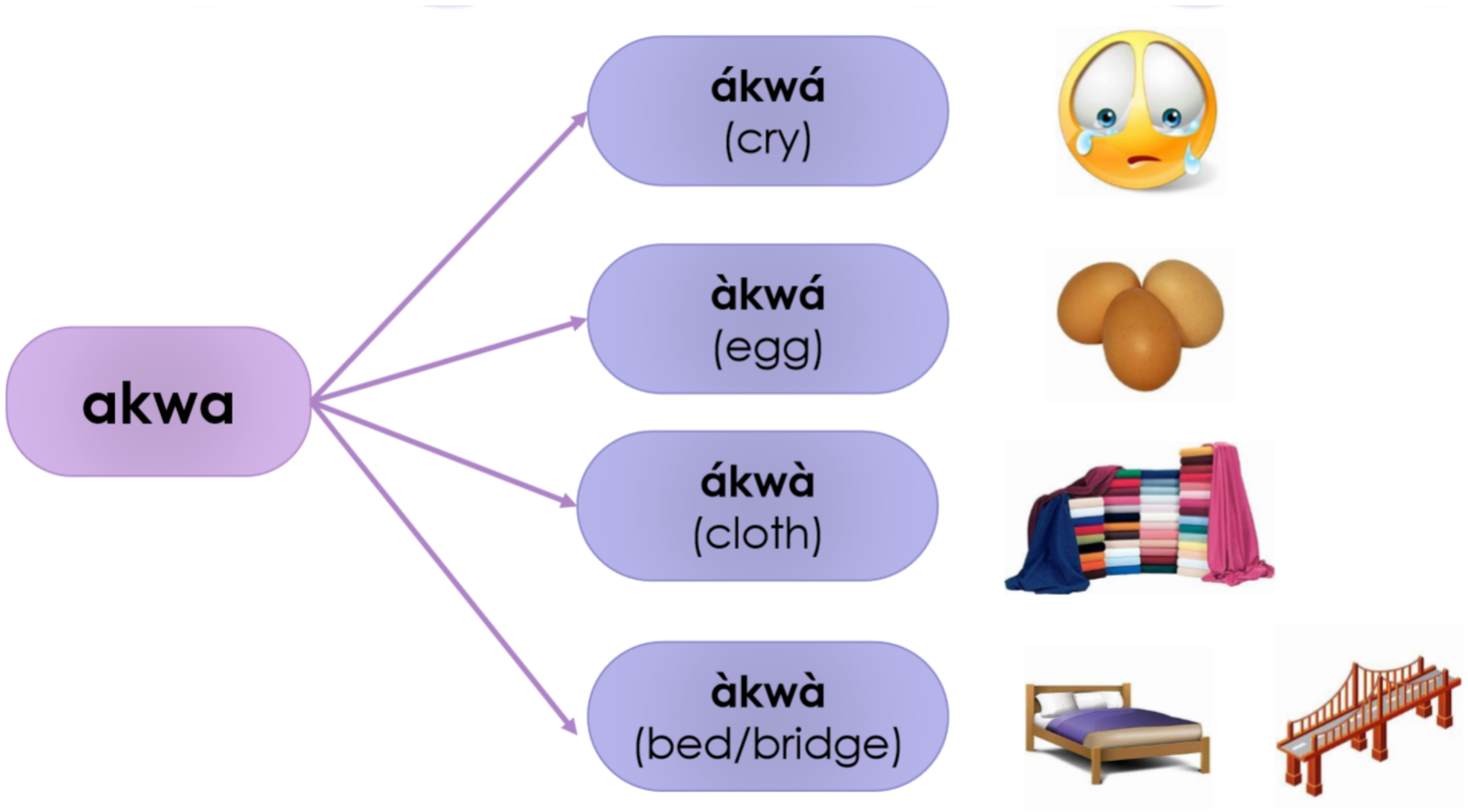

Fig. 4.1 Illustration of the diacritic ambiguities of the wordkey: *akwa*

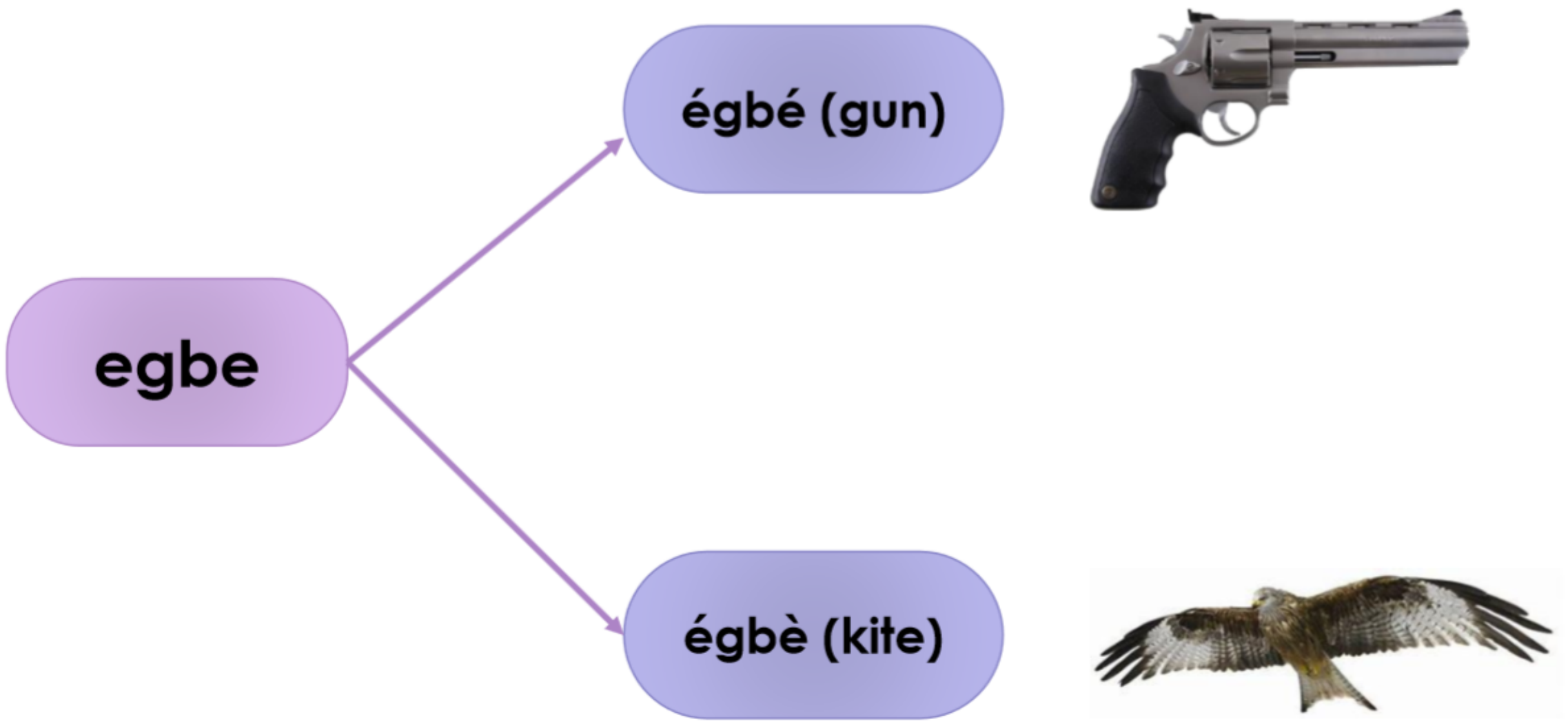

Fig. 4.2 Illustration of the diacritic ambiguities of the wordkey: *egbe*

### 4.1.1 Diacritic Wordkeys and Variants

Let us define a diacritic *wordkey* as the "latinized" form of a word i.e. a word stripped of its diacritics if it has any. *Wordkeys* could have multiple diacritic *variants*, one of which could be the same as the wordkey itself. In the examples below, the words *ugbo*, *olu*, *egbe*, *akwa*, and *egwu* are all wordkeys with different variants. Some of the common wordkeys and their variants are presented below:

| **Wordkey** | **Variants** |
|---|---|
| *akwa* | **ákwá**(cry), **ákwà**(cloth), **ákwà**(bed\|bridge),**àkwá**(egg) |
| *egbe* | **égbé**(kite), **égbè**(gun) |
| *ukwu* | **ụ́kwụ́**(leg), **úkwù**(waist), **ùkwù**(bunch) |
| *isi* | **ísí**(head), **ísì**(smell), **ìsì**(blindness), **ị́sị̄**(verb: to say) |
| *ugwu* | **úgwú** (mountain), **ùgwù**(respect), **úgwù**(circumcision) |

Table 4.1 Examples of Diacritic Wordkeys and Variants

The examples below show the effect of lack of appropriate diacritics for words on simple Igbo sentences:

### 4.1.2 Missing orthographic diacritics

1. *Nwanyi ahu <u>banyere</u> n'**ugbo** ya.* (The woman entered her [**farm**|**boat/craft**])

2. *O kwuru <u>banyere</u> **olu** ya.* (He/she talked about his/her [**neck**|**voice**|**work/job**])

### 4.1.3 Missing tonal diacritics

1. *Nwoke ahu nwere **egbe** n'ulo ya.* (That man has a [**gun**|**kite**] in his house)

2. *O dina n'elu **akwa**.* (He/she is lying on the [**cloth**|**bed,bridge**|**egg**|**cry**])

3. ***Egwu** ji ya aka.* (He/she is held/gripped by [**fear**|**song/dance/music**])

As seen above, ambiguities arise when diacritics – orthographic or tonal – are omitted in texts. In the first examples, **ugbo** (*farm*) and **ụgbọ** (*boat/craft*) as well as **olu** (*neck/voice*) and **ọlụ** (*work/job*) were candidates for diacritic replacements in their sentences.

In the second examples, **égbé** (*kite*) and **égbè** (gun); **ákwà** (cloth), **àkwà** (bed or bridge), **àkwá**(egg) (or even **ákwá**(cry) in a philosophical or artistic sense); as well as **égwù**(fear) and **égwú**(music) were all qualified to replace the ambiguous word in their respective sentences.

The ambiguities can span word classes. The examples above showed words that belong to the same class i.e. nouns. However, there are instances of wordkeys that represent actual forms that span different classes. For example, in the first two sentences, **banyere** could mean *bànyèrè* (*entered*, a verb) or *bànyéré* (*about*, a preposition).

### 4.1.4 The *Google Translate* Test

| **Statement** | ***Google Translate*** | **Comment** |
|---|---|---|
| O ji *egbe* ya gbuo *egbe* | He used his **gun** to kill *gun* | wrong |
| O ji **égbè** ya gbuo **égbé** | He used his **gun** to kill **kite** | correct |
| *Akwa* ya di n'elu *akwa* ya | It was on the **bed** in his room | wrong |
| **Ákwà** ya di n'elu **àkwà** ya | his **clothes** on his **bed** | correct |
| *Oke* riri *oke* ya | Her addiction | wrong |
| **Òké** riri **òkè** ya | **Mouse** ate his **share** | correct |
| O jiri *ugbo* ya bia | He came with his *farm* | wrong |
| O jiri **ụgbọ** ya bia | He came with his **car** | correct |

Table 4.2 *Google Translate* on Diacritic and Non-diacritic Texts

Diacritic restoration is important for other NLP systems such as speech recognition, text generation and machine translations systems. Although most translation systems are now very impressive, not a lot of them support Igbo language. However for *Google Translate*, which happens to be the only translation system we know of that supports Igbo, diacritic restoration plays a significant role in how well it performs.

If lack of diacritics introduces ambiguities in written text, it makes sense to assume that there may be a difference in performance of an existing NLP system, say a machine translation system on both diacritically and non-diacriticly marked texts. This assumption led us to inspect the translation outcomes on both types of texts by *Google Translate.*

As an extension to the challenge presented in §3.3, we gave the diacritic and non-diacritic versions of the following Igbo sentences to *Google Translate* to see if there will be any difference in the way they will be translated.

- **O ji égbè ya gbuo égbé.:** i.e. *He used his gun to kill kite.*
- **Ákwà ya di n'elu àkwà ya:** i.e. *His cloths are on his bed.*

- **Òké riri òkè ya:** i.e. *His share was eaten by a mouse.*
- **O jiri ụgbọ ya bịa:** i.e. *He came in his car.*

Table 4.2 shows that, contrary to the impression that the diacritic restoration task may not be so important hence the seeming lack of interest in it, the absence of diacritics can affect existing translation systems significantly in an adverse manner.

## 4.2 Data Types and Sources

Sourcing Igbo data was a problem since there is still very little effort toward gathering, pre-processing and annotating large collections of data Igbo. There are not large bodies of written text available in electronic format. This is mainly because Igbo is a largely spoken language, and not written as much and so it is often difficult to scrape Igbo texts from online platforms. Also, most Igbo literary works are not available in electronic format.

We present the descriptions of the data we used, their sources and the basic statistics on them. Depending on the approach applied to solving the IDR problem defined above, a range of considerations were made regarding the appropriateness of the dataset we use for different experiments. These considerations include the nature of the data we have, the level of ambiguity of each of the wordkeys, the percentage appearance of certain wordkeys and their variants as well as the distribution of the variants of each of the wordkeys. A quick description of the corpora we used in building the different IDR models presented in this work is presented below:

**Igbo Bible** This is the Bible data (***igBible***), which is an Igbo translation of the Bible available from the *Jehova Witness*[1] website. It is our major source of experimental data for this study and constitutes over 93% of our combined corpora with 902,150 word tokens and a vocabulary size of 16,061.

**Igbo Novels** These are two Igbo literary works. One of them, *Mmadụ Ka A Na-Arịa* (***igNovel1***), is a novel written by a Nigerian author Chuma Okeke, which has already been reported by Onyenwe in their work on the development of Igbo tagger [74]. This corpus contributes 35,401 (≈3.68%) word tokens to the experimental corpora. Its vocabulary size is 3,282.

The other novel is an Igbo translation of a popular high school novel *Eze Goes To School* (***igNovel2***), written in English by Onuora Nzekwu and Michael Crowder, translated by Joy Uzoalor, and published on the Nnamdi Azikiwe University, Open Educational Resources[2] website. It has only 23,004 word tokens and vocabulary size of 2,651.

[1] *https://www.jw.org/en/*

**Igbo UDHR** The smallest collection of Igbo text used in this work is the Igbo translation of the Universal Declaration of Human Rights (***igUDHR***) documents which is available from the Unicode Consortium[3]. It contains only 2,192 words with only 498 unique words.

### 4.2.1 Pre-processing and Statistics

The pre-processing of the corpus largely adopted the approach used by Onyenwe *et al.* [77]. This approach preserves certain linguistic properties of the Igbo language due to the special roles the tokens play in distinguishing the word classes. For example, given the root word *"**na**"*, we could decipher which of the linguistic functions it is performing in a sentence by just the way it is written, even without context. For example, *"na"* is mostly a conjunction (e.g. **Anyị riri ji na ede.** i.e. *We ate yam and coco-yam.*). When it is written with a hyphen, *"na-"*, it is a verb-auxiliary (e.g. **Ọ na-agba egwu.** i.e. *He is dancing.*). When it is abbreviated as *"n'"*, it is a preposition (e.g. **Ọdị n'elu tebulu.** i.e. *It is on the table.*)

The different documents used in this work rendered diacritic characters in different formats. The Bible used the *combining diacritics* such as the acute and grave accents or dot-belows, the *IgUDHR* used the Unicode versions of these characters, while the novels had a combination of these formats. Therefore, for consistency, diacritic formats are normalized using the Unicode's *Normalization Form Canonical* NFC composition. For example, the character *é* from the combined unicode characters *e* (u0065) and ´ (u0301) are decomposed and recombined as a single canonically equivalent character *é* (u00e9). Characters like *ñ* and *n̄* are generally replaced with *ṅ* where they are meant to be so.

Tables 4.3 and 4.4 show the relevant individual and combined statistics on the data we used for our experiments. In these counts, the case of each word in the corpus is preserved. For instance, *Ọtụtụ* and *ọtụtụ* have different counts. The table entries are explained as follows:

**Basic Stats** This section of the table gives the basic statistics on the corpus where

[2] *https://bit.ly/2vdGvKN*
[3] *https://unicode.org/udhr/d/udhr_ibo.html*

- *Lines:* indicates the number of *lines* in the data file. These "lines" refer to the verses in the Bible and the sentences in the other documents. This is because we think that the, with the Bible, information will be better preserved in verses.
- *All tokens:* gives the total counts of all the tokens found in the data file
- *Words only:* counts only the words excluding digits, punctuation marks and other non-word symbols
- *Vocab size:* gets the dictionary size which is set of all words that appeared in the data file.

**Diacritics** In this section, we present the statistics on the diacritics on the training data used in this work where:

- *All diac words* shows the total count of all words with at least one diacritic character
- *Unique diac words* gives the total count of all diacritized words that are unique to their wordkeys i.e. they do not share their wordkeys with any other word. These are generally easier to restore.
- *Amb diac words* gives the total counts of all diacritized words that share their wordkeys with at least one other word. These are relatively more challenging to restore and the bulk of our work revolves around restoring them.
- *Diac vocab size* As the name suggests, this gives the length of the dictionary of all diacritized words which is basically the set of all words with diacritics.

**Wordkeys and Variants** This section focuses on wordkeys and variants where

- *All wordkeys* shows the total count of all wordkeys which is in principle getting the set of all words with their diacritics removed. It is basically a shrunken dictionary of word tokens and so can include entries that can generate non-diacritized words.
- *Unique wordkeys* gets the total count of all wordkeys that are unambiguous i.e. they yield a *single* variant. This variant may still be a non-diacritized word. Again, during restoration, these are generally easy to fix.
- *Ambiguous wordkeys* gives the difference between *All wordkeys* and *Unique wordkeys*. Each of them has at least 2 diacritic variants and they have generated all the words in *Amb diac words.* Our experiments will focus mainly on replacing each of these keys with the appropriate variant during the restoration process.

| Item | IgBible | IgNovel1 | IgNovel2 | IgUDHR |
|---|---|---|---|---|
| *Basic Stats:* | | | | |
| – Lines | 32,416 | 2,024 | 1,155 | 90 |
| – All tokens | 1,070,429 | 39,754 | 25,467 | 2,386 |
| – Words only | 902,150 | 35,401 | 23,004 | 2,192 |
| – Vocab size | 16,061 | 3,282 | 2,651 | 498 |
| *Diacritics:* | | | | |
| – All diac words | 502,101 | 15,660 | 10,568 | 681 |
| – Unique diac words | 164,058 | 9,769 | 2,431 | 277 |
| – Amb diac words | 338,043 | 5,891 | 8,137 | 404 |
| – Diac vocab size | 9,146 | 1,795 | 1,406 | 217 |
| *Wordkeys and Variants:* | | | | |
| – All wordkeys | 15,454 | 3,134 | 2,426 | 463 |
| – Unique wordkeys | 14,905 | 2,998 | 2,236 | 436 |
| – Ambiguous wordkeys | 549 | 136 | 190 | 27 |
| – 2 variants | 515 | 127 | 162 | 22 |
| – 3 variants | 19 | 7 | 22 | 3 |
| – 4 variants | 9 | 1 | 5 | 1 |
| – 5 variants | 3 | 1 | 1 | 1 |
| – 6 variants | 3 | – | – | – |

Table 4.3 Individual Corpus Statistics

### 4.2.2 Generating the datasets

In this section, we discuss the structural differences between the two categories of datasets we used for our experiments: The *Basic* and the *Ambiguous* datasets.

#### Basic Dataset

Due to the nature of the task, generating the basic training and testing dataset is relatively simple if we have texts with diacritics. Building the basic dataset involves removing the diacritics on the text to create the non-diacritic version. A simple example of the diacritic and non-diacritic versions of some Igbo text with emphasis on the words that previously had diacritic characters is shown in Table 4.5.

| **Item** | **Combined Corpus** |
|---|---|
| *Basic Stats:* | |
| – Lines | 35,685 |
| – All tokens | 1,138,036 |
| – Words only | 962,747 |
| – Vocab size | 18,455 |
| *Diacritics:* | |
| – All diac words | 559,263 |
| – Unique diac words | 94,673 |
| – Amb diac words | 464,590 |
| – Diac vocab size | 10,655 |
| *Wordkeys and Variants:* | |
| – All wordkeys | 17,446 |
| – Unique wordkeys | 16,591 |
| – Ambiguous wordkeys | 855 |
| – 2 variants | 755 |
| – 3 variants | 67 |
| – 4 variants | 21 |
| – 5 variants | 5 |
| – 6 variants | 5 |
| – 7 variants | 2 |

Table 4.4 Combined Corpus Statistics

In this example, the first three tokens will be returned as they are during the diacritic restoration process. But words with diacritics, e.g. *sị*, *ìhè*, *dị*, will be replaced by the most probable diacritic variants. Our n-gram approach, presented in Chapter 5, will restore each sentence by replacing each token at a time from left to right.

In terms of structure, it retains the same structure as the original diacritic version but each word is basically replaced by its wordkey. This causes the non-diacritic text to maintain a vocabulary size of approximately 5% less that the original diacritic vocabulary size. This seems small but statistically, words with diacritics constitute more than 50% of the entire token size on the main experimental data, the **IgBible** and **IgNovel1**.

| | |
|---|---|
| **Diacritic version:** | 3 Chineke wee **sị** : “ Ka **ìhè dị** . ” **Ìhè** wee **dị** . |
| **Non-Diacritic version:** | 3 Chineke wee **si** : “ Ka **ihe di** . ” **Ihe** wee **di** . |

Table 4.5 Diacritic and Non-Diacritic Texts

Table 4.6 shows a summary of the files that contain the basic datasets. There is a file containing the diacritically marked text of each of the four data genres we used and another file containing the combination of all the texts. Each of these diacritically marked data files ends with "-**marked**.data" and has a file containing the diacritically stripped version of the text which ends with "-**stripped**.data". So in total we have 10 plain text files for the Basic datasets.

| **Filename** | **Lines** | **Tokens** | **Words** | **Wordtypes** |
|---|---|---|---|---|
| baịbụl_**marked**.data | 32,416 | 1,070,429 | 902,150 | 16,061 |
| baịbụl_**stripped**.data | 32,416 | 1,070,429 | 902,150 | 15,454 |
| eze_goes_to_school_**marked**.data | 1,155 | 25,467 | 23,004 | 2,651 |
| eze_goes_to_school_**stripped**.data | 1,155 0 | 25,467 0 | 23,004 0 | 2,426 |
| mmadụ_ka_a_na_arịa_**marked**.data | 2,024 | 39,754 | 35,401 | 3,282 |
| mmadụ_ka_a_na_arịa_**stripped**.data | 2,024 | 39,754 | 35,401 | 3,134 |
| UDHR_**marked**.data | 90 | 2,386 | 2,192 | 498 |
| UDHR_**stripped**.data | 90 | 2,386 | 2,192 | 463 |
| _combined_**marked**.data | 35,685 | 1,138,036 | 962,747 | 18,455 |
| _combined_**stripped**.data | 35,685 | 1,138,036 | 962,747 | 17,446 |

Table 4.6 Basic Dataset: File names and details of the diacritically marked and stripped texts

**Ambiguous Dataset**

The Basic dataset gives us a quick and relatively straight-forward approach to understanding the nature of the problem we are solving. In fact, our first experiments will apply this dataset as a proof of concept. But this dataset presents a few issues which we consider necessary to deal with at this stage. The first is that it generates a dataset that requires the restoration of *every single token* in the dataset. But this may not be required as a large amount of all the tokens in the dataset (≈50%) are either originally without diacritics (e.g. *Chineke*) or are unambiguous (e.g. *mmadụ*). These tokens clearly pose little or no restoration challenge.

Therefore, this approach is not only inefficient but it also returns an accuracy value that could be deceptively high giving the impression that our model is doing better that it actually does. This is because it gets a free score on each of the non-diacritic and unambiguous tokens. So it makes sense to create a dataset with only ambiguous wordkeys, their context and the correct diacritic variant as the classification label. Table 4.7 shows an example of the structure of our datasets with one of the ambiguous wordkeys, *akwa*, and the instances for each of the variants.

Like *akwa*, there are 855 ambiguous wordkeys in total from the combined corpus (see Table 4.4). However, in generating the datasets, we used the lower case version of the corpus which collapsed the wordkeys to 795. The counts of the top 25 most occurring wordkeys and their variants are presented in Table 4.8 which throws up a few irregularities. For instance, there some wordkeys with ambiguous sets with distributions skewed to one variant e.g. *gi*, *unu*, *anyi*, *umu*, *aka*, *ike*, *eze* and *ulo*.

| Meaning | Variant | Context |
|---|---|---|
| cloth | ákwà | o bu ______ isi e ji linin mee ka ha ga- eke n' isi ha ga- eyi uwe ukwu e ji linin mee .. |
| cloth | ákwà | ... o biliri n' ocheeze ya wee yipu uwe eze ya ma yiri ______ iru uju wee nodu ala na ntu |
| cloth | ákwà | ma lee ______ mgbochi nke ebe nso dowara uzo abuo malite n' elu ruo n' ala ala ... |
| cry | ákwá | oke mkpu ______ ga- ada n' ala ijipt dum nke udi ya na- adatubeghi nke o na- enweghikwa ... |
| cry | ákwá | nzuko ahu dum wee malite ibe akwa ha wee na- eti mkpu na- akwakwa ______ n' abali ahu dum ... |
| cry | ákwá | ma o no na- ebesara ya ______ ruo ubochi asaa a no na- emere ha oriri o wee ruo n' ubochi nke ... |
| egg | àkwá | ma o bu o buru na o rio ya ______ ya enye ya akpi |
| egg | àkwá | di ka okwa nke kpokotara ______ o na- eyighi otu ahu ka onye ji ikpe na- ezighi ezi ... |
| egg | àkwá | ... onya ududo onye o bula nke riri akwa ya ga- anwu ajuala ga- esikwa n' ______ ya nke kuwara akuwa puta |

Table 4.7 Sample instances for the wordkey *akwa* and its variants: *ákwà* (cloth), *ákwá* (cry) and àkwá (egg)

### Variant Distribution

With the heavily skewed wordkeys, choosing the most common variant will guarantee an accuracy of over 90% and therefore does not pose so much restoration challenge. Further inspection of our data reveals that while the training data contains a substantial number of diacritics, it is not entirely perfect. In some cases, there are obvious errors, for example the wordkeys *umu* and *ulo* should have **ụmụ** and **ụlọ** as the only valid variants but there are more invalid variants for both. The same goes for *gi*, *unu*, *anyi* and so on. However, for the wordkey *aka*, there are actually two variants **aka** (hand) and **ákà** (ivory bead), but the latter has only one instance in the dataset.

Therefore, in generating the datasets, we considered pruning our dataset by removing some high frequency, but low *entropy* ambiguous sets where using the most common class produces very high accuracies. Entropy is loosely used here to refer to the degree

**Most Common Wordkeys and Their Variants**

| *Wordkey* | *Count* | *#Varnts* | *Variant Counts* |
|---|---|---|---|
| **o** | 32842 | 6 | (ọ, 23123), (o, 8323), (ọ̀, 1053), (ọ́, 252), (ò, 83), (ó, 8) |
| **na** | 31713 | 2 | (na, 30364), (ná, 1349) |
| **ha** | 23818 | 4 | (ha, 23314), (hà, 477), (há, 26), (hā, 1) |
| **ndi** | 23727 | 2 | (ndị, 23688), (ndi, 39) |
| **m** | 19109 | 3 | (m, 19038), (m̀, 67), (ḿ, 4) |
| **bu** | 17288 | 7 | (bụ, 10949), (bọ́, 6050), (bu, 279), (bụ́, 6), (bú, 2), (bù, 1), (bụ̀, 1) |
| **ihe** | 16093 | 2 | (ihe, 15670), (ìhè, 423) |
| **di** | 14884 | 2 | (dị, 14573), (di, 311) |
| **a** | 14494 | 2 | (a, 14366), (à, 128) |
| **ahu** | 13736 | 3 | (ahụ, 12371), (ahụ́, 1361), (ahu, 4) |
| **onye** | 11343 | 3 | (onye, 10969), (ònye, 372), (ọnye, 2) |
| **gi** | 10104 | e 2 | (gị, 10100), (gi, 4) |
| **ma** | 9994 | 2 | (ma, 9993), (mà, 1) |
| **unu** | 9850 | 4 | (unu, 9753), (ùnu, 94), (unụ, 2), (ụnu, 1) |
| **si** | 9040 | 3 | (sị, 5182), (si, 3857), (sì, 1) |
| **ebe** | 7110 | 2 | (ebe, 7103), (èbè, 7) |
| **e** | 7075 | 2 | (e, 7041), (è, 34) |
| **anyi** | 6144 | 3 | (anyị, 6110), (ànyị, 31), (anyi, 3) |
| **otu** | 6004 | 6 | (otu, 3963), (otú, 1984), (òtù, 50), (òtú, 5), (òtụ́, 1), (ọtụ, 1) |
| **i** | 5572 | 8 | (ị, 3603), (i, 1744), (ị̀, 185), (ì, 23), (ị́, 13), (í, 2), (ị́, 1), (ị̄, 1) |
| **umu** | 4986 | 3 | (ụmụ, 4983), (umụ, 2), (ụmu, 1) |
| **aka** | 4052 | 2 | (aka, 4051), (ákà, 1) |
| **ike** | 3652 | 2 | (ike, 3637), (ikè, 15) |
| **eze** | 3546 | 2 | (eze, 3492), (ezé, 54) |
| **ulo** | 3539 | 4 | (ụlọ, 3536), (ulo, 1), (ulọ, 1), (ụlo, 1) |

Table 4.8 Counts of Top Most Common Wordkeys and Their Variants)

**Least Common Wordkeys and Their Variants**

| *Wordkey* | *Count* | *#Varnts* | *Variant Counts* |
|---|---|---|---|
| **ebezila** | 2 | 2 | (ebezila, 1), (ebezịla, 1) |
| **koshai** | 2 | 2 | (kọshai, 1), (kọshaị, 1) |
| **emesiwo** | 2 | 2 | (emesiwo, 1), (emesịwo, 1) |
| **kwugbuola** | 2 | 2 | (kwugbuola, 1), (kwụgbuola, 1) |
| **akwusikwa** | 2 | 2 | (akwụsikwa, 1), (akwụsịkwa, 1) |
| **zaghachi** | 2 | 2 | (zaghachi, 1), (zaghachị, 1) |
| **idu** | 2 | 2 | (idu, 1), (ịdụ, 1) |
| **sichara** | 2 | 2 | (sichara, 1), (sịchara, 1) |
| **kunyere** | 2 | 2 | (kunyere, 1), (kụnyere, 1) |
| **emesonu** | 2 | 2 | (emesonu, 1), (emesonụ, 1) |
| **enyezila** | 2 | 2 | (enyezila, 1), (enyezịla, 1) |
| **kuputara** | 2 | 2 | (kupụtara, 1), (kụpụtara, 1) |
| **sekpuuru** | 2 | 2 | (sekpuuru, 1), (sekpụụrụ, 1) |
| **asompi** | 2 | 2 | (asompi, 1), (asọmpi, 1) |
| **kwenyesiri** | 2 | 2 | (kwenyesiri, 1), (kwenyesịrị, 1) |
| **buruchaa** | 2 | 2 | (buruchaa, 1), (bụrụchaa, 1) |
| **bukarisiri** | 2 | 2 | (bukarịsịrị, 1), (bụkarịsịrị, 1) |
| **tulie** | 2 | 2 | (tulie, 1), (tụlie, 1) |
| **atuli** | 2 | 2 | (atuli, 1), (atụli, 1) |
| **ebidozi** | 2 | 2 | (ebidozi, 1), (ebidozị, 1) |
| **arubu** | 2 | 2 | (arụbu, 1), (arụbụ, 1) |
| **onweghi** | 2 | 2 | (onweghi, 1), (onweghị, 1) |
| **ntuli** | 2 | 2 | (ntuli, 1), (ntụli, 1) |
| **nnochite** | 2 | 2 | (nnọchite, 1), (nnọchịte, 1) |
| **nkuwuwaputa** | 2 | 2 | (nkuwuwaputa, 1), (nkuwuwapụta, 1) |

Table 4.9 Counts of Least Common Wordkeys and Their Variants)

of dominance of a particular class across the dataset and it is simply defined as:

$$entropy = 1 - \frac{max[Count(label_i)]}{len(dataset)}$$

where $i = 1..n$ and $n$ =number of distinct labels in the dataset.

### Appearance Threshold

Table 4.9 shows the list of the least common wordkeys and their variants. These ambiguous sets, though quite evenly distributed, are grossly under-represented in the corpus. In most of them, we also observed that obvious errors or dialectal inconsistencies contributed to creating ambiguous sets for some of the wordkeys that would have otherwise been unambiguous. For example, *ebezila* should be **ebezila** i.e. without diacritics, and *onweghi* should be **onweghị**.

Generally, such ambiguous sets do not present enough data to build a robust and reliable model and are therefore excluded from the training datasets. To do so, we defined an appearance threshold, *appThreshold*, which every wordkey and its variants will have to meet before being considered for inclusion in the dataset. Wordkeys with *appThreshold* below a certain stated value were removed from the experiment. The appearance threshold is defined as follows:

$$appThreshold = \frac{C(wordkeys)}{C(tokens)} * 100$$

There are also cases where a combination of the appearance threshold of a particular wordkey and its variant distributions may be necessary to prune some of the variants of the wordkey. For example, with a frequency of 9,040, the wordkey *si* and its three variants (***sị***=5182, ***si***=3857, ***sì***=1) are substantially represented in the data but the variant ***sì*** seems somewhat misplaced and is removed from the datasets.

In Table 4.10, we identified, after the pruning, only 29 ambiguous sets that met the requirements stipulated in §4.2.2. These ambiguous sets generated a total of 80,844 instances with the highest, **o**, giving *31,442* instances while the lowest, **ju**, gave only *97* instances.

Also, it can be observed (compare Tables 4.8 and 4.10) and the counts of the relevant variants for each of the ambiguous wordkeys dropped in some of wordkeys with high variant counts. For example, the wordkey **o** retained only 2 of its 6 variants (*ọ* and *o*) after the pruning process. Same goes for **bu**, **i** and others. 22 out of 29 of the remaining wordkeys in the dataset have only 2 variants; 6 wordkeys have 3 variants while only one, **igwe**, has 4 variants.

| Wordkeys | nVars | Counts | Variant counts |
|---|---|---|---|
| **o** | 2 | 31,446 | (o, 8323), (ọ, 23123) |
| **bu** | 2 | 16,999 | (bụ́, 6050), (bụ, 10949) |
| **si** | 2 | 9,039 | (si, 3857), (sị, 5182) |
| **otu** | 2 | 5,947 | (otu, 3963), (otú, 1984) |
| **i** | 2 | 5,347 | (i, 5,347), (ị, 3603) |
| **oke** | 3 | 2,267 | (oké, 1773), (òkè, 159), (ókè, 335) |
| **ukwu** | 3 | 1,432 | (ukwu, 537), (úkwù, 147), (ụkwụ, 748) |
| **igwe** | 4 | 1,392 | (igwe, 136), (ìgwè, 920), (ígwè, 129), (ígwé, 207) |
| **ama** | 3 | 1,353 | (ama, 570), (àmà, 279), (ámá, 504) |
| **akwa** | 3 | 1,191 | (akwa, 324), (ákwà, 504), (ákwá, 363) |
| **ibu** | 2 | 682 | (ibu, 438), (ịbụ, 244) |
| **ruru** | 2 | 488 | (ruru, 242), (rụrụ, 246) |
| **aku** | 3 | 384 | (akụ̀, 188), (akụ́, 116), (akụ, 80) |
| **iru** | 2 | 333 | (iru, 177), (ịrụ, 156) |
| **juru** | 2 | 306 | (juru, 142), (jụrụ, 164) |
| **nku** | 2 | 285 | (nku, 110), (nkụ, 175) |
| **okpukpu** | 2 | 211 | (okpukpu, 61), (ọkpụkpụ, 150) |
| **doro** | 2 | 205 | (doro, 131), (dọrọ, 74) |
| **iso** | 2 | 201 | (iso, 101), (isọ, 100) |
| **buuru** | 2 | 180 | (buuru, 60), (bụụrụ, 120) |
| **onya** | 3 | 160 | (ọnya, 9), (ọnyà, 85), (ọnyá, 66) |
| **inu** | 2 | 156 | (ịnụ, 90), (ịǹụ, 66) |
| **odo** | 2 | 154 | (odo, 112), (ọdọ, 42) |
| **ikpo** | 2 | 153 | (ikpo, 51), (ịkpọ, 82) |
| **too** | 2 | 125 | (too, 89), (tọọ, 36) |
| **doo** | 2 | 120 | (doo, 84), (dọọ, 36) |
| **wuru** | 2 | 112 | (wuru, 75), (wụrụ, 37) |
| **agbago** | 2 | 99 | (agbago, 53), (agbagọ, 46) |
| **ju** | 2 | 97 | (ju, 72), (jụ, 25) |
| **Total** | **80,844** | | |

Table 4.10 Ambiguous Dataset: Remaining wordkeys and their variants after pruning from the most frequent to the least frequent

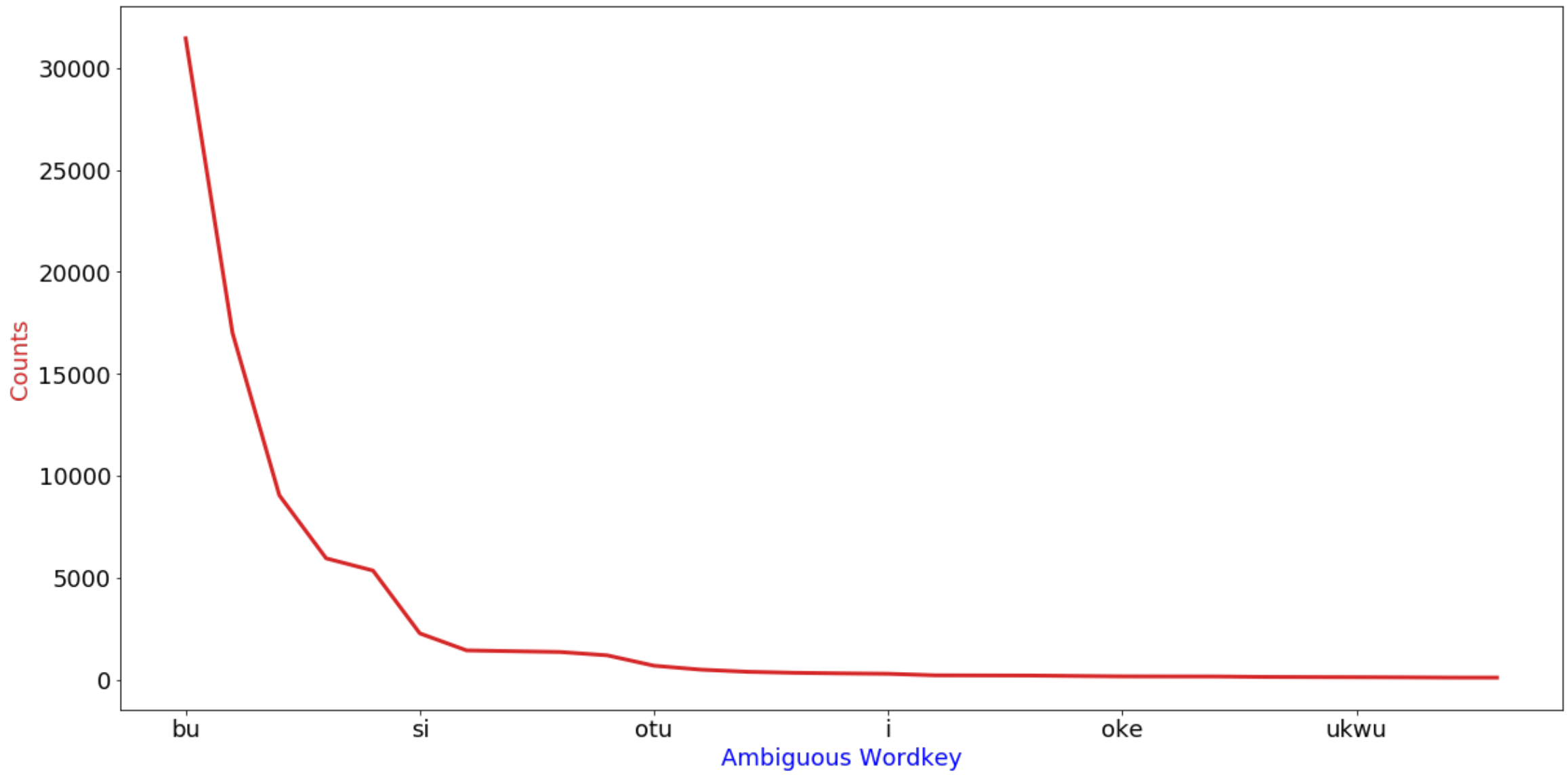

Fig. 4.3 A plot of the counts of the ambiguous wordkeys

### 4.2.3 Generic Framework for Data Processing

While processing the data, we realized that there is a need to build a framework for pre-processing the diacritized corpus for the purpose of generating the dataset. With this framework, we can easily scale up the data size, adjust the variant distribution (*varntDistrb*) and the appearance threshold settings during experiments depending on the nature of the task at hand.

In the framework, there are three basic parameters used determine how instances of the datasets are generated: *varntRep*, *wdkeyRep* and *varntDistrb*. Their descriptions and default settings are given as follows:

***varntRep*****:** This is variant representation which defines the minimum percentage contribution of a particular variant to the total count of its wordkey. Its default value is **0.05** which indicates that for a variant to be considered, it should have a minimum of 5% representation among other variants with the same wordkey. If a variant does not meet this requirement, it will be dropped and its count will be subtracted from the total count of the wordkey.

***wdkeyRep*****:** After pruning all wordkeys of under-represented variants, the wordkey representation, *wdkeyRep*, is applied. It defines the appearance threshold of a wordkey in the entire corpus. The default setting imposes an average minimum appearance of one variant of a wordkey per 10,000 word tokens of the corpus i.e. its default value is 0.0001. This is because we considered it appropriate to have $\approx$ 100 instances in the dataset of each wordkey given that we have about 1m tokens in total.

***varntDistrb*:** For the wordkeys that survive the above constraints, we will need to look at the variant distribution, *varntDistrb* in the dataset of the wordkey. In principle, we look at how evenly distributed the remaining variants are. If a variant of a wordkey dominates the others so much that by choosing to restore the wordkey with it in all cases, we get an accuracy of above 75%, we will consider the variant distribution as heavily skewed and the wordkey will be removed from the dataset generation.

This framework is built in Python and will be publicly available as part of this work. These parameters could be optimised depending on the language, the quality of training data and any other possible considerations that the user deems necessary. It simply takes a list of file names (or paths) and it generates both corpus statistics and datasets using the default parameter values.

Where necessary, it is also possible to pass other parameter values to override the default. However, it is important to note that the framework assumes that the data has been pre-processed and tokenised as it does not handle such tasks. This is necessary to keep it as language-independent as possible.

## 4.3 Evaluation: Method and Metrics

The IDR models we introduce in this work are trained and tested with the datasets we have presented above. In this section, we discuss the methods and the metrics we used in the evaluation of the performance of the models built.

### 4.3.1 Evaluation Method

Given a model and an instance (i.e. a sequence of non-diacritized tokens), we are interested in comparing its output with the label of the instance as contained in the dataset. Therefore, our general method for evaluating a model is to compute the average of the total counts of all (correctly and incorrectly) classified instances in our cross-validated test sets for each ambiguous wordkey.

With these counts, we can then construct a confusion-matrix as shown in Figure 4.4 for all the predicted labels against the true ones from which we can compute each of metric scores. This enables us to identify not just the counts of all correctly predicted labels but also to specifically see for each label, these four important counts: *true positives*, **TP**, *true negatives*, **TN**, *false positives*, **FP** and *false negatives*, **FN**.

Now remember that, as presented in Table 4.10, we have 29 ambiguous wordkeys. So the final score of the model is a weighted average of all the wordkey scores on each of the metrics. For our model evaluation, the metrics we will use include the following: *accuracy*, *precision*, *recall* and *F1-Score* and they will be computed using the confusion-matrix, an example of which is presented in Figure 4.4 for the **o** wordkey.

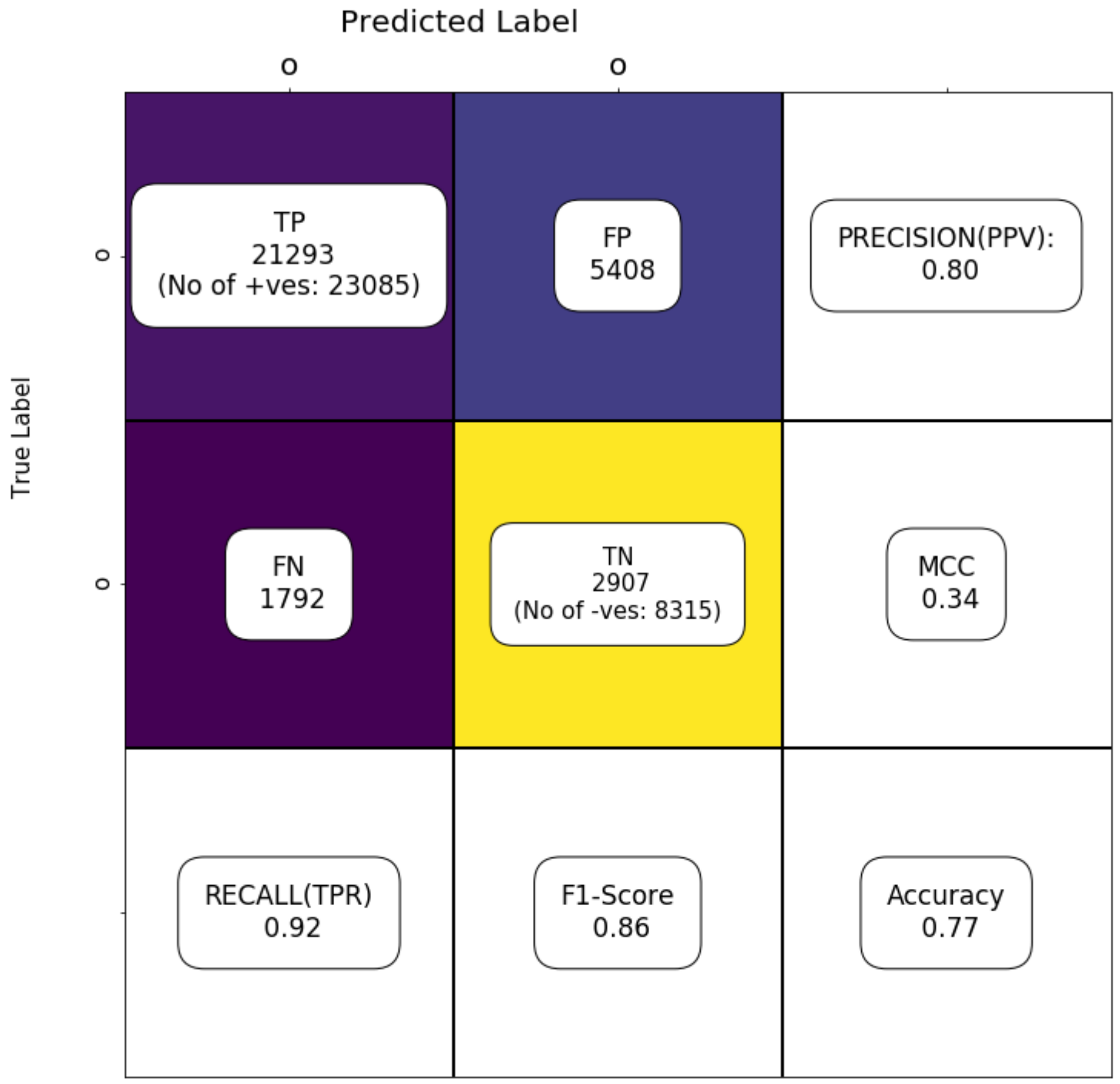

Fig. 4.4 The confusion matrix on the restoration results of the ambiguous set **o**:[**ọ**, **o**] using the 5-gram model.

### 4.3.2 Evaluation Metrics

The IDR task is a typical classification problem. Our evaluation approach compares the predicted variant of a wordkey with the instance label for each instance and accumulates the counts. With these counts we applied the standard evaluation metrics: *accuracy*, *precision*, *recall* as well as *F1* measures.

#### Accuracy

The performance of classification models is often evaluated by computing the basic accuracy. This is the proportion of the correctly predicted variants over the total predictions made. Simply put, it is the fraction of correct predictions over $n$-instances of a given wordkey.

Let's assume that the actual variant for a given $i$-th instance is $y_i$ and the corresponding predicted variant is $\hat{y}_i$, then given pairs of their actual and predicted variants, the definition of *accuracy* over $n$-instances is given by:

$$accuracy(y_i, \hat{y}_i) = \frac{1}{n_{instances}} \sum_{i=0}^{n_{instances}-1} 1(y_i = \hat{y}_i)$$

However, using the elements of the confusion matrix, it will be simply defined as:

$$accuracy = \frac{TP + TN}{TP + TN + FP + FN}$$

Although accuracy is a very simple and useful metric, it can also sometimes be very misleading. If a class is highly dominant, the model may return relatively high accuracy by always predicting this class correctly while making a lot of errors on the classes that are more critical to the task.

In general, accuracy measures are usually not the best for unbalanced or heavily skewed datasets[81]. Although we had included steps to ensure that we deal mostly with fairly balanced ambiguous sets, we will be applying some of the standard metrics to our results data to measure their performances.

#### Precision

Precision, as the name suggests, measures "how precise" the model is in predicting a given class. It returns the percentage of the true predictions for that label that are *really true.* It also known as the *positive predictive value* (**PPV**) and is defined as:

$$precision = \frac{TP}{TP + FP}$$

#### Recall

Recall is also referred to as *sensitivity*, *hit rate* or *true positive rate* (**TPR**) measures how many of the actual true positives in the dataset are captured in the **TP** counts. It is defined as:

$$recall = \frac{TP}{TP + FN}$$

#### F1-score

The F1-score returns the *Harmonic mean*[4] of the precision and recall. This is necessary because although one aims at maximizing the two metrics simultaneously, it is often hard to do that:

$$\text{F1-score} = 2 \cdot \frac{precision \cdot recall}{precision + recall} = \frac{2TP}{2TP + FP + FN}$$

### 4.3.3 Multi-class Averaging

The equations given above generally appear to assume that the task is a binary classification problem where we focus on splitting the test set into *true positives* and *true negatives.* Although we have over 75% (22 out of 29) of ambiguous datasets with only 2 labels, the rest of the datasets have 3 and 4 labels. Fortunately, there are implementations of the generalised versions of these metrics that support multi-class task evaluation on *scikit learn*[5] which we used to generate our scores.

In multi-class computation of the metric scores, three different types of averages are often used: *micro*, *macro* and *weighted*[6]. In micro-averaging, all the metrics – *accuracy*, *precision*, *recall* and *F1* – produce the same scores. We have also created a dataset containing only the wordkeys with fairly balanced distribution of variants and so the weighted-averaging may not also be necessary.

[4]https://en.wikipedia.org/wiki/Harmonic _mean#Harmonic_mean_of_two_numbers

[5]http://scikit-learn.org/stable/modules/model_evaluation.html

[6]In *macro-averaging*, we compute the metric score for each class independently and then take the average hence treating the contributions of all classes equally. A good explanation of *macro-*, *micro-* and *weighted-averages* is available on https://datascience.stackexchange.com/questions/15989/micro-average-vs-macro-average-performance-in-a-multiclass-classification-setting

Technically, micro-averaging produces the same results for all the metrics. So in this work, our reported metric scores are computed with macro-averaging on each of the variants of the individual wordkeys in the dataset. This gives us the macro-average scores on each of the 29 wordkeys in our dataset. However, our final score for each model is computed by aggregating these wordkey scores in a weighted-average form. This is achieved by summing the products of the wordkey scores and their instance counts and dividing the sum with the total instance count. It is important to mention that, in macro-averaging, the accuracy score is often equal to the recall score. Therefore, in most of our analysis, we will focus on only the precision, recall and F1.

## 4.4 Chapter Summary

The chapter sets the stage for the rest of experiments on this project. In this chapter, we presented a definition of the diacritic restoration problem with regards to Igbo. We also discussed the considerations that were involved in our data collection and pre-processing as well as dataset generation for the diacritic restoration tasks. Our data type and sources were presented (§4.2) with the analysis of their individual (Table 4.3) and combined statistics (Table 4.4).

In the dataset generation, we described the **Basic Dataset** with application and limitations as well as the need for the **Ambiguous Dataset** and the processes involved in creating them. A description of the standard generic framework for creating datasets for wordkeys from any corpus or corpora with diacritics irrespective of the language is also presented. The key parameters of the framework variant representation (*varRep*), wordkey representation (*wdkeyRep*) and variant distribution *varntDistrb* with default values of 0.05, 0.0001 and 0.7 respectively.

The method and metrics used in evaluating our restoration models were also presented. The evaluation method basically relies on constructing a confusion-matrix from the counts of the correctly and incorrectly predicted diacritic variants for each wordkey and taking a weighted average across all wordkeys. We also discussed the metrics we used which are some of the most common standard metrics for evaluating classification tasks e.g. *accuracy*, *precision*, *recall* and *f1-score*.

# Chapter 5

# IDR with N-Gram Models

The n-gram model is a simple but effective language model that uses the probability distributions of all words and sequences of words in the text to predict which word could come next. It has been successfully applied to speech and language processing tasks like speech recognition, handwriting recognition, spelling correction and even in machine translation.

In this chapter, we will describe the processes involved in developing the Igbo Diacritic Restoration (IDR) models using n-gram modelling techniques. We will evaluate the performance of $n$-grams (with $n = 1 \ldots 5$) using the metrics introduced in chapter 4.

## 5.1 Language Models

Language models (LMs) are probability distributions over all the words in the language and are used to assign probabilities to sequences of words. Like in most NLP tasks, language models play a key role in developing the methodology for solving diacritic restoration problems. The simplest language model of sentences and sequences of words is the $n$-gram model.

### 5.1.1 N-Gram Models

An $n$-gram is a sequence of $n$ words from a given text. For example, if we have the sentence: *Give me the book*, a 1-gram (unigram) is a one-word sequence of words i.e. “Give”, “me”, “the” or “book”; also a 2-gram (or bigram) is a two-word sequence like “Give me”, “me the” or “the book”; while a 3-gram (or trigram) is a three-word sequence like “Give me the” or “me the book”.

$N$-gram models can be used to estimate the probability of the last word of an $N$-gram given the previous words and also to assign probabilities to entire sequences. They have proven to be an effective technique in speech and language processing. In computing the probability $P(w|h)$, of a word, $w$ (say "*book*"), given a history, $h$ (say "*give me the*"), we have:

$$P(w|h) = \frac{C(h,w)}{C(h)}$$

i.e.

$$P(book|give\ me\ the) = \frac{C(give\ me\ the\ book)}{C(give\ me\ the)}$$

which simply divides the number of times the text *give me the* appeared in the corpus followed by *book* with the total number of times it appeared.

### 5.1.2 Hidden Markov Models

A background to our application of the $n$-gram model worth mentioning is the Hidden Markov Model, HMM which is a probabilistic function of a *Markov process*[1] that gives a higher level of abstraction. A version of it was applied to the Spanish diacritic restoration task by Crandall [20].

Hidden Markov Models (HMM) are a kind of Bayesian network. It is a tool for representing probability distributions over sequences of observations. It assumes that the observations are sampled at discrete, equally-spaced time intervals. It is a generative sequence model that creates a *hidden* structure which highlights the order of *categories* of words in a sequence. It is very commonly applied to tasks such as part-of-speech tagging, which is often an important step to building robust disambiguation systems.

The HMM is defined by two main properties:

1. the observation at time $t$ is generated by some process whose state $S_t$ is hidden from the observer

2. the state of this hidden process obeys the Markov assumption i.e. the current state $S_t$ is independent of all other states before $S_{t-1}$

So the basic definition of the HMM is based on the assumption that the previous state $S_{t-1}$ encapsulates all the required information about the history of the process to be able to predict the current state $S_t$. Therefore, suppose we have a sequence

[1] Markov processes - also referred to as *Markov chains* or *Markov models* - are based on the work by Andrei A Markov to model the sequences of letters in works of Russian literature [10] but has since become general statistical tools

$W = (w_1, \ldots, w_t)$ of random variables (e.g. words) each belonging to an element of some finites set $T = (t_1, \ldots, t_n)$ (e.g. tags), which is the state space, then $W$ is said to have the *Markov Properties* which are defined below as:

$$\begin{aligned} &P(w_{t+1} = t_k | w_1, \ldots, w_t) \\ &\qquad = P(w_{t+1} = t_k | w_t) \qquad\qquad (\textit{limited horizon}) \\ &\textit{and so}: \\ &\qquad = P(w_2 = t_k | w_1) \qquad\qquad (\textit{time invariant}) \end{aligned}$$

### 5.1.3 Maximum Likelihood Estimation (MLE)

Given that language is creative and produces new sentences all the time, it may be difficult to compute the probabilities of all the possible sequences of words in a sentence, [53]. So a better way is to represent a particular sentence as a sequence of words in the form: $w_1, w_2, \ldots, w_n$, or $w_1^n$. Applying the chain rule, $P(w_1, w_2, \ldots, w_n,)$ or $P(w_1^n)$ will be defined by:

$$\begin{aligned} P(w_1^n) &= P(w_1)P(w_2|w_1)P(w_3|w_1^2)\ldots P(w_n|w_1^{n-1}) \\ &= \prod_{k=1}^{n} P(w_k|w_1^{k-1}) \end{aligned}$$

Using *Markov Assumption*[2], the probability of a word given its entire history, can be approximated by just the last few words with the N-gram model. For example, with the bigram model,

$$P(\textit{book}|\textit{Please}\ ,\ \textit{can you give me the})$$

is approximately the same as

$$P(\textit{book}|\textit{the})$$

therefore,

$$P(w_k|w_1^{k-1}) \approx P(w_k|w_{k-1})$$

Maximum Likelihood Estimation (MLE) is an intuitive way to estimate $N$-gram probabilities. It is simply the count of the sequence divided (or normalized) by the count of the previous history. For example, to compute the bigram probability of a

[2]As highlighted in §5.1.2, Markov Assumption is the idea that a future event (in this case, the next word) can be predicted using a relatively short history (e.g. one or two words).

word $w_0$ given the previous word $w_{-1}$ using the MLE, we can use the formula:

$$P(w_0|w_{-1}) = \frac{C(w_0, w_{-1})}{C(w_0)}$$

### 5.1.4 MLE for Diacritic Restoration

The simple probability distributions of words and sequences of words play a key role in formulating models that restore diacritics in many of the works reviewed. In our work, we applied the MLE to estimate probabilities ranging from unigrams to 5-grams in building the n-gram models we used. For example, in a typical 5-gram restoration model of a sequence of wordkeys: $w_{-4}, w_{-3}, w_{-2}, w_{-1}, w_0$ for restoring $w_0$ to any of its variants, say $w_{0i}, \ldots, w_{0j}$, we will start by using the counts of the occurrences of the preceding words $w_{-4}, w_{-3}, w_{-2}, w_{-1}, \ldots$ with each of the variables. Each of these preceding words is also pre-restored with its context as described in §5.2.1.

Where these counts are equal (e.g. when none is greater than zero due to data sparsity), we step-down or *back-off* to the lower n-grams by reducing the context i.e. the previous words. For example, backing-off to 4-grams uses the following previous words: $w_{-3}, w_{-2}, w_{-1}, \ldots$, then to trigrams: $w_{-2}, w_{-1}, \ldots$, then to bigrams $w_{-1}, \ldots$ and finally, if necessary, to unigram counts of the contending variants. Although we currently think that our models could have improved if we had used both the left and right context in our restoration, this report contains the results of experiments with only the left context. This is because we want the models to work on a "generative" sense i.e. predicts the next word given the previous sequence of words.

### 5.1.5 Previous Work

The naïve restoration methods presented in the paper [32] assumed a closed-world scenario where the same set of words are seen at the training and testing stages. This is, of course, *not* standard practice, since the models will then 'over-fit' the test data such that the results are not a reliable indicator of the performance that can be expected. So while we still believe that we have created empirical methods for solving the diacritic restoration problem, we think that the early approach will not effectively accommodate a real world scenario where out-of-vocabulary words could be encountered.

Another point to consider in ensuring that we build a robust restoration system may be to measure our accuracy entirely on the performance of the models on the ambiguous wordkeys. This is because the unique wordkeys in our corpus constitute over 96% (14,905 out of 15,454) of the total wordkeys. So restoring and counting these unique wordkeys gives an illusion of high performance but the actual challenge lies with restoring the one of the remaining 549 ambiguous wordkeys.

The Igbo Bible is a good corpus to start this experiment with. It presents a sizeable amount of data (≈1m token) that is well-written and contains a good level of diacritic marks. But a major issue with it lies in the fact that it does not contain contemporary everyday Igbo word usage. We consider collecting and curating more Igbo text data across multiple genres a necessary task in building NLP resources for Igbo. However, given the time we had for this project, we had to strike a balance between developing relevant tools and corpus building.

In this experiment, we will apply the n-gram models to only the ambiguous sets generated in §4.2.2. Using this dataset ensures that we focus on building models that are more robust in dealing with wordkeys that are more ambiguous. In this experiment, to avoid over-fitting and better mimic the real-world, our approach will apply a cross-validation technique which uses a part of our data for training and the rest for testing.

## 5.2 Experimental Procedure

As we defined in §4.1, the diacritic restoration task is essential and promises to improve the performance of mainstream NLP tasks like machine translation (e.g. *Google Translate*). We therefore define a simple diacritic restoration process that uses a form of *n*-gram language model for Igbo language with the data described in §5.2.2.

As illustrated in Fig. 5.1, given a typical non-diacritic sequence of wordkeys (e.g. *Nwanyi ahu banyere n' ugbo ya .*), we generate, for each ambiguous wordkey, a set of candidate variants (*ahu*→{**àhụ́**, **áhụ̀**}; *banyere*→{**bànyéré**, **bànyèrè**}; *ugbo*→{**ugbo**, **ụgbọ**}) from which we select the most probable candidate based on the language model to replace the wordkey.

### 5.2.1 Restoration Process

As can be seen from the example in Figure 5.1 , we will be interested in restoring only the wordkeys: *ahu*, *banyere* and *ugbo.* Note that although *nwanyi* has a diacritic character, it is not ambiguous so its restoration is considered trivial as it simply replaces the wordkey with its diacritic version. In our ambiguous dataset §5.2.2, there will be a different instance of the same sentence which is labelled with the appropriate variant for each of the three wordkeys as shown. An example of the representation of the different versions of the sentence, *nwanyi ahu banyere n' ugbo ya .* in our dataset is given in Table table 5.1.

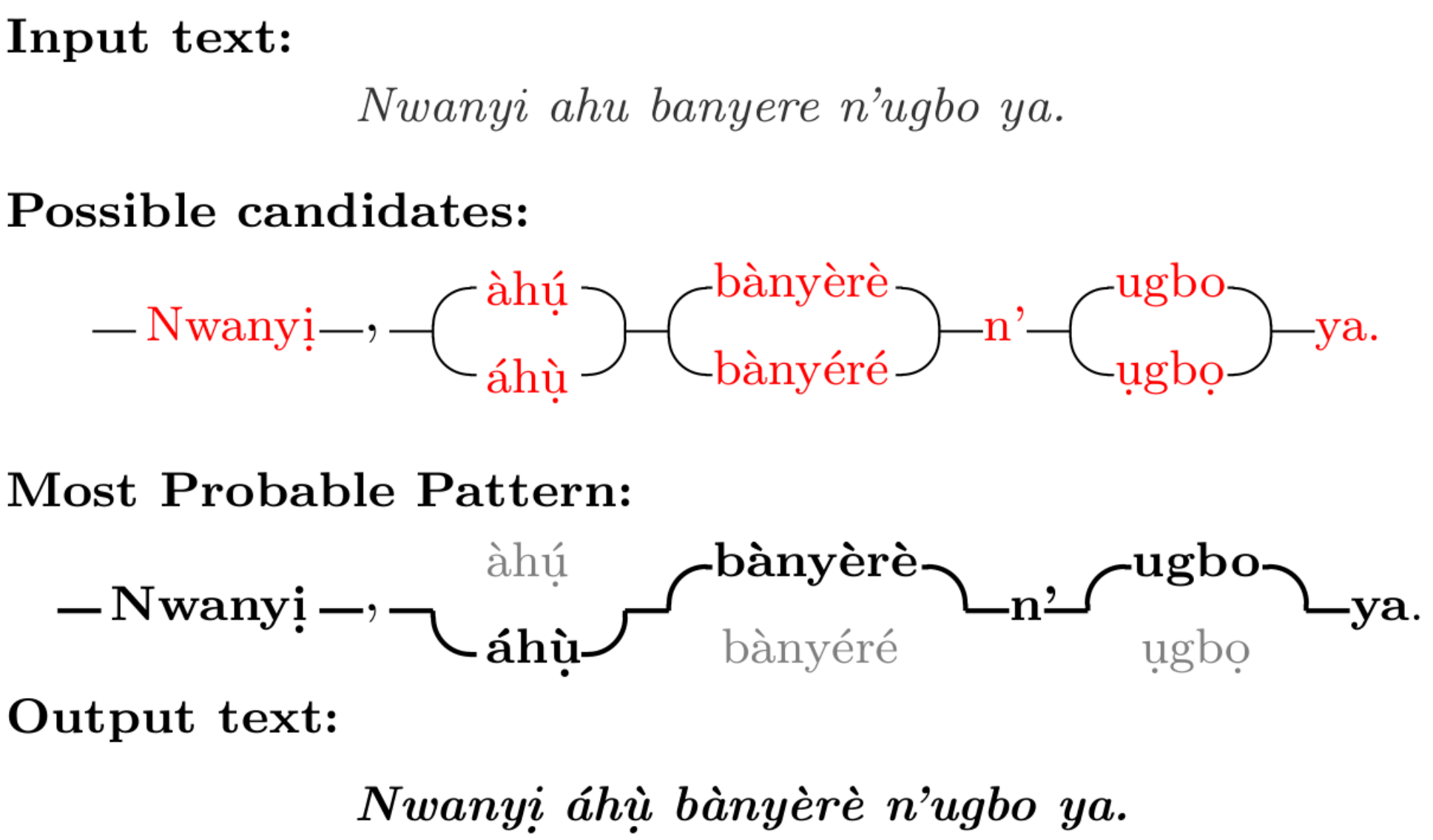

Fig. 5.1 Illustrative View of the Diacritic Restoration Process

| instance | label |
|---|---|
| ... | ... |
| *a. Nwanyi* ***ahu*** *banyere n' ugbo ya .* | **àhụ́** |
| *b. Nwanyi ahu* ***banyere*** *n' ugbo ya .* | **bànyèrè** |
| *c. Nwanyi ahu banyere n'* ***ugbo*** *ya .* | **ugbo** |
| ... | ... |

Table 5.1 Sample restoration instance for *ahu*, *banyere* and *ugbo* on the same sentence.

For example, in restoring an ambiguous wordkey like *ahu* which has two possible variants: **àhụ́** and **áhụ̀** using the unigram model, we simply have to compare the total counts of **àhụ́** and **áhụ̀** i.e. *C(àhụ́)* and *C(áhụ̀)* and then select the variant that yields the highest count. In restoring with the bigram model, we will compare the probabilities: *P(àhụ́ / nwanyị)* and *P(áhụ̀/nwanyị)* and choose the variant that gives the higher value. If they both happen to be equal (e.g. when both have 0 scores), our method defaults to the unigram model. The probability $P(\text{àhụ́}|nwanyị)$ is as defined in Equation 5.1.

$$P(\text{àhụ́}|\text{nwanyị}) = \frac{C(\text{nwanyị}, \text{àhụ́})}{C(\text{nwanyị})} \tag{5.1}$$

In a similar manner, we compare the counts *C(bànyèrè)* and *C(bànyéré)* and select the higher score for the restoration of *banyere* with the unigram model. With the bigram model, we will restore *ahu* to the most likely variant (say **áhụ̀**) with its own context (i.e. nwanyị), and then proceed to restoring *banyere* with **áhụ̀** as the context. If we have zero counts at any stage, we default to a lower n-gram as our back-off strategy.

### 5.2.2 The Ambiguous Dataset

As presented in §4.2.2 (see Table 4.10), the ambiguous dataset is produced by passing the data through the processes of balancing variant distributions as well as having each variant or wordkey meeting the stipulated appearance thresholds within the ambiguous set or the entire corpus. The dataset consists of only 29 ambiguous wordkeys from which 80,844 data instances were generated. As shown in Table 4.7, each instance is made of a sequence of non-diacritic tokens (i.e. a Bible verse or a sentence) with a wordkey to be replaced and labelled with the correct variant for wordkey.

In creating the instances, we represent each wordkey with its own instance i.e. even if there is more than one wordkey of a variant in a sentence, we represent them differently. For example, the sentence *akwa ya di n' elu akwa* (*his/her cloth is the bed*) has two **akwa** wordkeys and therefore will produce two different instances, as shown in Table 5.2, with different labels each aiming to replace a different place-holder. The restoration process identifies the target wordkey and builds its context accordingly.

| **instance** | **label** |
|---|---|
| ... | ... |
| ***akwa*** *ya di n' elu akwa ya* | **ákwà** |
| *akwa ya di n' elu* ***akwa*** *ya* | **àkwà** |
| ... | ... |

Table 5.2 Example of *one-variant-per-instance*

## 5.3 Evaluation and Results

Our evaluation method compares the predicted variant of a wordkey with the instance label for each instance and accumulates the counts. With these counts, we apply the standard evaluation metrics. *accuracy*, *precision*, *recall* as well as *F1* measures. Macro-averaging was used for the computation of the precision, recall and F1 scores on all the variants of each wordkey.

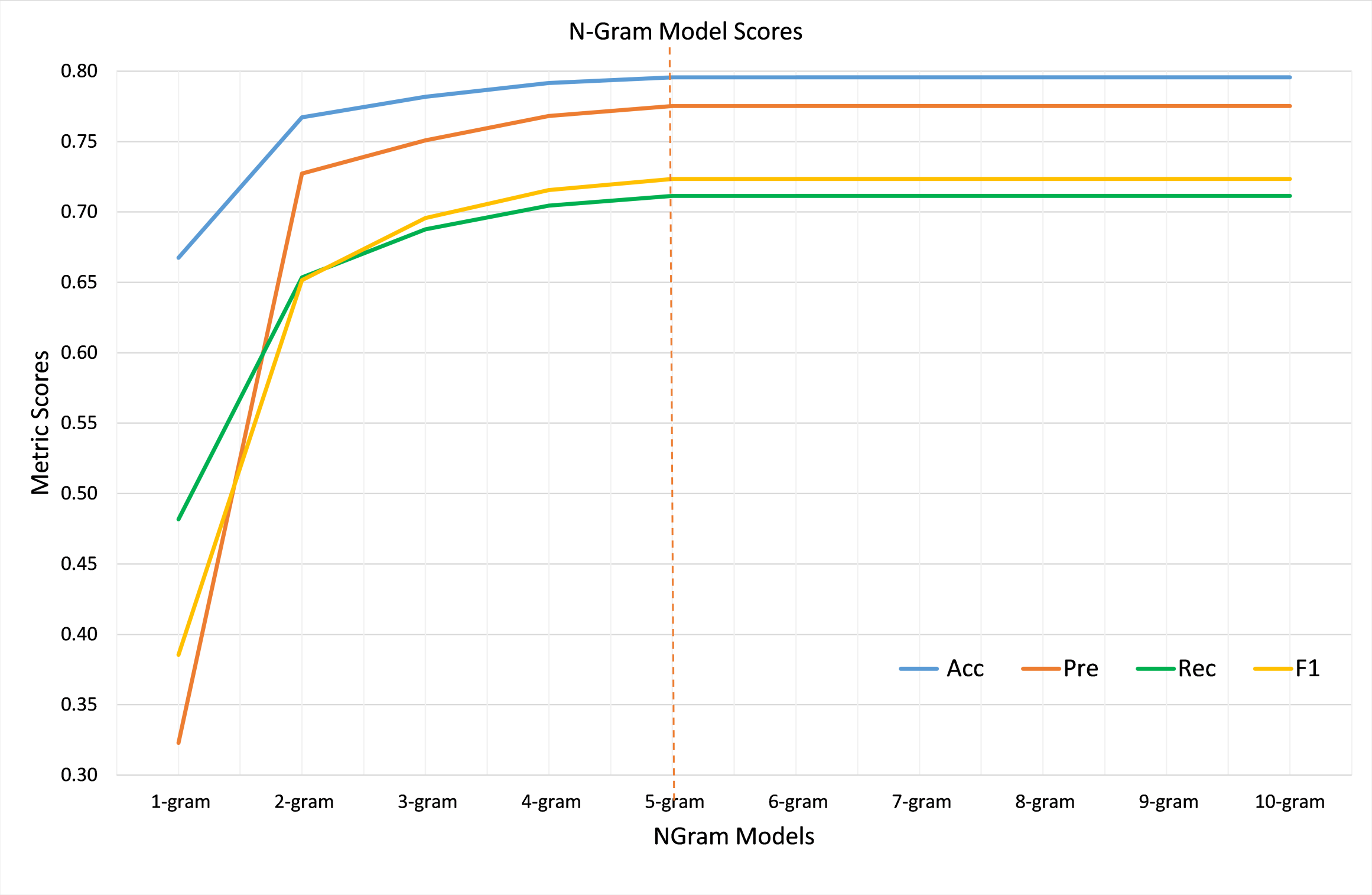

Fig. 5.2 Graph showing the convergence of the accuracy as n-grams increases.

On the n-grams, our experiments indicate, as shown in Figure 5.2, that beyond *5-grams*, there is no further significant improvement in the score. Therefore, though we have experimented with higher n-grams, this report presents only results for up to and including 5-grams across all the wordkeys in our dataset. We used 10-fold cross-validation to avoid over-fitting. The unigram (1-gram) model which simply returns the most common variant is used as the baseline model for all the metrics.

### 5.3.1 Summary of Results

Table 5.3 contains the accuracy, precision, recall and F1 scores of the n-gram models used in this experiment. The results for the **Best** model showed in the table is the systematic aggregation of all models which allows us to use the best model for the restoration of each wordkey. For accuracy, we shall present other details and analysis in §5.3.2. For the other metrics – precision, recall and F1 – there will be a combined analysis of the results in comparison with accuracy in §5.3.3, however most of the details of their results will be presented in Appendix A.

| Model | Acc | Pre | Rec | F1 |
|---|---|---|---|---|
| *1-Gram* | 66.75 | 32.30 | 48.17 | 38.55 |
| *2-Gram* | 77.21 | 73.52 | 66.67 | 66.45 |
| *3-Gram* | 78.68 | 75.82 | 69.94 | 70.66 |
| *4-Gram* | 79.62 | 77.48 | 71.55 | 72.55 |
| *5-Gram* | 80.01 | 78.15 | 72.20 | 73.30 |
| *Best* | 80.05 | 78.19 | 72.21 | 73.31 |

Table 5.3 **Summary:** Table showing the summary of results from all N-Gram models on accuracy, precision, recall and F1. The *Best* model refers to the combination of the best performing models on the wordkeys.

### 5.3.2 Analysis: Accuracy

Accuracy is basically the percentage of correctly predicted variants for each ambiguous set. In Table 5.4, we present the full details of the raw accuracy scores from all n-gram models on all wordkeys as well as the analysis of the results. This table shows the weighted average of the unigram accuracies on all wordkeys which is considered as the global baseline score for evaluating the performance of the model on the wordkeys.

We also have the wordkey baseline which is simply the unigram score on the wordkey itself. The best performance score for each wordkey is the highest score it got across all n-ngram models recorded for the first model that got it. For example, if the n-gram scores for a wordkey is, say, 1-gram=66.75%, 2-gram=72.21%, 3-gram=76.34%, 4-gram=76.34%, 4-gram=76.34%, then the best score is **76.34%** and the best model is **3-gram**. The unigram accuracy score (i.e. **66.75%**) will be our working baseline accuracy score. In the same way, we will use the unigram scores on precision, recall and F1 as the baseline score for each metric.

The graphs below present different levels of clarity and depth to the details that are embedded in Table 5.4. Figure 5.3 shows the best accuracy result achieved on each wordkey as well as the model that performed best.

In terms of performance improvement from the baseline, we reported two types of performance improvements as shown in Table 5.4 and plotted in Figure 5.5. The first is the improvement on the wordkey unigram i.e. how the score of the best model for the wordkey compares with its unigram score. The other is the improvement on the global baseline which compares the score of the best model for a wordkey with the global baseline of **66.75%**.

| *Wordkey* | *Counts* | *No of Variants* | 1-gram | 2-gram | 3-gram | 4-gram | 5-gram | *BestScore (BS)* | *Best Model* | *Wdkey Improvement* | *Baseline Improvement* | *Error Reduction* |
|---|---|---|---|---|---|---|---|---|---|---|---|---|
| **agbago** | 99 | 2 | 0.5354 | 0.5354 | 0.5354 | 0.5354 | 0.5354 | 0.5354 | 1-gram | 0.00% | -13.21% | 0.00% |
| **juru** | 306 | 2 | 0.5359 | 0.5948 | 0.6046 | 0.6046 | 0.6046 | 0.6046 | 3-gram | 6.87% | -6.29% | 14.80% |
| **ama** | 1353 | 3 | 0.4213 | 0.6186 | 0.6253 | 0.6260 | 0.6260 | 0.6260 | 4-gram | 20.47% | -4.15% | 35.37% |
| **buuru** | 180 | 2 | 0.6667 | 0.5222 | 0.5444 | 0.5444 | 0.5444 | 0.6667 | 1-gram | 0.00% | -0.08% | 0.00% |
| **onya** | 160 | 3 | 0.5312 | 0.6438 | 0.6750 | 0.6750 | 0.6750 | 0.6750 | 3-gram | 14.38% | 0.75% | 30.67% |
| **inu** | 156 | 2 | 0.5769 | 0.6923 | 0.6859 | 0.6859 | 0.6859 | 0.6923 | 2-gram | 11.54% | 2.48% | 27.27% |
| **iru** | 333 | 2 | 0.5315 | 0.6997 | 0.7147 | 0.7147 | 0.7147 | 0.7147 | 3-gram | 18.32% | 4.72% | 39.10% |
| **ibu** | 682 | 2 | 0.6422 | 0.7243 | 0.7317 | 0.7302 | 0.7331 | 0.7331 | 5-gram | 9.09% | 6.56% | 25.41% |
| **ruru** | 488 | 2 | 0.5041 | 0.7316 | 0.7418 | 0.7439 | 0.7439 | 0.7439 | 4-gram | 23.98% | 7.64% | 48.36% |
| **ikpo** | 133 | 2 | 0.6165 | 0.7143 | 0.7444 | 0.7444 | 0.7444 | 0.7444 | 3-gram | 12.79% | 7.69% | 33.35% |
| **aku** | 384 | 3 | 0.4896 | 0.7370 | 0.7448 | 0.7474 | 0.7474 | 0.7474 | 4-gram | 25.78% | 7.99% | 50.51% |
| **iso** | 201 | 2 | 0.5025 | 0.7463 | 0.7512 | 0.7512 | 0.7512 | 0.7512 | 3-gram | 24.87% | 8.37% | 49.99% |
| **i** | 5347 | 2 | 0.6738 | 0.7099 | 0.7312 | 0.7490 | 0.7514 | 0.7514 | 5-gram | 7.76% | 8.39% | 23.79% |
| **o** | 31446 | 2 | 0.7353 | 0.7343 | 0.7467 | 0.7623 | 0.7707 | 0.7707 | 5-gram | 3.54% | 10.32% | 13.37% |
| **igwe** | 1392 | 4 | 0.6609 | 0.7507 | 0.7680 | 0.7737 | 0.7723 | 0.7737 | 4-gram | 11.28% | 10.62% | 33.26% |
| **doo** | 120 | 2 | 0.7000 | 0.7750 | 0.7583 | 0.7500 | 0.7500 | 0.7750 | 2-gram | 7.50% | 10.75% | 25.00% |
| **ukwu** | 1432 | 3 | 0.5223 | 0.7661 | 0.7961 | 0.7996 | 0.7996 | 0.7996 | 4-gram | 27.73% | 13.21% | 58.05% |
| **akwa** | 1191 | 3 | 0.4232 | 0.7935 | 0.8069 | 0.8144 | 0.8144 | 0.8144 | 4-gram | 39.12% | 14.69% | 67.82% |
| **otu** | 5947 | 2 | 0.6664 | 0.7963 | 0.8111 | 0.8167 | 0.8177 | 0.8177 | 5-gram | 15.13% | 15.02% | 45.35% |
| **okpukpu** | 211 | 2 | 0.7109 | 0.8057 | 0.8199 | 0.8199 | 0.8199 | 0.8199 | 3-gram | 10.90% | 15.24% | 37.70% |
| **nku** | 285 | 2 | 0.6140 | 0.8246 | 0.8316 | 0.8316 | 0.8316 | 0.8316 | 3-gram | 21.76% | 16.41% | 56.37% |
| **ju** | 97 | 2 | 0.7423 | 0.8351 | 0.8247 | 0.8247 | 0.8247 | 0.8351 | 2-gram | 9.28% | 16.76% | 36.01% |
| **si** | 9039 | 2 | 0.5733 | 0.8339 | 0.8411 | 0.8431 | 0.8448 | 0.8448 | 5-gram | 27.15% | 17.73% | 63.63% |
| **bu** | 16999 | 2 | 0.6441 | 0.8347 | 0.8533 | 0.8584 | 0.8593 | 0.8593 | 5-gram | 21.52% | 19.18% | 60.47% |
| **too** | 125 | 2 | 0.7120 | 0.8320 | 0.8640 | 0.8640 | 0.8640 | 0.8640 | 3-gram | 15.20% | 19.65% | 52.78% |
| **doro** | 205 | 2 | 0.6390 | 0.8732 | 0.8732 | 0.8732 | 0.8732 | 0.8732 | 2-gram | 23.42% | 20.57% | 64.88% |
| **oke** | 2267 | 3 | 0.7821 | 0.8310 | 0.8672 | 0.8738 | 0.8729 | 0.8738 | 4-gram | 9.17% | 20.63% | 42.08% |
| **odo** | 154 | 2 | 0.7273 | 0.8896 | 0.8961 | 0.8961 | 0.8961 | 0.8961 | 3-gram | 16.88% | 22.86% | 61.90% |
| **wuru** | 112 | 2 | 0.6696 | 1.0000 | 1.0000 | 1.0000 | 1.0000 | 1.0000 | 2-gram | 33.04% | 33.25% | 100.00% |
| **Baseline = 1gram; %Error = 33.25%** | | | **66.75%** | **77.21%** | **78.68%** | **79.62%** | **80.01%** | **80.05%** | **Performance Analysis** | Model | Improvement | Error Reduction |
| **Best Model Counts** | | | **2** | **5** | **9** | **7** | **6** | | *Best Score* | ***5-gram*** | **13.25%** | **39.87%** |
| **Model Error Reduction:** | | | 0.00% | 31.45% | 35.87% | 38.71% | 39.87% | 39.98% | *Best Counts* | ***3-gram*** | **11.93%** | **35.87%** |

Table 5.4 **N-Gram Accuracy:** Table showing the full raw accuracy result data and analysis for all n-gram models. The color code indicates the worst(red)-to-best(green) based on the accuracy performances and improvements on wordkey and global baselines

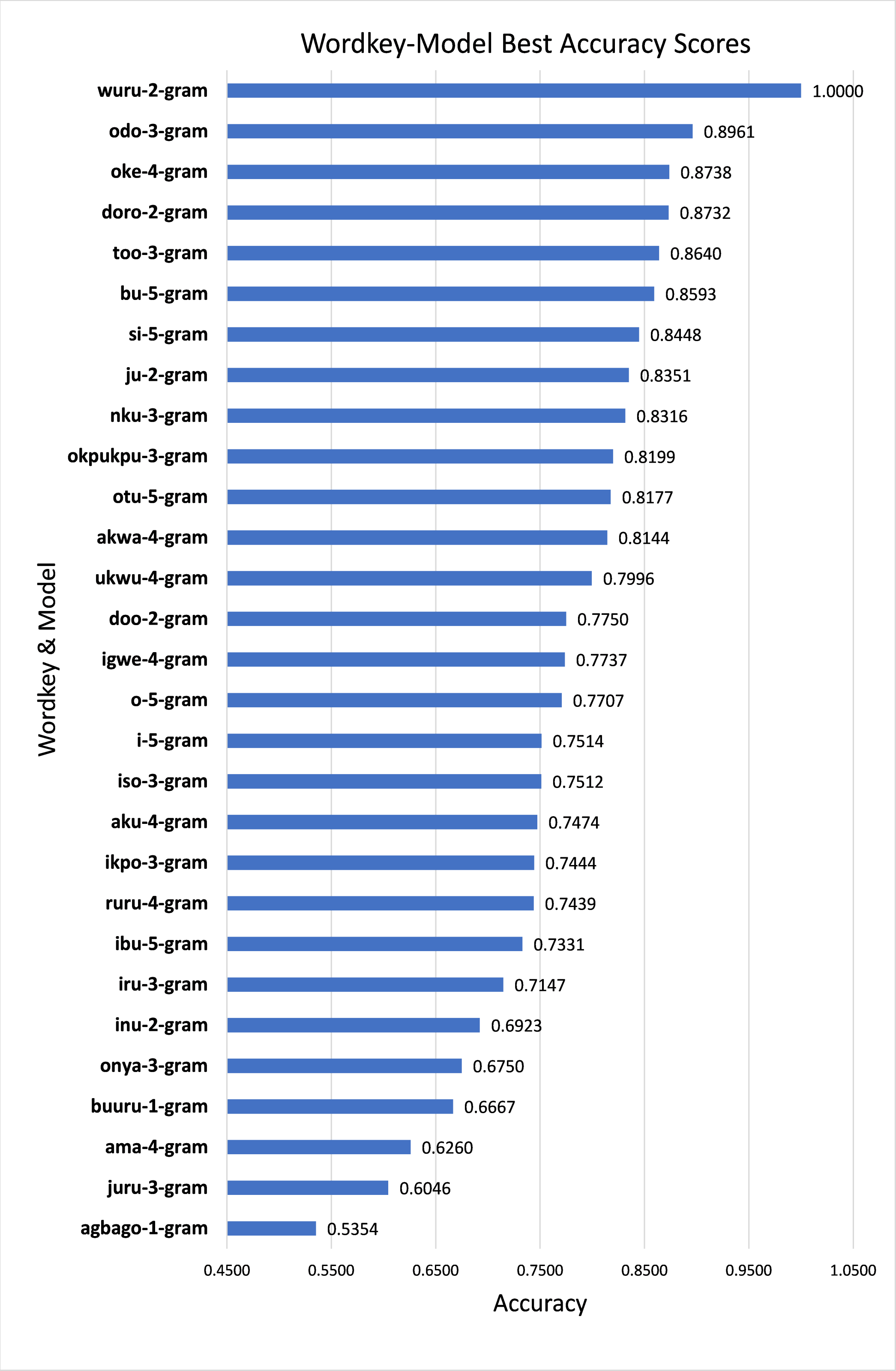

Fig. 5.3 **N-Gram Accuracy:** Graph showing the best accuracy score for each wordkey and the n-gram models that produced it in their descending order of performance.

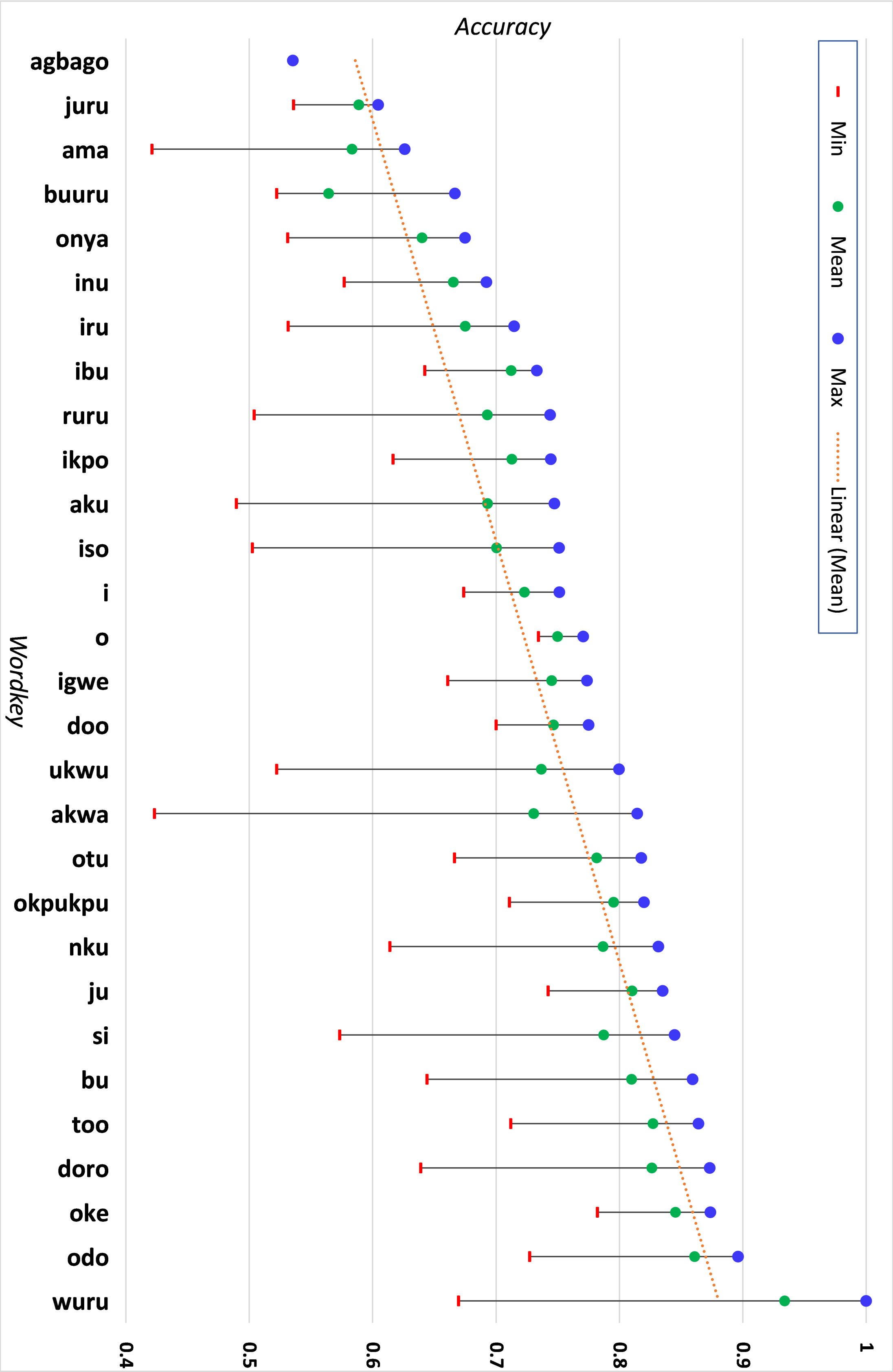

Fig. 5.4 **N-Gram Accuracy:** Graph showing the position of the best accuracy score of the n-gram models on the minimum-to-maximum bar for each wordkey

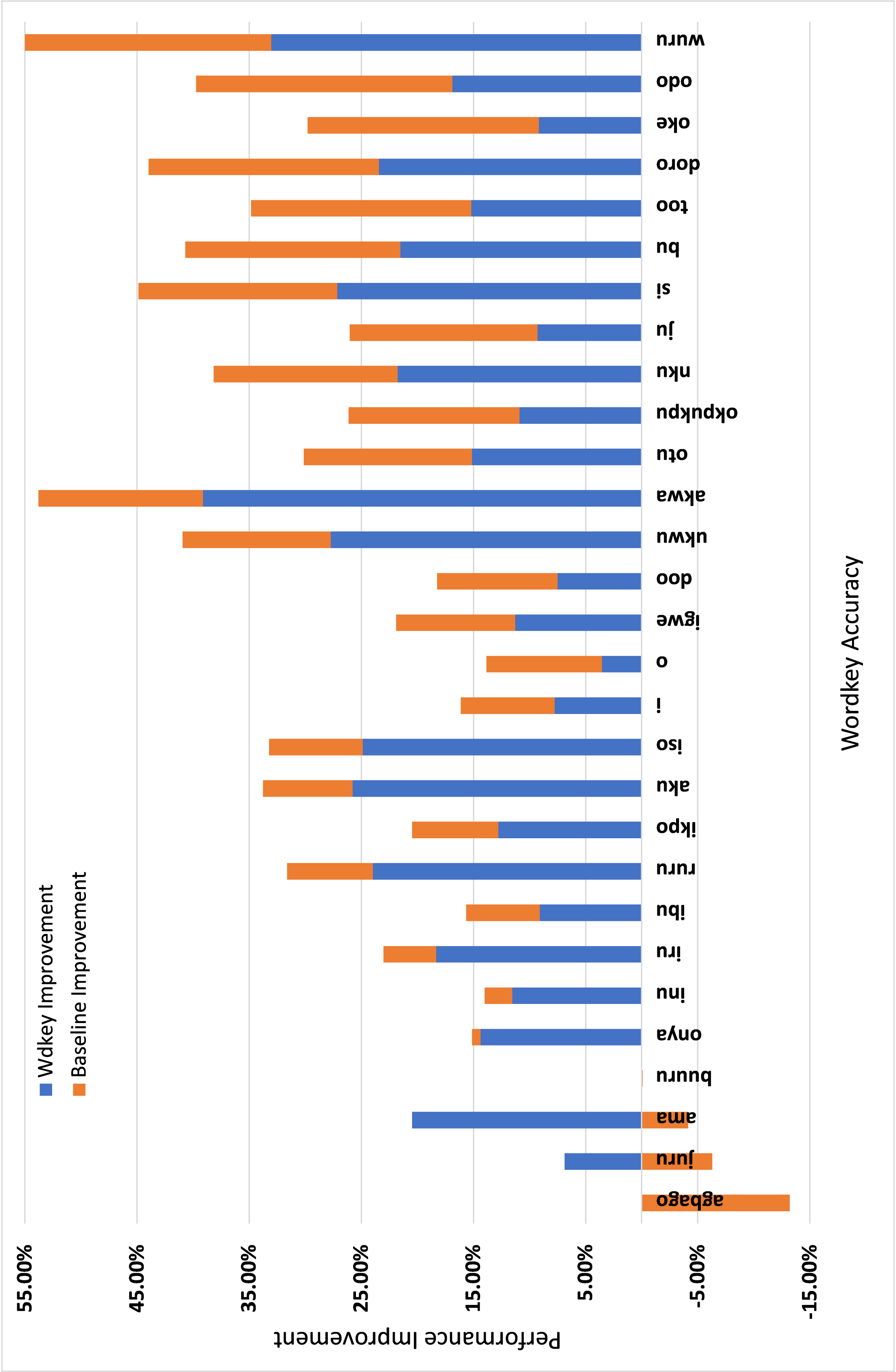

Fig. 5.5 **N-Gram Accuracy:** Stacked bar-chart showing the performance improvement scores achieved on both the wordkey and global baselines by the best performing model for each wordkey sorted from worst to best.

In a similar manner, we also present the percentage error reduction graph in Figure 5.6. This graph shows the extent of error reduction achieved by the best wordkey model with respect to the initial errors made with the wordkey unigram model. Also Figure 5.7 shows the model level error reduction on the global baseline.

Remember that our ngram restoration method backs-off to lower n-gram models during restoration and also restores the context of a wordkey before using it for the restoration of the wordkey. So given that the score obtained by a lower ngram may not improved by a higher one, we considered it necessary to keep counts of best performance of each of the n-gram models. Figures 5.8 and 5.9 indicate that beyond accuracy, the 3-gram and 4-gram models contested strongly for the best performing models with individual performances of 31% and 24% as well as the cumulative scores of 55.17% (16/29) and 79.31% (23/29) respectively.

With the accuracy scores presented in Table 5.4 and the different details extracted from the raw data which are represented in Figures 5.3, 5.4, 5.5, 5.6, 5.7, 5.8 and 5.9, we present the following observations which will serve as a basis for reading and interpreting subsequent results on other metrics and from other classes of restoration models discussed in this work:

- compared to the global baseline, the models struggled most with the following wordkeys starting with the worst: *agbago*, *buuru*, *juru*, *ama* and *onya*, *inu*, *iru*, *ibu*, *i*, *ikpo*, *ruru*, *aku*, *iso*, *o*, *doo*, *igwe*, *ukwu* and *akwa*, achieving less than 10% baseline improvement on each of them.

  These wordkeys will be observed more closely as we consider our result data on other metrics and also when we compare the n-grams models with other restoration models.

- in some of the wordkeys,(*agbago*, *buuru*, *juru*, *ama*, *onya* and *inu*) the models actually achieved below baseline performance in spite of improving their wordkey baseline performance significantly in some cases.

- with higher n-grams, *agbago* did not improve at all (0%) thereby becoming 13.22% less than the global baseline. *buuru*, on the other hand actually got worse with higher n-grams thereby getting an average accuracy score that is 10.22% less than both its baseline and 10.31% the global baseline.

- *ama* and *akwa* started out with the worst wordkey baseline scores of 42.13%. However, while *akwa* made a huge improvement (30.73%) on its baseline achieving an overall improvement of 6.29%, *ama* made only a modest 16.22% increase on its baseline thereby scoring below the global baseline.

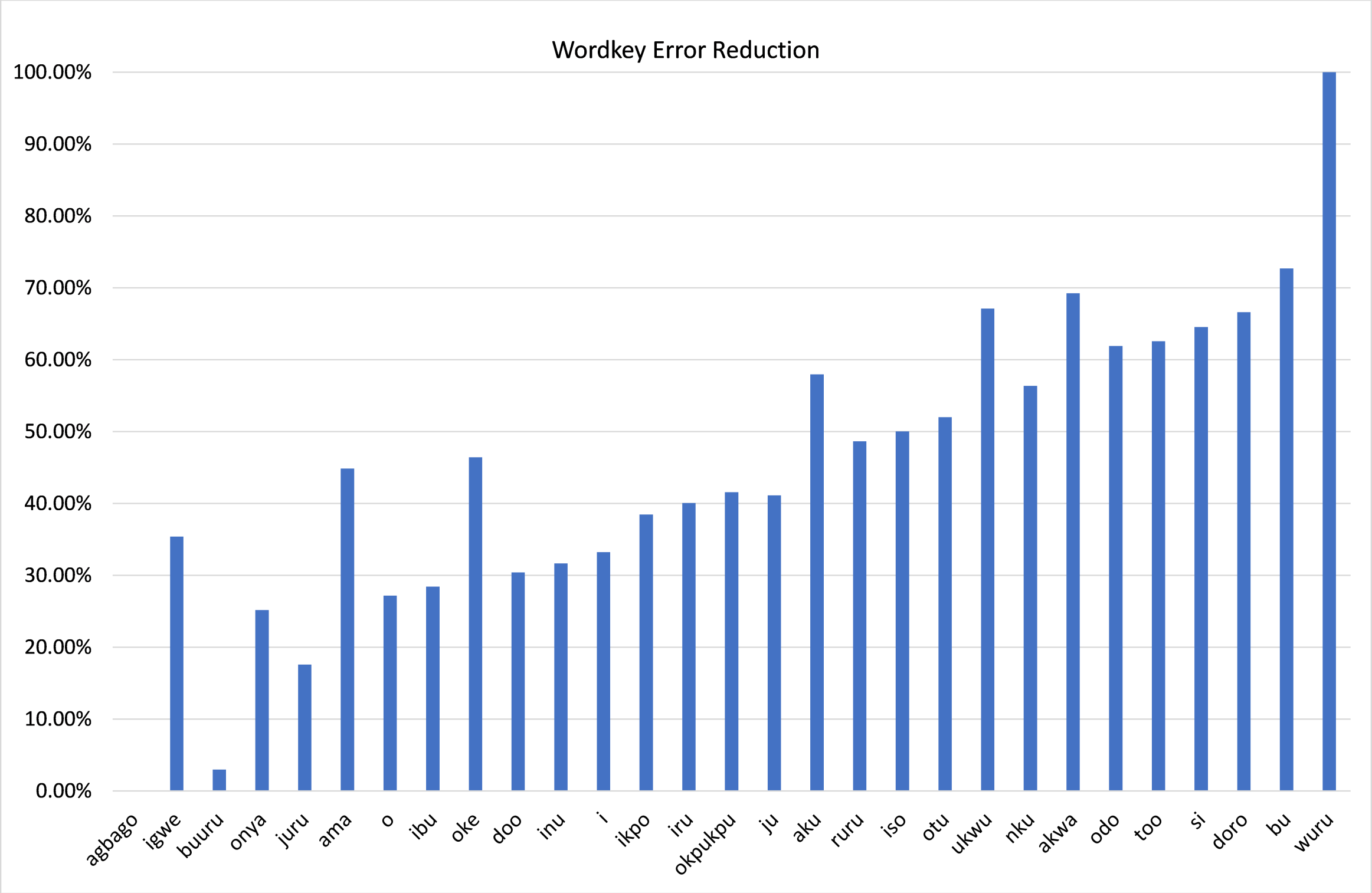

Fig. 5.6 **N-Gram Accuracy:** Column-chart showing the percentage error reduction achieved on each wordkey unigram model.

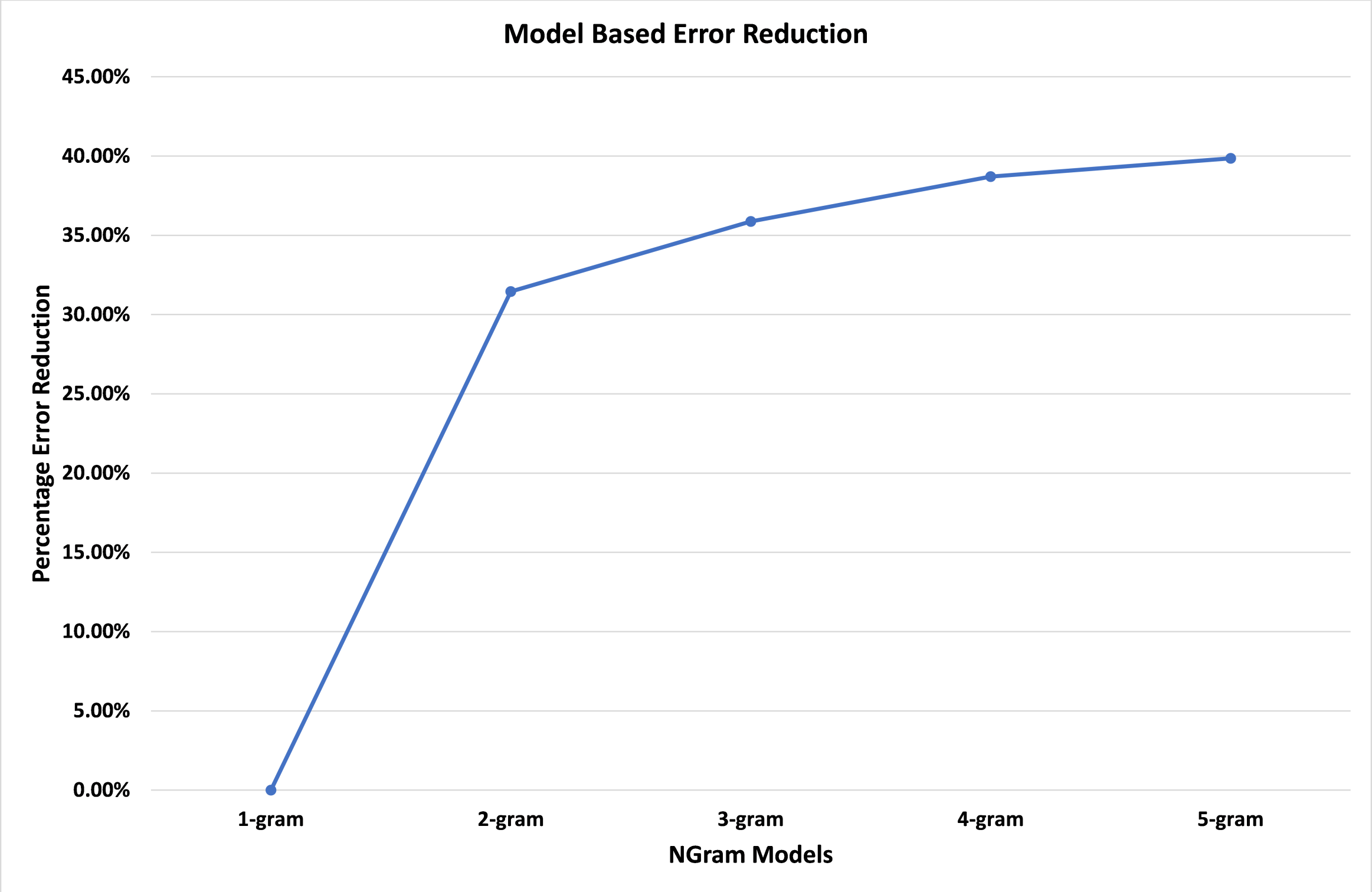

Fig. 5.7 **N-Gram Accuracy:** Plot showing the percentage model level error reduction achieved on the global baseline (i.e. **66.75%**) by each of the modal global score.

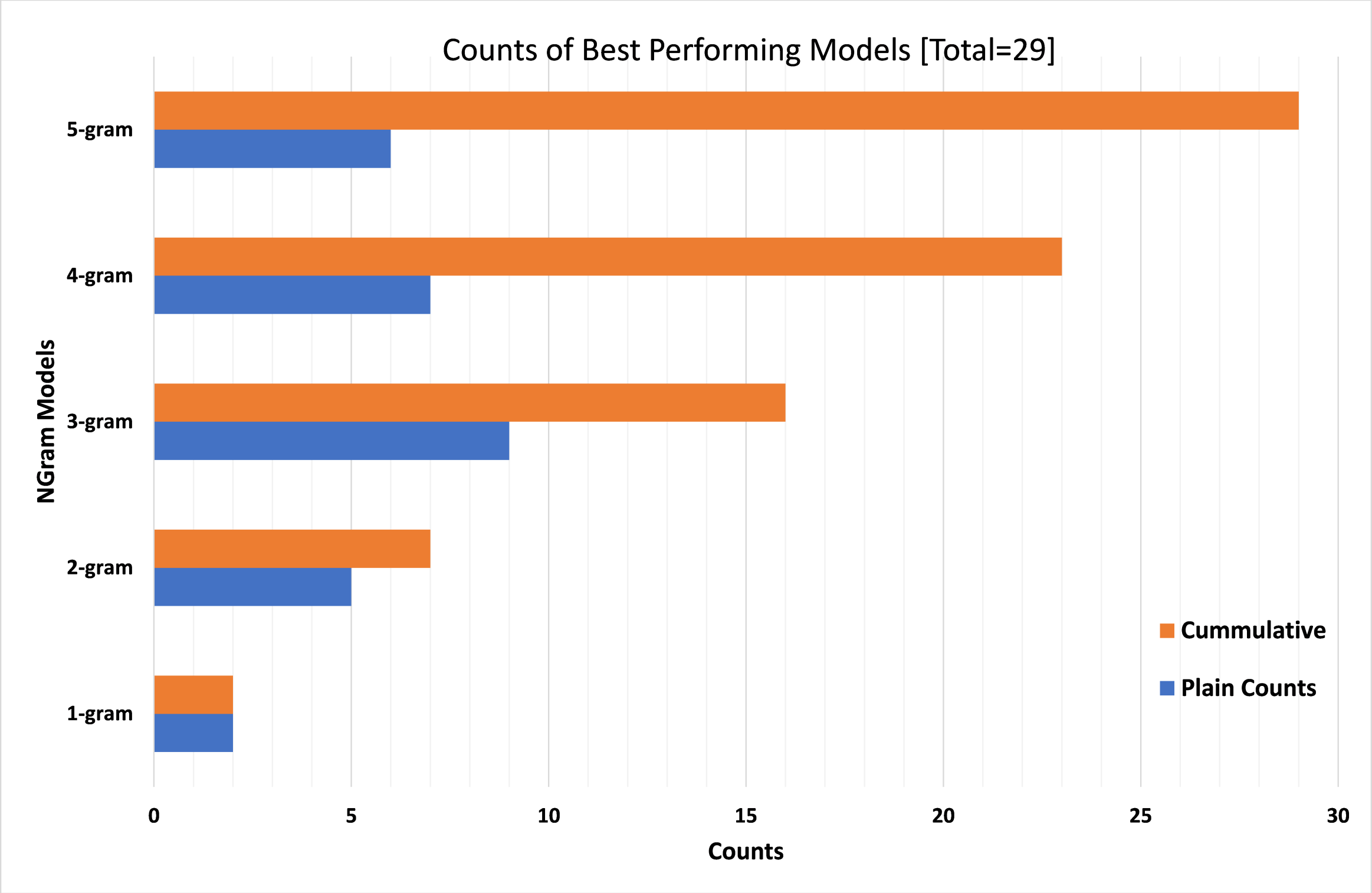

Fig. 5.8 **N-Gram Accuracy:** Bar-chart showing the normal and cumulative distributions of best performance by each of the n-gram models on all wordkeys.

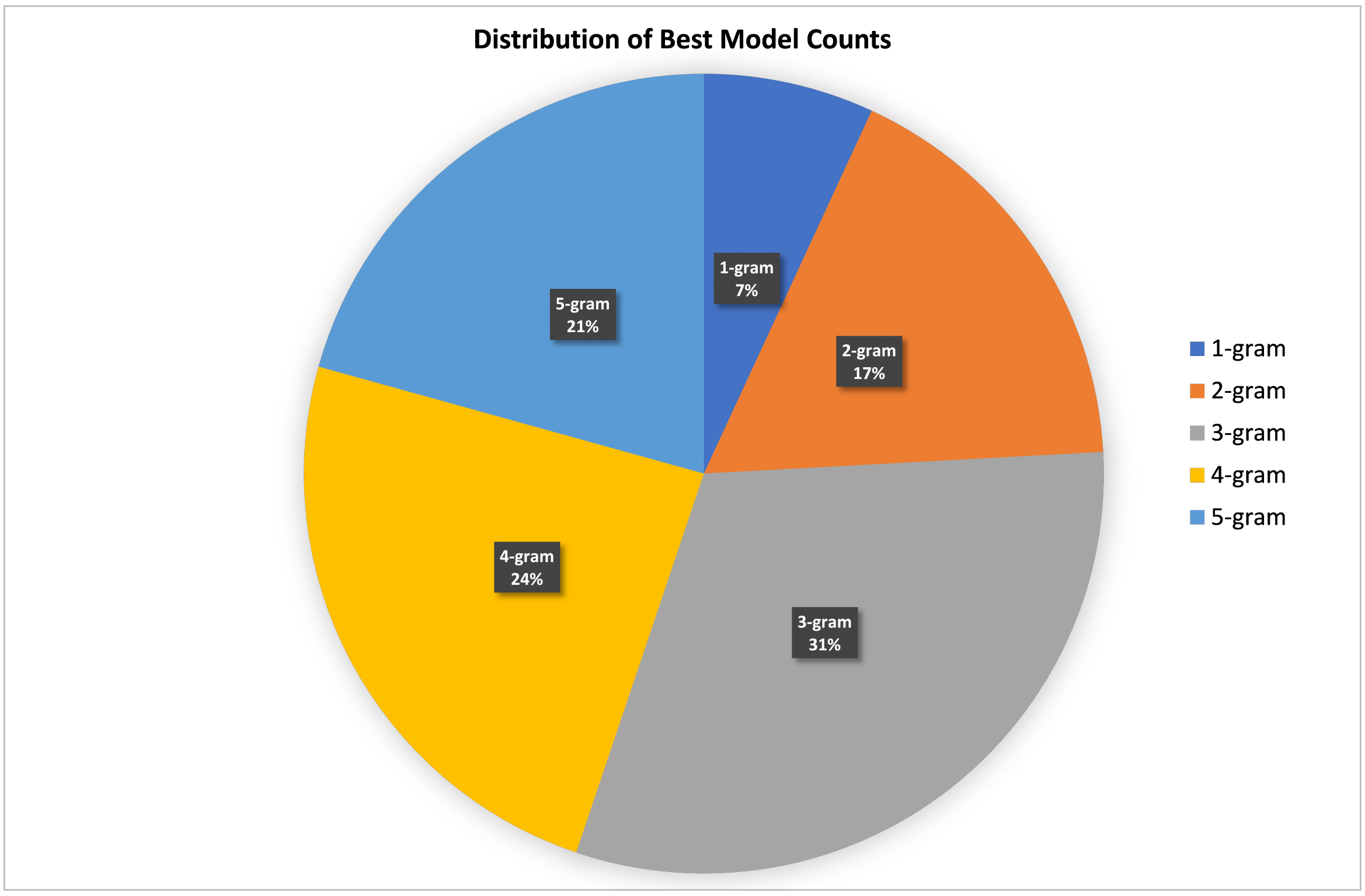

Fig. 5.9 **N-Gram Accuracy:** Pie-chart showing the percentage distribution of best performances achieved by all models across all wordkeys.

- *oke* got the best wordkey baseline (78.21%) which dwarfed its wordkey improvement score in spite of good improvement score (17.79%) on the global baseline.
- the models achieved the highest averaged performance on *wuru* (93.39%) thereby getting an improvement of 26.64% on the global baseline.
- besides *wuru*, the other high accuracy wordkeys that achieved above 15% improvement on the global baseline include: *doro* (15.88%), *too* (15.97%), *oke* (17.79%), *odo* (19.35%)
- the overall weighted average accuracy of the *Best* model, which uses the best model for each wordkey, is 80.05% which is an improvement 13.25% on the global baseline performance and thereby achieving a percentage error reduction of approximately 40%.

Figure 5.4 shows that wordkeys like *akwa*, *ukwu* and *wuru* made substantial improvements on their accuracy from their baseline unigram accuracies. However, some wordkeys like *juru*, *o* and *agbago* made comparatively slight improvements.

### 5.3.3 Analysis: Precision, Recall and F1 Scores

Having presented the raw scores and some analysis graphs on the accuracy of our n-gram models, this section presents a view of the model performances on the other standard metrics used for our evaluation i.e. precision, recall and F1 measures. Along with the n-gram model scores on these metrics, we shall also present the score from the combination of the best models for the wordkeys i.e. the combination of the scores on the *Best Score (BS)* column of the corresponding tables.

We earlier presented the basic formula for computing each of the metrics in §4.3.2 and also indicated that the *macro-averaging* of all individual variant scores reported for each wordkey. The full set of analysis graphs for each metric is presented in Appendix A.

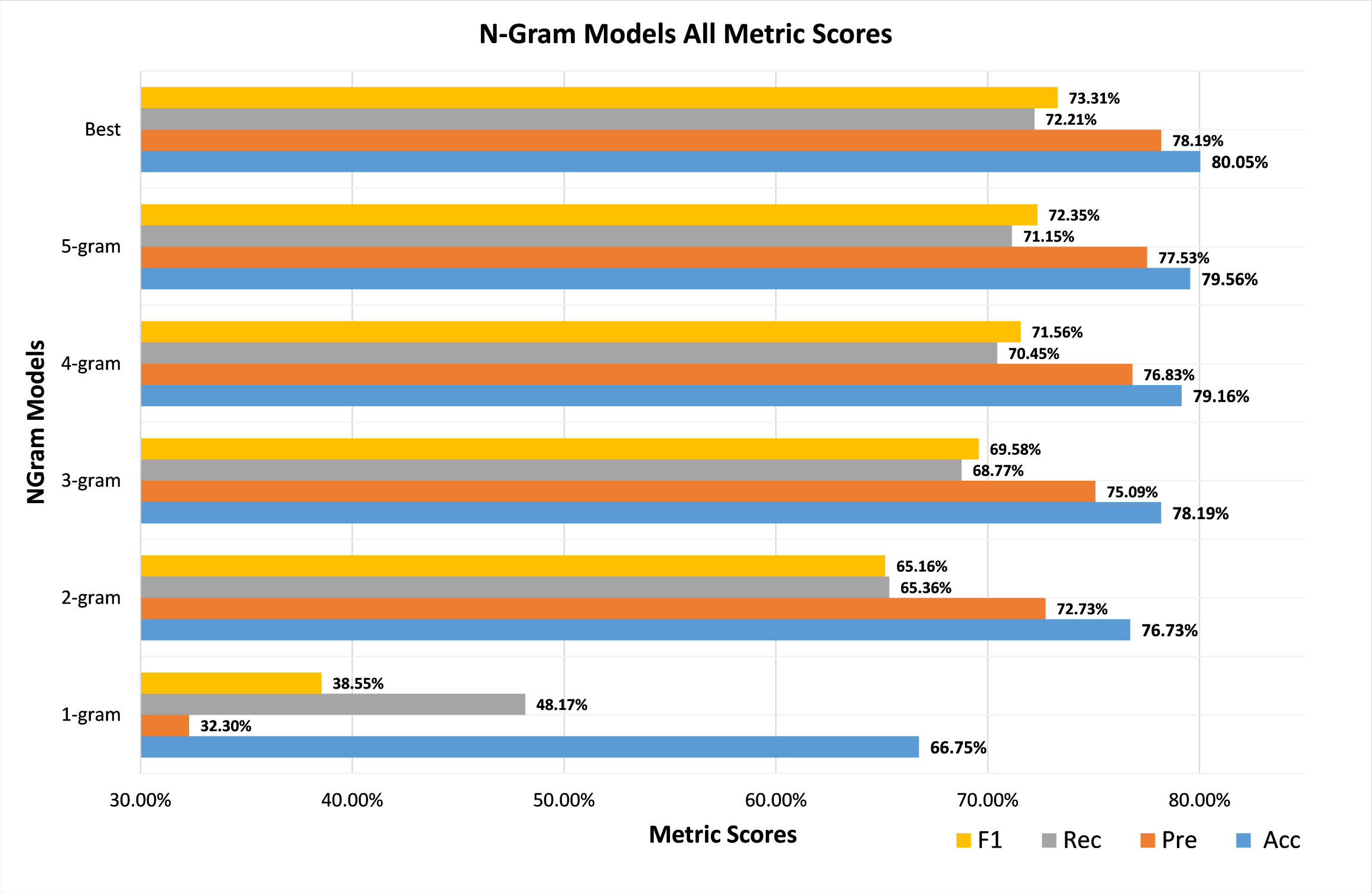

Fig. 5.10 **Summary:** Bar-chart showing the overall performance of all ngram models, as well as the combination of all best models, on all metrics.

### Why *Best* is better than 5-gram

Figure 5.10 indicates that there is a progressive improvement in the metric scores from the unigram to higher ngram models. Contrary to what may be expected, the *Best* model, whose score is given by the combination of the scores of the best model[3] for each wordkey, actually performed better than 5-gram. This is because, not only did some of the lower ngrams get the best score, the 5-gram model actually performed worse than the lower n-gram models on some of the wordkeys across the metrics. Table 5.5 shows a list of wordkeys, for each metric, in which the 5-gram model actually performed worse than the lower ngram model that got the best score.

[3]Here, the best model is defined at the lowest ngram score to achieve the best score.

| **Accuracy [6]** | | | |
|---|---|---|---|
| **wordkey** | **5grm Score** | **Best Score** | **Best Model** |
| *buuru* | 0.5444 | 0.6667 | 1-gram |
| *inu* | 0.6859 | 0.6923 | 2-gram |
| *igwe* | 0.7723 | 0.7737 | 4-gram |
| *doo* | 0.7500 | 0.7750 | 2-gram |
| *ju* | 0.8247 | 0.8351 | 2-gram |
| *oke* | 0.8729 | 0.8738 | 2-gram |
| **Precision [6]** | | | |
| *inu* | 0.6895 | 0.7030 | 2-gram |
| *doo* | 0.7059 | 0.7857 | 2-gram |
| *oke* | 0.8076 | 0.8114 | 4-gram |
| *igwe* | 0.8220 | 0.8252 | 4-gram |
| *ju* | 0.8284 | 0.8397 | 2-gram |
| *si* | 0.8745 | 0.8753 | 3-gram |
| **Recall [4]** | | | |
| *ibu* | 0.6416 | 0.6422 | 3-gram |
| *doo* | 0.6468 | 0.6528 | 3-gram |
| *inu* | 0.6551 | 0.6586 | 2-gram |
| *ju* | 0.6861 | 0.7061 | 2-gram |
| **F1 [5]** | | | |
| *igwe* | 0.5904 | 0.5905 | 4-gram |
| *ibu* | 0.6421 | 0.6436 | 3-gram |
| *inu* | 0.6545 | 0.6571 | 2-gram |
| *doo* | 0.6590 | 0.6664 | 3-gram |
| *ju* | 0.7161 | 0.7382 | 2-gram |

Table 5.5 **Summary:** Table showing the 5-Gram Model scores below the highest score on some wordkeys across all metrics.

### Distribution of *Best Scores* across models

Again, as shown in Figure 5.12 the counts of the best metric performance for each wordkey across the models could have followed a normal distribution curve if extended beyond the 5-gram. For example, the 3-gram model gets most of best score counts by itself across the metrics. That means that it gets the best recall and F1 scores for the highest number of wordkeys (12 out of 29) which keeps a cumulative best count with other lower ngrams that accounts for the best scores on 17 out of 29 wordkeys i.e. 58.62%. For the 4-grams, although the individual counts dropped, its cumulative count of best performance on, say F1, is 23 out of 29, i.e. 79.31%.

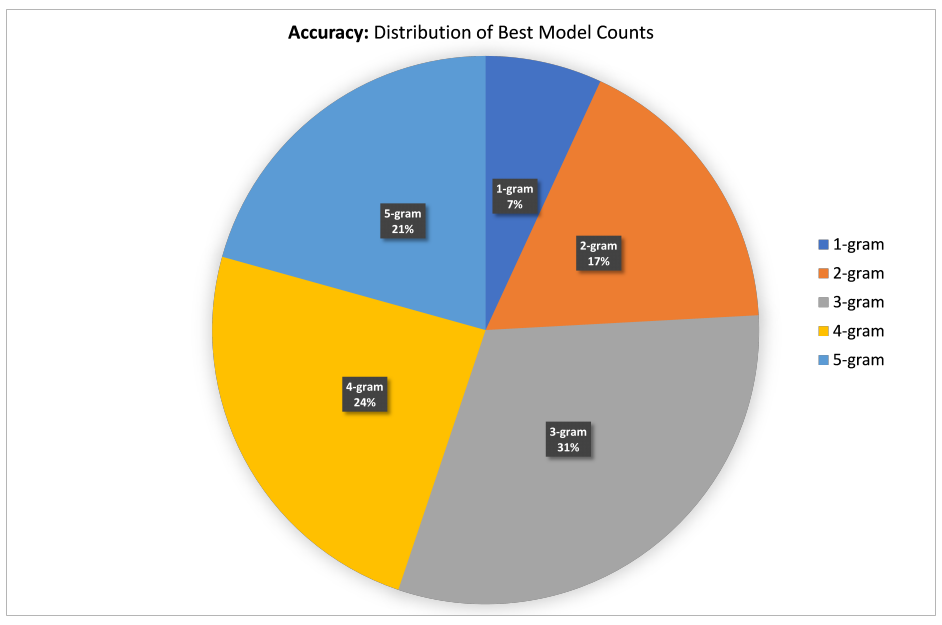

(a) Accuracy

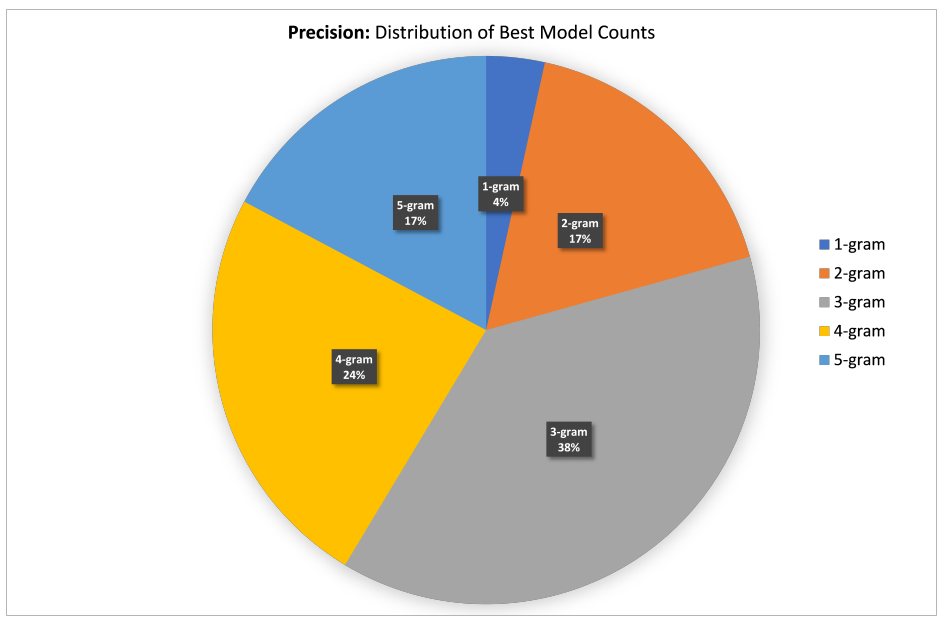

(b) Precision

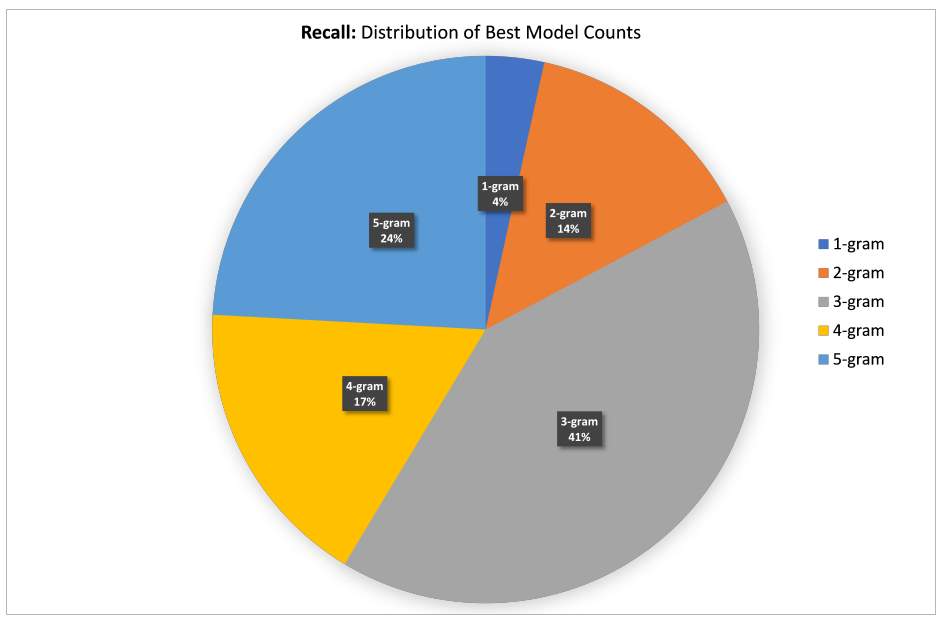

(c) Recall

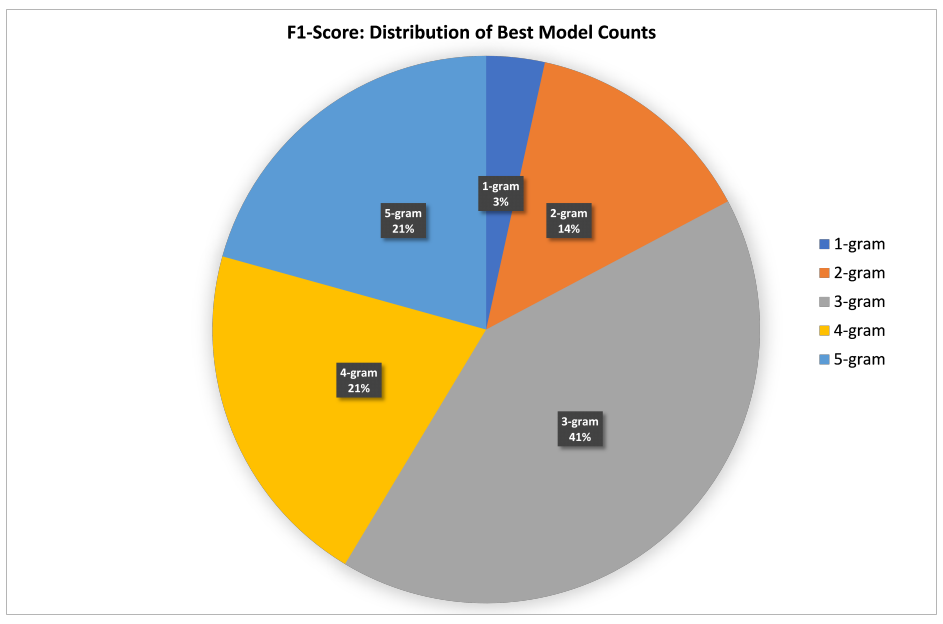

(d) F1-Score

Fig. 5.11 **Summary:** Pie-charts showing the distribution of best performance counts for each of the metrics across all models.

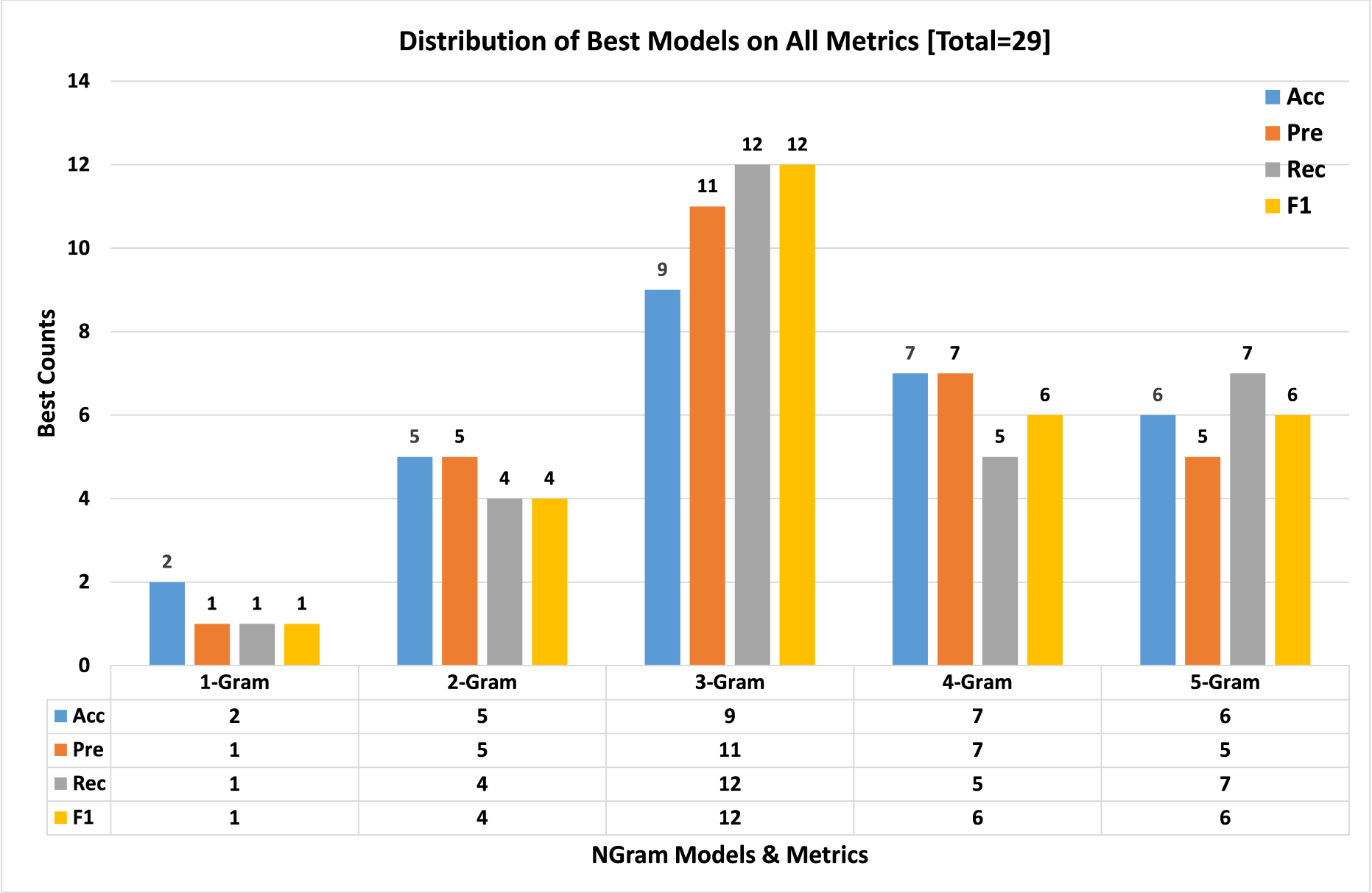

Fig. 5.12 **Summary:** Column-chart combining the pie-charts in Figure 5.11, shows the distribution of best performing counts on wordkeys across all models for each metric.

Figure 5.13 shows the graph of the percentage error reduction by the models starting from the assumed initial 100% error on all metrics. As can be seen from the graph, while each model contributed to reducing the remaining error from the previous model, the 3-gram achieved a much higher reduction in the previous model's error. The amount of error reduction from previous model tapers as higher n-grams are approached indicating the inability of the higher n-gram models to improve the results.

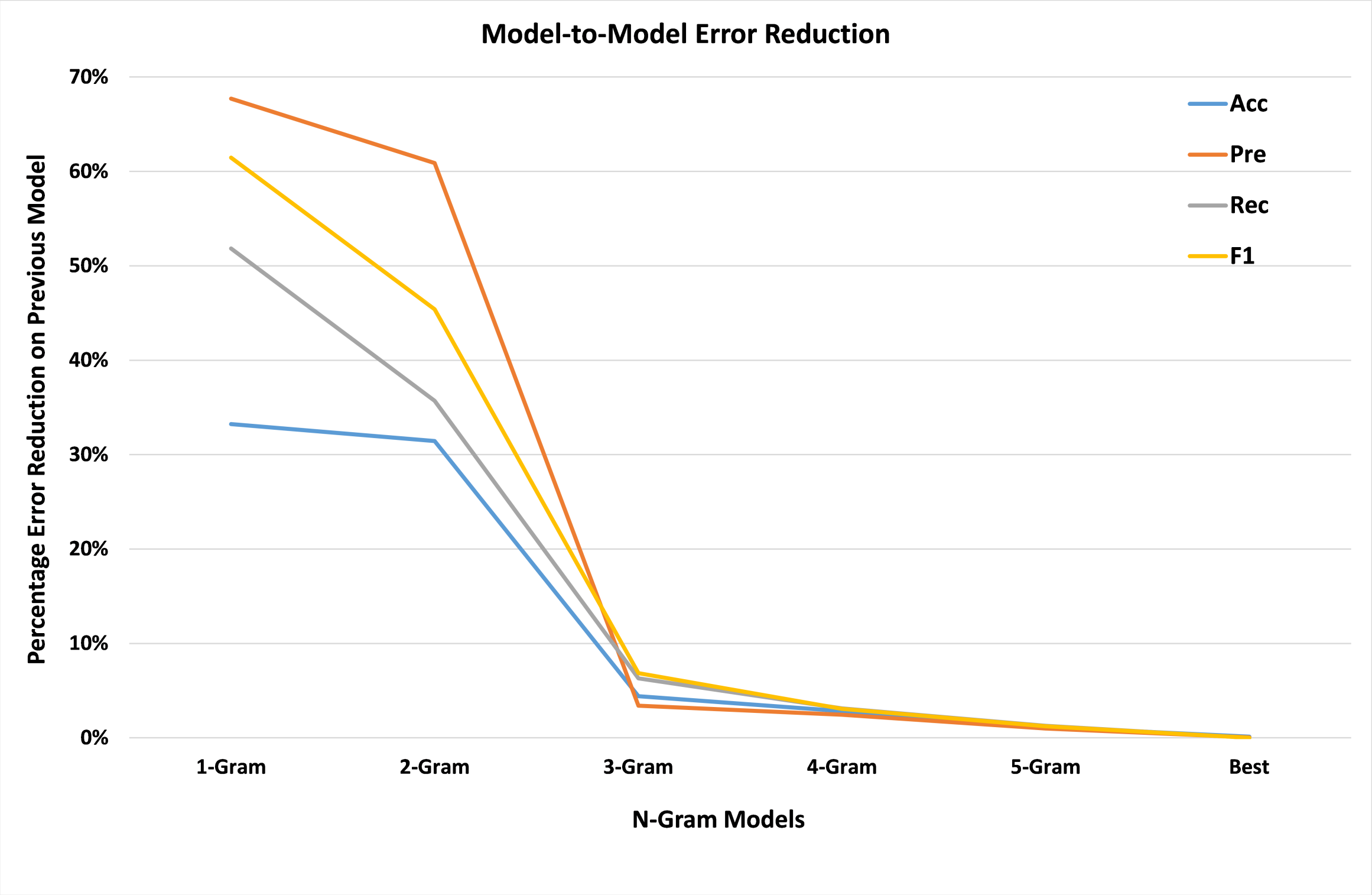

Fig. 5.13 **Summary:** Graph showing the percentage model-to-model reduction of performance errors on all metrics with the initial error as the unigram model error for each metric.

### 5.3.4 Wordkeys with special characteristics

We also felt the need to keep track of the behaviours of the models with respect to 7 wordkeys with specific characteristics and see if we could identify any pattern with regards to the nature of their instances in the dataset or the performance of the models on them. The wordkeys are listed and described below:

**agbago** showed an interesting pattern with its results because none of the models could perform better than the unigram model and so there was absolutely no performance improvement on this wordkey.

| **wordkey** | **variants and counts** |
|---|---|
| *agbago* | agbagọ=46; agbago=53 |
| *akwa* | ákwá=363; ákwà=504; akwa=324 |
| *buuru* | buuru=60; bụụrụ=120 |
| *igwe* | igwe=136; ígwé=207; ìgwè=920; ígwè=129 |
| *ju* | jụ=25; ju=72 |
| *o* | o=8323; ọ=23123 |
| *wuru* | wuru=75; wụrụ=37 |

Table 5.6 **Summary:** Distribution of variants of the *special-7* wordkeys in the dataset.

**akwa** is a wordkey with many commonly confused variants. Although our corpus does not contain all its variants in substantial proportions to be included in the dataset, it has 3 fairly distributed variants in out dataset.

**buuru** showed an even more interesting behaviour which sees the higher n-grams actually producing *worse* results than the unigram model.

**igwe** has the highest number of variants (i.e. 4) and whether that has any specific impact on the performance of the models as we progress.

**ju** has the least number of instances (only 97) and it will be interesting to see the effect of its tiny representation on the model performances

**o** is a single-letter word with variants that are very commonly used. It has the highest number of instances in our dataset accounting for almost 39% of the entire dataset.

**wuru** is on the flip-side of *agbago* and *buuru* above. Apart from the unigram model, all others were able to cleanly separate the variants with 100% accuracy.

In addition to the raw results presented in Table 5.4, Figures 5.14 and 5.15 show the confusion matrices of the gold and predicted variants for all the *special-7* wordkeys by the 1-gram and the 5-gram models. We shall compare the scores on these wordkeys with those of the models in the machine learning experiments presented in chapter 6. Table 5.6 shows the original distribution of the variants in the test data.

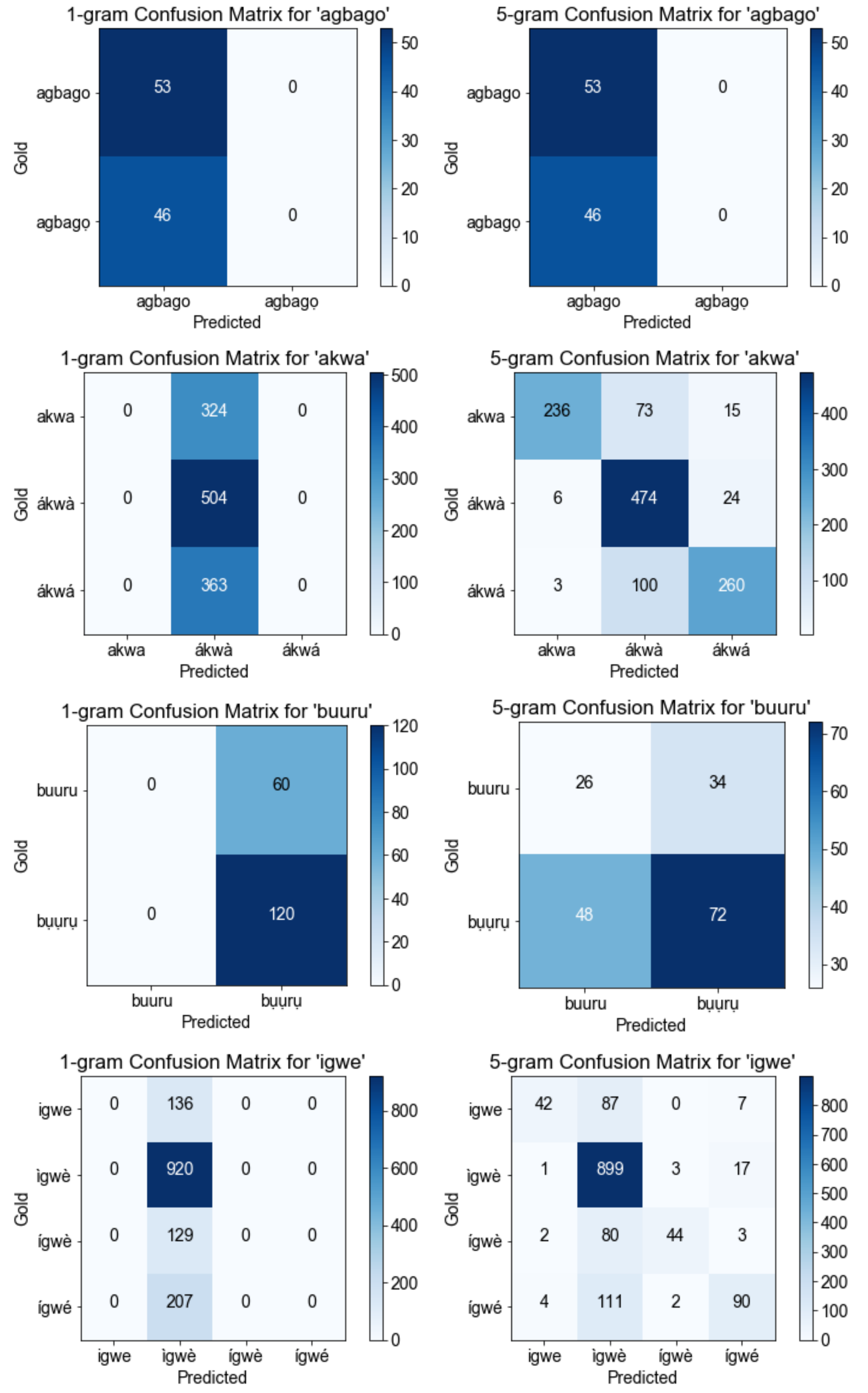

Fig. 5.14 (a) **Summary:** Confusion matrix of the results from all the n-grams on the *special-7* wordkeys.

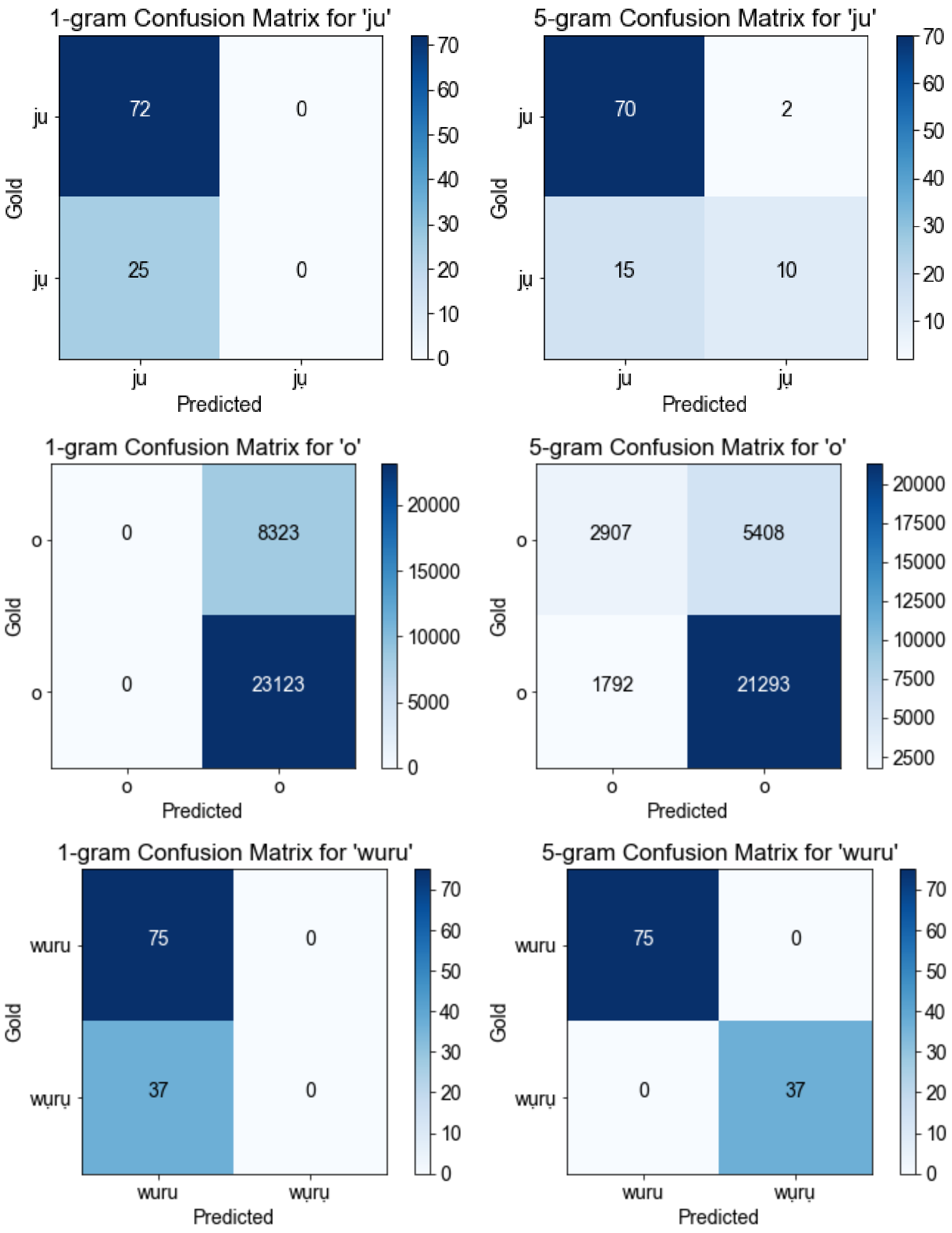

Fig. 5.15 (b) **Summary:** Confusion matrix of the results from all the n-grams on the *special-7* wordkeys.

### Observations from the Confusion Matrices

The confusion matrices compare the performance of the baseline unigram model and the 5-gram models on the special wordkeys. On the left column are the unigram confusion matrices which show all instances being classified as belonging to the variant with the most instances, hence column appearance, while those of the 5-gram models are on the right column.

It is obvious that for most of the ambiguous sets, there is a good attempt by the 5-gram model to classify the instances into their variants hence the diagonal appearance. However, there is a varying degree of the improvement on the baseline which ranges from no improvement at all, to achieving a clean split i.e. 100% accuracy. For instance, as noted in §5.3.4, the wordkey *wuru* was clearly separated into its two distinct variants while for the wordkey *agbago*, there was no improvement at all as it still returned the most common variant for all instances.

The other wordkeys have varying levels of improvement. For example, only a little improvement in the overall counts of correctly classified instances (920 vs 975 out of 1389) of the wordkey *igwe* was achieved by the 5-gram. This is because the majority of the instances of all of its variants were still being classified as the most common variant.

## 5.4 Chapter Summary

Having introduced the Igbo diacritic restoration problem in chapter 4, this chapter presented the design and implementation of an enhanced version of the $n$-gram based solution reported in [32]. For this experiment, the restoration task is performed on the dataset created in §4.2.2 which contains only the instances of 29 wordkeys extracted from the corpus. The restoration process basically involves the extraction of candidate variants for each wordkey and selection of the most likely variant using maximum likelihood estimation, MLE.

The MLE approach and its application to diacritic restoration have been discussed in §5.1.3 and §5.1.4 respectively. In this experiment, a back-off method that adopts the immediate lower n-gram model for the restoration is applied where necessary. Only the left context of wordkey being restored is used in this experiment and the context words, which are non-diacritic themselves are restored prior to the restoration of the main wordkey. This thesis contains only the results from the unigram (1-gram) to the 5-gram because higher n-gram models did not achieve any significant improvement in the results. In addition, results from the combination of scores from the models that achieved the best results for specific wordkeys were also presented in this work.

Overall the models achieved good results on the restoration task as indicated from the summary of results presented in §5.3.1. The baseline scores were those of the unigram model which is the most naïve of all the models. There is a progressive improvement on all performance metrics by the n-gram models as $n$ increases from 1 to 5. Although the 5-gram performs best on it own as a model, it's generally lower (though only very slightly) than the combination of the best models on the workeys. We also identified in §5.3.4, some of the wordkeys that produced interesting outcomes which we will be observing in subsequent experiments.

So in general, at the end of the experiments on using the n-gram for diacritic restoration, the 5-gram model gave the best individual results: $accuracy = 80.01$, $precision = 78.15$, $recall = 72.20$, $F1 = 73.30$ while the combination of models gave a slightly higher set of results for i.e. 80.05, 78.19, 72.21, 73.31 respectively. Both sets of results indicate that both the 5-gram and the *Best* models improved the global baseline model by 13.25% thereby reducing the error by almost 40%.

In the next chapter, we shall present our next set of experiments in which we will apply machine learning classification algorithms that will use the immediate context words of the wordkey as features.

sorte

# Chapter 6

# IDR with Classification Models

In chapter 5, we presented the report of the experiments with applying n-gram models to diacritic restoration. The best results on the 4 key metrics were reported as follows: $accuracy = 80.01$, $precision = 78.15$, $recall = 72.20$, $F1 = 73.30$; while the combined effect of the best models on wordkeys gave 80.05, 78.19, 72.21, 73.31 respectively. Although the $n$-gram models performed considerably well, they mostly plateau after the 5-gram model as shown in Figure 5.2 i.e. extending beyond the 5-gram does not improve the results any further despite being more computationally expensive.

We consider the inability of the n-gram models to make use of a wider context in its classification a handicap given that beyond the left 5-gram context, there may still be features that can suggest the proper diacritic. We think that there are patterns in the training data that can help us improve the performance of our system beyond the n-gram. So we consider using classification techniques that could automatically learn models to classify instances with extended context. In this chapter, we will apply classification functions derived with machine learning algorithms to solving the Igbo diacritic restoration task.

In this experiment, we will apply a list of 12 commonly used learning algorithms for training classification models. This is basically an exploratory experiment and so only the default parameters of the algorithms will be used in the model training and no assumptions will be made on the data. For a closer comparison with the n-gram models, we will combine the results from the top-3 best performing models using performance scores and efficiency.

## 6.1 Overview of Machine Learning

As a classification task, diacritic restoration is very well suited for and will be more generalisable with machine learning. Machine learning algorithms use computational methods to *"learn"* information directly from data and adaptively improve their performance with increased sample points without assuming a predetermined equation as a model. Rivest [85] noted that the goal of machine learning (ML) research is basically to "identify the largest possible class of concepts" that can be learned from examples.

Learning can be *supervised* or *unsupervised*. A supervised learning process is built based on a training data set with the correct class label for each input and is primarily applied to classification or regression problems. In classification, the aim is to learn from the training data how to predict the class of unlabelled data given a finite list of discrete categories. Regression learns how to predict output that are continuous (or non-discrete) in nature e.g temperature or height, by studying the training data.

In the general definition of a classification model, we aim to define $y = f(X)$ that will take $X$, which a set of values in the form $x_1, x_2, \ldots, x_n$, as input and predict an output $y \in Y$ where $Y$ is a set of all possible output values (also referred to as *classes* or *labels*). A learning algorithm takes the *training set* and applies an iterative process to identify the best coefficients (*weights*) for each of the values in $X$. The training set is a set of $n$ samples of $X$s along with their desired $y$s generally in the form $S_{train} = \{(X_1, y_1), (X_2, y_2), \ldots, (X_n, y_n)\}$.

When the training is done and an optimal set of *weights* is identified, the model built is applied to the *test set* i.e. a held-out set of $X$ values, without including their actual $y$ values i.e. $S_{test} = \{X_1, X_2, \ldots, X_n)\}$. The *test set* was not seen by the learning algorithm during model training so that the models performance could be measured on how its predicted outputs compare with the actual $y$ values. There are different types of learning algorithms for classification models and they work in different ways with various degrees of similarities in their principles.

In unsupervised learning, the training data consists of a set of input vectors $X$ (where $X = x_1, x_2, \ldots, x_n$), without any corresponding target values. The aim in such problems may be to discover groups (or clusters) of similar examples within the data which is referred to as *clustering*, or to determine the density estimation i.e the distribution of data within the input space or to reduce the dimensionality of the data space for the purpose of visualization

Bird *et al.* [11] defined *classification* as the task of choosing the correct *class label* for a given input. Examples of classification tasks include: determining whether an

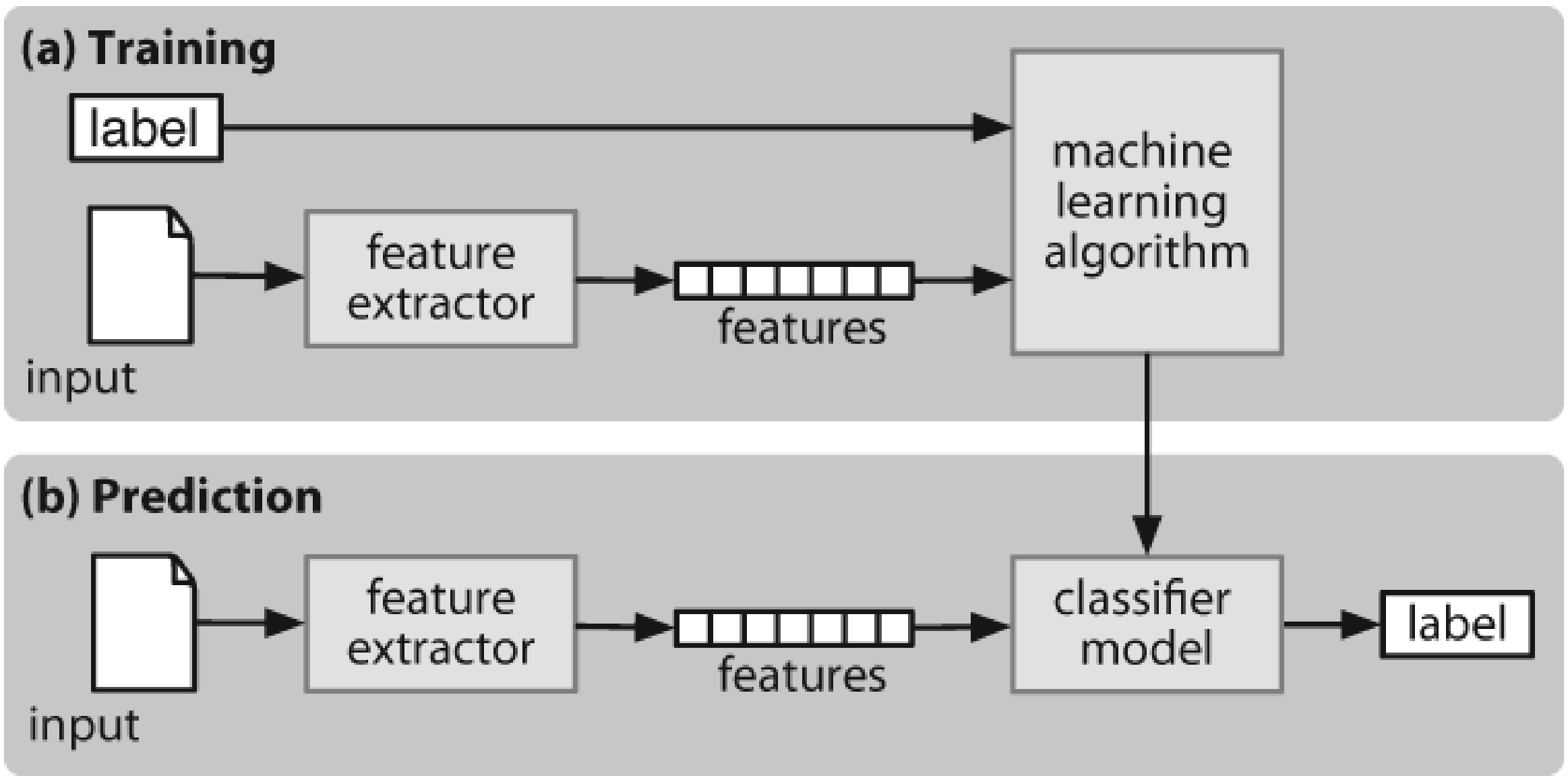

Fig. 6.1 Framework for Developing a Supervised Classifier. (*Source*: Bird *et al.* [11])

email is spam or not, assigning a topic area (e.g. "sports," "technology," or "politics.") to a news article and deciding the sense of a word, say *bank*, as used in a context (e.g. river bank, a financial institution, the act of tilting to the side, or making a bank desposit to a financial institution.

Figure 6.1 above describes the framework for building a supervised classification system. The training process, (a), uses a feature extractor to convert each input value to a feature set that captures the relevant classification information about the input. A model is then generated with the pairs of feature sets and labels which serves as input to a machine learning algorithm. In the prediction part, the feature extractor is again used to convert unseen inputs to feature sets which are passed to the model to generate predicted labels.

Naïve bayes and decision tree classifiers discussed in §6.2 and other techniques are often used in machine learning to build automatic classification models. Some other examples of the commonly used ones are: *maximum entropy classifiers*, *perceptrons* and *k-nearest neighbour*. Most of them can be used as black boxes to simply train models and use them without worrying about the internal structures. But a good understanding of how they work may help in the selection of appropriate features that can improve the classification accuracy.

## 6.2 Training Algorithms

As explained above, diacritic restoration is a classification task which assigns a variant i.e. the *label*, to a wordkey given its context which is basically a sequence of non-diacritic words. These words in the context of the wordkey are regarded as *features* that are used in predicting the right diacritic variant for the wordkey.

In this experiment, we used the same dataset we created in chapter 3, §4.2.2, which was also used for the n-gram models. As a summary, it has 29 ambiguous wordkeys as shown in 4.10. The wordkeys generated a total of 80,844 instances for the training and testing of our classification models using 12 of the most common machine learning classification algorithms implemented in the Python *Scikit-Learn* libraries.

Some of the most commonly applied ones in NLP research have been used in building the models for our diacritic restoration task and their basic definitions are presented below. No parameter optimization was implemented at this stage for any of the models i.e in nearly all cases, the default parameters as implemented in *Scikit-Learn* were used. In this section, an attempt will be made to present brief descriptions of the training algorithms used in this experiment starting with a deeper insight into the basic training process using a simple perceptron.

### 6.2.1 PCPT: Perceptron

The perceptron is a well known algorithm for learning classification models which has become quite popular in downstream NLP tasks such as part-of-speech tagging and syntactic parsing [17]. It was originally invented by Frank Rosenblatt in 1957 [86]. It is basically a single-layer neural network that consists of four fundamental parts: the *inputs and bias*[1], the *weights*, the *weighted sum* and the *activation (or step) function* as shown in Figure 6.2.

In supervised training, the perceptron algorithm usually starts by initialising the weights to 0 or a small random value. For each $X$ and its actual $y$ in $S_{train}$, a predicted $y$ (often written as $\hat{y}$) is computed as follows:

$$\hat{y}(t) = \begin{cases} 1 & \text{if } \left\{ \sum_{i=1}^{n} x_i \cdot w_i(t) \right\} + b > 0 \\ 0 & \text{Otherwise} \end{cases} \tag{6.1}$$

$$\text{where } x_i \in X \text{ and } n = |X|$$

[1]The *bias* is basically a constant that is usually initialized to 1.

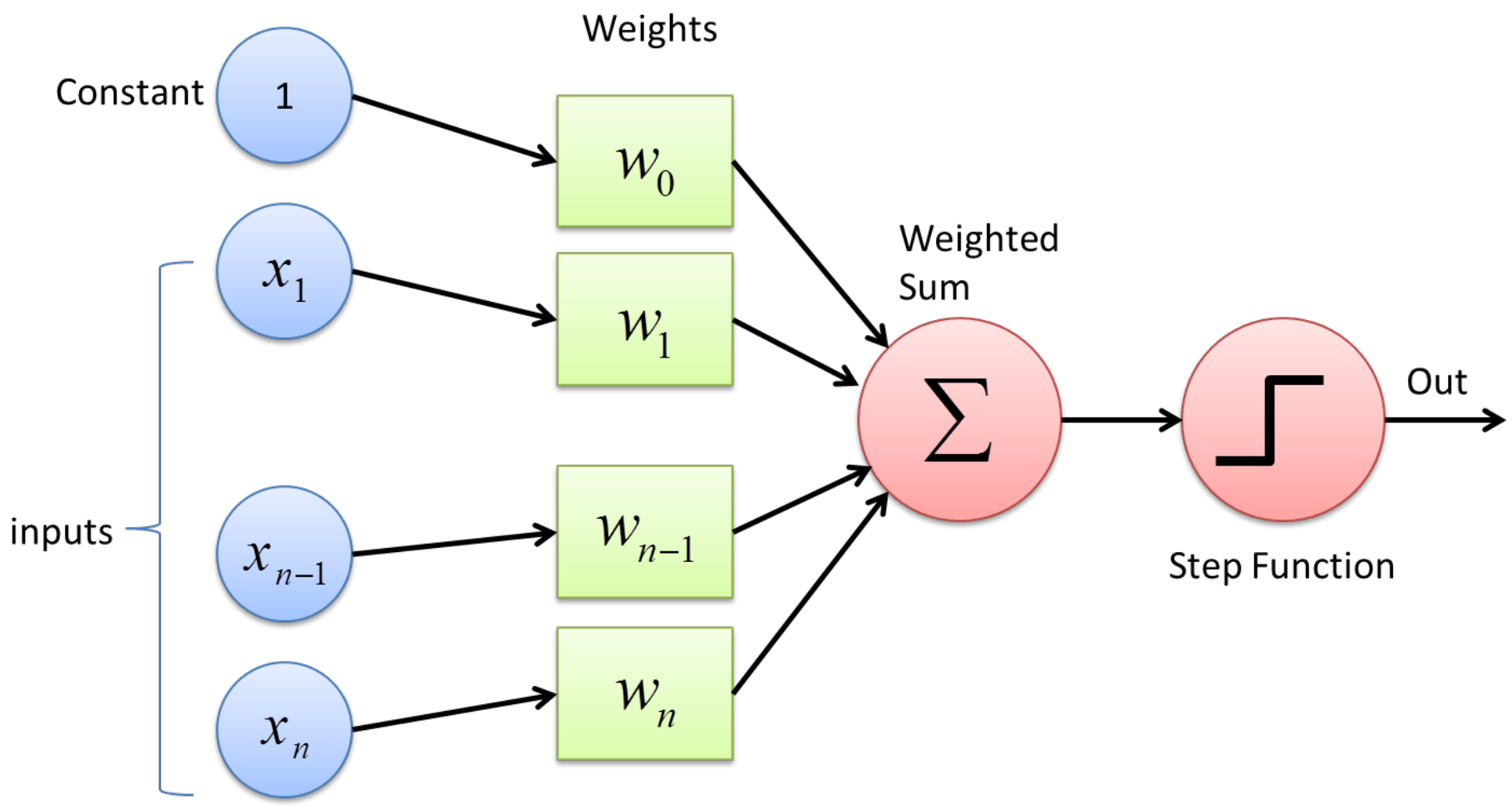

Fig. 6.2 An illustration of a Simple Perceptron

During training, the values of the *weights* are time-dependent and with $w_i(t)$, we refer to the $i$th weight at time $t$ (i.e. an iteration or epoch), which represents a complete pass through all the samples in $S_{train}$. The values of $y \in Y$ and their corresponding predicted values, $\hat{y}$, are then compared and where they are not the same, a classification error is deemed to have occurred. The goal of the training is to minimize the total percentage classification error, $\varepsilon_{train}$, over the training set given by:

$$\begin{aligned} \varepsilon_{train} =& \frac{1}{n}\sum_{j=1}^{n}|\hat{y}_j(t) - y_j| \\ &\text{where } y_j \in Y \text{ and } n = |Y| \end{aligned} \tag{6.2}$$

To achieve this, the weights are adjusted to ensure a reduction in the training error, $\varepsilon_{train}$, with a learning rate, **r**, that enforces only a small change in the weights. The new weights, $w_i(t+1)$; $0 \leq i \leq n$, are then applied in the next iteration.

$$\begin{aligned} w_i(t+1) =& w_i(t) + r \cdot (y_j - y_j(t)) \cdot x_{j,i} \\ &\text{for } 0 \leq i \leq n \text{ and} \\ &r \text{ is the learning rate.} \end{aligned} \tag{6.3}$$

This process is repeated until the training error, $\varepsilon_{train}$, is below a specified error threshold or a pre-defined number of iterations have been reached. At that point, the trained model is ready to the applied to the test set, $S_{test}$. Generally, the final evaluation error or $\varepsilon_{test}$ is expected to be higher than the $\varepsilon_{train}$, but the lower the $\varepsilon_{test}$ the better the model.

### 6.2.2 LRCV: Logistic Regression

Another popular classification model is the logistic regression which was developed in 1958 by David Cox [18]. Despite its name, logistic regression is actually a linear classification model. It is also generally referred to as *log-linear classifier*, *maximum entropy* (max-ent) or *logit-regression.* It uses a logistic function to model the probabilities of the possible outcomes of an event.

In its basic binary form, there will be a dependent variable, $y$, with two possible values (e.g. 0 and 1). As with the perceptron, if we have a set of predictors (or input values), say $x_1, x_2, \dots, x_n$, each of which could take continuous or discrete values, the aim is to compute the coefficients $\beta_0, \beta_1, \beta_2, \dots, \beta_n$ such that:

$$\hat{y} = \begin{cases} 1 & \beta_0 + \beta_1 x_1 + \beta_2 x_2 + \cdots + \beta_n x_n + \varepsilon > 0 \\ 0 & \text{else} \end{cases} \tag{6.4}$$

where:

the coefficients $\beta_i$ are the parameters and

$\varepsilon$ is the standard logistic distribution error

### 6.2.3 SGDC: Stochastic Gradient

Stochastic Gradient Descent (SGD) is a popular technique for discriminative learning. Though SGD is well known in machine learning, it has recently received more attention especially in large-scale learning because of its simplicity and efficiency. For example, text classification and natural language processing where the data is often sparse, SGD can effectively scale to problems with over $10^5$ training instances and more than $10^5$ variables [58]. Basically, in statistical modelling, we generally aim to minimize a loss function, $F$:

$$F(p) = \frac{1}{n} \sum_{i=1}^{n} F_i(p)$$

where $p$ i.e. the parameters that minimize $F(p)$ are to be estimated and $F_i$ refers to the $i-$th example in the training set. So generally, the SGD training process takes the following steps:

1. Randomly initialize the set of parameters, $p$, and a learning rate, $\eta$

2. Repeat the following steps until an approximate minimum is reached:
   - Randomly shuffle examples in the training set
   - For each example, $i$, in training set do:
     - Adjust the parameters: $p \leftarrow p - \eta \nabla Q_i(p)$

As can be observed, these parameters $p$, which are adjusted when applied to training examples, are similar to the coefficients of the input variables in the general definition of the classification models.

### 6.2.4 MNNB & BNNB: Multinomial and Bernoulli Naïve Bayes

As we have earlier mentioned in §3.4.1, a *naïve Bayes* model, which is also referred to as the *idiot Bayes* [89], assumes that its features are conditionally independent of each other given a class. So under the naïve Bayes assumption, to compute the probability of a set of features $x_1, x_2, \ldots, x_n$ given a particular label $l$, we have:

$$p(x_1, ..., x_n|l) = \prod_{i=1}^{n} p(x_i|l) \tag{6.5}$$

Therefore, using a naïve Bayes model to classify a new example, the posterior probability can be computed in this form:

$$p(l|x_1, ..., x_n) \propto p(l) \cdot p(x_1|l) \cdot p(x_2|l)...p(x_n|l) \tag{6.6}$$

Multinomial naïve Bayes presents the $p(x_i|l)$ as a multinomial distribution which is a generalisation of the binomial distribution. In binomial distribution, only two labels can be assigned to each example. But multinomial distributions allow more than two possible labels. It is usually good for multi-class labels where distributions can be represented in discrete forms such as word counts.

For the Bernoulli naïve Bayes model, the features are also assumed to be independent but are boolean in type i.e. they are represented as binary variables. It is better in strict modelling of the occurrence or otherwise of a feature rather than the extent (or frequency) of its occurrence.

### 6.2.5 DCTC: Decision Tree

Decision tree algorithms are commonly applied in both classification and regression tasks. They are used in building predictive models that classify data instances into their target labels by using simple decision rules that are learned from the training data.

In principle a decision tree is a recursive function that partitions the training samples $x_i \in R^n$ where $n$ is the number of features, $i = \{1, \ldots, l\}$, and a $y \in R^l$, in such a manner that the samples with the same labels are grouped together. So if at node $m$ we have the samples $S$, we perform a split $\theta$ on $S$ using a particular feature $j$ and a threshold $t_m$, i.e. $\theta = (j, t_m)$, which then produces the subsets $S_{left}(\theta)$ and $S_{right}(\theta)$ which are defined by:

$$\begin{aligned} S_{left}(\theta) &= (x, y)|x_j \leq t_m \\ S_{right}(\theta) &= S \setminus S_{left}(\theta) \end{aligned} \tag{6.7}$$

Depending on the nature of the task, common impurity measures such as *gini* or *cross-entropy* and *misclassification* are then used to compute the impurity at node $m$. The $\theta$ that minimises the impurities is selected for the next recursive split until either the maximum depth is reached, the node has a minimum number of samples or just 1 sample.

The *Iterative Dichotomiser 3* (ID3), which was developed by Ross Quinlan [82], was the foremost algorithm for building decision tree models. It focused on building a multi-way tree that identifies the categorical feature that gives the largest information gain for categorical targets. ID3 was later succeeded by versions C4.5 and C5.0 with improvements in numerical data support and better memory management. A commonly used algorithms is *Classification and Regression Trees* (CART) which is an enhanced version of C4.5 with its optimised version implemented in Scikit-learn.

### 6.2.6 LSVC & SVEC: Linear and Non-linear Support Vector Machines

Support vector machines (SVMs) have been successfully applied to learning models for classification, regression and outlier detection. Because it uses *support vectors* (a subset of the training points) in its decision function, it is very memory efficient. It is also very versatile and effective even when the size of the feature set is higher than the training sample size. SVM provides common kernels but also allows for the definition of custom kernels. Support vector machines perform classification, regression or other tasks by drawing hyper-planes in the data space.

With our usual set of training data $S_{train} = \{(X_1, y_1), (X_2, y_2), \ldots, (X_n, y_n)\}$, the output $y_i$ can either be 1 or -1 and $X_i$ is a $p-$dimensional real vector. The functional margin (i.e. the hyper-plane that has the largest distance to the nearest data point of any of the classes in the training data) divides $S_{train}$ into the two classes. In general, the wider the functional margin, the better the generalization. Figure 6.3 is a simple diagram showing the main components of the SVM algorithm. There are linear and non-linear versions of SVM implementation to account for data that are clearly separable and those that are not.

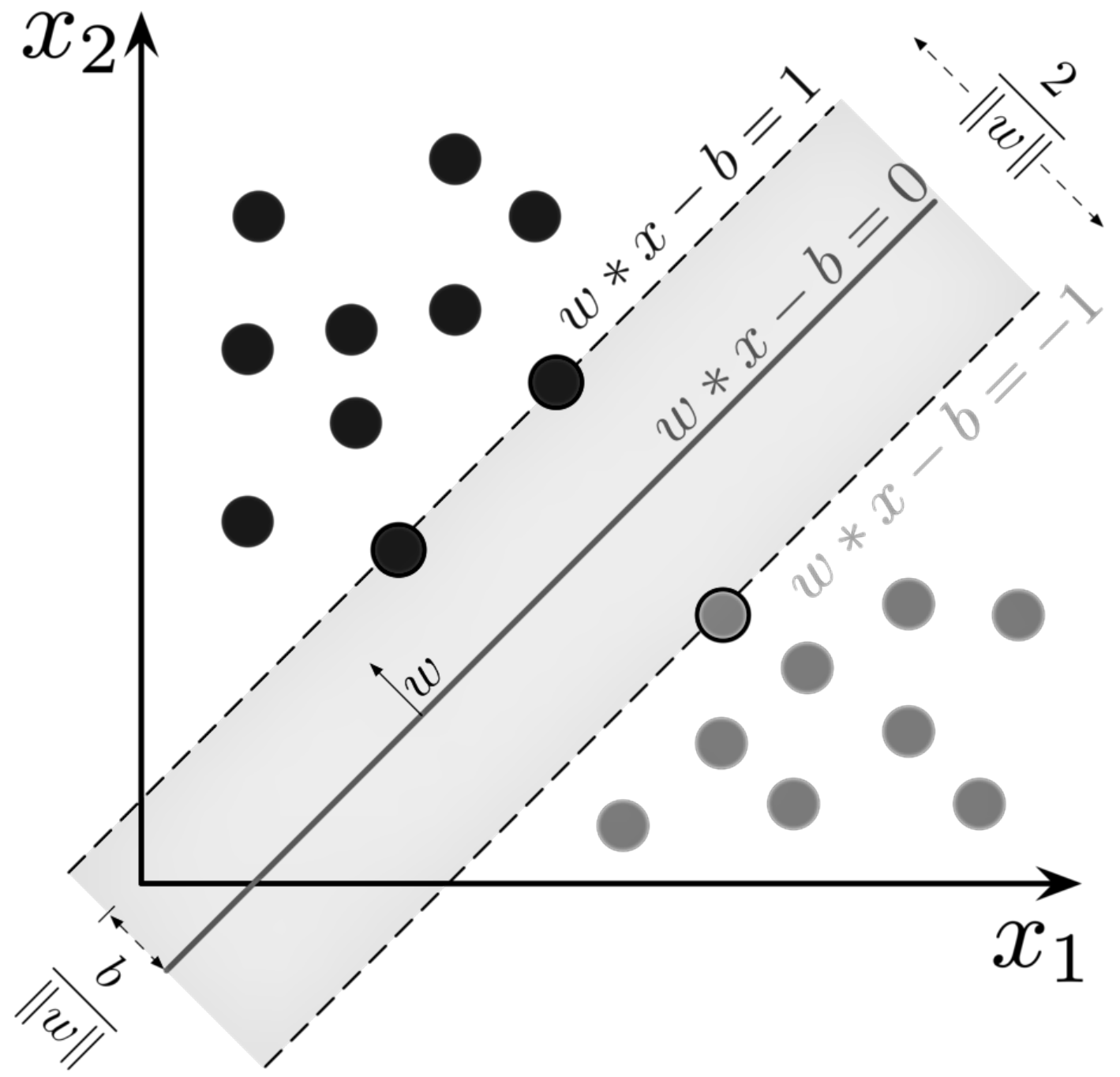

Fig. 6.3 A Simple Description of the Components of SVM Algorithm

### 6.2.7 KNNC: K-Nearest Neighbours

K-nearest neighbours, KNN, is a simple, non-parametric, instance-based learning method that locally approximates the function and performs the actual computation during classification. In the KNN algorithm for training classification models, a data point is classified by the majority vote of its $k$ closest neighbours i.e. the data point is allocated to the class that most of its neighbours belong to.

The training data and their labels $S_{train} = \{(X_1, y_1), (X_2, y_2), \ldots, (X_n, y_n)\}$ form the main input to the algorithm. Depending on the nature of the task, distance measures such as *Euclidean distance* or *Hamming distance* are often adopted [54]. The value, $k$, is defined by the user and has to be carefully selected in order to avoid noise or lack of clear class distinction.

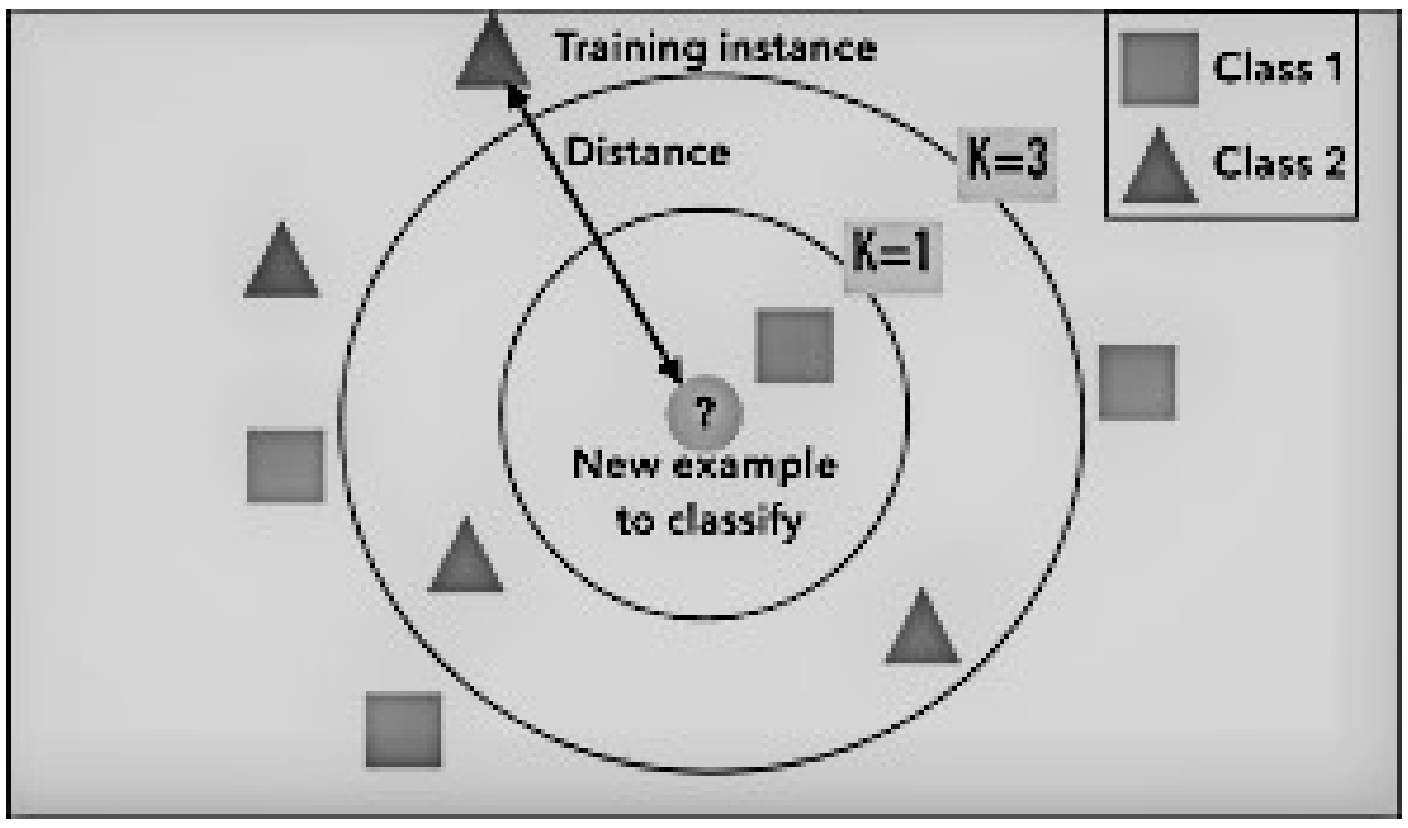

Fig. 6.4 A Diagram showing the K-Nearest Neighbour Algorithm

### 6.2.8 ADAB, BAGG & RFCL: Ada-Boost, Bagging and Random Forest

In machine learning, the adaptive boosting, *AdaBoost*, is a member of the family of ensemble learners. It is a meta-classifier developed by Freund and Schapire in 1995 [37] which trains a sequence of basic classification models (e.g. models that are barely better than random selection) on different versions of the data with weighted instances. It produces the final prediction by combining the predictions of all the models through a weighted majority vote or a weighted sum for classification or regression respectively.

The process begins by training a model with the original dataset and then proceeds with training copies of the same model on the same dataset but with the weights adjusted such that subsequent models focus more on incorrectly classified instances. The weights, $w_1, w_2, \ldots, w_N$, for the training samples are initially set to $w_i = \frac{1}{N}$. The instance weight adjustments occur at each iteration and consist of increasing the weights of misclassified instances as well as reducing those of correctly classified instances.

Bootstrap aggregation, otherwise known as *Bagging*, is another ensemble learning method that basically averages predictions from different models with the aim of reducing the variance. Again, the predictions from a set of $M$ base models could be aggregated with bagging by voting or averaging depending on whether it is a classification or regression task as shown in Equation 6.8.

$$f(x) = \frac{1}{M} \sum_{m=1}^{M} f_m(x) \tag{6.8}$$

*Random Forests*, also referred to as the forest of randomized trees, are a popular type of ensemble algorithms that combines decision tree base learners. Each of the base trees is applied to a random sample of the training data with a random subset of the features. In some extreme cases, instead of choosing the most discriminative splitting threshold, the best out of a randomly generated set of thresholds from candidate features is selected. The combination of these less correlated trees tend to give an overall better model.

## 6.3 Experimental Method

Having described the training algorithms used above, it is important to restate that we adopted a black-box approach to applying the basic scikit-learn implementations of the models using a common interface for presenting data and receiving classification results and analysis. We did not embark on any form of parameter optimisation for any of the training algorithms we used.

### 6.3.1 High-Level Process Flow

Figure 6.5 show a process flow that we followed in the experiments which focused more on the performance of each of the models across the range of wordkeys present in our dataset.

### 6.3.2 Training Data and Instances

As with the experiments presented in chapter 5, we used our standard dataset presented in Table 4.10, which has 29 ambiguous sets as defined in §4.2.2. There are a total of 80,844 instances. The extraction of the feature sets for the classification models was originally inspired by the works of Scannell [91] on character-based diacritic restoration which was extended by Cocks & Keegan [16] to deal with word-based restoration. In both cases, they applied only one training algorithm, the naïve Bayes which they had to manually implement for training their classifiers.

As shown in §3.4.1 and §3.4.2, the features used in [91] and [16] consist of a combination of $n$-grams that are adjacent to the target word and positioned to both sides of the target word. In our case, the words surrounding the wordkey define the context and are used to build the features used in training our models. The details of the feature extraction process are discussed in §6.3.3.

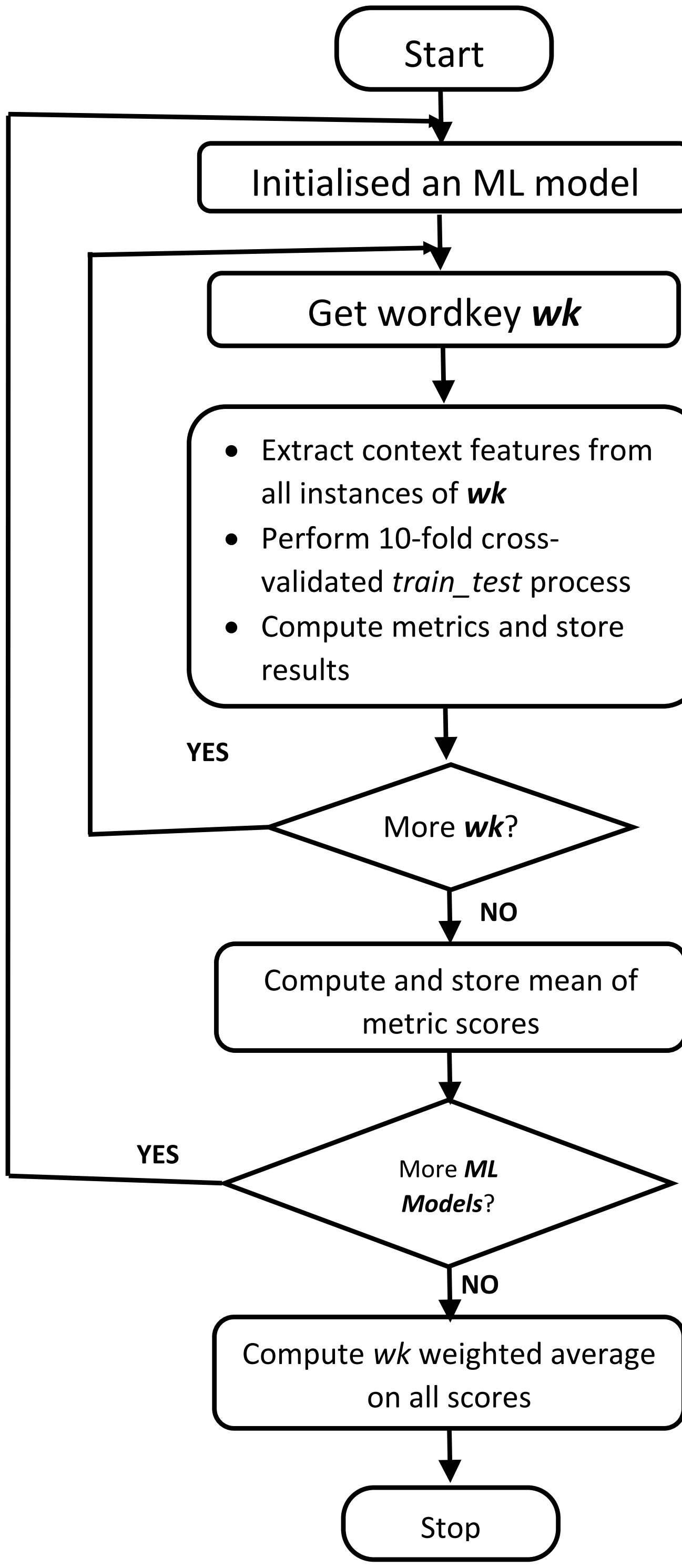

Fig. 6.5 High-Level Process Diagram of Diacritic Restoration with Machine Learning Classification

### 6.3.3 Feature Extraction

Unlike in [91] and [16] above, we *do not* use the n-grams on both sides as the actual features of our models but rather to define the context window of interest for each instance in our dataset. In our feature extraction method, we start by defining a maximum context window size of $n$ on the target wordkey such that the wordkey is *best centered* within the sequence of $n$ words that form the context. These context words will then be *vectorized* as explained in §6.3.3 i.e. represented as vector in terms of the words that appear in their context and their frequencies.

#### Defining the 'sticky' context window

For example, assuming we have defined a context window of 11 words i.e. 5 words before and after the wordkey as well as the wordkey itself, every instance (or sentence) in our dataset, no matter how many words it has originally, is reduced to only these 11 words. That is actually the maximum because some instances may not even have up to 11 words. The wordkey itself is not included in building the features for model training.

Now on the sticky window, our method ensures that if a target wordkey appears at the beginning or end of its sentence (i.e. a sequence of words), the first or the last sequence of 11 words respectively becomes the context. The context window slides across the sentence in an attempt to position the target wordkey $\boldsymbol{w_t}$ in its center but '*sticks*' (i.e. does not go beyond) both ends of the sentences hence the term 'sticky window'. Table 6.1 shows four examples of context words $\boldsymbol{c_i}, \ldots, \boldsymbol{c_{i+n-1}}$ (in **bold**), where $n = 7$, extracted from four different arbitrary 10-word sentences give the positions of each of the target wordkeys $\boldsymbol{w_t}$.

In our experiment only *actual words* in context, with the exception of the wordkey itself, are used as features. Punctuations, special symbols and numbers are removed before the extraction. The correct variant for the target wordkey is used as the label (or class) for the extracted context. Different window sizes, $n$ were used but $n = 9$, which roughly returns 4 words from each side of the target wordkey, produced the best scores for the performance metrics.

#### Feature Vectorization and Normalization

Having extracted the relevant context words for each instance, we build a vector-based representation of all the instances. Vectorization is a type of *bag-of-word* text representation technique in which a *document* is basically a vector of the counts of each of the words occurs in the entire corpus. In our case, the 'corpus' is a combination of all extracted contexts (not the entire original instance) while each context represents a 'document'. In other words, vectorizing a context implies creating a vector of counts of all the corpus words that appeared in the context. As expected, each context vector is sparse with the counts of its words in few places and zeros anywhere else.

| Examples | $w_0$ | $w_1$ | $w_2$ | $w_3$ | $w_4$ | $w_5$ | $w_6$ | $w_7$ | $w_8$ | $w_9$ |
|---|---|---|---|---|---|---|---|---|---|---|
| *1:* | $c_0$ | $\boldsymbol{c_1}$ | $\boldsymbol{c_2}$ | $\boldsymbol{c_3}$ | $\boldsymbol{w_t}$ | $\boldsymbol{c_5}$ | $\boldsymbol{c_6}$ | $\boldsymbol{c_7}$ | $c_8$ | $c_9$ |
| *2:* | $\boldsymbol{c_0}$ | $\boldsymbol{w_t}$ | $\boldsymbol{c_2}$ | $\boldsymbol{c_3}$ | $\boldsymbol{c_4}$ | $\boldsymbol{c_5}$ | $\boldsymbol{c_6}$ | $c_7$ | $c_8$ | $c_9$ |
| *3:* | $c_0$ | $c_1$ | $c_2$ | $\boldsymbol{c_3}$ | $\boldsymbol{c_4}$ | $\boldsymbol{c_5}$ | $\boldsymbol{c_6}$ | $\boldsymbol{c_7}$ | $\boldsymbol{c_8}$ | $\boldsymbol{w_t}$ |
| *4:* | $\boldsymbol{c_0}$ | $\boldsymbol{c_1}$ | $\boldsymbol{w_t}$ | $\boldsymbol{c_3}$ | $\boldsymbol{c_4}$ | $\boldsymbol{c_5}$ | $\boldsymbol{c_6}$ | $c_7$ | $c_8$ | $c_9$ |

Table 6.1 Feature Extraction: An illustration of the 'sticky window' Approach using the $n = 7$

To standardize this process and be able to use a common interface to build and compare models with the algorithms described in §6.2, we used the Scikit-learn set of libraries. Also for efficiency gain, Scikit-learn prefers features to be *NumPy*[2] arrays, we had to present our contexts in vectorized format.

Beyond vectorization, Scikit-learn provides different normalization functions including the *TfidfVectorizer* which also transforms the resultant vectors into a matrix of tf-idf values thereby reducing the dominance of the super high frequency words with little meaningful information. tf-idf stands for *term-frequency, inverse document-frequency* which is the product

$$\text{tf-idf(t,d)} = \text{tf(t,d)} \times \text{idf(t)}$$

of the *term-frequency* tf(t,d) i.e. the count of a term or word, in the document or instance, and the *inverse document-frequency* idf(t) i.e. how widespread the word is across instances. tf-idf is a term-weighting scheme for information retrieval but has recently been applied with success to document classification.

### 6.3.4 Restoration Process

As shown in the high-level process diagram, Figure 6.5, the restoration process in this experiment is a 10-fold cross-validated train-test run on each classification model for all wordkeys. The performance of each model is given by the weighted aggregation of its performances on the wordkeys. In cross-validation, we adopted the stratified approach to ensure that the distributions of the variants are roughly the same in all the folds and 9:1 train-test split.

[2]NumPy is a popular scientific computing library in Python with numerous routines for fast operations on arrays. See http://www.numpy.org/

This process is then repeated for each of the models trained with the 12 algorithms presented in §6.2 above and the model-based performances across the evaluation metrics are reported. Performance is assessed on how the predicted variants compare with the actual variants. As earlier defined in §4.3.2, the evaluation metrics are the accuracy, precision, recall and F1.

## 6.4 Evaluation of Results

In this section, we will be comparing the classifiers among themselves not only based on the defined performance metrics – accuracy, precision, recall and F1 – but also on their training efficiencies e.g. as it relates to the window-sizes. The best models will then be compared with best ones from the n-gram experiment. Also the behaviour of the *special-7* wordkeys identified in the last chapter will be monitored in this experiment as well.

### 6.4.1 Effect of context window sizes

In this experiment, context windows of different lengths were used in the training and testing of all the models. This is in order to observe the effect of varying windows lengths on the performance of the models with the aim of selecting the best performing window size for the rest of the experiment. Table 6.2 shows the average accuracy score obtained by using each of the window sizes across all the models. The scores are sorted in the order of the performance of the models but it indicates that, across the models, a window size of **9** appears to be the best though only slightly better than **7** and **11**.

| | Window Sizes | | | | | | |
|---|---|---|---|---|---|---|---|
| | **5** | **7** | **9** | **11** | **21** | **31** | *Avg* |
| **SVEC** | 0.6675 | 0.6675 | 0.6676 | 0.6676 | 0.6675 | 0.6676 | *0.6616* |
| **KNNC** | 0.7357 | 0.7427 | 0.7396 | 0.7348 | 0.6962 | 0.6733 | *0.7204* |
| **PCPT** | 0.7244 | 0.7426 | 0.7487 | 0.7487 | 0.7036 | 0.6702 | *0.7230* |
| **MNNB** | 0.7557 | 0.7555 | 0.7501 | 0.7433 | 0.7160 | 0.7034 | *0.7373* |
| **DCTC** | 0.7343 | 0.7576 | 0.7668 | 0.7617 | 0.7240 | 0.6854 | *0.7383* |
| **RFCL** | 0.7650 | 0.7861 | 0.7875 | 0.7833 | 0.7401 | 0.7127 | *0.7624* |
| **BNNB** | 0.7730 | 0.7883 | 0.7914 | 0.7886 | 0.7507 | 0.7259 | *0.7697* |
| **ADAB** | 0.7646 | 0.7859 | 0.7943 | 0.7956 | 0.7753 | 0.7582 | *0.7790* |
| **BAGG** | 0.7596 | 0.7842 | 0.7945 | 0.7946 | 0.7634 | 0.7347 | *0.7718* |
| **LSVC** | 0.7764 | 0.7987 | 0.8074 | 0.8070 | 0.7694 | 0.7419 | *0.7835* |
| **SGDC** | 0.7751 | 0.7991 | 0.8098 | 0.8075 | 0.7802 | 0.7575 | *0.7882* |
| **LRCV** | 0.7851 | 0.8073 | 0.8155 | 0.8163 | 0.7853 | 0.7638 | *0.7956* |
| *Avg* | *0.7514* | *0.7680* | **0.7728** | *0.7708* | *0.7393* | *0.7162* | *0.7531* |

Table 6.2 **ML:** Table showing the average accuracies of all models with contexts of different sizes arbitrarily selected from the range *5* to *31*. The result table is presented the order of increasing performance of the models from top to bottom. A window size of 9 produced the best results.

Also Figure 6.6 also shows that, besides ADAB and BAGG that got the same results window sizes **9** and **11**, all the other models got their best scores with the window size of **9**, i.e approximately 4 words on both sides of the wordkey, and so the scores and the analysis presented in the subsequent sections will be based on the window size of **9**.

### 6.4.2 Analysis: Accuracy

As with the n-gram model, we base our evaluation of the models trained in this experiment on their performances with respect to the simple accuracy measure that gives the percentage of the variants in the dataset for each wordkey that has been correctly predicted. We report the micro-averaged scores across the wordkeys i.e. the scores are weighted on the frequency of each wordkey in the dataset.

Table 6.3 shows the raw accuracy scores of the 3 *best* classification models (LSVC, SGDC and LRVC) along with the baseline (the unigram) and the best n-gram model (5-gram). The selected 'best' in this case are simply those that gave higher accuracy scores than the 5-gram models[3]. The full raw scores from all the classification models are shown in Table 4 and it may be observed from there that the SVEC model basically got the same score as unigram model.

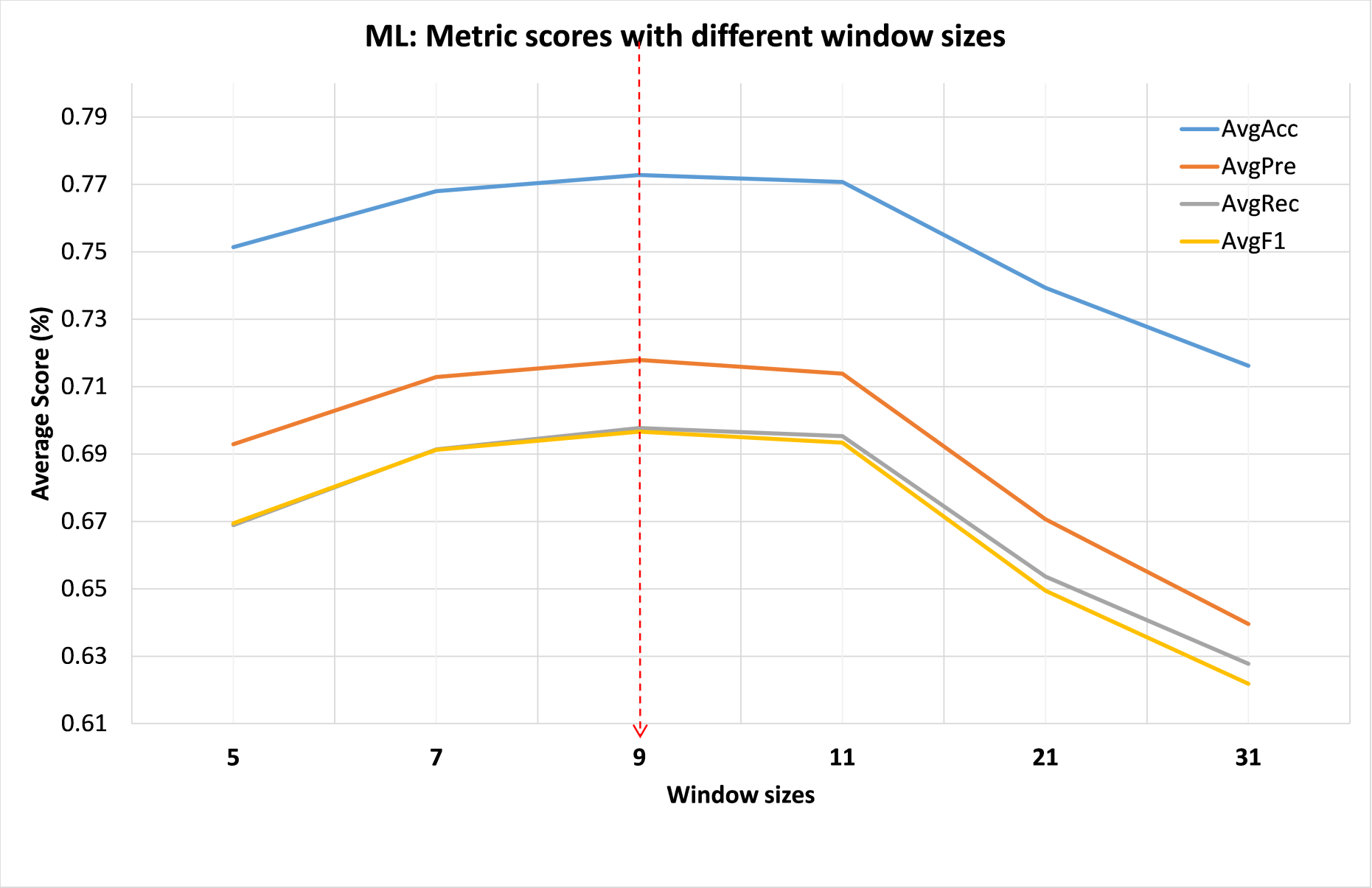

Fig. 6.6 **ML Winsizes:** Graph showing the effect of the different window sizes on the averages of all metric scores across the models

Table 6.3, as well as the vertical axis of Figure 6.7 indicates the models that got the *Best Score* which is similar to the one in ngram models described in §5.3.2, but is computed differently for ease of comparison across the models. The computation for the best model considers any model that has a score *greater than or equal to* the maximum score got by the classification models. So the 5gram score is not taken into account in computing the maximum but gets its best count if it gets the same maximum or higher.

So unlike the ngram model *Best Score (BS)* column, there could be multiple *best models* for each wordkey. One immediate implication of this method of selecting the best model for each wordkey is that the *best count* graph or piechart in Figures 6.8 and 6.9 does not indicate a mutually exclusive plot for each of the models i.e. models are likely to get counts on wordkeys that have been counted for others. Column *Best Model* in Tables 4 and 6.7 show a concatenation of all models that got the best results for each wordkey.

[3]This is merely a convenient choice because accuracy is the key factor for this comparison. If other factors such as efficiency were to be considered, LRCV may find it difficult to make the list.

| Wordkey | Counts | No of Variants | 1-gram | 5-gram | LSVC | SGDC | LRCV | BestScore (BS) | Best Model | Wdkey Imprvt | Baseline Imprvt | Error Reduction |
|---|---|---|---|---|---|---|---|---|---|---|---|---|
| buuru | 180 | 2 | 0.6667 | 0.5444 | 0.7000 | 0.6889 | 0.6889 | 0.7000 | LSVC | 3.33% | 3.25% | 9.99% |
| onya | 160 | 3 | 0.5312 | 0.6750 | 0.7125 | 0.6813 | 0.6687 | 0.7125 | LSVC | 18.13% | 4.50% | 38.67% |
| bu | 16999 | 2 | 0.6441 | 0.8593 | 0.7015 | 0.7212 | 0.7300 | 0.7300 | 5-gramLRCV | 8.59% | 6.25% | 24.14% |
| ukwu | 1432 | 3 | 0.5223 | 0.7996 | 0.7486 | 0.7165 | 0.7416 | 0.7486 | 5-gramLSVC | 22.63% | 8.11% | 47.37% |
| akwa | 1191 | 3 | 0.4232 | 0.8144 | 0.7498 | 0.7254 | 0.7322 | 0.7498 | 5-gramLSVC | 32.66% | 8.23% | 56.62% |
| aku | 384 | 3 | 0.4896 | 0.7474 | 0.7630 | 0.7500 | 0.7604 | 0.7630 | LSVC | 27.34% | 9.55% | 53.57% |
| iso | 201 | 2 | 0.5025 | 0.7512 | 0.7910 | 0.7711 | 0.7910 | 0.7910 | LSVCLRCV | 28.85% | 12.35% | 57.99% |
| i | 5347 | 2 | 0.6738 | 0.7514 | 0.7945 | 0.8001 | 0.7986 | 0.8001 | SGDC | 12.63% | 13.26% | 38.72% |
| agbago | 99 | 2 | 0.5354 | 0.5354 | 0.7980 | 0.7677 | 0.8081 | 0.8081 | LRCV | 27.27% | 14.06% | 58.70% |
| ibu | 682 | 2 | 0.6422 | 0.7331 | 0.8138 | 0.8079 | 0.8167 | 0.8167 | LRCV | 17.45% | 14.92% | 48.77% |
| ama | 1353 | 3 | 0.4213 | 0.6260 | 0.8219 | 0.8152 | 0.8211 | 0.8219 | LSVC | 40.06% | 15.44% | 69.22% |
| nku | 285 | 2 | 0.6140 | 0.8316 | 0.8281 | 0.8211 | 0.8211 | 0.8281 | 5-gramLSVC | 21.41% | 16.06% | 55.47% |
| si | 9039 | 2 | 0.5733 | 0.8448 | 0.8219 | 0.8289 | 0.8302 | 0.8302 | 5-gramLRCV | 25.69% | 16.27% | 60.21% |
| ju | 97 | 2 | 0.7423 | 0.8247 | 0.8144 | 0.8351 | 0.8041 | 0.8351 | SGDC | 9.28% | 16.76% | 36.01% |
| otu | 5947 | 2 | 0.6664 | 0.8177 | 0.8261 | 0.8298 | 0.8371 | 0.8371 | LRCV | 17.07% | 16.96% | 51.17% |
| juru | 306 | 2 | 0.5359 | 0.6046 | 0.8399 | 0.8366 | 0.8431 | 0.8431 | LRCV | 30.72% | 17.56% | 66.19% |
| ruru | 488 | 2 | 0.5041 | 0.7439 | 0.8422 | 0.8422 | 0.8443 | 0.8443 | LRCV | 34.02% | 17.68% | 68.60% |
| inu | 156 | 2 | 0.5769 | 0.6859 | 0.8462 | 0.8141 | 0.8333 | 0.8462 | LSVC | 26.93% | 17.87% | 63.65% |
| o | 31446 | 2 | 0.7353 | 0.7707 | 0.8490 | 0.8454 | 0.8514 | 0.8514 | LRCV | 11.61% | 18.39% | 43.86% |
| too | 125 | 2 | 0.7120 | 0.8640 | 0.8560 | 0.8560 | 0.8480 | 0.8560 | 5-gramLSVCSGDC | 14.40% | 18.85% | 50.00% |
| ikpo | 133 | 2 | 0.6165 | 0.7444 | 0.8571 | 0.8647 | 0.8346 | 0.8647 | SGDC | 24.82% | 19.72% | 64.72% |
| oke | 2267 | 3 | 0.7821 | 0.8729 | 0.8844 | 0.8778 | 0.8778 | 0.8844 | LSVC | 10.23% | 21.69% | 46.95% |
| igwe | 1392 | 4 | 0.6609 | 0.7723 | 0.8994 | 0.8879 | 0.8966 | 0.8994 | LSVC | 23.85% | 23.19% | 70.33% |
| okpukpu | 211 | 2 | 0.7109 | 0.8199 | 0.9052 | 0.9100 | 0.8957 | 0.9100 | SGDC | 19.91% | 24.25% | 68.87% |
| iru | 333 | 2 | 0.5315 | 0.7147 | 0.9189 | 0.9069 | 0.9279 | 0.9279 | LRCV | 39.64% | 26.04% | 84.61% |
| doo | 120 | 2 | 0.7000 | 0.7500 | 0.9417 | 0.9417 | 0.9417 | 0.9417 | LSVCSGDCLRCV | 24.17% | 27.42% | 80.57% |
| wuru | 112 | 2 | 0.6696 | 1.0000 | 0.9643 | 0.9643 | 0.9643 | 0.9643 | 5-gramLSVCSGDCLRCV | 29.47% | 29.68% | 89.19% |
| doro | 205 | 2 | 0.6390 | 0.8732 | 0.9561 | 0.9659 | 0.9561 | 0.9659 | SGDC | 32.69% | 29.84% | 90.55% |
| odo | 154 | 2 | 0.7273 | 0.8961 | 0.9610 | 0.9675 | 0.9610 | 0.9675 | SGDC | 24.02% | 30.00% | 88.08% |
| **Baseline = 1gram; %Error = 33.25%** | | | **66.75%** | **80.01%** | **80.74%** | **80.98%** | **81.55%** | **81.66%** | **Performance Analysis** | **Model** | **Imprvnt** | **Error Reduction** |
| **Best Model Counts** | | | **0** | **7** | **14** | **9** | **12** | | *Best Score* | ***LRCV*** | **14.80%** | **44.51%** |
| **Model Error Reduction:** | | | 0.00% | 39.87% | 42.06% | 42.78% | 44.51% | 44.83% | *Best Counts* | ***LSVC*** | **14.22%** | **42.78%** |

Table 6.3 **ML Accuracy:** Table showing the raw accuracy scores of the best 3 machine learning models as compared with the baseline (1-gram) and the best n-gram model (5-gram). [*NB:* The color code indicates the worst(red)-to-best(green) based on the accuracy performances and improvements on wordkey and global baselines].

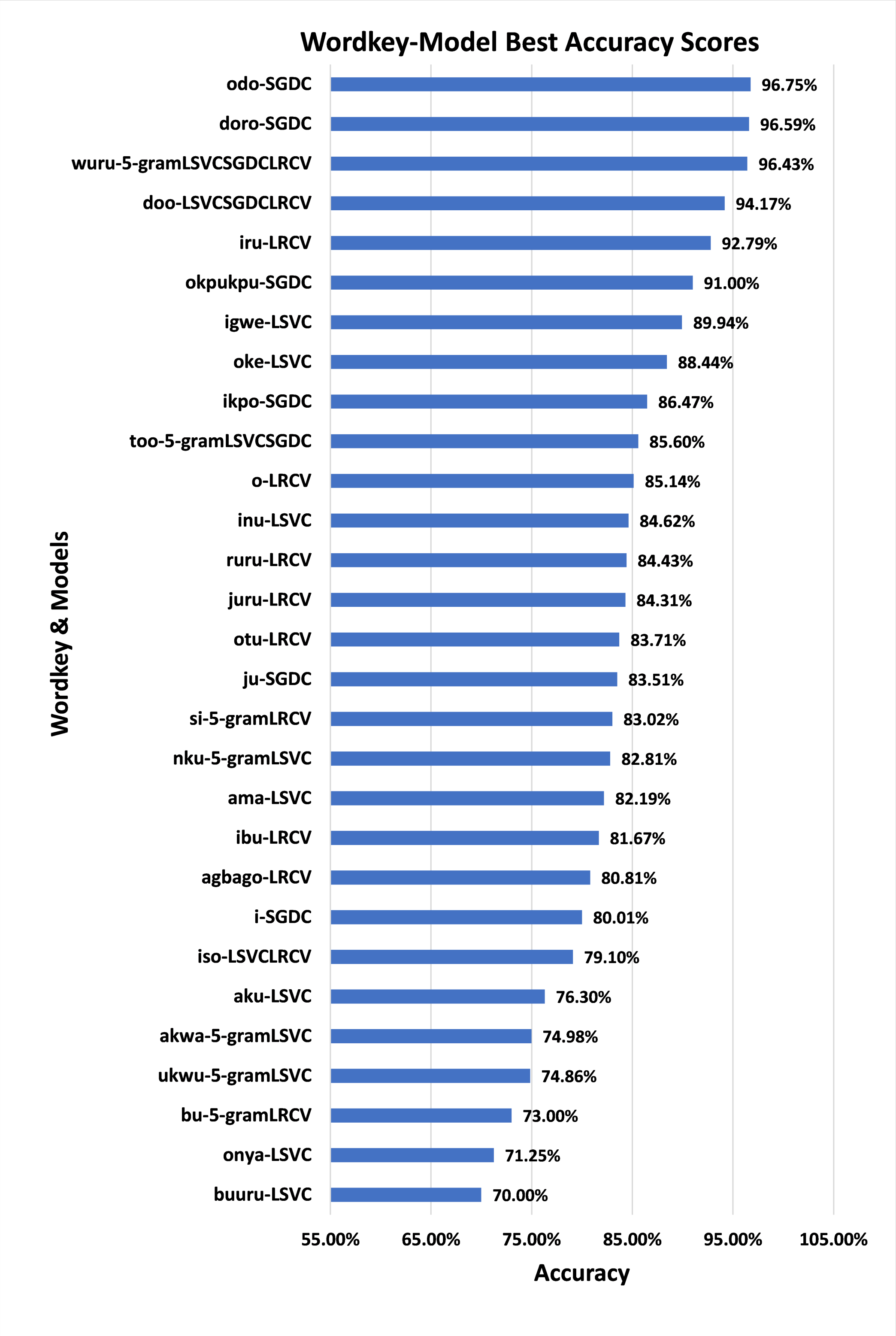

Fig. 6.7 **ML Accuracy:** A bar-chart showing the plot of the wordkey best score as well as the models that got those scores in their descending order of performance.

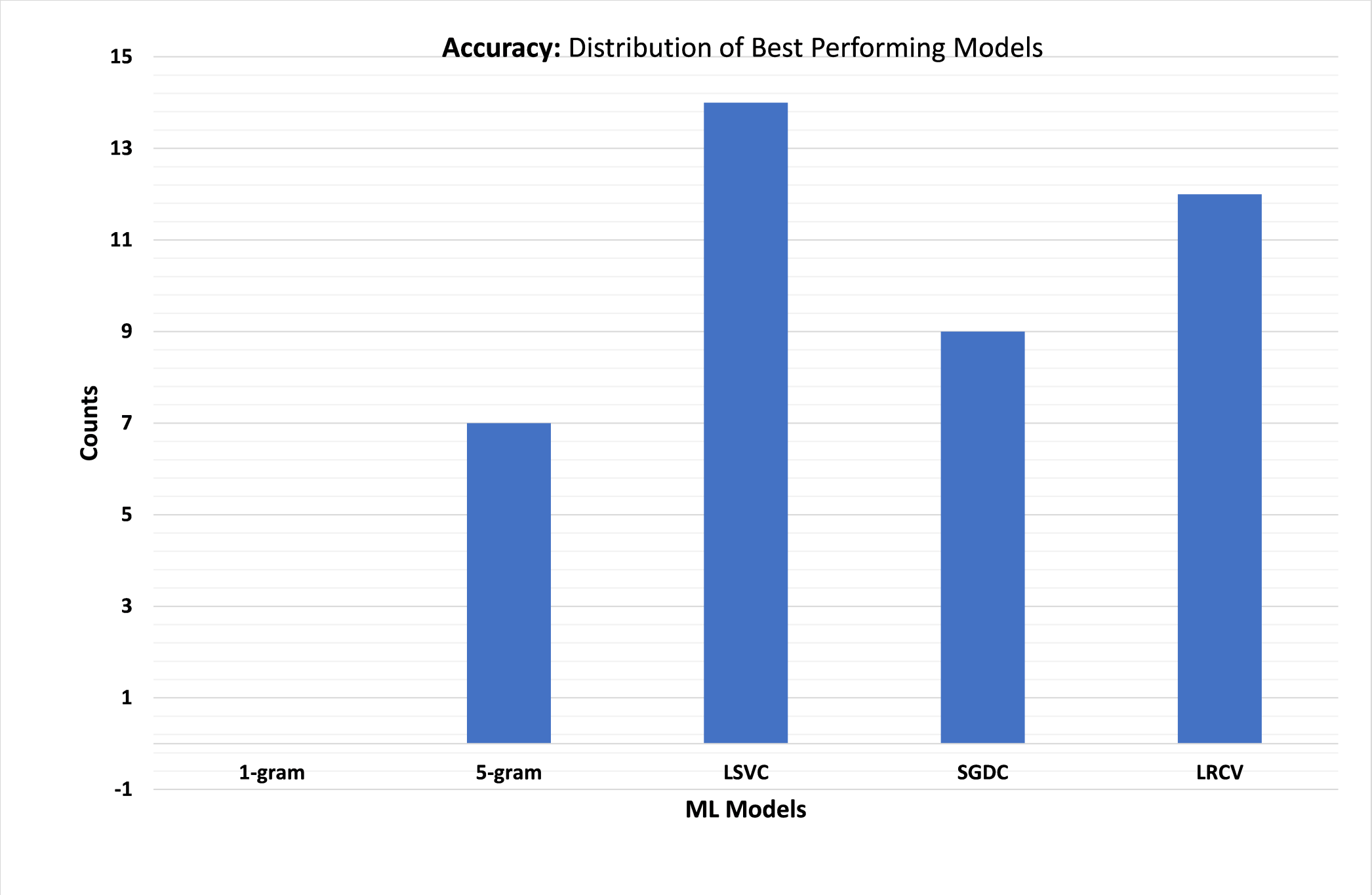

Fig. 6.8 **ML Accuracy:** Graph comparing the frequency of getting the best score on wordkeys by each of the models.

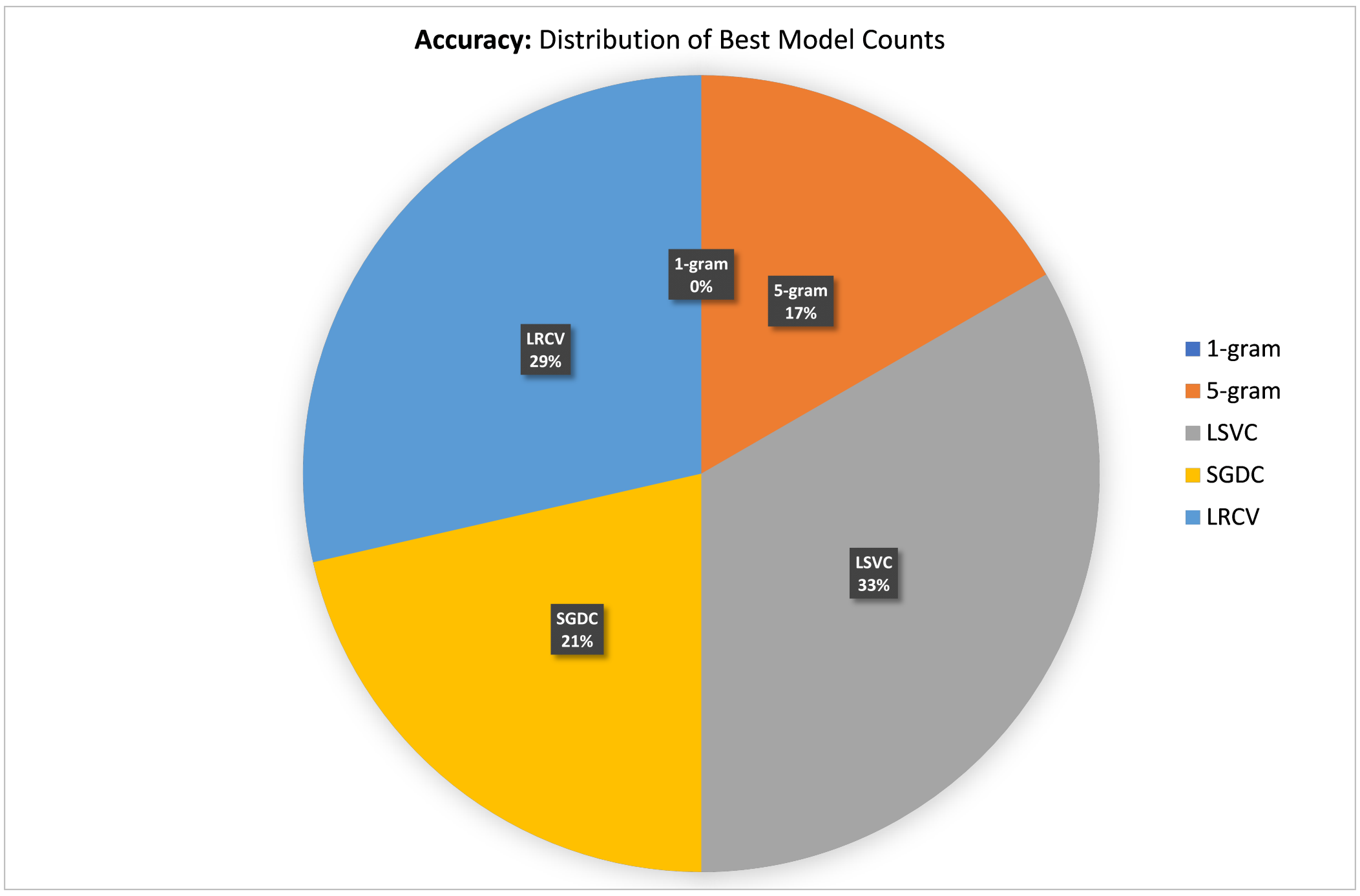

Fig. 6.9 **ML Accuracy:** Pie-chart showing the relative percentage distribution of the *best score* frequencies shown in Figure 6.8 across the models.

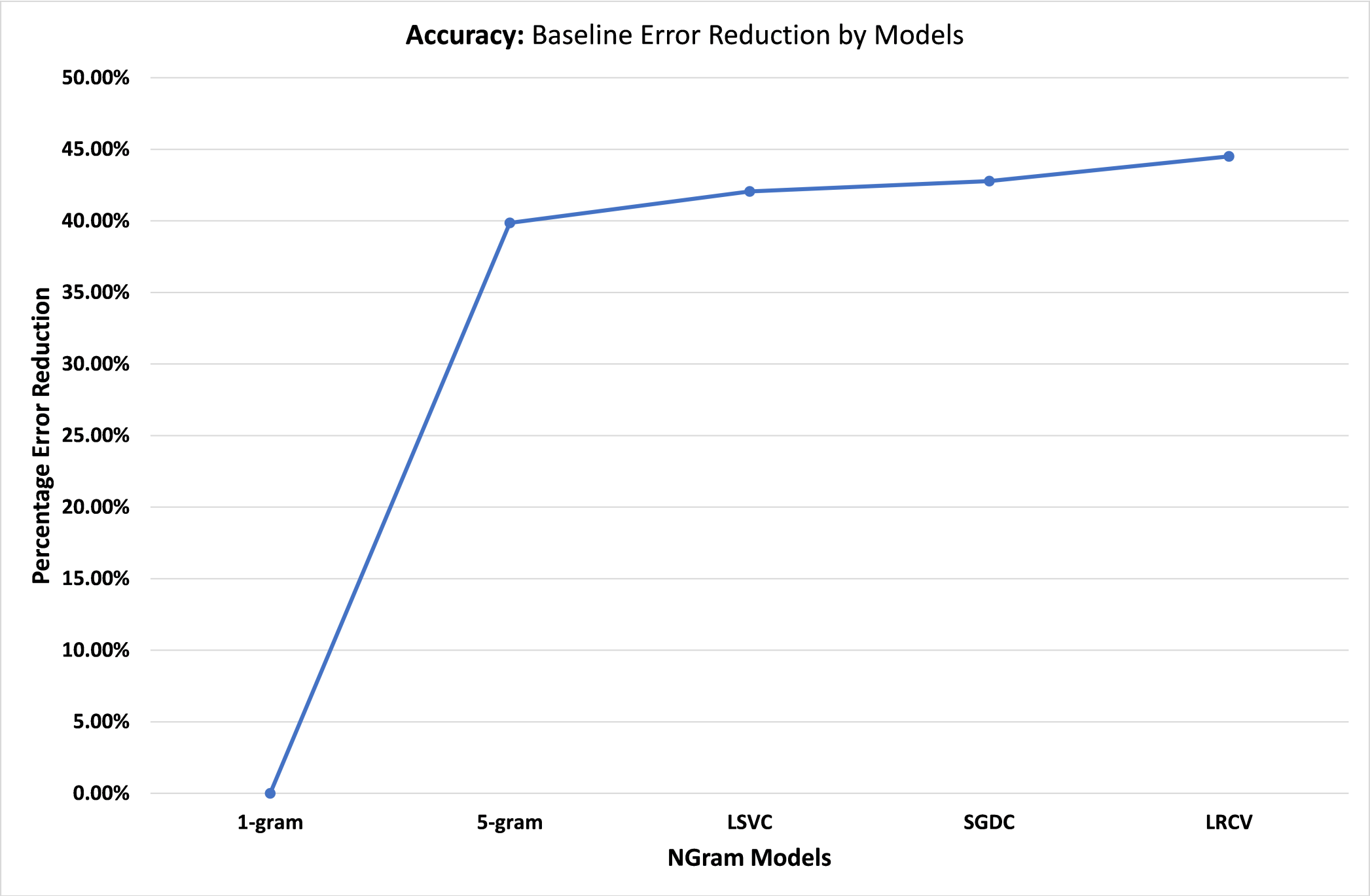

Fig. 6.10 **ML Accuracy:** Graph showing the percentage reduction of the global baseline error by all the models.

It will be noticed that the 5-gram model got the same or better result in 7 out of the 29 wordkeys. Indeed the models struggled to out-perform the the 5-gram model such that the best model got only 1.65% better than the 5gram model on the global baseline which gains only approximately 4.64% in the baseline error reduction as shown in 6.10.

Recall that with the ngram models, it was difficult to improve the accuracies of *agbago* and **buuru** even with higher ngrams. However with the classification models, there is an increase in the wordkey baseline accuracy scores. Figure 6.11 shows that, unlike with the ngram experiment, every wordkey made some positive improvement on both its baseline score and the global average score. This suggests that the classification algorithms seem to pick up some predictive properties which could not be identified by the ngram models.

Also, if the 5Gram and LRCV (best models from both experiments) are compared on their abilities to reduce the individual wordkey classification errors, Figure 6.12 shows that the 5Gram did better in only 8 out of the 29 wordkeys and worse in the rest. Indeed, for *buuru*, it increased the error instead of reducing it. This probably explains why there is a wider gap (6.40%) between the wordkey macro-averaged[4] score of the 5-Gram (76.79%) and that of the LRCV (83.19%) and a 16.51% difference in the error reduction from the macro-averaged baseline error of 38.79% by both models..

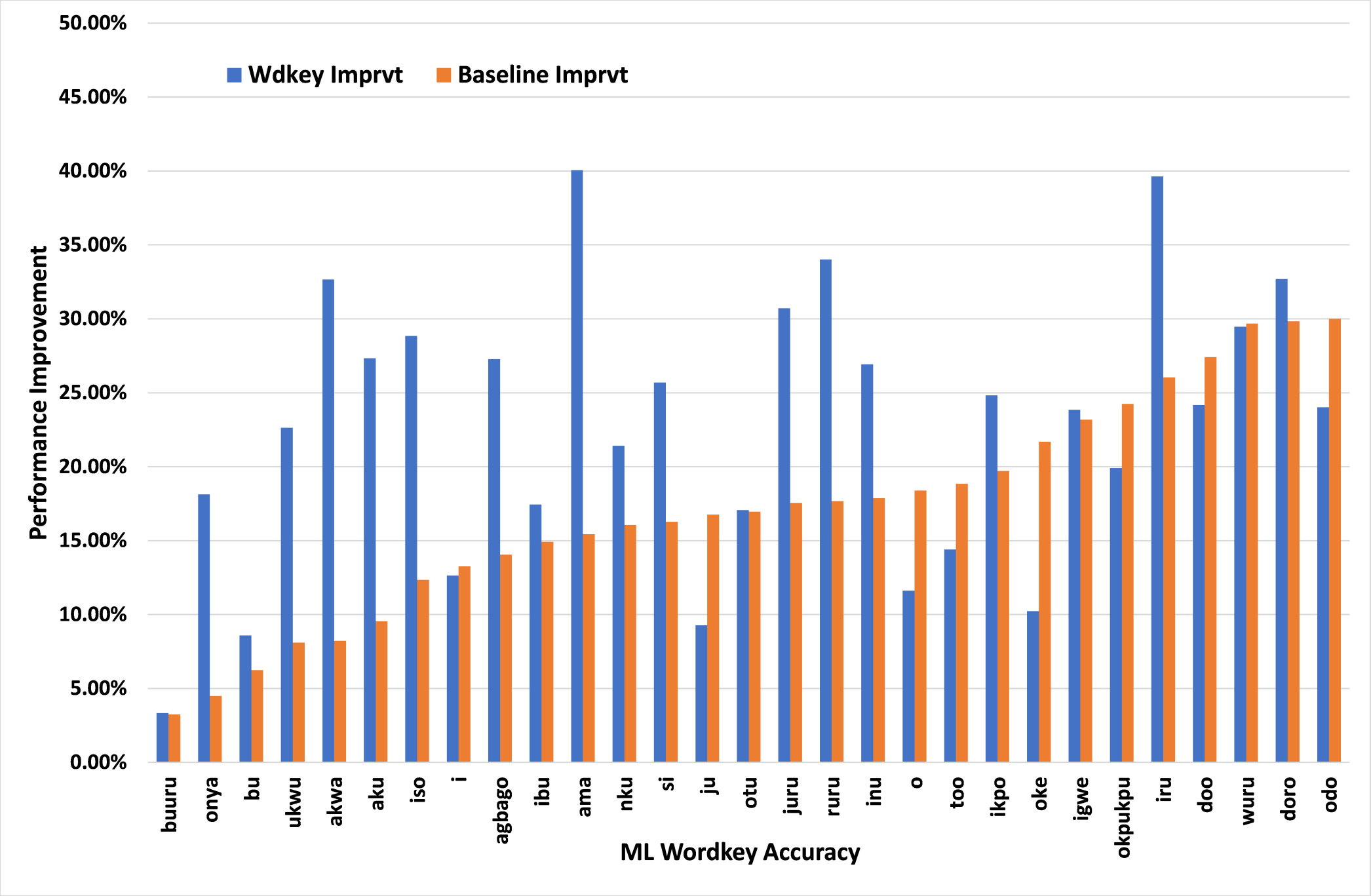

Fig. 6.11 **ML Accuracy:** Graph showing the maximum improvement in the baseline accuracy scores on the individual wordkeys .

### 6.4.3 Analysis: Precision, Recall and F1

In addition to the accuracy scores, the comparison of the models with respect to their precision, recall and F1 scores is presented in Figure 6.13 and a few details became obvious. The first is that LSVC does better than the rest on precision. This means that that it is more likely to correctly predict the individual variants of a wordkey than the others.

Another observation is that the 5Gram model is comparatively lower on the precision, recall and F1 than the other models in spite of having a good accuracy score. This means that, compared to the classification models, the 5gram model tends to get a lower score on the number of predicted “correct” variants that are truly correct. Also, in general, the 5gram model predicts less of the actual correct total correct variants than the classification models, hence the low recall score.

[4]This is the simple average of the wordkey scores that treats all wordkeys equally without considering their frequency in the dataset

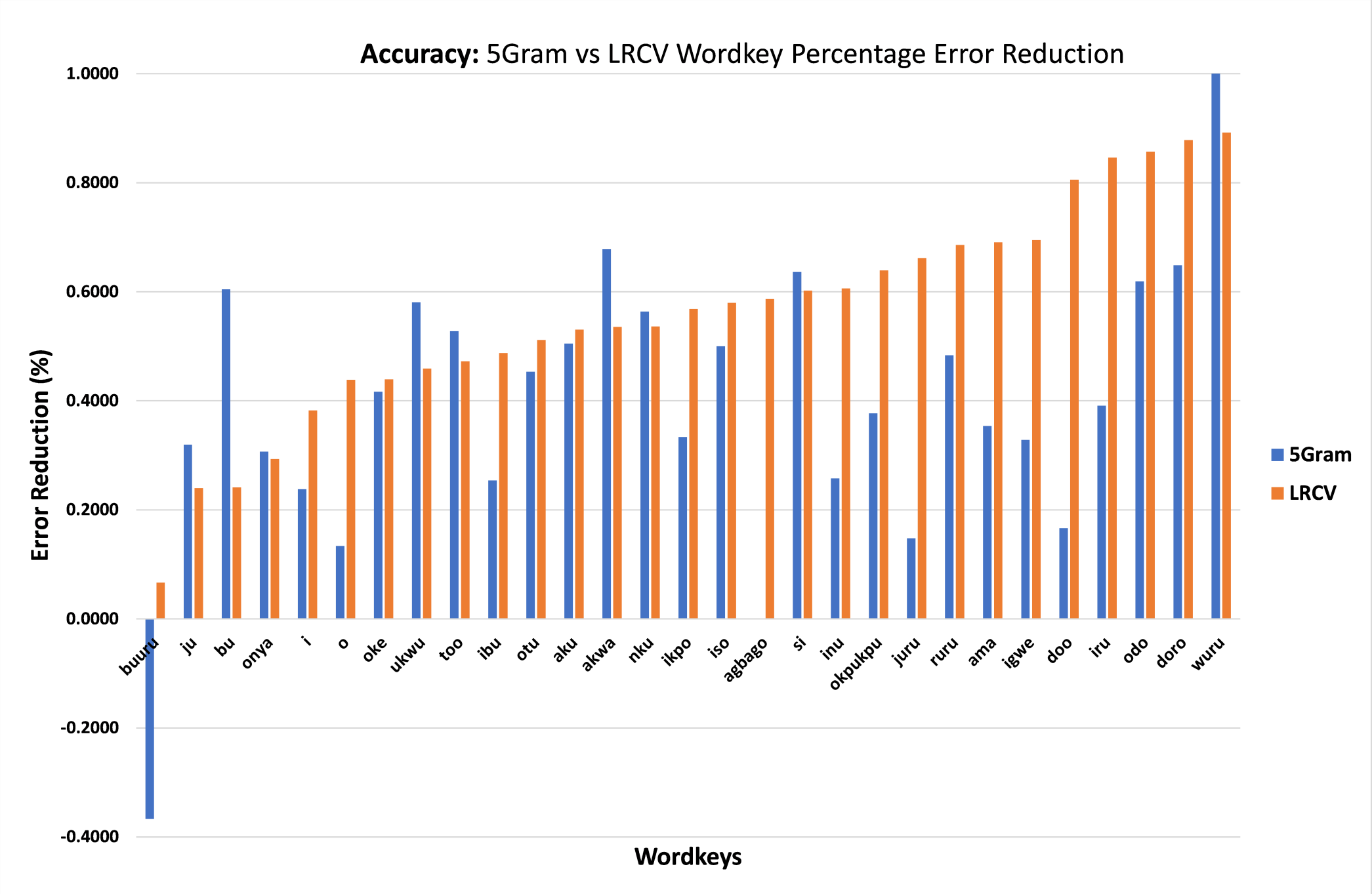

Fig. 6.12 **ML Accuracy:** Graph showing the comparison of the percentage reduction of the wordkey baseline errors by the 5Gram and the LRCV.

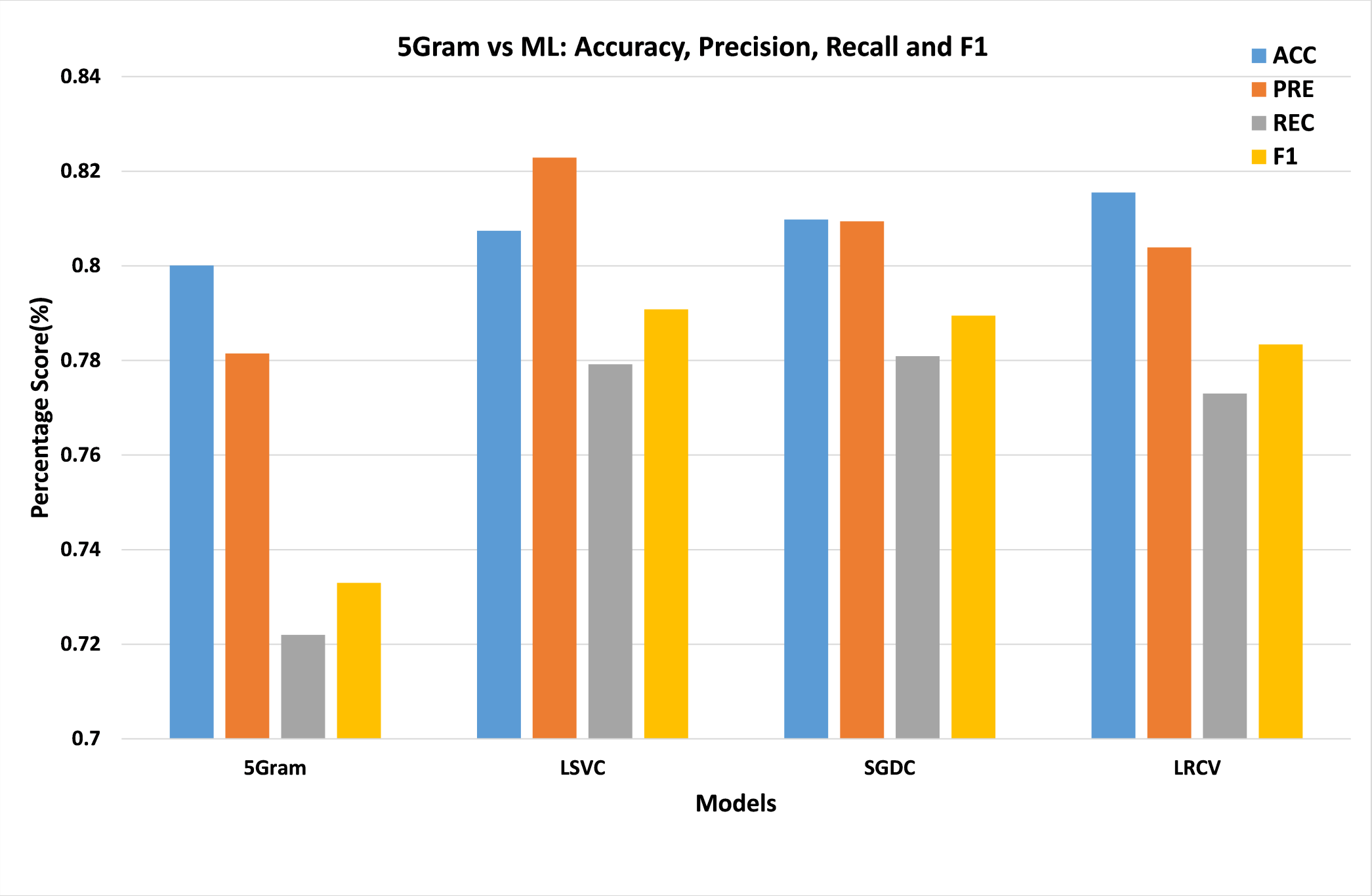

Fig. 6.13 **ML APRF:** Graph showing the comparison of the 5gram model and the 3 best classification models on their accuracy, precision, recall and F1 scores.

### 6.4.4 Wordkeys with special characteristics

Again, we looked at the special wordkeys described in §5.3.4 to compare how the 5Gram and LRCV models performed on them. On average, there seem to be not much improvement given that LRCV did better only on approximately half of the set i.e. *buuru*, *agbago*, *o* and *igwe*. However, a key take away is that the two most challenging, *buuru* and *agbago* got improved scores not just by LRCV also by other models with *agbago* getting above 80% accuracy.

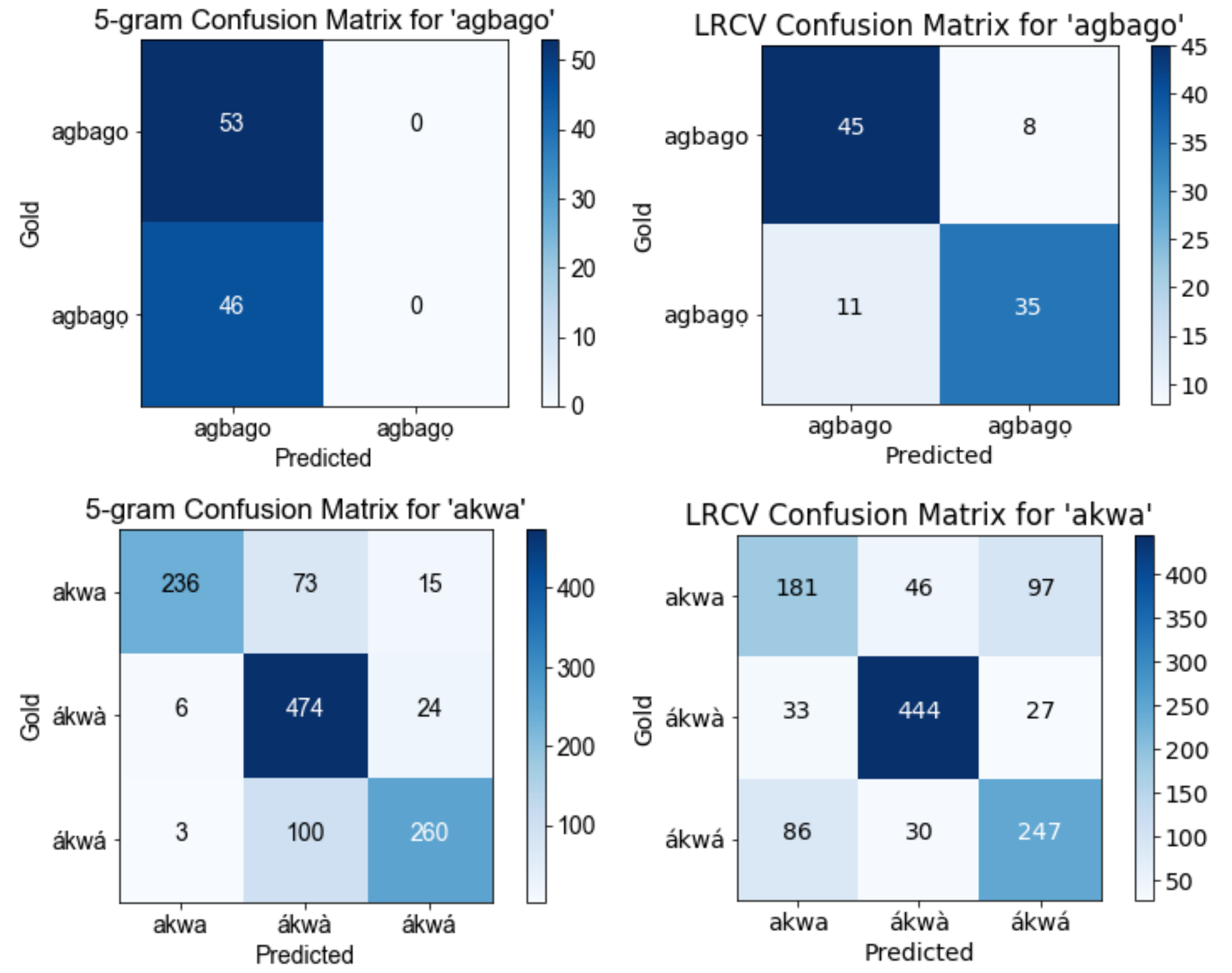

Fig. 6.14 (a) **Summary:** *Special-7* wordkeys - Confusion matrix of the results 5Gram and LRCV.

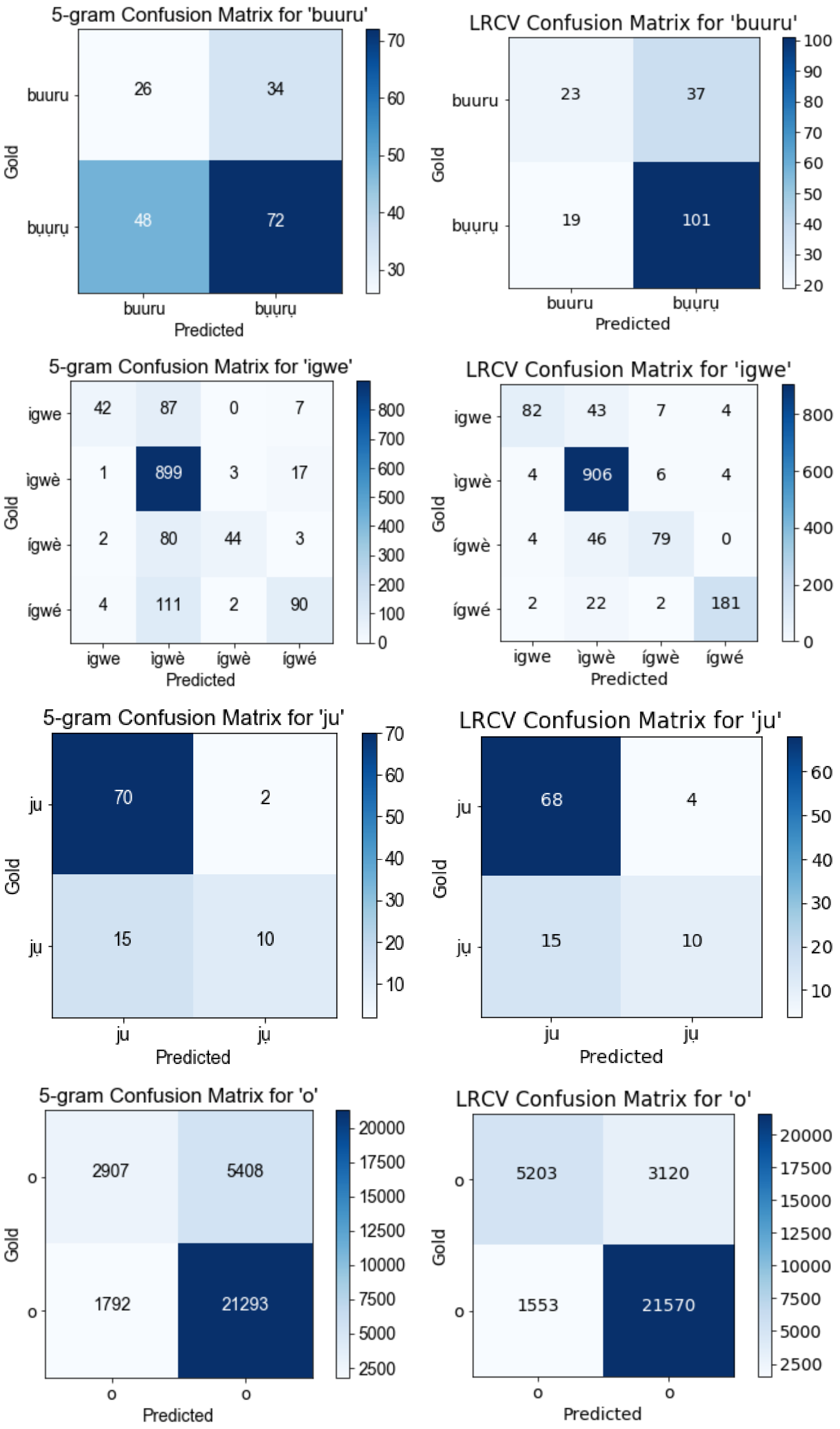

Fig. 6.15 (b) **Summary:** *Special-7* wordkeys - Confusion matrix of the results 5Gram and LRCV.

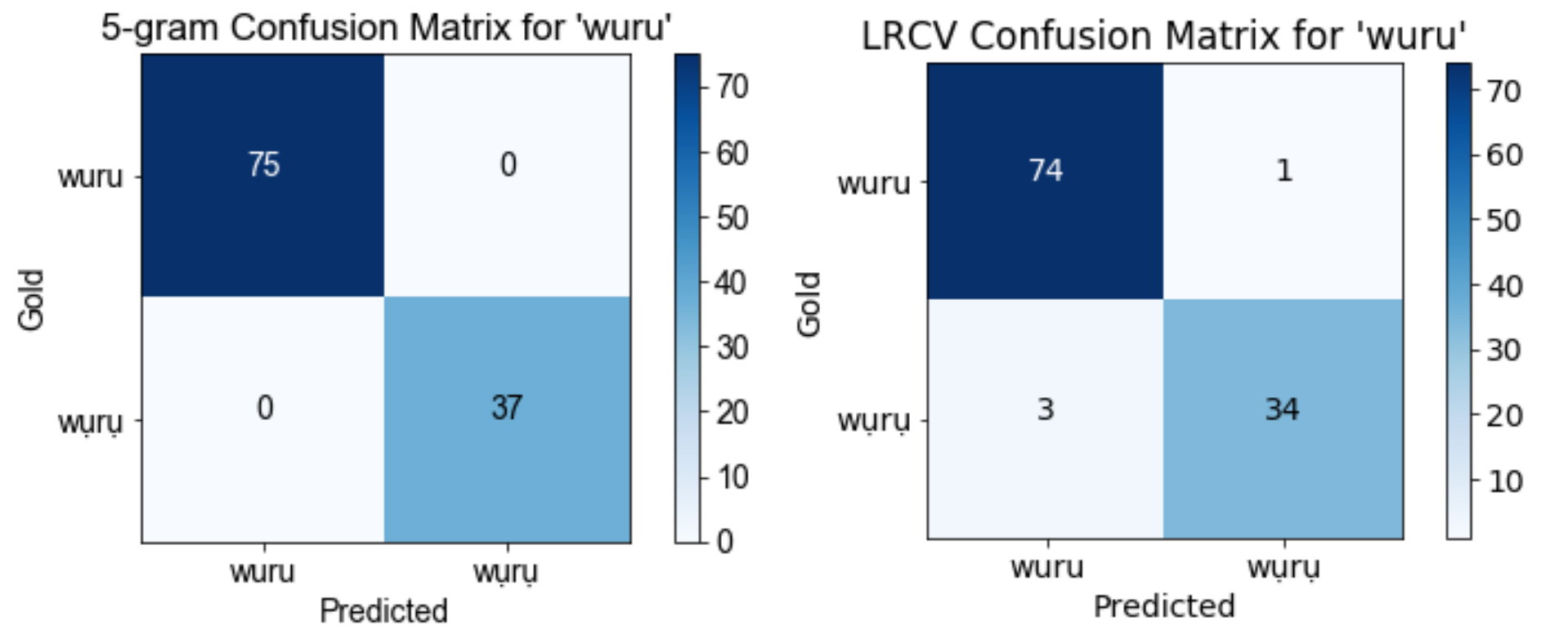

Fig. 6.16 (c) **Summary:** *Special-7* wordkeys - Confusion matrix of the results 5Gram and LRCV.

## 6.5 Chapter Summary

In this section, we designed an experiment that trained, tested and compared 12 classification models with the default parameters of different learning algorithms different in §6.2. The training features that worked best for the trainings were the vectorized words around each wordkey which are within the 'sticky' (see §6.3.3) context window of 9 i.e. approximately 4 words on both sides of the wordkey. The training and testing process was as described in the high-level process diagram presented in Figure 6.5 and applied a 10 fold cross-validation.

The window sizes tested were 5, 7, 9, 11, 21, and 31. While the efficiency dropped as window sizes increased, performances peaked at window size of 9 and 11 but declined afterwards. Our experiments show that that **MNNB**, **BNNB** and **LSVC** were the most efficient to train and produced considerably good results as shown in Table 6.2; while the SVEC, despite getting the worst results, took the longest time to train.

Three models that out-performed the 5gram are LSVC, SGDC and LRCV with the accuracy scores of 80.74%, 80.98% and 81.55% respectively and their raw wordkey scores are compared with those of the 5gram model in Table 6.3. As shown in Table 6.3, on 7 out of the 29 wordkeys (almost 25%), the 5-gram performed better than all the models but did worse in the rest. Despite not getting the best score, the LSVC got the highest score on almost 50% of the wordkeys making it the model with the highest wordkey. The best model, LRCV reduced the baseline error by 44.51% on the weighted average scores that considers that size of the wordkey in the dataset. However, on the simple average of the wordkey scores, the LRCV got 83.19% against the baseline score of 61.21% and the 5-gram score of 76.79% there reducing the error by 56.67% while the 5gram error reduction is 40.16% i.e. a difference of 16.51%.

Beyond the accuracy scores, it is clear that the 5-gram model did not measure up to the machine learning models on the other metrics. The LSVC has relatively better results with respect to precision, recall and F1. On the special wordkeys, although the machine learning models did not perform as well as the ngram models in 3 out of 7 of the wordkeys (i.e. *akwa*, *ju* and *wuru*), they improved the results of the rest including the *agbago* that did not improve across all ngram models and *buuru* that got worse results with higher ngrams.

Overall we got better results with the machine learning models. Depending on which one matters more to the user between accuracy and speed, one can choose any of the best three models LSVC, SGDC and LRCV.

# Chapter 7

# IDR with Embedding Models

In the last two chapters, we presented two broad approaches to diacritic restoration: the *n*-gram approach and the machine learning approaches. In this chapter, we shall be introducing yet another approach that will be based on denser vector representation of words referred to as *embedding models*.

Given that our training data is comparatively smaller than those used in other works §3.7, we shall explore a few transfer learning techniques to create Igbo word embeddings from a variety of existing English trained embedding models. These techniques, which will be explained in §7.3, will enable us to leverage the resources of mainstream languages and see how well they will do in some IgboNLP tasks especially diacritic restoration.

As an aside, we shall also conduct experiments to validate the embedding models created in this process by developing some standard intrinsic evaluation tasks – *odd word*, *analogy* and *word similarity* – to see how well the models will do on them. We start by introducing the key concepts of word embeddings and their application to diacritic restoration as well as the transfer learning methods that will be applied in this work.

## 7.1 Word Embedding Models

Word embedding models gained a lot of traction since the introduction of *word2vec* in 2013 by Mikolov et al [65], which was also almost immediately followed by another popular model, *Glove*, developed by Pennington et al. [79]. They are generic semantic representations from the corpus that highlight the concept of distributional hypothesis [45] and count-based distributional vectors [9]. Word embeddings provide an alternative to the *one task, one model* approach in which a model is specifically trained to solve a particular problem. Their application areas span most NLP tasks and other fields such as biomedical, psychiatry, psychology, philology, cognitive science and social science [4].

There are many approaches to training embedding models but the most common are approaches are:

*count-based* models [79] in which the semantic similarity between words are learned by counting the co-occurrence frequencies of the words i.e. counting the number of times both words appear within the same context.

*predictive* models [65], [67] in which the vectors are learnt by improving the "predictive ability" of the model i.e. minimizing the loss between the target word and the context word.

### 7.1.1 Word vectors

The key concepts in these models are similar to the vectorized models we introduced in chapter 6 where sentences (or word sequences) are represented as vectors. However, our earlier models have vectors that are largely sparse and their lengths are determined by the size of the vocabulary of the training data. This is because such models often rely on the addition of the *one-hot* encoding of the words in the sentence.

#### One-hot vectors

A *One-hot* (or 1-*of*-*N*) vector encodes a word in a vocabulary of size $N$ by setting its corresponding element to 1 and the rest to zero. Figure 7.1 shows the one-hot encoding for the word *queen* in a 5-word vocabulary that contains: *king*, *queen*, *man*, *woman*, *child*.

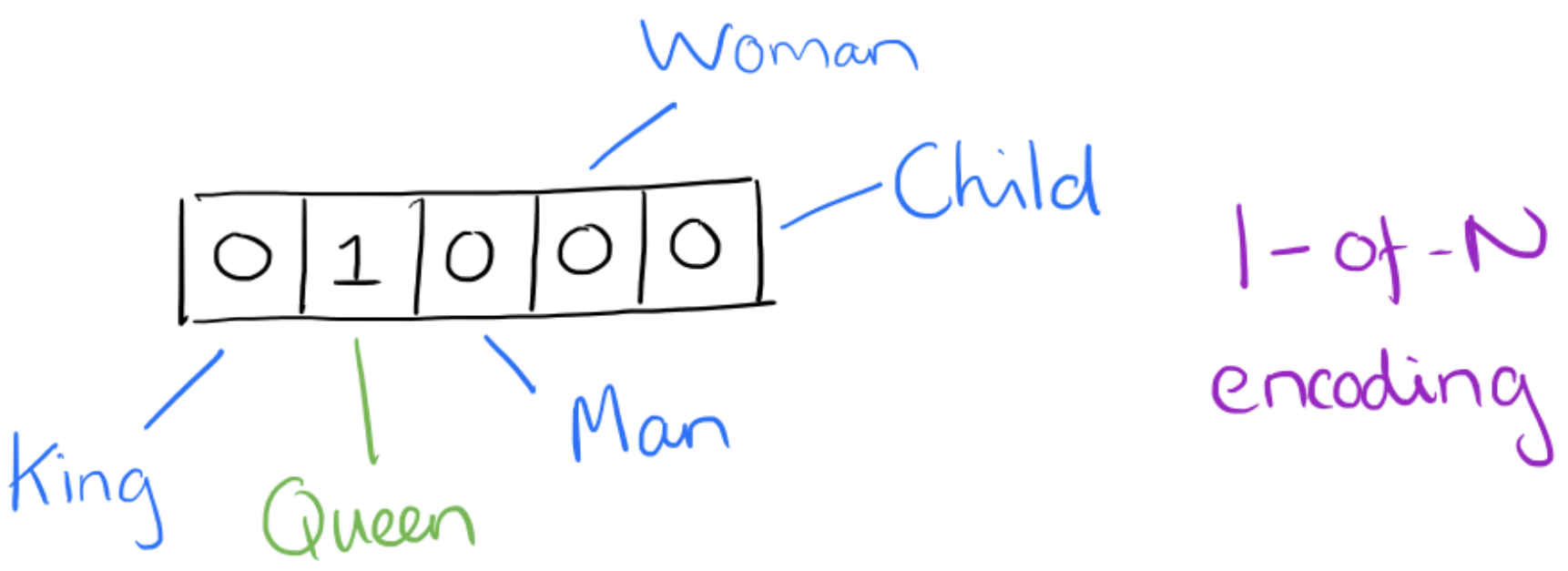

Fig. 7.1 Example of one-hot vector encoding, *Source*: [19].

As can be observed from the above, when dealing with this type of encoding, *equality* – i.e. whether any two words (or vectors) are exactly the same – is the only test possible. It does not allow any meaningful comparison between different word vectors. Therefore, although we know that there is a relationship between *queen* and *king* or between *queen* and *woman*, we cannot measure that as their vectors are orthogonal and so the dot product of any two different vectors gives 0. With one-hot representation, nothing suggests that a *queen* is a *woman* or that *king* and *queen* are both royalty. Also, in spite of encoding only one position, the length of each vector is the size of the vocabulary(!) and so it is too sparse to be efficient.

### Distributed representation

Word embedding uses a distributed representation of words. In distributed representation, the algorithm learns from the training corpus a vector representation for the word across $N$-dimensions with each element as a *weight*, i.e. the degree of closeness or relevance with respect to each of the dimensions.

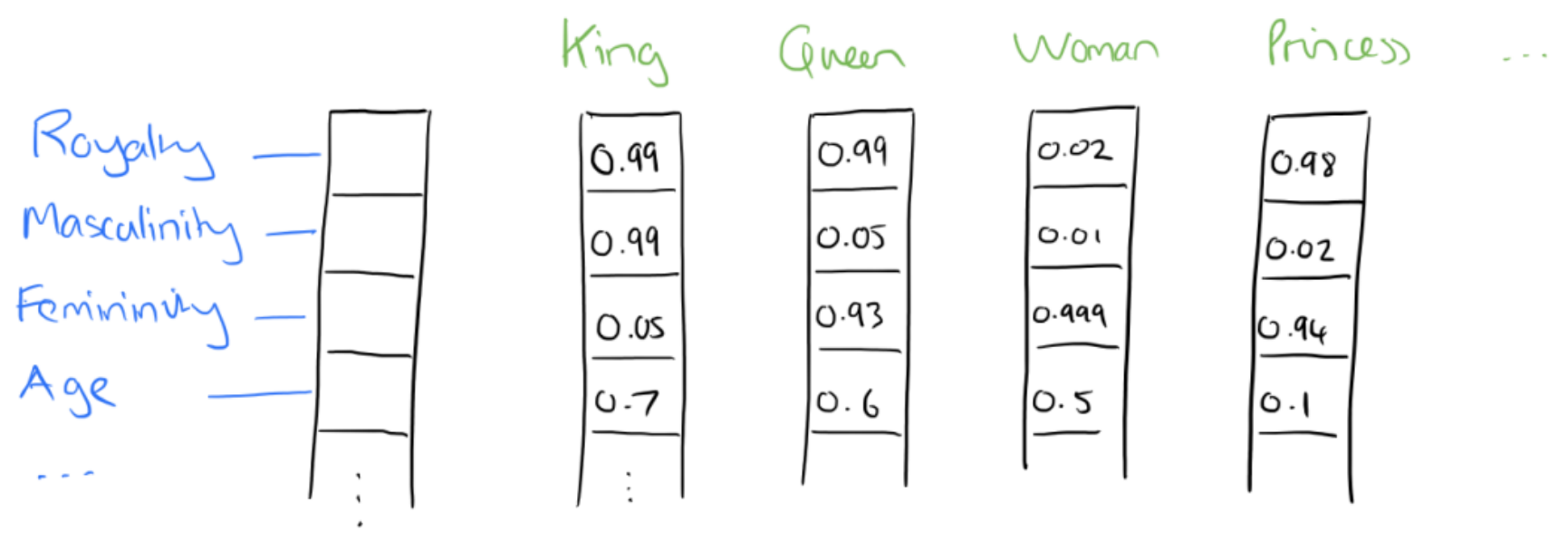

Fig. 7.2 Example of distributed representation, *Source*: [19].

In this case, the number of dimensions, $N$, is predetermined at training time leaving the training algorithm to identify $N$-features whose weights will form the elements of a given word vector. Figure 7.2 shows learned feature labels which could depict such concepts as *royalty*, *masculinity*, *femininity* or *age*[1]. This approach is more efficient as $N$ is often much lower, say 300 or even less, than the length of the dictionary which typically runs into tens or hundreds of thousands of unique words from the corpus. It is also more effective and informative because it allows room for other kinds of measures beyond just equality e.g. the similarity between words or the relatedness of a word to a feature.

[1]This illustration is for clarity. In practice, we do not know what the identified features actually mean or what they refer to. Unsupervised learning algorithms are used to figure out which features are distinct and relevant.

### word2vec

*word2vec* [65] is an efficient Python implementation of the *continuous bag-of-words* (CBOW) and *skip-gram* (SG) architectures for computing vector representations of words. It takes a text corpus as input and produces the word vectors as output.

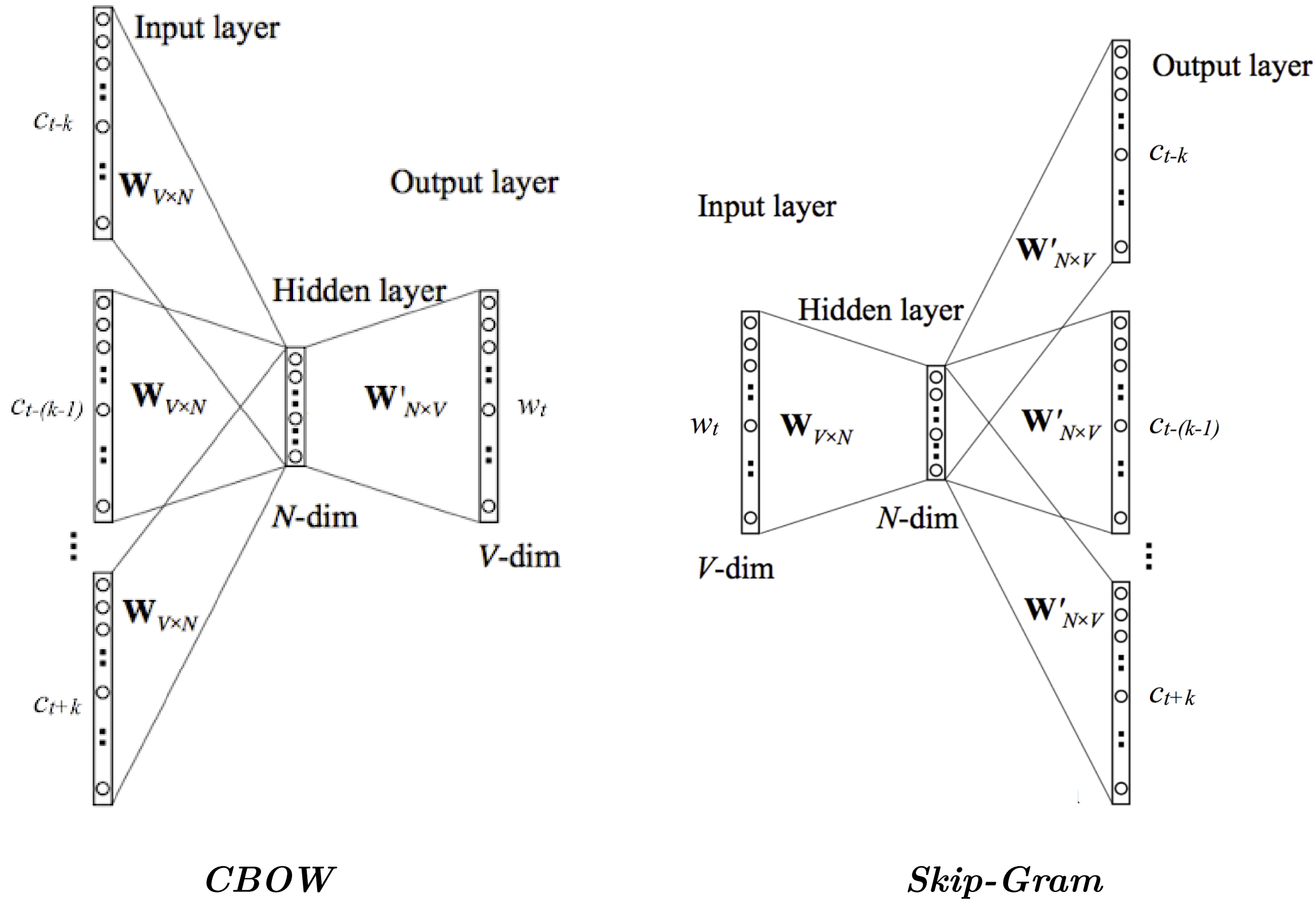

Fig. 7.3 Continuous bag-of-word (CBOW) and Skip-gram (SG) architectures

Figure 7.3 illustrates the CBOW and the SG architectures. While we do not need to delve into details of the training process, it may be necessary to highlight the fundamental training process as well as the core difference between these two main architectures. Looking at Figure 7.3, the CBOW aims to predict a target word $w_0$ given its context $w_{-i}, \ldots, w_{-i} \setminus w_0$ i.e. $i$ words on both sides excluding the target word.

Given the example in Figure 7.4, CBOW produces a model that maximizes the chance of the output *learning* given the set of context words *"an", "efficient", "method", "for", "high", "quality", "distributed", "vector"*. The sequence of occurrence does not really matter, hence "bag-of-words".

SG is basically the reverse of the CBOW in that it aims to predict the most frequently co-occurring set of words to a given target word. If we consider the example in Figure 7.4, with the SG architecture, we will aim to build a model that maximizes the chance of outputting the set *"an", "efficient", "method", "for", "high", "quality", "distributed", "vector"* when the input is *learning*. Incidentally, the word2vec models capture such very interesting semantic representations of words that they could be applied to downstream NLP tasks like analogy and word similarity and we will present our adaptation of these tasks for Igbo language later in this chapter.

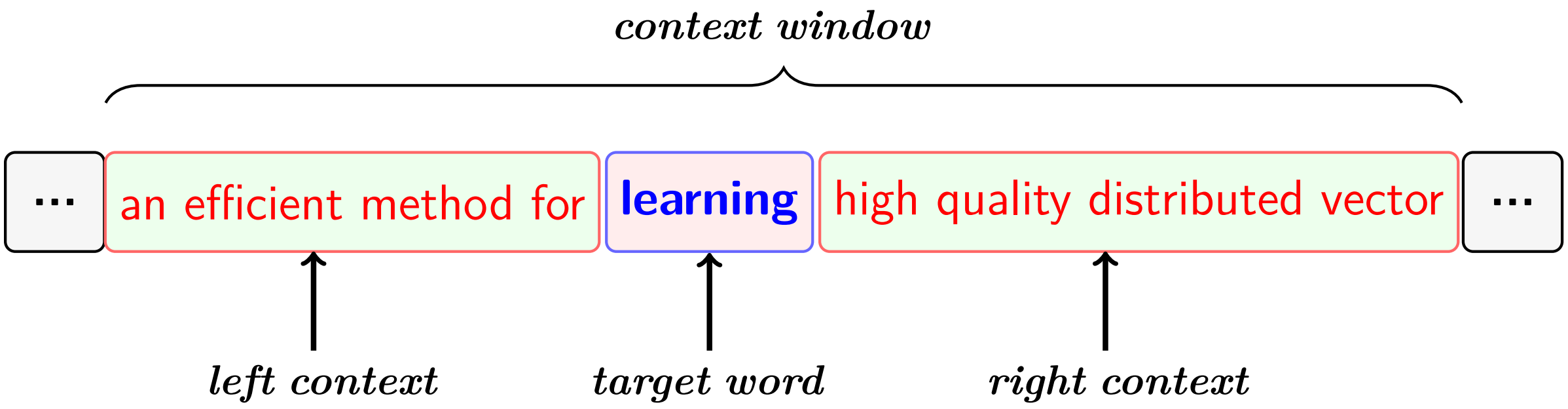

Fig. 7.4 Context words as used in the *word2vec* CBOW and SG architectures [65].

## 7.2 IDR Model Description

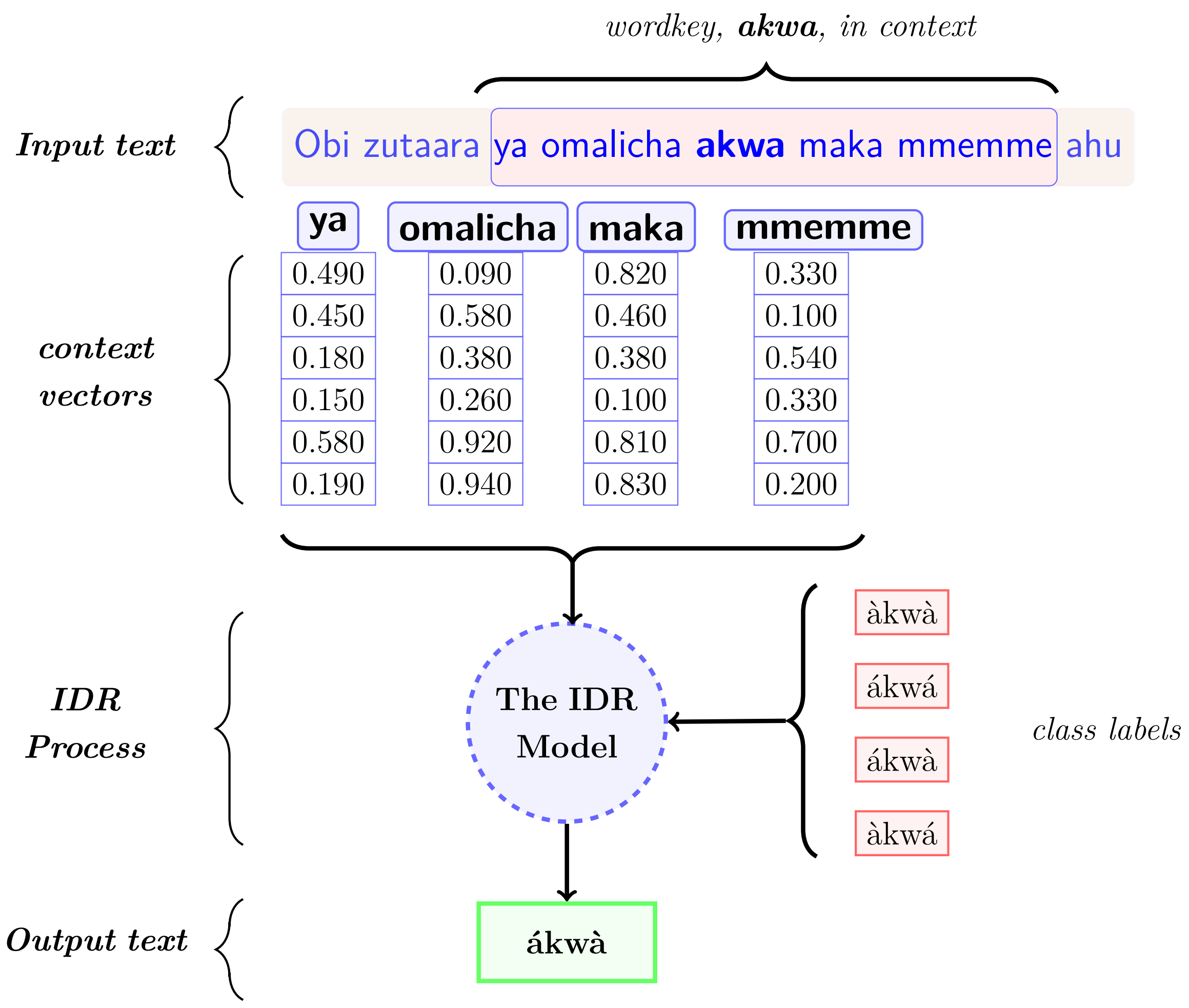

Fig. 7.5 The illustration of a high-level description of the application of the embedding models to the IDR task.

Inspired by the capacity of unsupervised embedding models to capture meaning representations in a universal sense, we decided to consider ways we could apply them to our diacritic restoration task. In a typical scenario, embedding models trained with Igbo language data should not only represent the variants of a wordkey with different vectors but should also use different context words for predicting each of them.

In our proposed approach, we expect that given the wordkey **akwa**, the variants *àkwá* (egg) and *ákwá* (cry) will have different vectors. However, while the former will predict (or be predicted) by such words as *chicken*, *poultry*, *toast* etc, the latter will have *weep*, *sad*, *mourn* etc. as its context words. Figure 7.5 captures our general concept of diacritic restoration using the embeddings of the immediate context words of target *akwa* in sentence: "Obi zutaara *ya ọmaịcha* akwa *maka mmemme* ahụ" (Obi bought [him/her] a lovely dress for the event).

In our overview of embedding models and their architectures in §7.1, we did highlight the fact that, in order to learn meaningful details and build a reliable model for any language, training embedding models requires a huge amount of data. This often runs into hundreds of billions of texts [65, 79, 12]. This is a major challenge for low resource languages, most of which have comparatively little amounts of data available.

Although training embedding models does not require a lot of data pre-processing or annotation, embarking on collecting a sizeable amount of quality Igbo data will drain substantially the resources available for this project. Therefore we have only two options: to train our models on the data we have ($\approx 1m$ tokens) or to find a way to transfer the knowledge from existing trained off-the-shelf word embeddings from a 'big' language, say English, to some 'empty' Igbo embedding space. In practice, we chose to compare both types of models in this experiment.

Figure 7.6 describes a high-level model of our proposed 3-stage pipeline with the following key stages: 1. creating the Igbo word embedding model – this could be either by training or projection from existing models; 2. deriving the "diacritic embedding model" and finally 3. defining the restoration process.

## 7.3 Transferring Embeddings

Before we proceed, we will give a background to the concept of transfer learning as it applies to our task. Transfer learning generally refers to the transfer of knowledge acquired in one domain to solving a problem in another domain. It is commonly applied when the target domain training data is limited [107]. With transfer learning for instance, we could take advantage of parallel data that exists across languages in the form of word-aligned data, sentence-aligned data (e.g. Europarl corpus), document-aligned data (e.g. Wikipedia), lexicon (bilingual or cross-lingual dictionary) or even zero-shot learning with no parallel data.

In a survey of cross-lingual embedding models [87], four different approaches were identified: *monolingual mapping* [66, 35, 42] which trains embeddings on large monolingual corpora and then linearly maps a target language word to its corresponding source language embedding vectors; *pseudo-cross-lingual* [28, 41, 109] which trains embeddings with a pseudo-cross-lingual corpus, i.e mixing contexts from different languages; *cross-lingual* [46, 47, 56] which trains embeddings on a parallel corpus constraining similar words to be close to each other in a shared vector space; *joint*

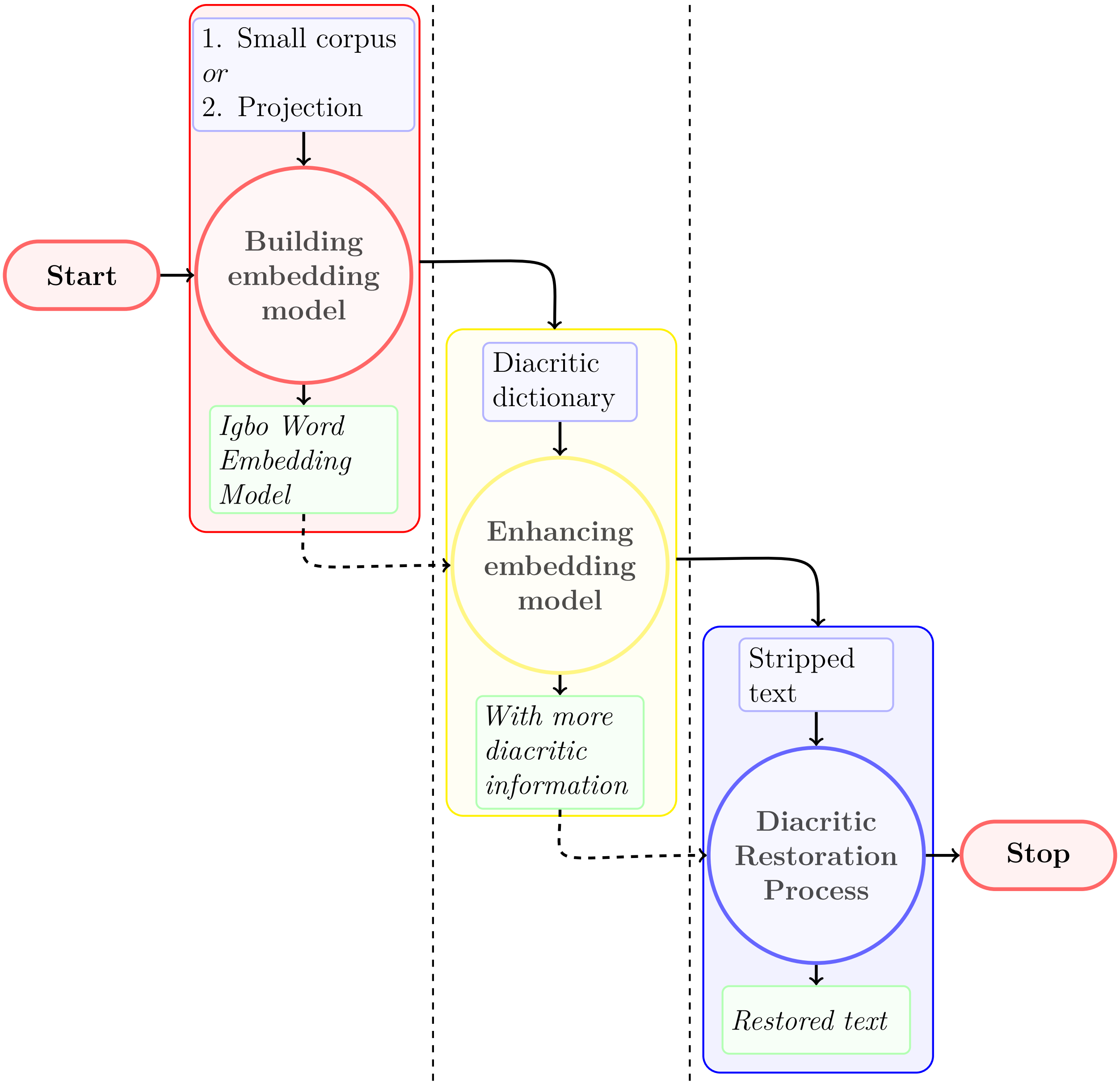

Fig. 7.6 Experimental pipeline for embedding-based IDR

*optimization* [55, 59, 40] which trains models on parallel or monolingual data but jointly optimise a combination of monolingual and cross-lingual losses. In this experiment, we adopted a monolingual mapping approach similar to the one described in [42] and we will elaborate on the process in §7.4.2.

The intuition for transferring word embeddings from one language to another hinges on the universality of the representation of meaning which suggests that objects and ideas have the same meaning across languages. Semantic representation models built for one language may likely capture similar concepts and structures in another language. This concept is at the heart of many research works on multilingual embedding models for NLP [5]. We will present the processes involved in building both the trained and projected (or transferred) embeddings for Igbo in §7.4.2.

## 7.4 Building Embedding Models

As mentioned earlier, we will compare the performances of two categories of embedding models: the *trained* and *projected* models. Also worthy of mention is a training-and-projection task in which we train some small English text from our little collection of Igbo-English bi-text (mainly the Bible and the UDHR document) and also project it into the Igbo embedding space. This section will describe how the training and projection processes are implemented.

### 7.4.1 Model Training

In this work, we compared the performances of both trained and transferred embeddings for supporting the restoration tasks. The Igbo trained embedding **igTnModel** was built with a small Igbo corpus using the *gensim*[2] Python implementation of the *word2vec* algorithm [84]. Again, we used the default set of parameter values for our training. Particularly, we used the default architecture (i.e. *CBOW*) which seems somewhat more related to the "predict-a-variant-given-the-context" approach we adopted in the previous experiments.

The dimensionality parameter, *size* was set to 300 mainly to conform with the dimensions of most off-the-shelf trained embeddings. The window size parameter *window* in *gensim* refers to the number of words on each side of the target word. Therefore we set that to 5 which actually implies a total window length of 11 (including the center word) and is similar to what we used in the previous experiment. Other details on the embedding model produced, **igTnModel**, and its training data are as presented in Table 7.2.

For a broader comparison, we applied a similar method to the one above in training an English embedding model with only the English corpus from our Igbo-English bitext i.e. the English Bible and the UDHR texts. This will later be projected (as described in §7.4.2) into the Igbo embedding space and the model produced will be referred to as **igEnModel**.

[2]Gensim offers an open-source Python library for efficient automatic extraction of semantic topics from data.

### 7.4.2 Model Projection

As shown in Figure 7.7, our aim here is to create the Igbo embedding space by transferring (i.e. literally copying) the embeddings of English words to their Igbo translations. Not all of the translations are one-to-one (some Igbo words translate to multiple English words and vice versa) and that poses another challenge that needs to be resolved.

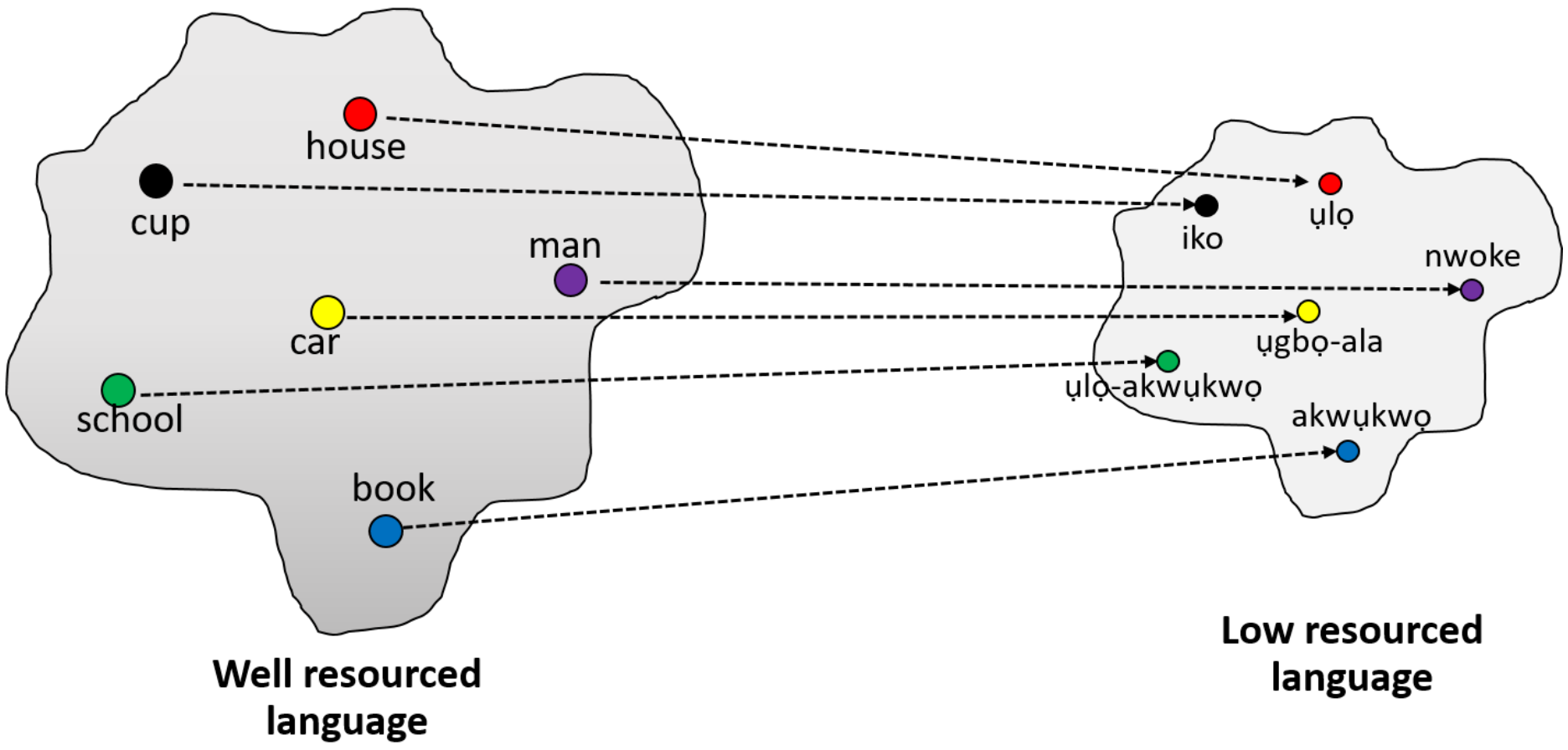

Fig. 7.7 A high-level illustration of embedding model projection

Therefore, in the embedding projection (or transfer) process, we applied an alignment-based method similar to the approach described in [42]. This projection process is preceded by a word-alignment of the Igbo-English bitext from English Bible and the UDHR texts using the *fast_align* tool described in [29]. It works by creating Igbo-English alignment dictionary $\mathbf{A}^{I|E}$ from the aligned texts with structure and examples as shown in Table 7.1. The dictionary basically uses the function $\boldsymbol{f}(w_i^I)$ to map each Igbo word $w_i^I$ to all its most co-aligned English words $w_{i,j}^E$ and their counts $c_{i,j}$ as defined in Equation 7.1. $|V^I|$ is the vocabulary size of Igbo and $n$ is the number of co-aligned English words.

| Igbo word | Co-aligned English words |
|---|---|
| *ákwá* | weeping [94], weep [63], outcry [35], over [25], howl [13], tears [12], cry [9], wept [7], gave [5], crying [5], lamentations [5] |
| *ákwà* | cloth [29], cloths [24], put [19], upon [18], clothing [9], headdress [9], fabric [8], veil [7], garments [5], wrap [5] |
| *àkwá* | eggs [7] |
| *óké* | perfect [38], sound [10], mind [10], soundness [5] |
| *òkè* | share [50], portion [38], allowance [10], lot [9], part [9], portions [6], have [5], shares [5], sharer [5] |
| *ókè* | boundary [129], border [56], and [22], boundaries [8], to [7], extent [7], point [6], excess [5], mark [5] |
| égbé | kite [5] |
| égbè | thunders [18], thunder [10], voices [5] |

Table 7.1 The structure and sample entries in the Igbo-English alignment dictionary used for the embedding projections.
The embedding of the Igbo word is the weighted combination of the embeddings of the co-aligning English words from the model being projected.

$$\begin{aligned} \mathbf{A}^{I|E} &= \{< w_i^I, \boldsymbol{f}(w_i^I) >\}; i = 1..|V^I| \\ \boldsymbol{f}(w_i^I) &= \{< w_{i,j}^E, c_{i,j} >\}; j = 1..n \end{aligned} \quad (7.1)$$

The projection is formalised as assigning the weighted average of the embeddings of the co-aligned English words $w_{i,j}^E$ to the Igbo word embeddings $\mathbf{vec}(w_i^I)$ [42]:

$$\boldsymbol{vec}(w_i^I) \leftarrow \frac{1}{C} \sum_{w_{i,j}^E, c_{i,j} \in f(w_i^I)} \boldsymbol{vec}(w_{i,j}^E) \cdot c_{i,j} \quad (7.2)$$

where $C \leftarrow \sum_{w_{i,j}^E, c_{i,j} \in f(w_i^I)} c_{i,j}$

Using this projection method, we built 5 additional embedding models for Igbo:

- **igEnModel** the model we trained with the English from our bitext collection described in §7.4.1.
- **igGglNews** from the pre-trained *Google News*[3] *word2vec* model.

[3]https://code.google.com/archive/p/word2vec/

- **igWikNews** from *fastText* Wikipedia 2017, UMBC webbase corpus [44] and statmt.org news dataset.
- **igWikSbwd** from same as **igWkNews** but with subword information.
- **igWikCrwl** from *fastText* Common Crawl dataset

| model | vocab_len | dimensions | en_vocab | data |
|---|---|---|---|---|
| *igTnModel* | 4968 | 300 | – | 902.5k |
| *igEnModel* | 4057 | 300 | 6.3k | 881.8k |
| *igGglNews* | 3046 | 300 | 3m | 100bn |
| *igWikNews* | 3460 | 300 | 1m | 16bn |
| *igWikSbwd* | 3460 | 300 | 1m | 16bn |
| *igWikCrwl* | 3510 | 300 | 2m | 600bn |

Table 7.2 Details of the trained and projected Igbo embedding Models:where *vocab_len* = length of the vocabulary; *dimensions* = vector size; *en_vocab* = size of the English model if projected and *data* = size of the training data

### 7.4.3 Intrinsic Model Evaluation

Embedding models are mostly generic products of some unsupervised learning process and often not tied to any downstream NLP tasks. Therefore, there is often the need to assess their usefulness by applying them to some general tasks such as *odd word*, *analogy* and *word similarity*. Having trained or projected the Igbo embedding models (as described in §7.4.1 and §7.4.2 respectively) which we will use for our diacritic restoration experiment, we then subject them to such general tasks as an aside to this experiment. Besides the fact that these tasks are commonly used to intrinsically evaluate embedding models [93], they also apply similar techniques as used in the diacritic restoration task to identify the relatedness (or otherwise) of a target word given its context.

Incidentally, there are no standard test datasets on any of these tasks for Igbo. We therefore embarked on building the datasets for these key tasks by auto-generating test instances from our data and transferring existing ones from English using Igbo native speakers to refine and validate instances of the dataset and methods used. Below is a brief description of each of the tasks and the simple methods used in creating test datasets for Igbo. However, we intend to pursue a more rigorous and extended research on that in our future work.

### Odd Word

In this task, the model is used to identify the *odd word* from a list of words e.g. *breakfast, cereal, dinner, lunch → "cereal"*. We created four simple categories of Igbo words (Table 7.3) that should naturally be mutually exclusive. Test instances were built by randomly selecting and shuffling three words from one category and one from another e.g. *ọkpara, nna, ọgaranya, nwanne → ọgaranya.* For this experiment, we generated 50,000 unique test instances for the odd word experiment each of which is a random combination of three words from one category and one word from another category.

| **Category** | **Igbo words** |
|---|---|
| nouns(family) *e.g. father, mother* | ada, ọkpara, nna, nne, nwanna, nwanne, di, nwunye |
| adjectives *e.g. tall, rich* | ọcha, ọgaranya, ogbenye, ogologo, oji, ọjọọ, okenye, ọma |
| nouns(humans) *e.g. man, woman* | nwaanyị, nwoke, nwata, nwatakịrị, agbọghọ, okorobịa |
| numbers *e.g. one, seven* | otu, abụọ, atọ, anọ, ise, isii, asaa, asatọ, itoolu, iri |

Table 7.3 Word categories for *odd word* dataset. 50,000 unique test instances were generated.

Because the selection of these category words is mostly subjective, we thought right to objectively estimate the measure of human confidence in selecting the odd words from the generated instances. 100 instances were randomly sampled from the entire test instances and their answers were crowd-sourced from 300 most reliable responders[4] on the then *CrowdFlower* (now *Figure Eight*) platform.

Participants were subjected to a 10-question test before the main questions and each question was attempted by at least 3 participants. Over 90% of the 100 sample instances got an aggregate confidence[5] score of 0.5 and above while up to 60% got the maximum confidence score of 1.

[4]Igbo speakers were specifically requested. A total of 784 contributors responded to the pre-task assessment test and the top 300 were chosen.

[5]The confidence score defines each combined result in a *CrowdFlower* (aka *Figure Eight*) job. It measures the agreement level of multiple participants which is also determined by each participant's trust score. It shows the "confidence" in the reliability of the result. The combined score is mostly determined by the response with the highest confidence score. For more on how it is computed, see [6].

Each of the 50,000 instances is a tuple containing a list of four words and the actual odd word (*true_odd*) among them. The evaluation for this task involves passing each list of words to the *doesnt_match* function of the embedding model to get the predicted odd word (*pred_odd*) which is then compared with the actual odd word. This process returns a binary score ($True$ or $False$) for each instance in our dataset. The performance of the model is given by the percentage of the $True$ outcomes that it produced over the entire test instances.

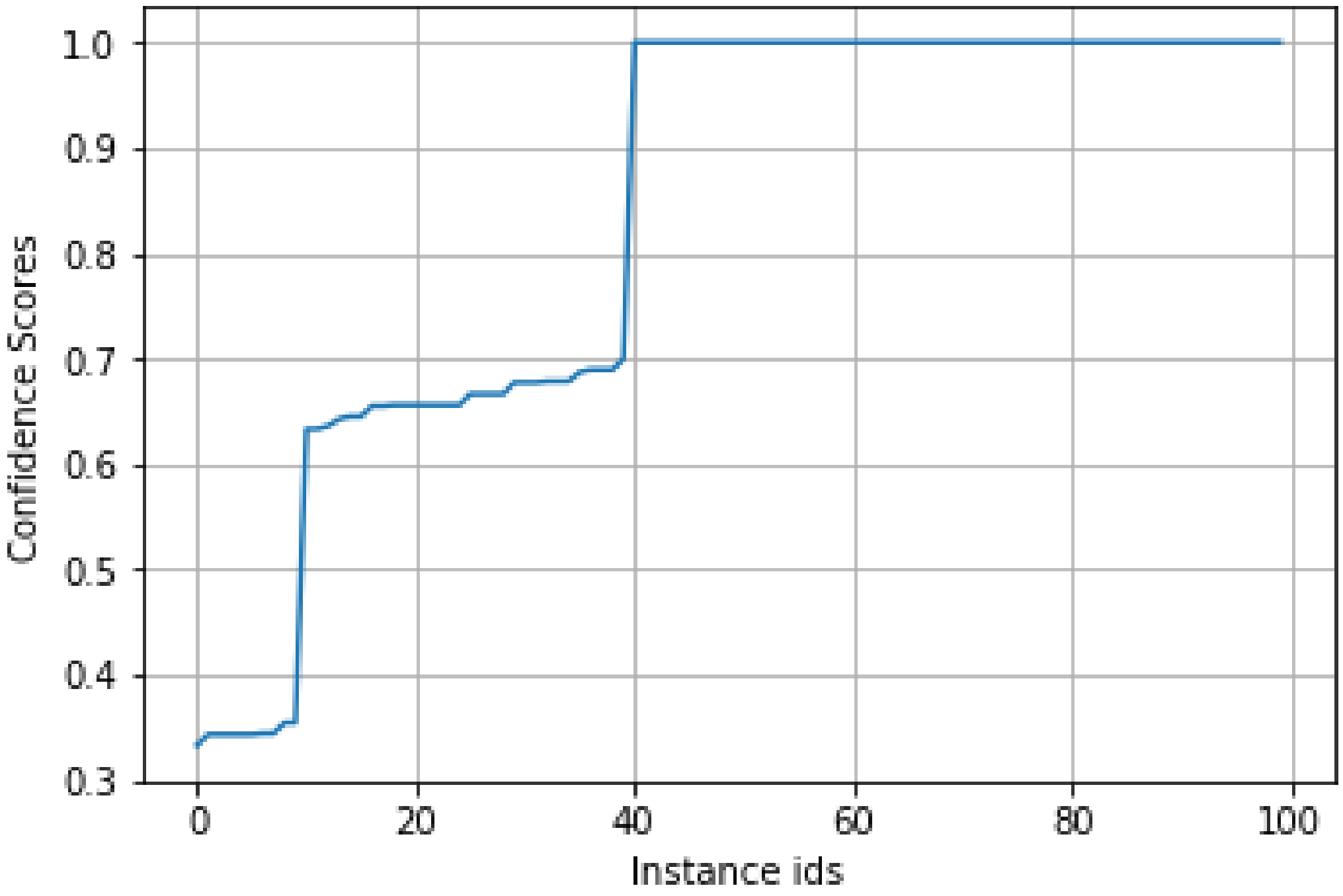

Fig. 7.8 Least-to-Most Confident: Plot of the aggregate crowd-sourced response confidences on each of the 100 sample instances used to validate the *odd word* test data.

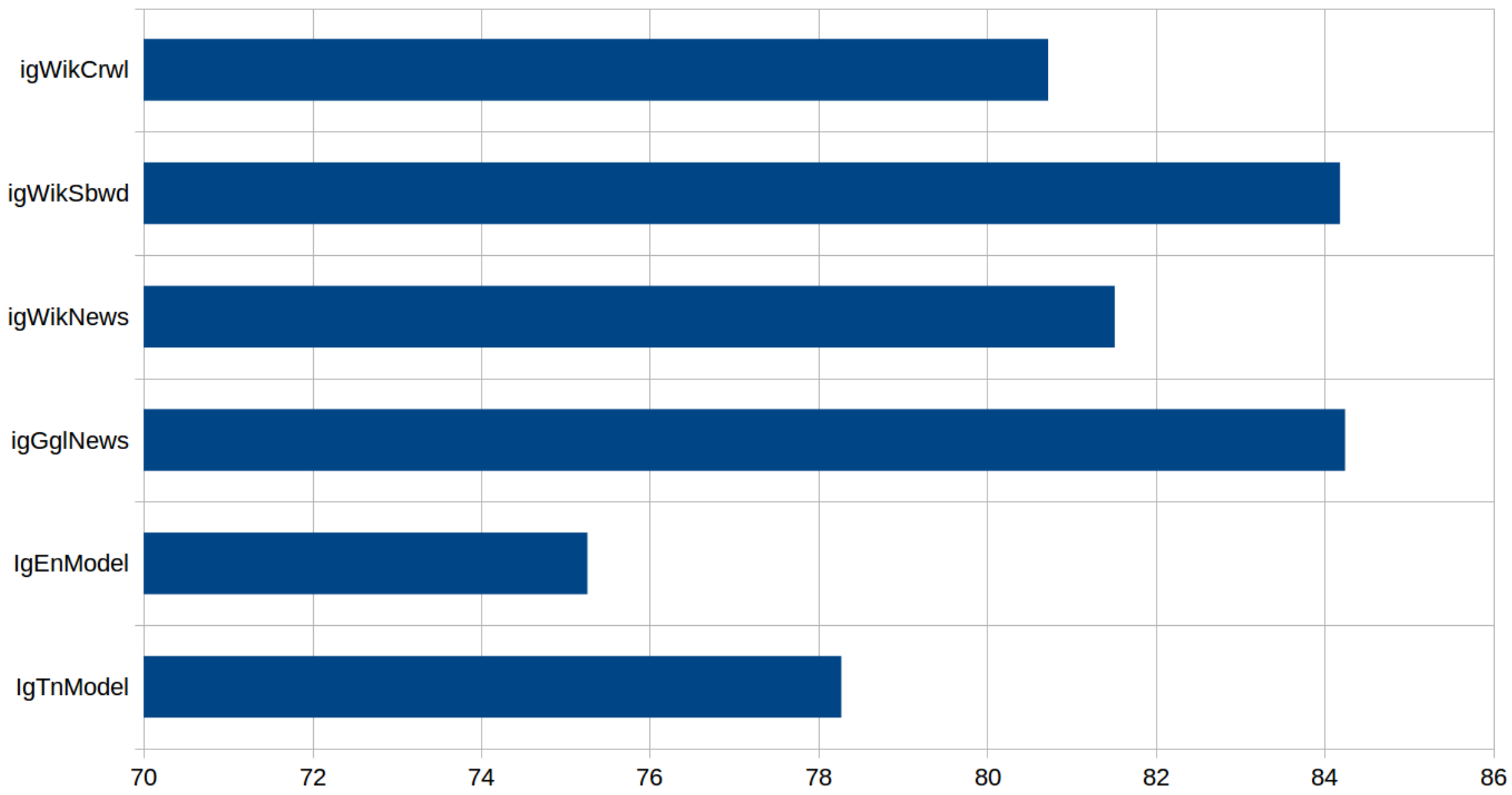

| igTnModel | igEnModel | igGglNews | igWikNews | igWkSbwd | igWkCrwl |
|---|---|---|---|---|---|
| 78.27 | 75.26 | 84.24 | 81.51 | 84.18 | 80.72 |

Fig. 7.9 Plot of the performance of the embedding models on the odd-word task. The results show that the models projected from large trained English models perform better on this task.

### Analogy

This is based on the concept of analogy as defined by [65] which tries to find $y_2$ in the relationship: $x_1 : y_1$ as $x_2 : y_2$ using vector arithmetic e.g $king - man + woman \approx queen$. We created pairs of opposites for some common noun and adjectives (Table 7.4) and randomly combined them to build the analogy data e.g. *di* (husband) – *nwoke* (man) + *nwaanyị* (woman) $\approx$ *nwunye* (wife) ?

| **category** | **opposites** |
|---|---|
| oppos-nouns | nwoke:nwaanyị, di:nwunye, okorobịa:agbọghọ, nna:nne, ọkpara:ada |
| oppos-adjs | agadi:nwata, ọcha:oji, ogologo:mkpụmkpụ, ọgaranya:ogbenye |

Table 7.4 Word pair categories for *analogy* dataset. In total, there are 72 entries or instances generated for this task: 42 entries in the *oppos-nouns* category and 30 entries in the *oppos-adjs* category

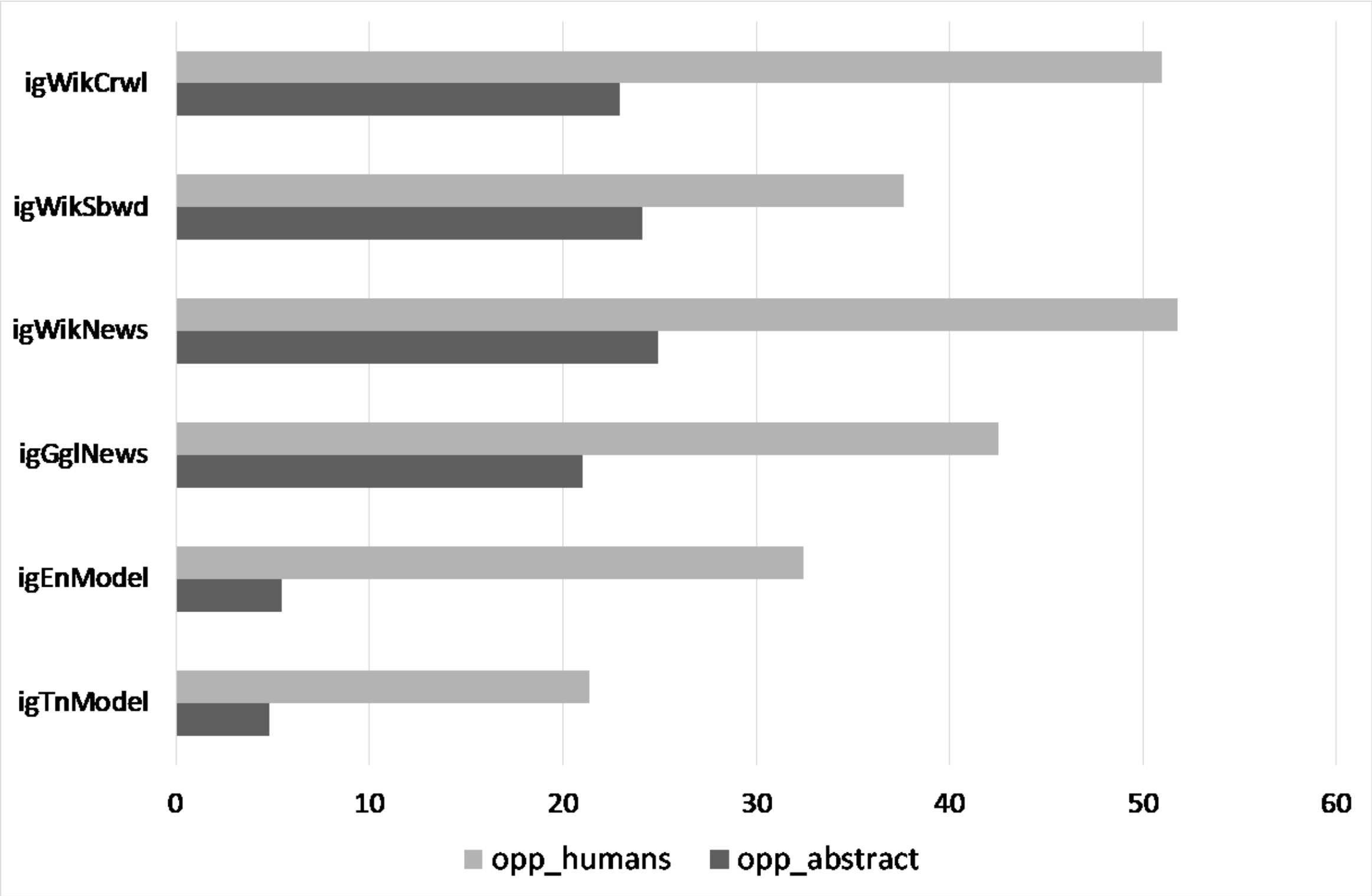

| Models | igTnMdl | igEnMdl | igGglNw | igWikNw | igWkSbd | igWkCrl |
|---|---|---|---|---|---|---|
| **opp-abstract** | 4.83 | 5.44 | 21.03 | 24.91 | 24.14 | 22.97 |
| **opp-humans** | 21.36 | 32.42 | 42.53 | 51.77 | 37.62 | 51.00 |

Fig. 7.10 Plot of the performance of the embedding models on the analogy task. As in Figure 7.9, the result shows that the projected models capture the underlying relationships between words and concepts as contained in our datasets better than the models trained (or projected) from the small Igbo and bitext data.

The evaluation method for this task does not really use the actual similarity scores. So for each analogy question, a number of possible answers, ranked in the order of similarity scores (highest to lowest), are presented by the models. Often the expected answer may not be the topmost in rank and evaluating our models based on that seems harsh. On the other hand, giving a full score to an answer at the bottom of the rank may also be too lenient.

For evaluation, we used the *mean reciprocal ranking* method which gives a score that is equivalent to the inverse of the rank of the matching answer if it is found in the solution list, or zero otherwise. For example, if the correct answer is at the top of the list (i.e. rank 1), we give the full score of 1. If however it's found, say, in position 5, then the score will be 0.2 i.e. $frac15$. If we do not find it in the list, we give 0. For this experiment, we have set the length of the solution list to 100 in order to increase the chance of our model finding the correct answer and the results are as shown in Figure 7.10.

### Word Similarity

Word similarity is the most popular intrinsic evaluation task for embedding models and it typically uses the standard *wordsim353* dataset created by [36]. It computes the correlation of the human assigned similarity scores and the *cosine similarity* scores of the embeddings word pairs in the *wordsim353* dataset. Higher cosine similarity scores indicate better embedding model in terms of the semantic relationships of the word pairs. The cosine similarity of the embeddings of any two words, $a$ and $b$, is defined as follows:

$$\cos(\mathbf{a}, \mathbf{b}) = \frac{\mathbf{a} \cdot \mathbf{b}}{\|\mathbf{a}\| \|\mathbf{b}\|} = \frac{\sum_{i=1}^{n} \mathbf{a}_i \mathbf{b}_i}{\sqrt{\sum_{i=1}^{n} (\mathbf{a}_i)^2} \sqrt{\sum_{i=1}^{n} (\mathbf{b}_i)^2}} \tag{7.3}$$

To build the Igbo word similarity dataset, we transferred the word pairs in the standard *wordsim353* dataset [36] to Igbo by using *Google Translate* to translate the individual word pairs in the combined dataset and return their human similarity scores. The similarity scores are numerical values that range between 0 (words that are totally unrelated) and 10 (words that are very closely related).

Minor corrections were performed by Igbo speakers on the generated datasets. In some cases, we removed instances with words that could not be translated, (e.g. the pair "cell→*cell* & phone→*ekwentị*, 7.81" where the word ***cell*** could not be translated) and those with translations that yield compound words (e.g. the pair "situation→*ọnọdụ* & conclusion→*nkwubi okwu*,4.81" where conclusion produced ***nkwubi okwu***)[7].

For the evaluation of this task, we compute the linear correlation between the human similarity scores which we adapted from the *wordsim353* dataset and the model-predicted similarity scores using the *Pearson correlation coefficient* [8]. As may be observed from the table in Figure 7.11, none of the models contained all the words in our dataset. Therefore, only the pairs that exist in the models were used. Apart from the trained model **igTnModel** that could get 88 word pairs, the projected models contain only 75 each. Our results show that there is a strong correlation ($> 0.5$) between the human scores and the model-predicted scores.

[7] An alternative considered is to combine the word e.g. *nkwubi okwu* → **nkwubi-okwu** and update the model with a projected vector or a combination of the vectors of constituting words.

[8] For more details on this coefficient, see https://en.wikipedia.org/wiki/Pearson_correlation_coefficient

| Word1 | | Word2 | | |
|---|---|---|---|---|
| **English** | **Igbo** | **English** | **Igbo** | **Similarity** |
| announcement | ọkwa | effort | mgbalị | 2.75 |
| investigation | nyocha | effort | mgbalị | 4.59 |
| money | ego | bank | ụlọ-akụ | 8.50 |
| love | ịhụnanya | sex | mmekọahụ | 6.77 |
| tiger | agụ | cat | pusi | 7.35 |
| plane | ụgbọ-elu | car | ụgbọ-ala | 5.77 |
| train | ụgbọ-oloko | car | ụgbọ-ala | 6.31 |
| television | telivishọn | radio | redio | 6.77 |
| tiger | agụ | animal | anụmanụ | 7.00 |
| psychology | akparamagwa | mind | uche | 7.69 |
| planet | ụwa | moon | ọnwa | 8.08 |
| news | akụkọ | report | akụkọ | 8.16 |
| canyon | kaniyon | landscape | ọdịda-obodo | 7.53 |
| image | oyiyi | surface | elu | 4.56 |
| discovery | nchọpụta | space | ohere | 6.34 |
| stock | ngwaahịa | market | ahịa | 8.08 |
| stock | ngwaahịa | egg | àkwá | 1.81 |
| fertility | ọmụmụ | egg | àkwá | 6.69 |
| life | ndụ | term | okwu | 4.50 |
| governor | gọvanọ | interview | nyocha | 3.25 |

Table 7.5 **igbwordsim164**: Examples of Igbo word pairs and their similarity scores as adapted from the *wordsim353* dataset [36]. Only 164 entries were extracted using our simple approach.

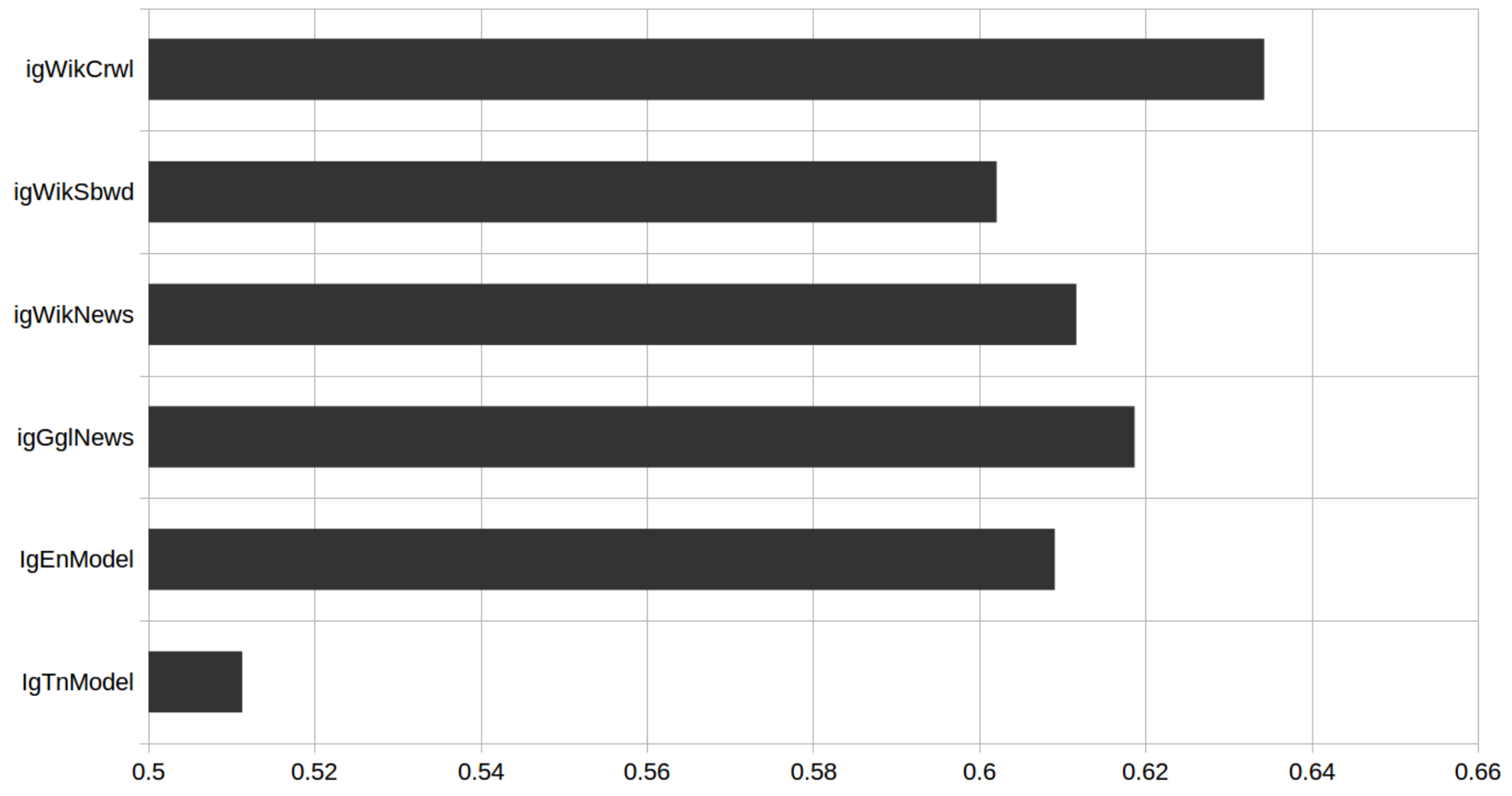

| | **igTnMdl** | **igEnMdl** | **igGglNw** | **igWkNw** | **igWkSbd** | **igWkCrl** |
|---|---|---|---|---|---|---|
| *Rem Pairs* | 88 | 75 | 75 | 75 | 75 | 75 |
| *Correlation* | 0.5113 | 0.6091 | 0.6187 | 0.6117 | 0.6021 | 0.6343 |

Fig. 7.11 Plot of the performance of the embedding models on the word similarity task. All models produced scores that have strong correlation with human scores. However, the Igbo bible trained model is still lagging behind on the projected models on this task

## 7.5 The Diacritic Restoration Process

Having built the embedding models and performed some intrinsic evaluations on them, we proceed to our focus task: using them for diacritic restoration. This restoration process, as we will present it, may (or may not) involve a stage we earlier defined as the *model enhancement* (see Figure 7.6) which will be further discussed in this section.

The restoration process simply computes the cosine similarity (see equation 7.3 above) of the variant and context vectors and selects the most "similar" candidate i.e. the one with the highest similarity score. For each wordkey, $wk$, candidate vectors, $D^{wk} = \{d_1, ..., d_n\}$, are extracted from the embedding model on-the-fly. $C$ is defined as the context words (i.e. words within a given window or all the words in the sentence) and $vec_C$ is the context vector of $C$ (Equation (7.4)).

$$\mathbf{vec_C} \leftarrow \frac{1}{|C|} \sum_{w \in C} vec_w \tag{7.4}$$

$$\mathbf{res_{wk}} \leftarrow \underset{d_i \in D^{wk}}{\mathbf{argmax}} sim(\mathbf{vec_C}, \mathbf{vec_{d_i}}) \tag{7.5}$$

Figures 7.12 shows a high-level description of the steps involved in the actual restoration process with with an original embedding model (i.e. without any form of enhancement as described in §7.5.1) for each of the instances of a given wordkey.

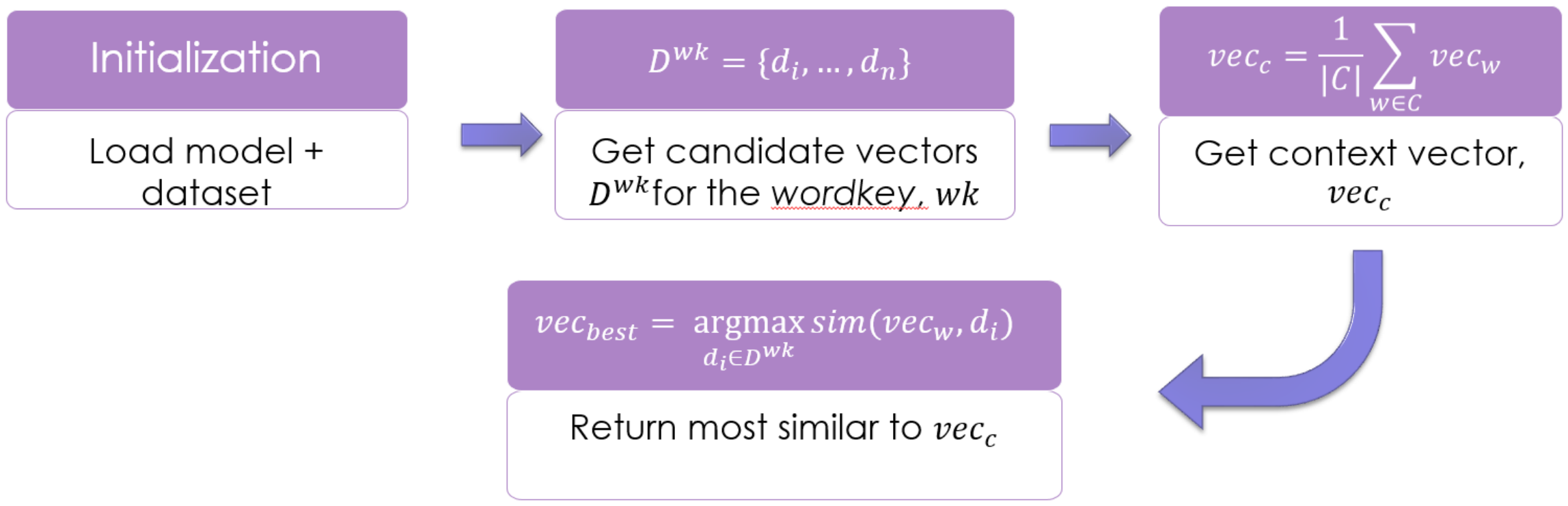

Fig. 7.12 Embedding-based diacritic restoration *without* model enhancement

We could have used the embedding models described above *as is* for the restoration task and got some decent results. We refer to this as the *basic* approach. However, we considered (and indeed implemented) other approaches that could possibly "enhance" the models in some way, thereby improving our results. These enhancement methods and their motivations are discussed in §7.5.1.

### 7.5.1 Enhancing Embedding Models

The idea of "enhancement"[9], as we originally thought of it, basically puts an extra layer on the process to ensure that the embedding of each variant is nudged as close as possible to those of its top $n$ most co-occurring words. This is expected to be more beneficial to the projected models whose original embeddings were not learned with the Igbo language structure.

In practical terms, the enhancement of a model is achieved by first identifying the subset of its top-$n$ co-occurring words[10] of each variant that do not appear in the contexts of the other variants. As shown in Figure 7.13, only the embeddings of the words in the green area will be used to update the embedding of the variant *ákwá*.

[9]We originally referred to the process as *deriving diacritic embedding* but "enhancement" was suggested by a reviewer of one of our papers [34].

[10]$n$ is arbitrarily chosen and could be optimized but for this experiment, we used $n = 50$

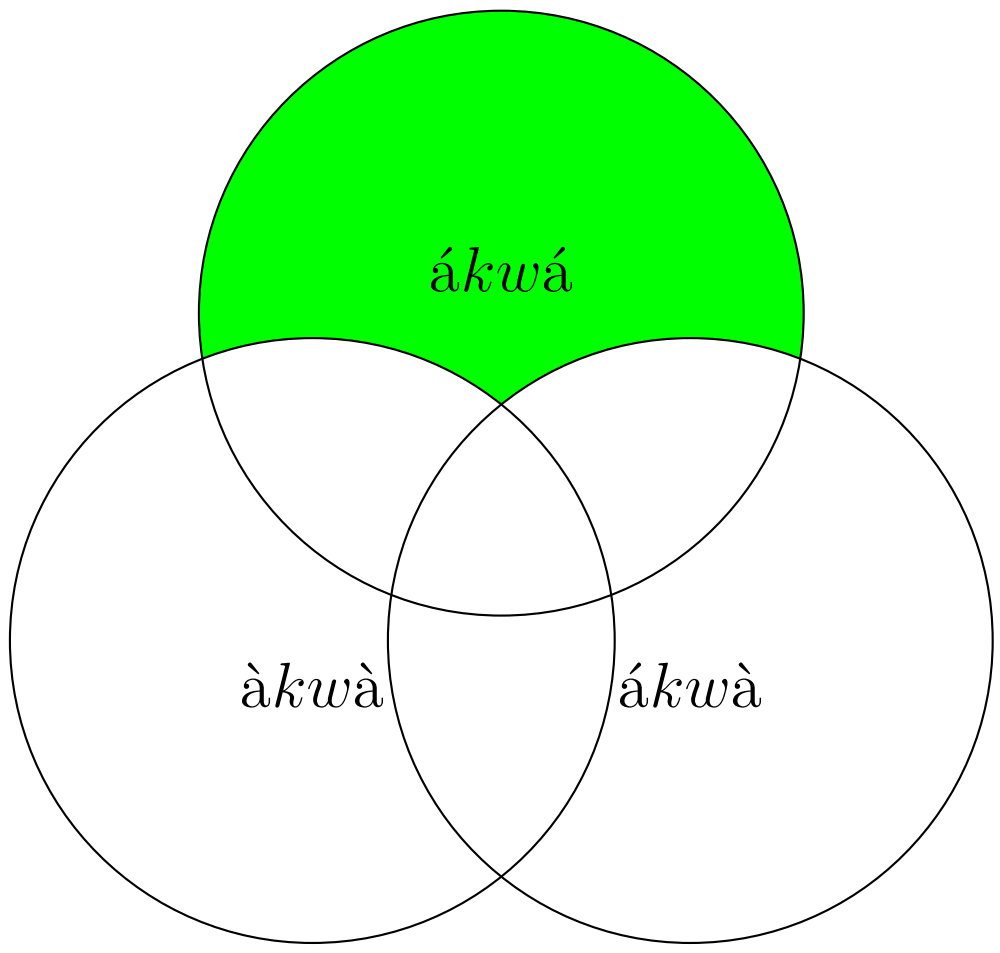

Fig. 7.13 Diagram showing the interactions of the sets of the co-occurring words for each of the variants of the wordkey *akwa*. The weighted combination of the words in the green area is used to update the variant *ákwá*.

This update could be a combination of the existing variant embedding and those of the co-occurring words or an outright replacement by those co-occurring embeddings.

Ultimately a modified version of the original model is created which is the same with the original at every point except points representing the variants of all the wordkeys that have been adjusted to become a sort of centroid for its unique co-occurring words. Figure 7.14 shows a slight modification of the original process to integrate the enhancement process.

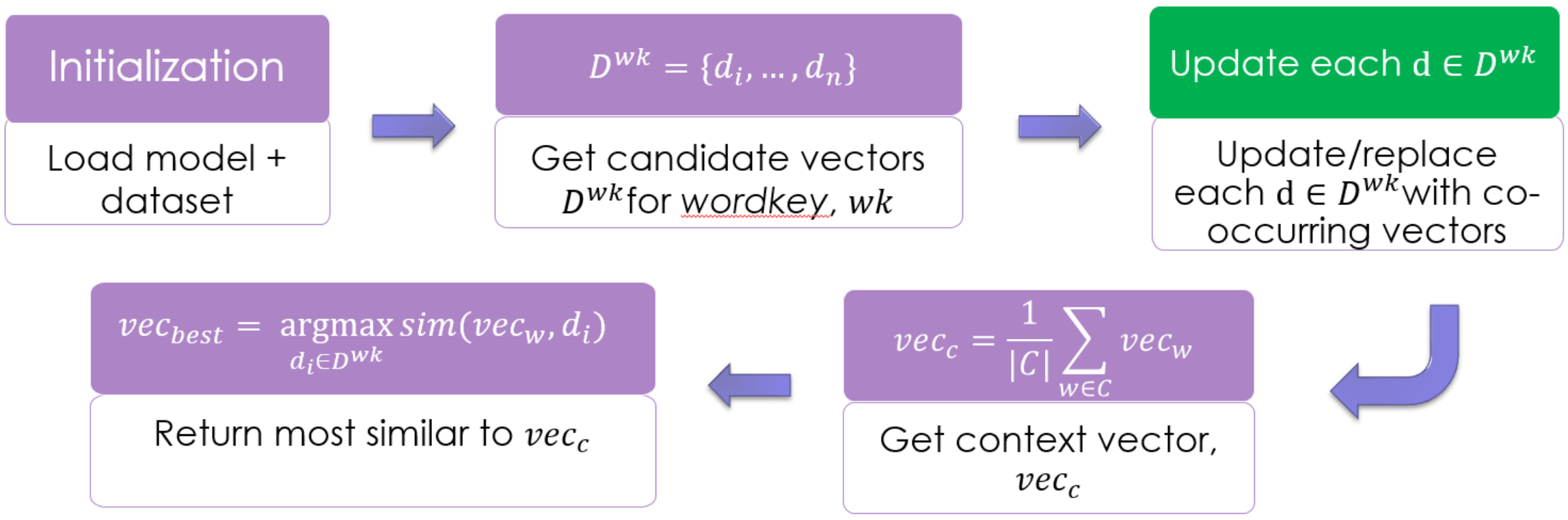

Fig. 7.14 Embedding-based diacritic restoration *with* model enhancement

For our experiments, we compared the restoration processes with and without the enhancement strategy described in §7.5.1.

### 7.5.2 The *Basic* Restoring Diacritics

The restoration process of an instance, with or without model enhancement, involves the last two stages shown in each of Figures 7.12 and 7.14 i.e computing the context vector by taking the average of the vectors of the words in context, and returning the variant whose vector that is the "most similar" to the context vector. Again, we applied the cosine similarity measure here because it captures better the relationship between the vectors by considering how the individual dimensions compare.

Now recall that in both the training and projection methods of building the embedding model, vectors are already assigned to each word in our working[11] dictionary, and that includes each diacritic variant of a wordkey. The process of using the model, trained or projected, as it is without any enhancement is what we refer to as the *basic* restoration process.

### 7.5.3 Restoration with Enhancements (or *Tweaks*)

Having explained the concepts of "enhancement" (§7.5.1) and "restoration" (§7.5.1) as well as how the *basic* restoration process (§7.5.2) works, this section will present three different schemes that apply some enhancement strategies (or tweaks) to the restoration process:

**Tweak1:** During enhancement, this scheme "updates" (i.e. performs vector addition on a 50:50 basis) each initial diacritic variant vector with the weighted average of the vectors of its most co-occurring words (or *cowords*). The counts of these cowords are computed from the data. At restoration time, *all* the words appearing in context are used to build the context vector.

**Tweak2:** This scheme uses the same enhancement method i.e updating each variant with the weighted sum of its coword vectors. However in this restoration process, each variant vector is compared with its own context vector which is built with the vectors of *only* the words in its coword set that appear in context. Common words that generally appear everywhere are excluded.

**Tweak3:** The enhancement method used in this scheme *replaces* (not updates) each of the variant vectors with the weighted sum of its coword vectors. In principle, it throws away whatever the model originally knows about each variant and positions it in the embedding space as a centroid of its cowords. The restoration process in this scheme is similar to that of **Tweak2** i.e. for each variant, it uses only the words in context that uniquely co-occur with the variant.

[11] The content and size of the dictionary depends on the approach used. For example, the trained Igbo model **IgTnModel** contains more entries than the ones projected from other English models. This is mainly because a lot of the Bible terms and names simply do not exist in those models

| Models | Basic | Tweak1 | Tweak2 | Tweak3 |
|---|---|---|---|---|
| igTnModel | 64.50\|62.24 | 71.24\|66.58 | 69.68\|63.98 | 69.69\|63.96 |
| igEnModel | 57.90\|57.99 | 62.20\|62.12 | 60.68\|58.99 | 61.07\|59.23 |
| igGglNews | 59.26\|59.10 | 63.96\|62.96 | 63.24\|61.53 | 63.36\|61.56 |
| igWikNews | 57.92\|55.79 | 61.46\|59.80 | 61.09\|60.00 | 61.31\|59.99 |
| igWikSbwd | 60.52\|58.00 | 59.97\|58.06 | 60.42\|60.75 | 60.59\|60.67 |
| igWikCrwl | 59.53\|56.75 | 59.90\|58.38 | 60.43\|61.40 | 60.54\|61.31 |
| Average | **59.94\|58.31** | **63.12\|61.32** | **62.59\|61.11** | **62.76\|61.12** |

Table 7.6 Comparison of using a scoped context (i.e. with the window size of 11) and using the entire sentence. The entry in the table is the corresponding pair (i.e. *scoped;not-scoped*) of scores for the model and the scheme in focus

## 7.6 Evaluation of Results

For the evaluation, we compared performances of all the models on all of the previously defined metrics while applying the different enhancement and restoration schemes. The scores, as defined in the previous experiments, are the weighted-average scores across all the wordkeys. The preliminary analysis in the following sections intends to show the impact of the different context window sizes on the average scores across the models used in the experiments and the enhancement methods applied on them.

### 7.6.1 Effect of context window sizes

Clearly the choice of the window size used in the experiment has an impact on the overall performance of not just the models but also the enhancement methods (or tweaks) used. Figures fig. 7.15 and fig. 7.17 show that the best average scores are achieved between the window sizes of 9 and 11 (i.e. 4 or 5 words on both sides of the wordkey). Both window sizes were going head to head across the methods as shown in Figure fig. 7.15. It is also easy to see that the enhancement methods significantly improved the model performances across the different window sizes. Table 7.6 compares pairs of results with scoped and non-scoped contexts by all models and enhancements methods.

Tracing the same average scores across the models, we could see that the overall best average scores are achieved by window sizes 9 and 11 as indicated earlier. Figure 7.17 shows that the performance of **igTnModel**, which happens to be the best performing model, peaks at around the window size of 7. Besides the **igTnModel**, it is interesting to observe that the **igGglNews** did consistently better than the other projected models. So given the insight we got from this section, we shall be presenting the results of the experiment with only the window size of 11 i.e. 5 words on both sides of the wordkey.

### 7.6.2 Effect of enhancement methods (or *Tweaks*)

Figure 7.18 shows the accuracy scores of all the models tested as well as the different enhancement techniques. The figure indicates that in nearly all cases, the enhancements improved the performance of the models. Overall, all the enhanced models achieved clear improvements from the basic model but the *Tweak1* did slightly better than the others. This scheme updates the initial variant vectors with the weighted average of the vectors of its most frequent co-occurring words *in the corpus* while using *all* the words in the sentence containing the wordkey to be restored to build the context vector.

### 7.6.3 Analysis: Accuracy

As with the previous experiments, our evaluation and comparison of the embedding models is based on the weighted-average of the performances of both trained and projected models on individual wordkeys. In this section, we present the raw accuracy scores obtained by the models as well as comparative analysis of the models among themselves and with reference to the baseline unigram model. We note however that while building word embeddings is a very useful approach to modelling semantic relationships and concepts in languages, embedding models as applied to the Igbo diacritic restoration task did not compare well with previous techniques.

The raw accuracy scores of all the 6 embedding models tested in this experiments as well as the baseline unigram model are shown in Table 7.7. The table shows that besides the **igTnModel** i.e. the model trained with the small amount of Igbo data, no other model could beat the unigram baseline.

We also showed the counts of the number of times each model got the best score as indicated in column *Best Score* of Table 7.7 and on the graph shown in Figure 7.19.

| *Wordkey* | *Counts* | *No of Variants* | 1-gram | igWikCrwl | igWikSbwd | igWikNews | igEnModel | igGglNews | igTnModel | *BestScore (BS)* | *Best Model* | *Wdkey Improvement* | *Baseline Improvement* | *Error Reduction* |
|---|---|---|---|---|---|---|---|---|---|---|---|---|---|---|
| **ukwu** | 1432 | 3 | 0.5223 | 0.4001 | 0.3939 | 0.3980 | 0.3799 | 0.4106 | 0.5405 | 0.5405 | igTnModel | 1.82% | -12.70% | 3.81% |
| **ama** | 1353 | 3 | 0.4213 | 0.3001 | 0.3126 | 0.3237 | 0.4191 | 0.4309 | 0.5669 | 0.5669 | igTnModel | 14.56% | -10.06% | 25.16% |
| **akwa** | 1191 | 3 | 0.4232 | 0.4828 | 0.3610 | 0.4165 | 0.5382 | 0.4333 | 0.5861 | 0.5861 | igTnModel | 16.29% | -8.15% | 28.24% |
| **aku** | 384 | 3 | 0.4896 | 0.4115 | 0.3412 | 0.3125 | 0.4453 | 0.3125 | 0.5990 | 0.5990 | igTnModel | 10.94% | -6.86% | 21.43% |
| **agbago** | 99 | 2 | 0.5354 | 0.6263 | 0.5152 | 0.6162 | 0.5455 | 0.5960 | 0.6263 | 0.6263 | igWikCrwl | 9.09% | -4.13% | 19.56% |
| **onya** | 160 | 3 | 0.5312 | 0.5875 | 0.5625 | 0.5750 | 0.6125 | 0.5500 | 0.6313 | 0.6313 | igTnModel | 10.01% | -3.63% | 21.34% |
| **ibu** | 682 | 2 | 0.6422 | 0.5689 | 0.5425 | 0.6202 | 0.6305 | 0.6217 | 0.5938 | 0.6422 | 1-gram | 0.00% | -2.53% | 0.00% |
| **bu** | 16999 | 2 | 0.6441 | 0.4298 | 0.4807 | 0.5299 | 0.6160 | 0.6463 | 0.6536 | 0.6536 | igTnModel | 0.95% | -1.40% | 2.66% |
| **juru** | 306 | 2 | 0.5359 | 0.6078 | 0.5654 | 0.5784 | 0.5882 | 0.6209 | 0.6569 | 0.6569 | igTnModel | 12.10% | -1.07% | 26.06% |
| **otu** | 5947 | 2 | 0.6664 | 0.3851 | 0.4132 | 0.3723 | 0.6232 | 0.3518 | 0.6538 | 0.6664 | 1-gram | 0.00% | -0.11% | 0.00% |
| **i** | 5347 | 2 | 0.6738 | 0.6374 | 0.5147 | 0.6385 | 0.4868 | 0.6587 | 0.6843 | 0.6843 | igTnModel | 1.05% | 1.68% | 3.22% |
| **nku** | 285 | 2 | 0.6140 | 0.6526 | 0.6456 | 0.6386 | 0.6983 | 0.5579 | 0.6632 | 0.6983 | igEnModel | 8.43% | 3.07% | 21.83% |
| **si** | 9039 | 2 | 0.5733 | 0.6218 | 0.6503 | 0.5954 | 0.6683 | 0.6000 | 0.7172 | 0.7172 | igTnModel | 14.39% | 4.97% | 33.73% |
| **igwe** | 1392 | 4 | 0.6609 | 0.7012 | 0.6444 | 0.7220 | 0.4799 | 0.6020 | 0.5589 | 0.7220 | igWikNews | 6.11% | 5.44% | 18.01% |
| **ikpo** | 133 | 2 | 0.6165 | 0.6692 | 0.7143 | 0.7368 | 0.7143 | 0.7143 | 0.7143 | 0.7368 | igWikNews | 12.03% | 6.93% | 31.38% |
| **buuru** | 180 | 2 | 0.6667 | 0.6889 | 0.6111 | 0.6278 | 0.7556 | 0.6889 | 0.7333 | 0.7556 | igEnModel | 8.89% | 8.80% | 26.66% |
| **iso** | 201 | 2 | 0.5025 | 0.5871 | 0.5025 | 0.5025 | 0.7512 | 0.5025 | 0.7612 | 0.7612 | igTnModel | 25.87% | 9.37% | 52.00% |
| **doo** | 120 | 2 | 0.7000 | 0.6500 | 0.3833 | 0.5667 | 0.7333 | 0.3000 | 0.7667 | 0.7667 | igTnModel | 6.67% | 9.91% | 22.22% |
| **ruru** | 488 | 2 | 0.5041 | 0.6066 | 0.6680 | 0.6742 | 0.7152 | 0.6291 | 0.7705 | 0.7705 | igTnModel | 26.64% | 10.30% | 53.72% |
| **okpukpu** | 211 | 2 | 0.7109 | 0.4028 | 0.4171 | 0.4218 | 0.7299 | 0.4455 | 0.7725 | 0.7725 | igTnModel | 6.16% | 10.50% | 21.31% |
| **too** | 125 | 2 | 0.7120 | 0.7520 | 0.7280 | 0.7440 | 0.7360 | 0.7760 | 0.7680 | 0.7760 | igGglNews | 6.40% | 10.85% | 22.22% |
| **o** | 31446 | 2 | 0.7353 | 0.7251 | 0.7248 | 0.7239 | 0.6565 | 0.7218 | 0.7765 | 0.7765 | igTnModel | 4.12% | 10.90% | 15.57% |
| **oke** | 2267 | 3 | 0.7821 | 0.7512 | 0.6956 | 0.7503 | 0.5990 | 0.7843 | 0.7931 | 0.7931 | igTnModel | 1.10% | 12.56% | 5.06% |
| **ju** | 97 | 2 | 0.7423 | 0.7423 | 0.7423 | 0.7423 | 0.7423 | 0.7423 | 0.7938 | 0.7938 | igTnModel | 5.15% | 12.63% | 19.99% |
| **doro** | 205 | 2 | 0.6390 | 0.7317 | 0.5854 | 0.7366 | 0.7805 | 0.7073 | 0.8293 | 0.8293 | igTnModel | 19.03% | 16.17% | 52.71% |
| **inu** | 156 | 2 | 0.5769 | 0.7244 | 0.7051 | 0.7051 | 0.7885 | 0.7115 | 0.8462 | 0.8462 | igTnModel | 26.93% | 17.86% | 63.64% |
| **wuru** | 112 | 2 | 0.6696 | 0.6161 | 0.5446 | 0.5446 | 0.7143 | 0.5179 | 0.8839 | 0.8839 | igTnModel | 21.43% | 21.64% | 64.87% |
| **iru** | 333 | 2 | 0.5315 | 0.8198 | 0.6457 | 0.7267 | 0.7808 | 0.8619 | 0.9099 | 0.9099 | igTnModel | 37.84% | 24.24% | 80.77% |
| **odo** | 154 | 2 | 0.7273 | 0.8701 | 0.6948 | 0.7273 | 0.9416 | 0.7403 | 0.9416 | 0.9416 | igEnModel | 21.43% | 27.40% | 78.57% |
| **Baseline = 1gram; %Error = 33.25%** | | | **66.75%** | **59.90%** | **59.97%** | **61.46%** | **62.20%** | **63.96%** | **71.24%** | **71.68%** | **Performance Analysis** | Model | Improvement | Error Reduction |
| **Best Model Counts** | | | **2** | **1** | **0** | **2** | **3** | **1** | **20** | | *Best Score* | ***igTnModel*** | **4.49%** | **13.50%** |
| **Model Error Reduction:** | | | 0.00% | -20.61% | -20.40% | -15.92% | -13.71% | -8.41% | 13.50% | 14.81% | *Best Counts* | ***igTnModel*** | **4.49%** | **13.50%** |

Table 7.7 **Emb Accuracy:** Table showing the raw accuracy scores of all the 6 embedding models and compared with the baseline unigram model. These scores are obtained using the context window size of 11 with the *Tweak1* enhancement method. [*NB:* The color code indicates the worst(red)-to-best(green) based on the accuracy performances and improvements on wordkey and global baselines].

The **igTnModel** being the best model as well as the only model that beat the baseline, got the best score on 20 out of 29 wordkeys (i.e 69%) as shown in Figures 7.20 and 7.21 while **igWikSbwd** did not get the best score on any wordkey.

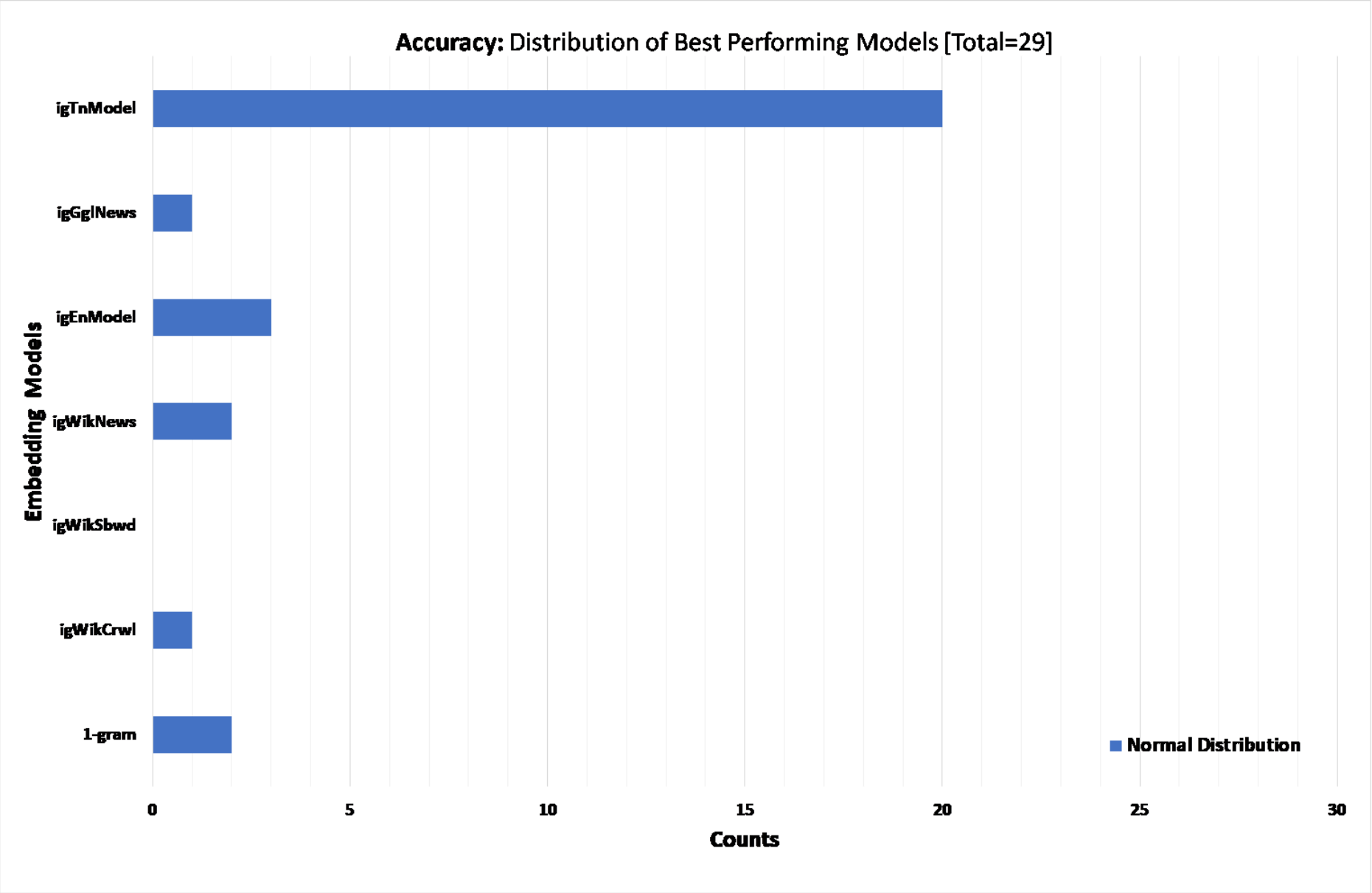

Fig. 7.20 **Emb Accuracy:** Graph comparing the frequency of getting the best score on wordkeys by each of the models.

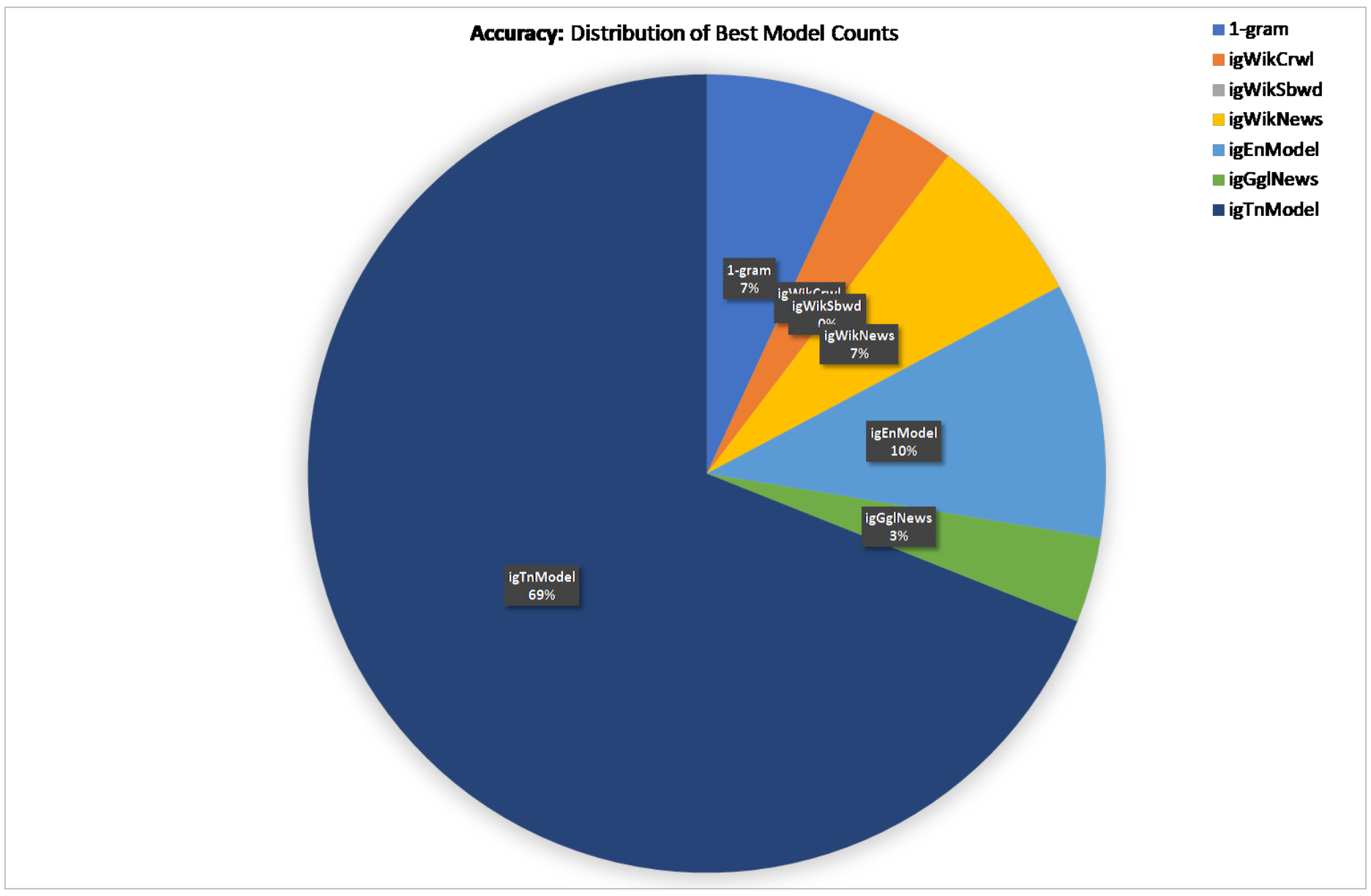

Fig. 7.21 **Emb Accuracy:** Pie-chart showing the relative percentage distribution of the *best score* frequencies shown in Figure 7.20 across the models.

The plot of the error reduction as shown in Figure 7.22 looks interesting as almost all the embedding models increased the error apart from **igTnModel**.

On the wordkey-level improvements, Figure 7.23 shows that while a few wordkeys achieved scores below the global baseline, almost all of them improved their individual baselines.

### 7.6.4 Analysis: Precision, Recall and F1

In this section we present along with the accuracy scores, the precision, recall and F1 scores of the embedding models in Figure 7.24. Clearly, **igTnModel** got the best scores across the metrics. **igGglNews** achieved the next best accuracy score but the did not do as well with the other metrics.

### 7.6.5 Analysis: Result Summary

The summary of the performances of the best models from the different experiments – **5-gram**, **LRCV** and **igTnModel** – is presented in Figure 7.25. Also a trend of the worst to best performing wordkeys for these best is shown in Figure 7.26.

## 7.7 Chapter Summary

As a quick recap, in chapter 5, we defined the baseline for the diacritic restoration task using the unigram model and applied higher n-gram models up to 5-gram to the task. The best score was achieved with the 5-gram beyond which there was no further performance improvement. In Chapter 6, we compared 12 models built with different machine learning algorithms and then presented a comparison of the scores of the best 3 models – LSVC, LRCV, SGDC – with those of the 5-gram scores.

In this chapter, we introduced word embedding models and developed schemes to apply them to the diacritic restoration task. On the positive side, word embedding models are good in capturing the relationships between words in a very simplistic manner and they are relatively easy to train using unsupervised learning algorithms. They are also generic and so off-the-shelf word embedding models can easily be adapted to most NLP tasks.

For our work with Igbo language there is a fundamental downside. Training a word embedding model with our little Igbo data (barely a million words) is obviously an option, and we did that. But to get the most from these models, one needs to train them with a huge amount (often in billions of words) of data. The available models are mostly trained with the English language text and so could not be directly applied to our task in Igbo language. So we had to devise some transfer learning techniques to project some of the existing English word embedding models to some kind of Igbo embedding space.

Additionally, we also created datasets and conducted experiments to intrinsically evaluate the models we built on three key tasks: *odd word*, *analogy* and *word similarity*. The results, as reported in §7.4.3, highlight the fact that not only did the projected models do very well on those tasks, but also that transfer learning techniques could be very useful in building resources for Igbo. Figures 7.9, 7.10 and 7.11 show that the projected models are doing better with these analogy tasks. We think that lack of adequate data may have affected the ability of the trained models to better separate the terms in each set. The projected models, on the other hand, benefited from a large amount of training data which makes it possible for them to capture more accurate feature values for each term thereby getting a better similarity score in comparison.

On our core task of diacritic restoration, the embedding models – both trained and projected – did fairly well but performed worse than the ngram and machine learning models. In our process, proper diacritic variants are selected using a simple restoration function that compares the aggregated embeddings of the context words of an instance with each variant of a given wordkey. The basic cosine similarity measure was used for the comparison. Schemes to modify the embeddings of variants with the vectors of their most co-occurring words in the training data were explored and we observed some improvement in the results. Again, this process is exploratory as only default parameters were used in the embedding model training. The projection process also adopted a simple alignment-based approach. Although these scores do not generally compare very well against the scores from other techniques, there is a lot of potential opportunities such as using them as features for training models for other NLP tasks: part-of-speech and semantic tagging, word sense disambiguation, machine translation and others.

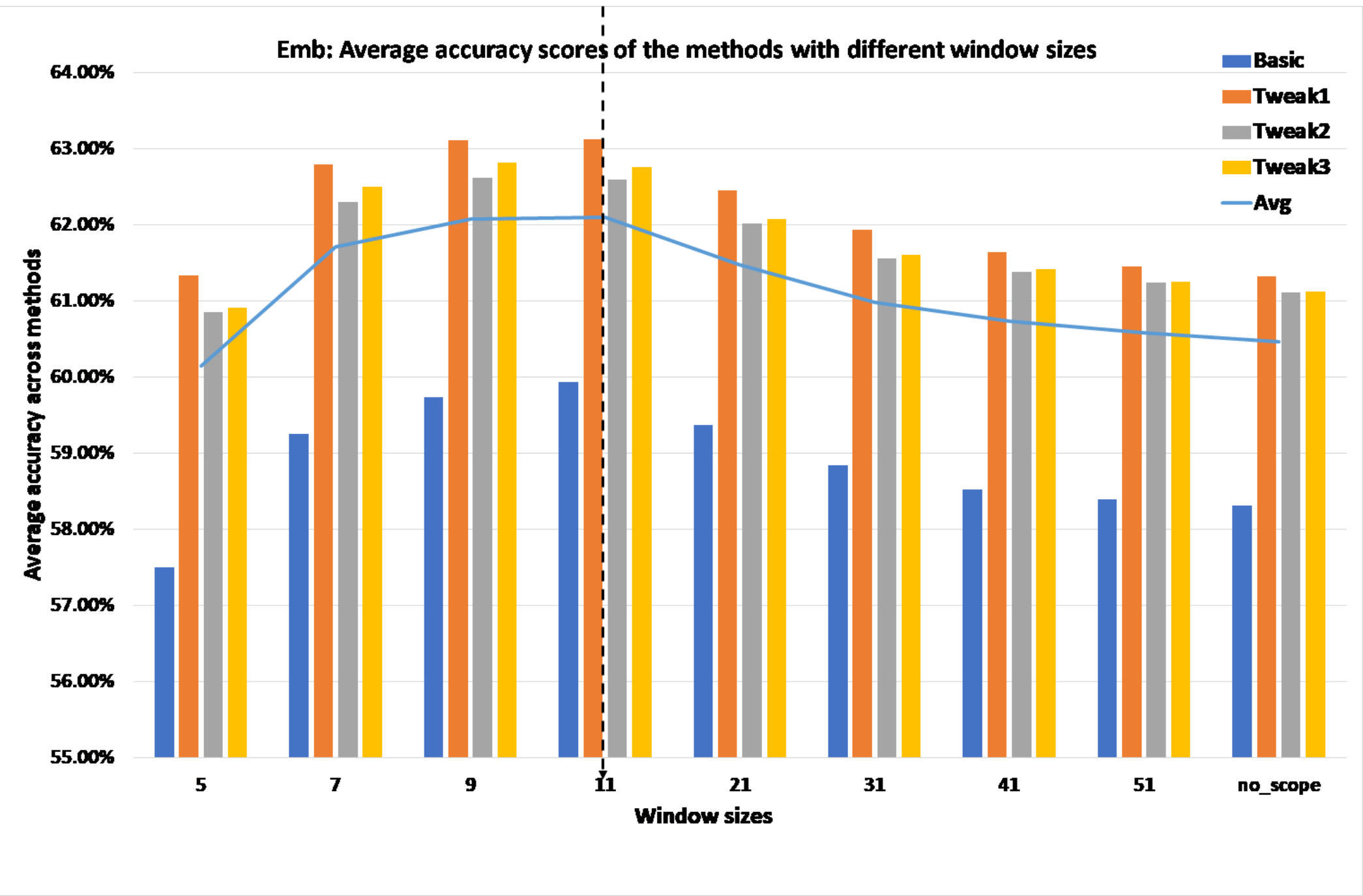

Fig. 7.15 **Embedding Model:** Graph showing the average accuracy scores achieved with different enhancement methods over all models using different window sizes. Overall, the window sizes 9 and 11 appear to have performed better than the others.

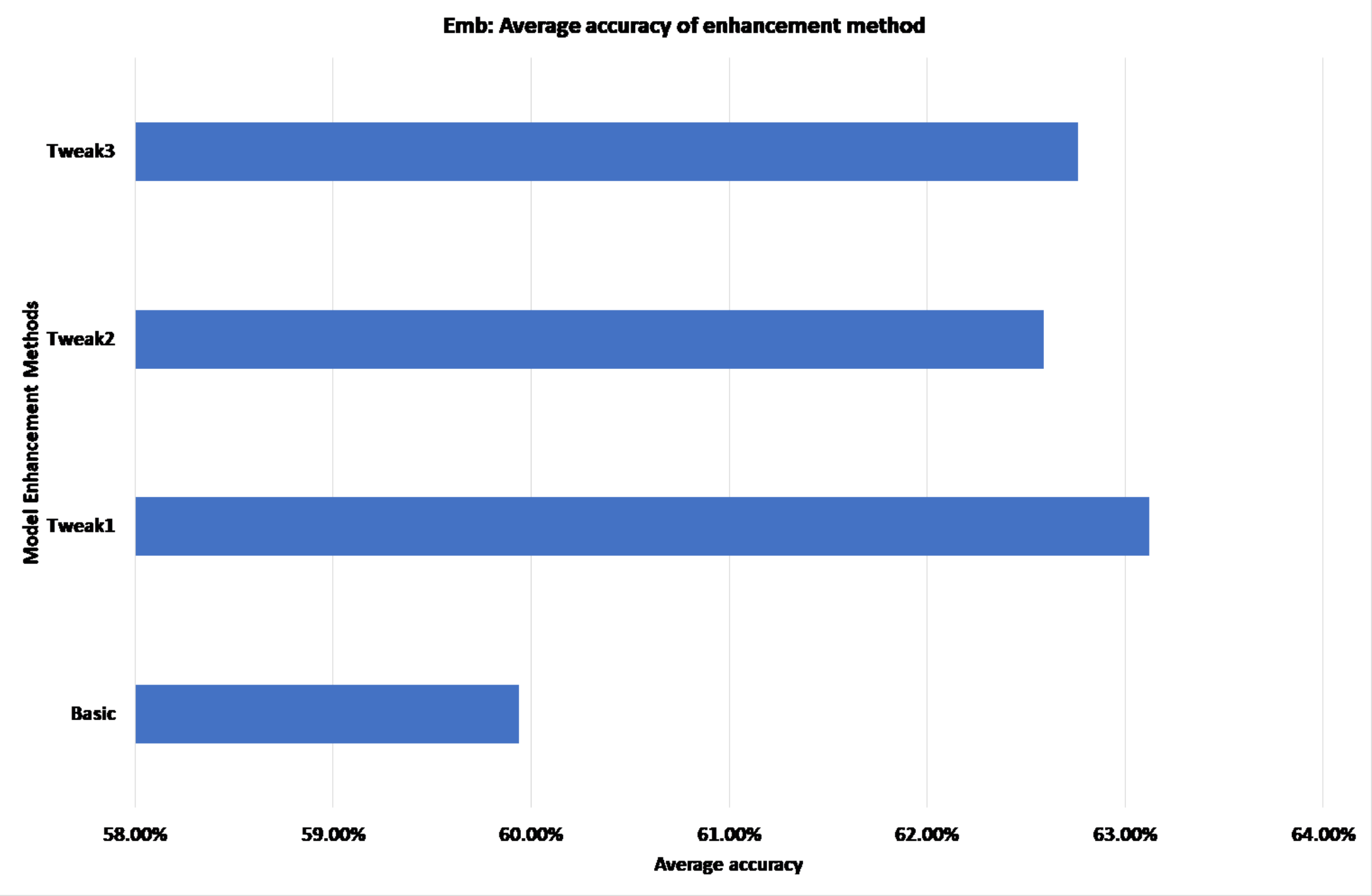

Fig. 7.16 **Embedding Model:** Graph showing the summary of the performance of the enhancement methods as compared with using the *Basic* set-up.
The values plotted are average scores for each enhancement method over all models.

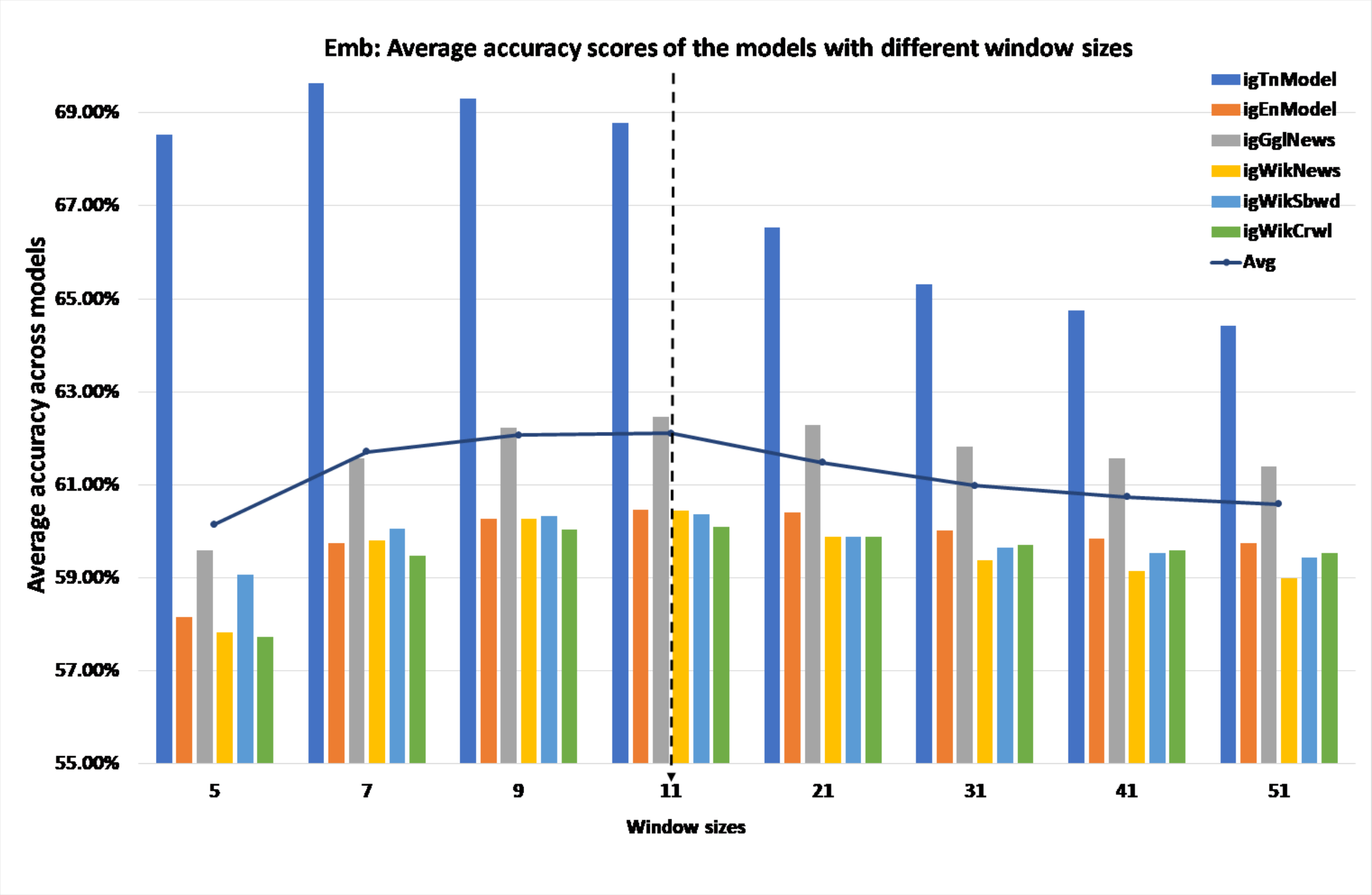

Fig. 7.17 **Embedding Model:** Graph showing the average accuracy scores achieved by all models across all tweaks using different window sizes. Again, the window sizes followed the same trend as in Figure 7.15 but **igTnModel** performed best with window size of 7.

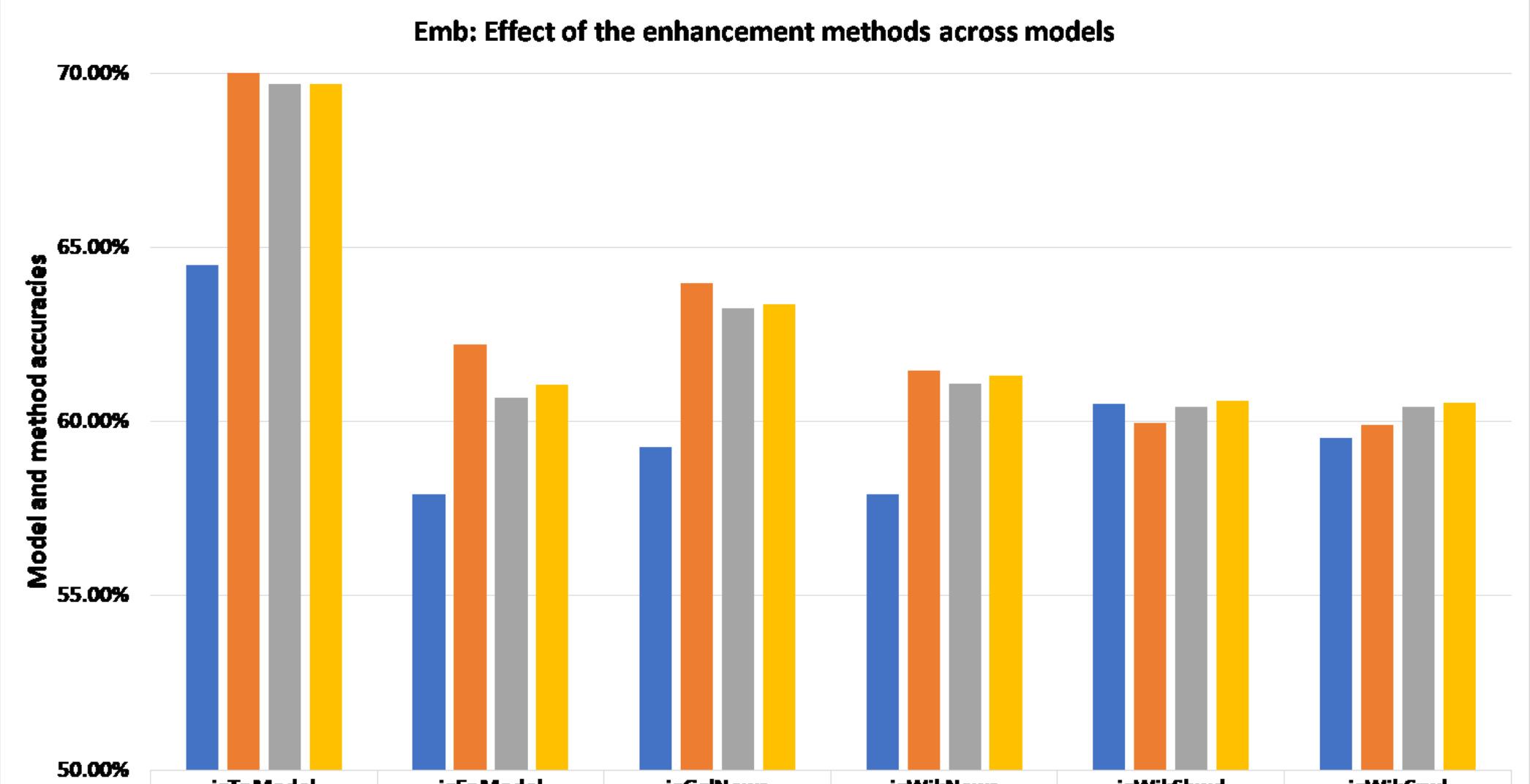

| | igTnModel | igEnModel | igGglNews | igWikNews | igWikSbwd | igWikCrwl |
|---|---|---|---|---|---|---|
| Basic | 64.50% | 57.90% | 59.26% | 57.92% | 60.52% | 59.53% |
| Tweak1 | 71.24% | 62.20% | 63.96% | 61.46% | 59.97% | 59.90% |
| Tweak2 | 69.68% | 60.68% | 63.24% | 61.09% | 60.42% | 60.43% |
| Tweak3 | 69.69% | 61.07% | 63.36% | 61.31% | 60.59% | 60.54% |

Fig. 7.18 **Embedding Model:** Graph showing the accuracy scores by all the models with all enhancement methods highlighting the positive effect of enhancements. *Tweak1* performed slightly better.

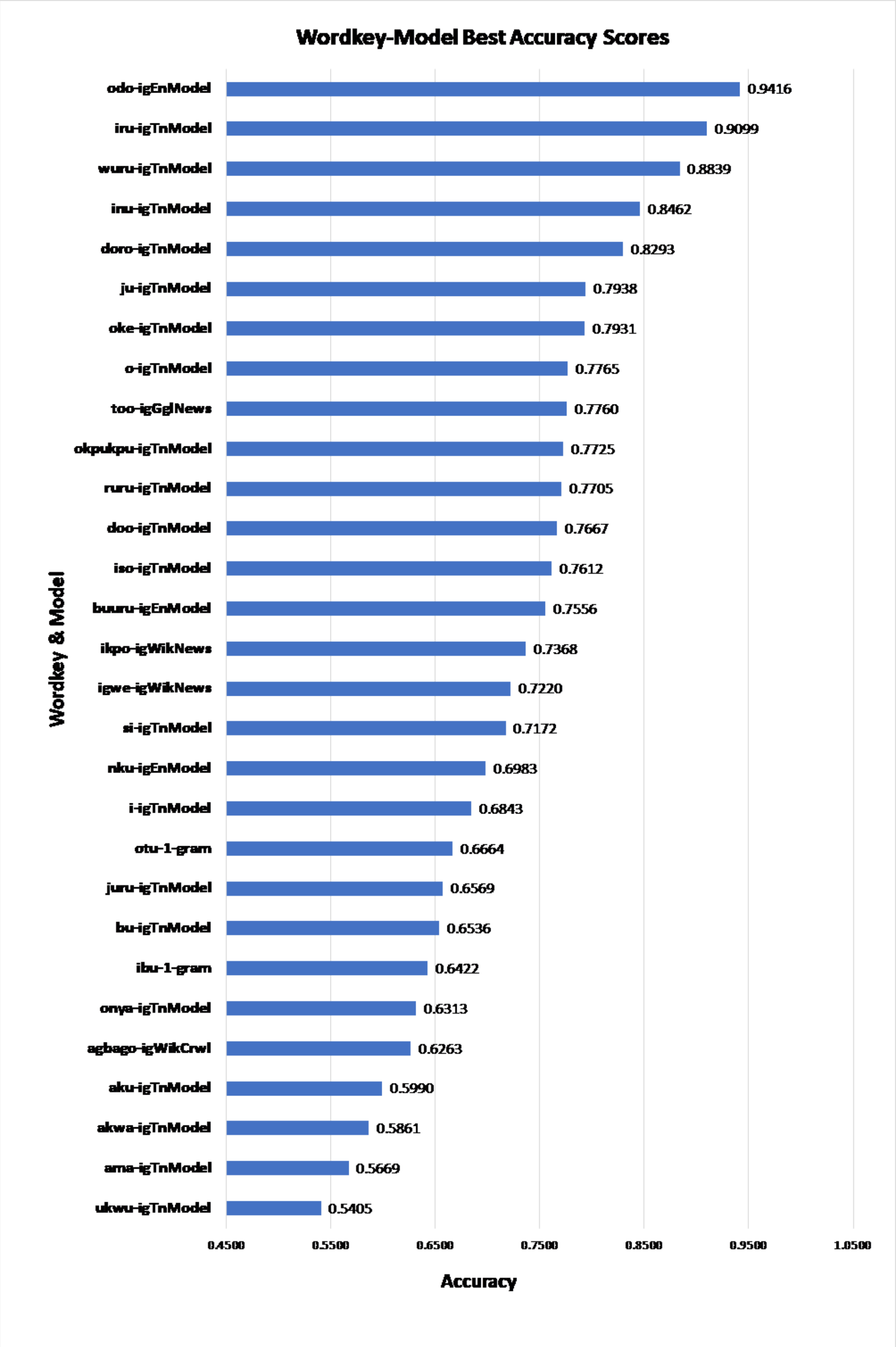

Fig. 7.19 **Emb Accuracy:** A bar-chart showing the plot of the wordkey best score as well as the models that got those scores in their descending order of performance.

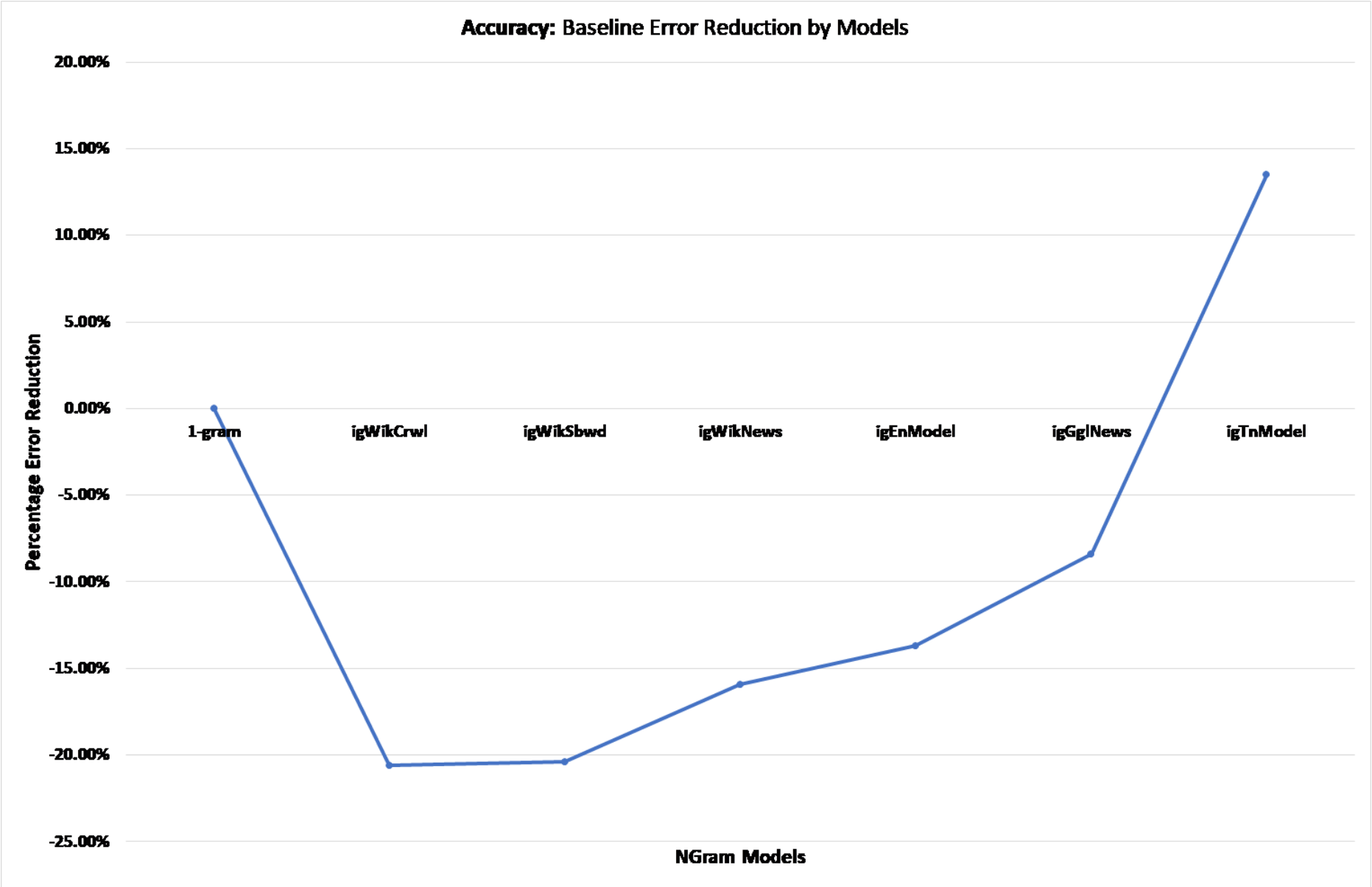

Fig. 7.22 **Emb Accuracy:** Graph showing the percentage reduction of the global baseline error by all the embedding models.

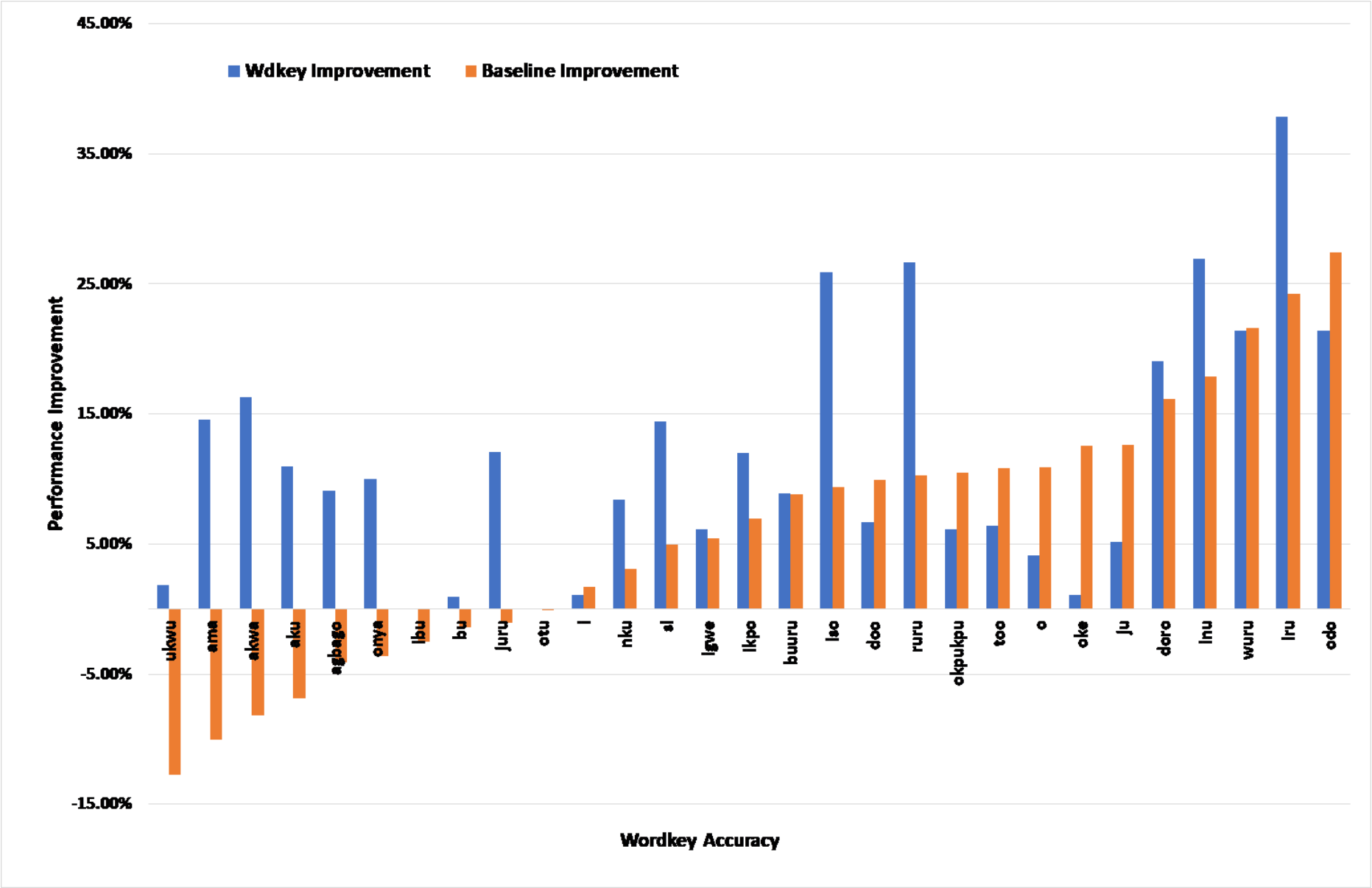

Fig. 7.23 **Emb Accuracy:** Graph showing the maximum improvement in the baseline accuracy scores on the individual wordkeys.

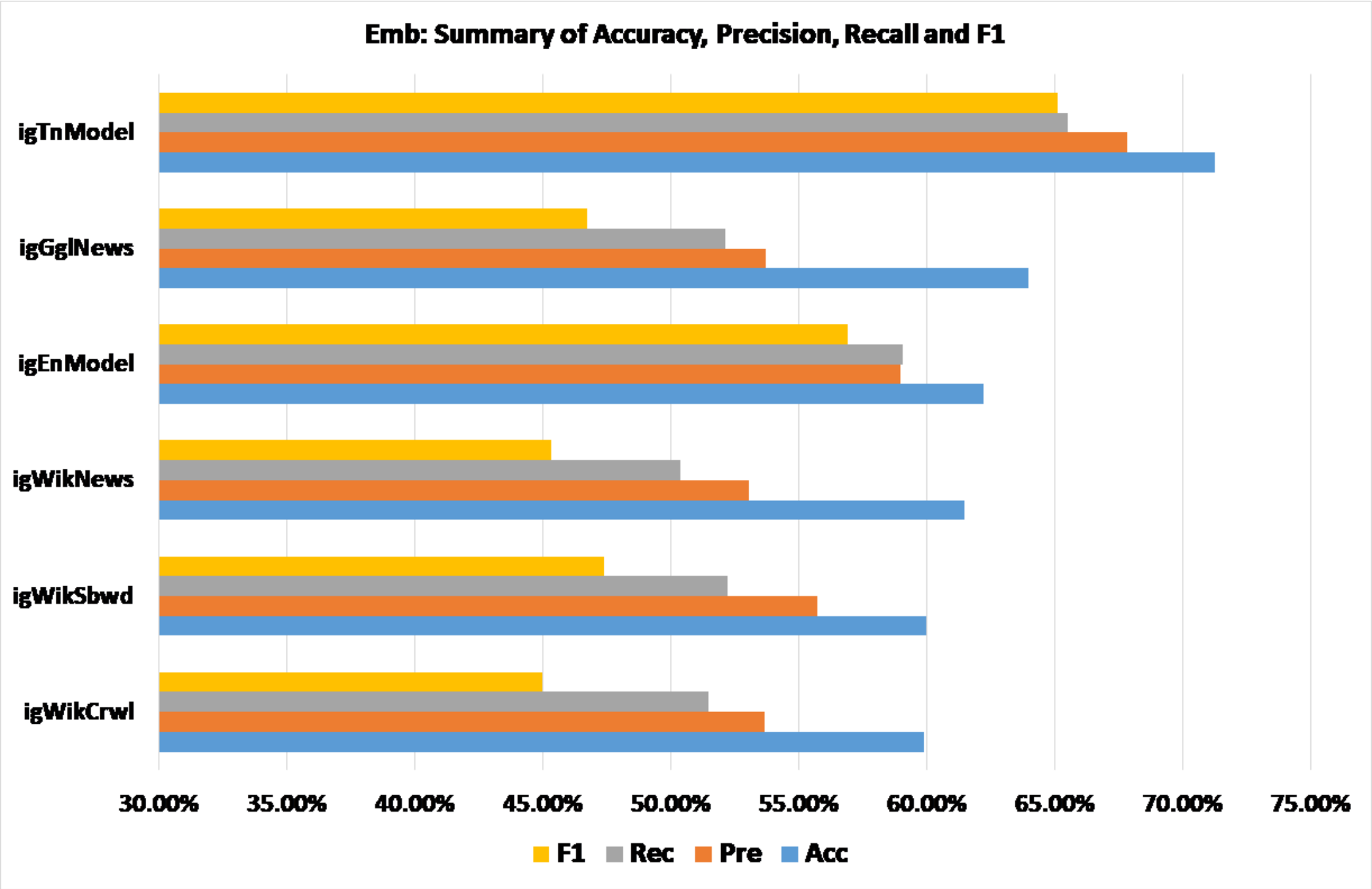

Fig. 7.24 **Emb APRF:** Graph showing the comparison of the embedding models on their accuracy, precision, recall and F1 scores.

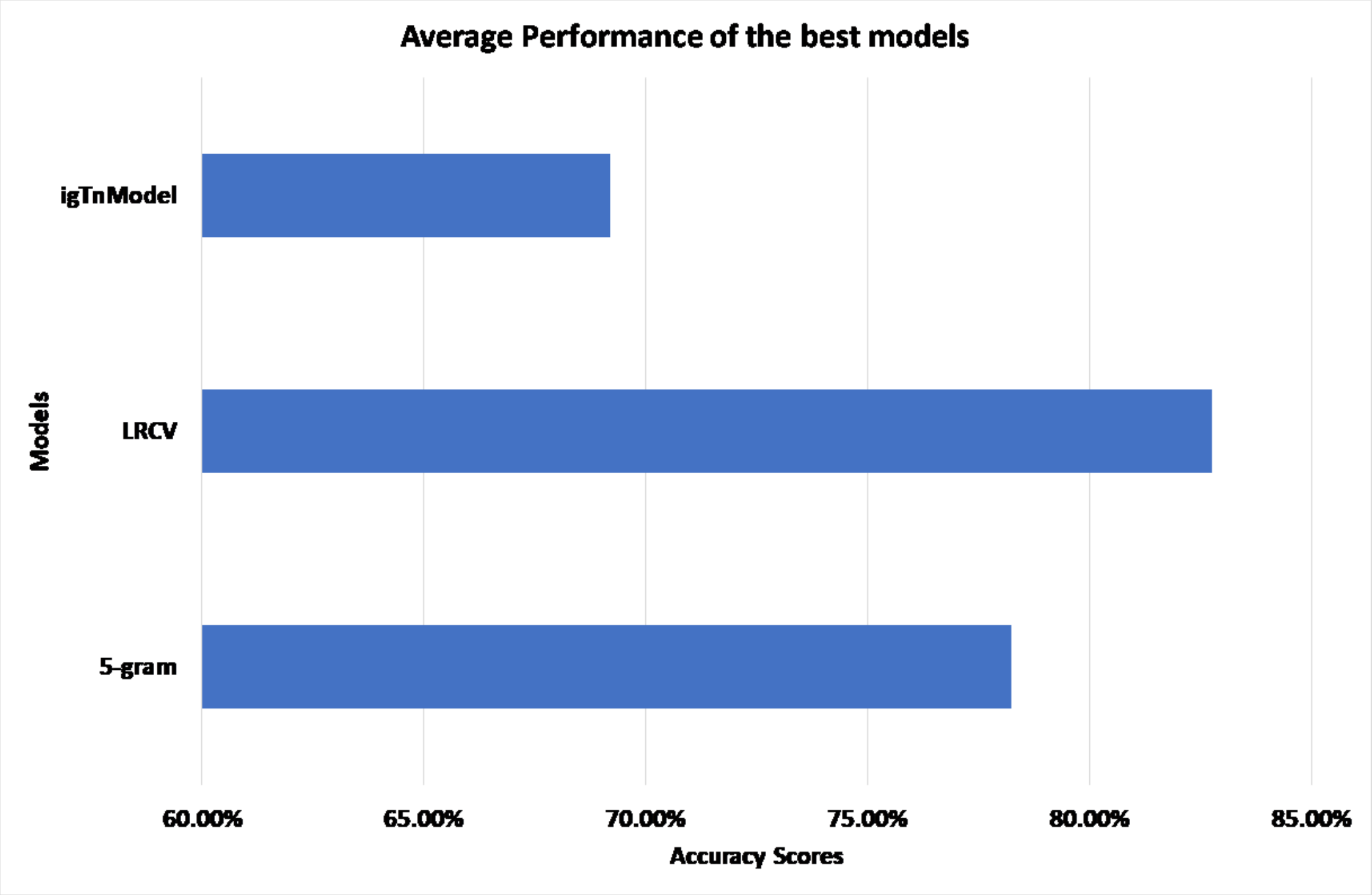

Fig. 7.25 **Best models:** Graph showing the average accuracies of the best performing models – **5-gram**, **LRCV** and **igTnModel** – from each of our experiments.

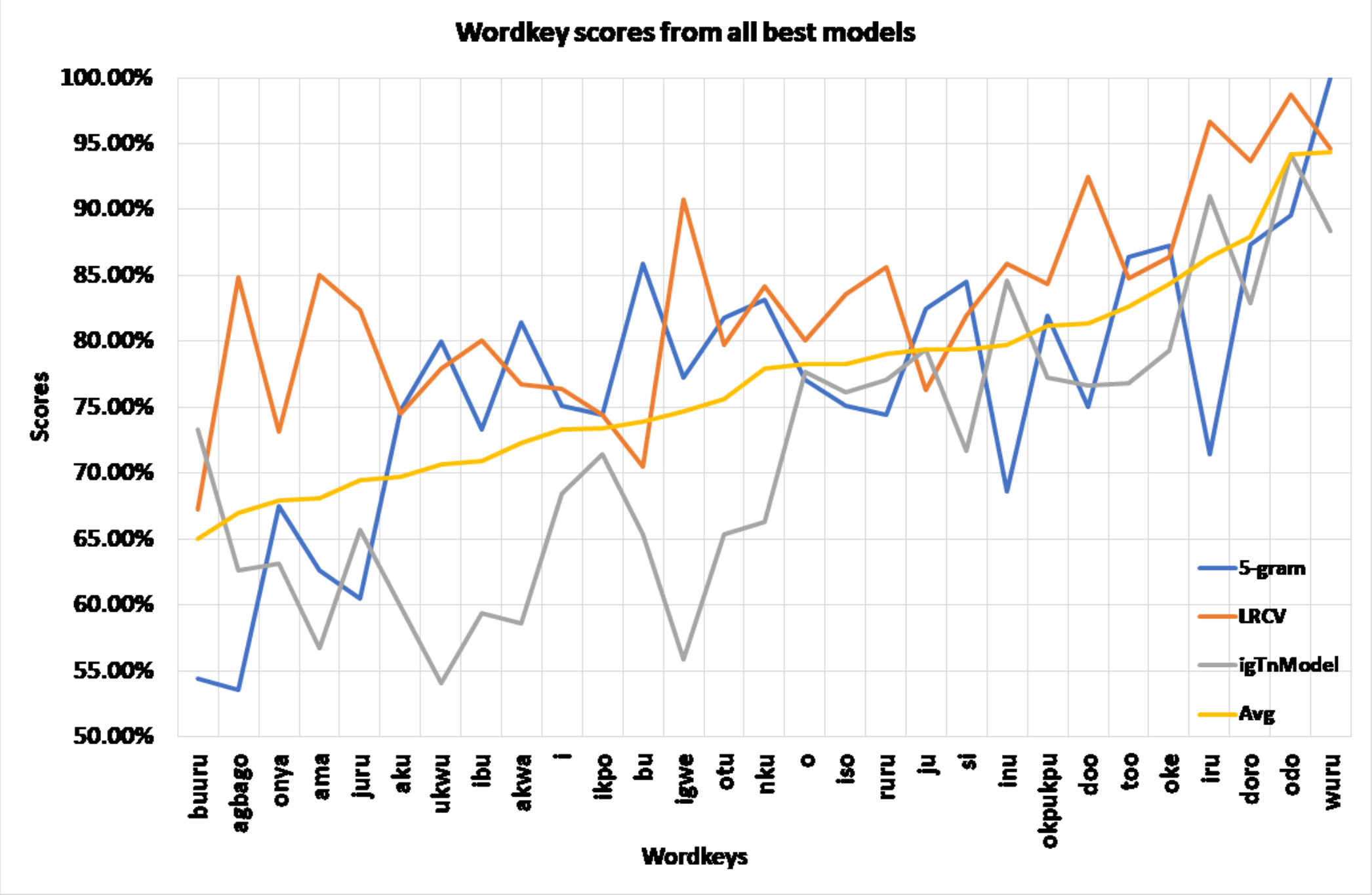

Fig. 7.26 **Best models:** Graph tracing the performances of each of the wordkeys by the models shown in Figure 7.25.

# Chapter 8

# Testing Restoration Models

In Chapters 5, 6 and 7, we discussed the design and implementation of the experiments on the three approaches to diacritic restoration used in this work. We also presented and compared the results from the different models in each of the methods on a selection of ambiguous sets defined in §4.2.3. In this chapter, we will define a complete restoration pipeline and build a restoration system using the best performing model from each of the methods. Also, we will test our system with a sample text from the jw.org website which is more diverse and contemporary than the Bible but has a similar level of diacritic marks.

## 8.1 Summary of Experimental Results

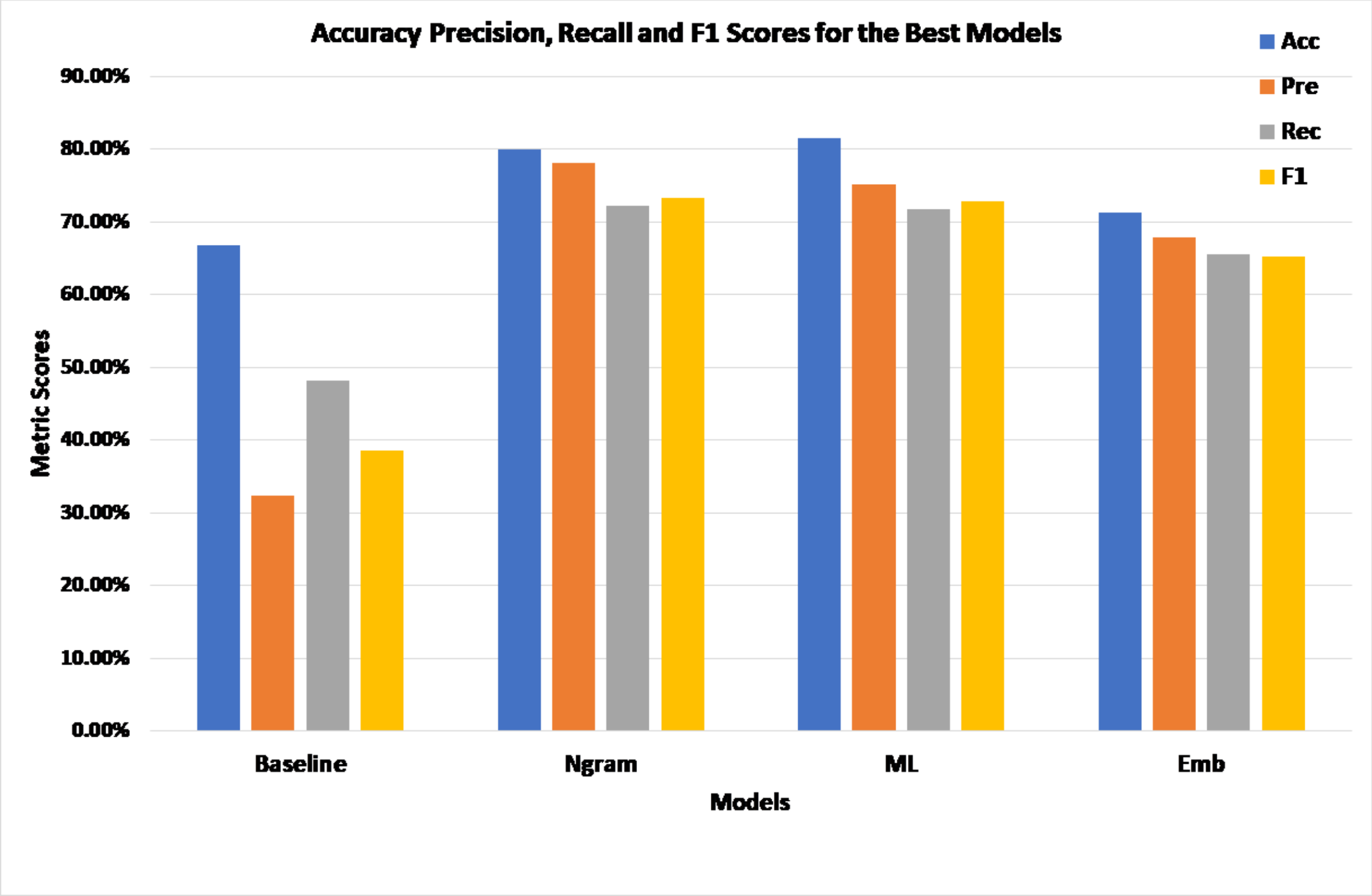

| Metrics | Unigram | N-gram | ML | Emb |
|---|---|---|---|---|
| *Accuracy* | 66.75 | 80.01 | 81.55 | 71.24 |
| *Precision* | 32.30 | 78.15 | 75.11 | 67.85 |
| *Recall* | 48.17 | 72.20 | 71.74 | 65.51 |
| *F1-Score* | 38.55 | 73.30 | 72.86 | 65.19 |

Table 8.1 Graph and Table showing the average performances of the best models from 3 key techniques: *n-gram models*, *classification* and *embedding models* on diacritic restoration. These scores are as obtained by the best models and configurations for each technique.

## 8.2 Restoration Pipeline

The restoration pipeline defined in Figure 8.1 models the process described in §5.2.1. It basically accepts a stripped Igbo text and pre-processes, i.e. tokenises, it with the *Igbo tokeniser*. The tokenised text will then be passed to the restoration model, **Restorer**, which will further prepare the text as expected by the model. Then with the tokens taken one at a time (and with the relevant context), the model produces and stores the most probable diacritic variant. These restored texts will then be prepared (simply concatenated) and delivered as the output from the restoration.

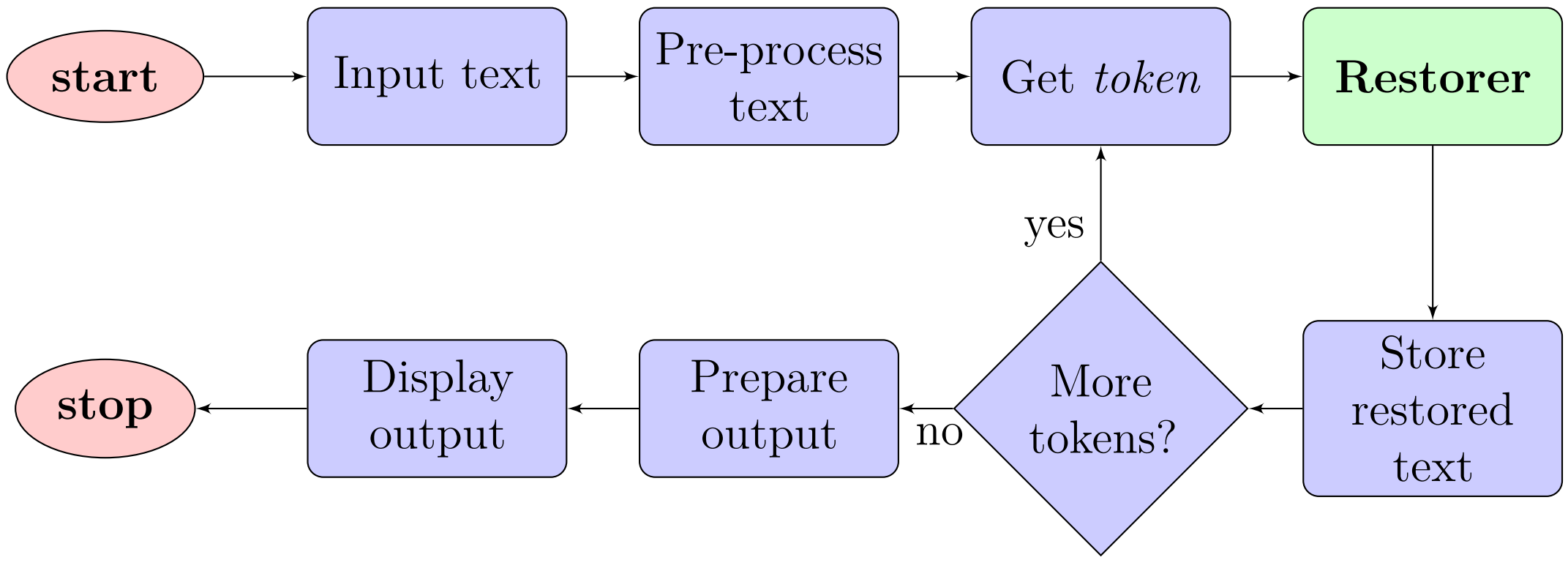

Fig. 8.1 The full diacritic restoration pipeline with the **Restorer** being any of the models previously defined in Chapters 5, 6 and 7

The diagram is presented in a high-level format for clarity. In practice, the **Restorer** encompasses more than just the model. It includes the processes that identify non-diacritic tokens, punctuations, digits and other special characters and return them as they are. Unambiguous tokens are also replaced with their diacritic versions. So the major focus of the restoration process is on the ambiguous tokens.

## 8.3 Datasets

### 8.3.1 Training data

Statistically, up to 70% of the word tokens (not including punctuation, digits and special characters) in the data used are either without diacritics or unambiguous. Models tested on *all* tokens are likely to get very high scores by doing very little. Therefore, in the experiments reported, we selected only 29 diacritically ambiguous sets as earlier explained in §4.2.3 with the following key conditions in mind:

1. the *wordkey representation* (i.e. all occurrences of its variants) in the text should constitute, at the minimum, one per 10000 word tokens (0.01%).

2. only variants that make up, at least, 5% of the wordkey representation in the text are considered for the set

3. heavily skewed wordkeys (i.e. with one variant accounting for up to 75% of the entire set) are also dropped

Though the above conditions seem harsh, we wanted to ensure the task was challenging enough for the models we were building. Also, a quick manual inspection indicated that some of the ambiguous sets removed were actually noise in the data due to spelling errors or diacritic mark inconsistencies. Details of these experiments and their results were presented in Chapters 5, 6 and 7.

### 8.3.2 Evaluation Dataset

The test conducted in this chapter includes all the word tokens in the evaluation scores. Punctuation marks, digits and special characters are replaced as they are and therefore not counted in the scoring. The complete pipeline is run on two datasets: the *all data* dataset i.e. original Bible dataset described in §4.2.2 and a *small Igbo dataset* built from a 1766-word text randomly sampled from the online Igbo versions of the Jehova Witness's *Awake* and *Watch Tower* magazines on the jw.org website. Although this new dataset has a similar level of diacritic marks to the Bible text that dominated our training original dataset, its content is more contemporary and diverse.

We evaluated the models on the *Small Igbo Dataset* referred to above which was built from 96 sentences randomly extracted from the sample text collected from the jw.org website. The analysis of the text is presented in Table 8.2.

| **Item Desc** | **Counts** |
|---|---|
| *No of Sentence* | 96 |
| *Total no of word* | 1766 |
| *All wordkeys* | 448 |
| *Ambiguous words* | 361 |
| *Unambiguous words* | 1405 |
| *Ambiguous wordkeys* | 45 (12 in this text) |

**Summary:**
– 48 keys
– 1766 words
– 20.44% (361/1766) ambiguous
– 79.56% (1405/1766) unambiguous

Table 8.2 Basic statistical analysis for the *Small Igbo Data.*

It is important to mention that although the reported number of ambiguous wordkeys is small compared to the 855 we saw in the larger training data, the real level of ambiguity is much higher. So in counting the number of ambiguous words, we included wordkeys that we know are ambiguous from the previous experiments, even if they have only one variant in the small evaluation dataset. The full count of ambiguous wordkeys and their variants is shown in Table 8.3 with only 12 of them featuring multiple variants.

| **Wordkeys** | **Variant counts** | **Wordkeys** | **Variant counts** |
|---:|---|---:|---|
| **o** | ọ=52, o=15, ọ́=1 | **na** | na=65, ná= 3 |
| **ahu** | ahụ=37, ahụ́=4 | **bu** | bụ=25, bọ́=6 |
| **otu** | otụ́=14, otu=9 | **di** | dị=13, di=1, dọ́=1 |
| **i** | i=7, ị=6 | *otutu* | ọtụtụ=12 |
| **si** | si=10, sị=1 | **buru** | bụrụ=8, buru=3 |
| *juu* | juu=5 | *ibu* | ibu=4 |
| **ikwu** | ikwu=3, ịkwụ=1 | *inu* | ịṅụ=4 |
| **bi** | bi=3, bí=1 | *isi* | isi=3 |
| *oku* | ọkụ=3 | **siri** | sịrị=2, siri=1 |
| *ama* | ama=2 | *akwa* | ákwà=2 |
| *bia* | bịa=2 | *azu* | azụ=2 |
| *kwuo* | kwuo=2 | *iso* | iso=2 |
| *igba* | ịgba=2 | *kuru* | kuru=2 |
| *onu* | ọnụ=1 | *ukwu* | ukwu=1 |
| *atu* | atụ=1 | *udo* | udo=1 |
| *agwa* | àgwà=1 | *iru* | iru=1 |
| *ekpe* | ekpe=1 | *nweghi* | nweghị=1 |
| *iga* | ịga=1 | *enyi* | enyi=1 |
| *arusi* | arụsi=1 | *ntu* | ntụ=1 |
| *too* | too=1 | *ndo* | ndo=1 |
| *igu* | ịgụ=1 | *mbo* | mbọ=1 |
| *amasi* | amasị=1 | *nukwara* | ṅụkwara=1 |
| *kuziiri* | kụziiri=1 | *dochie* | dochie=1 |
| *naghikwa* | naghịkwa=1 | *akuzi* | akụzi=1 |

Table 8.3 Wordkey and variant counts for the *Small Igbo Data.*

## 8.4 Results

This section shows the summary of the evaluation scores for accuracy, precision, recall, and F1 (Table 8.4) of the restoration models trained using the three different methods discussed in this work: n-grams, machine learning and embedding models. For each of the methods, we used the best performing model: *5-grams* for the n-gram methods, *Logistic Regression* for the machine learning methods and *igTrain* i.e. the embedding model trained with Igbo text.

Given that the evaluation is on all the wordkeys and not only the ambiguous ones, we used a baseline accuracy that just compares the marked text with the stripped version. Table 8.4 shows the baseline scores as well as the scores of the other models on the *all data* and small *eval set* datasets. The analysis of the key errors observed in the restored text is also presented in Table 8.5.

### 8.4.1 Evaluation scores

| **Model** | **All data** | | | | **Eval set** | | | |
|---|---|---|---|---|---|---|---|---|
| | ACC | PRE | REC | F1 | ACC | PRE | REC | F1 |
| **Baseline** | 66.99 | 30.79 | 29.53 | 29.72 | 62.65 | 39.35 | 38.60 | 38.77 |
| **5-Gram** | 96.27 | **96.61** | **96.84** | **96.50** | 91.78 | 79.17 | 79.45 | 79.14 |
| **LRCV** | **98.71** | 94.23 | 93.57 | 93.16 | **93.22** | 75.84 | 74.76 | 74.29 |
| **IgTrain** | 90.44 | 89.13 | 89.43 | 88.95 | 88.35 | 77.51 | 76.19 | 75.63 |

Table 8.4 Summary of the evaluation results for different models on different datasets

Table 8.4 shows a marked improvement of the model scores from the baseline (stripped text) score. All the metrics recorded substantial improvement across all the metrics. There is an improvement of almost 30% on the accuracy, dropping the error rate by over 80%. The precision and recall scores also doubled indicating that the model is better both in getting most of the wordkeys right and choosing the right variant for each of the wordkeys.

Although the machine learning model performs best overall in terms of accuracy, the precision and recall scores of the n-gram model are generally better. This is an indication that the n-gram models predict the variant with higher confidence than the other models. The scores for the embedding models are trailing behind the others for both the *all data* and *eval set* datasets.

In general, given that each of the models is trained on the *all data* dataset which is predominantly the Bible, they all seem to have modelled the language well enough to get fairly high scores on more contemporary text. However, the fact that the majority of the words do not contain diacritics helped boost the score and this is why evaluating on the stripped text can get as high as almost 70% accuracy.

### 8.4.2 Analysis of evaluation errors

The result of the restoration on the evaluation dataset indicates that 145 (out of 1766) words were incorrectly restored. Table 8.5 shows the analysis of the result with some of the common types of errors found

| **Details/error type** | **Stats** *all data* | *Eval set* |
|---|---|---|
| **Sentences with errors** | 20028 (out of 35685) | 68 (out of 96) |
| **Incorrectly restored** | 37084 (out of 995308) | 145 (1766) |
| **Out of vocabulary** | 0 | 39 (e.g *amerịka*) |
| **Encoding variations** | 0 | *approx* 10 (e.g *bụ́:bụ´ vs bụ́* ) |

Table 8.5 Analysis of the errors from the evaluation dataset

In the error analysis shown in Table 8.5, we see that although we have approximately 82 errors per 1000 words, 67% of the sentences have at least one wrongly restored word. It is not surprising that about 39 of the words were not found in the training data which is mostly made up of the Bible. However, because *all data* was the same data that was used for the model training, there was no *out-of-vocabulary* (OOV) word found during evaluation.

Again, due to the differences in the unicode encodings between the *all data* and the *eval set*, there were errors reported for even words that were correctly restored. This happens because different encoding styles can sometimes present the same marked word in different forms especially where there are multiple diacritics on a character in the word.

For example the marked word, **bụ́** which has a **u** with a dot-below (̣) and an acute accent (´) can be encoded as **ụ** + **´** (i.e. a dot-below ‘u’ + an acute accent) or **ú** + **̣** (i.e. an acute accent ‘u’ + a dot-below). In practice, comparing the two encoded words will always return false and therefore count as incorrectly restored but in reality the restoration is successful.

## 8.5 *Google Translate* Test on Evaluation Dataset

As presented in §4.1.4, the impact of diacritics may be subtle but the lack of it on text does affect the performance of NLP systems adversely. In this section, we performed another test using *Google Translate*'s Igbo-English translation platform on the *Eval set* to see the extent to which quality of the translations will be affected. The original marked text was submitted as well as the stripped version.

The translation results were scored by a native speaker using a simple point-based system defined as follows:

**0** - Completely wrong

**1** - Mostly wrong

**2** - Mostly right

**3** - Completely right

Some of the sample sentences translated and their outputs are presented in Table 8.7. All the 96 lines in the evaluation data were passed to *Google Translate*. Interestingly, the stripped version is often translated to a sufficiently good English equivalent. In some cases, it actually surpassed the diacritic version.

Overall, the diacritic version translations scored a total of **200** points out of a possible total of **288** (i.e. $96 * 3$) which is equivalent to a percentage score of **69.44%**. On the other hand, the translations of the stripped version scored a total of 163 points which amounts to **56.59%**.

| **Details** | **Stats** | |
|---|---|---|
| | *Marked* | *Stripped* |
| **No of points** | 200 (out of 288) | 163 (out of 288) |
| **Accuracy** | 69.44% | 56.59% |

Table 8.6 *GTranslate* translation scores on the evaluation data

Table 8.6 clearly shows that, although *GTranslate* performs well on the stripped text, there is a 13% performance improvement with diacritics on the text. This indicates that diacritic restoration can enhance the performance of a machine translation system. One can also infer that the good performance of *Google Translate* on the stripped version may be as a result of its training with mostly unmarked data on the web.

| | |
|---|---|
| *Text* | gịnị bụ àgwà ọma ? |
| *GT-Diac* | **what is a good attitude ? [3]** |
| *GT-Strip* | what is the good guy? [1] |
| *Text* | a mụrụ m n' amerịka n' afọ 1979 . |
| *GT-Diac* | I was born in 1979.[2] |
| *GT-Strip* | **I was born in America in 1979.[3]** |
| *Text* | onye ahụ chọọ ikwu okwu , gee ya ntị nke ọma . |
| *GT-Diac* | **The person wants to talk, listen attentively. [3]** |
| *GT-Strip* | **he wants to speak, listen carefully. [3]** |
| *Text* | i nwedịrị ike ịchọpụta na nke a ga- eme ka gị na ya dịkwuo ná mma . |
| *GT-Diac* | **you may even find that this will make you feel closer to him. [3]** |
| *GT-Strip* | It's hard to find that this will make you feel better. [0] |

Table 8.7 Sample translation output from Google Translate on Evaluation dataset

## 8.6 Chapter Summary

In our experiments, we applied three major categories of models: *n-grams*, *machine learning* and *word embeddings* to the diacritic restoration task defined in Chapter 4. In each of the categories, we designed experiments that compared the performances of a number of models on the key metrics of accuracy, precision, recall and F1. The average results across the models from each of the techniques were presented in the relevant chapters, while the summary of the performances on all the metrics is presented in Table 8.1.

### 8.6.1 NGram models

N-gram models did generally very well in these experiments as the next best category after the machine learning classification models. A key advantage of the n-gram models is the reasonable ease to conceptualise and implement them. However, while they can be scaled up to higher n-grams, there are not a lot of options to optimise their performance in this case, as their scores plateau at some point.

### 8.6.2 Machine learning models

On the other hand, it is slightly harder to design, build and tune machine learning classification models but they appear to offer more latitude for exploration of the solution space. For instance, we got the best results for the task set out in these experiments with machine learning classification models in spite of our using only the default parameters for the model training algorithms. Another interesting aspect is that, though we did not exploit it in this work, there is still a lot of room to improve the robustness of these classification models through better feature engineering and algorithm parameter optimisation even with the same size of data.

### 8.6.3 Embedding models

The experiments with the embedding models present a lot of materials for future work with fairly good potential for success. Although its results are lagging behind others in these experiments, the opportunities it provides for future research especially in transfer learning experiments, which is very essential for low resource languages, is quite massive. As a class of models, they are also very generalisable and could easily be applied to various kinds of tasks, as we did in 7.4.3, with a good reputation for improving the performance of NLP systems [88].

This is because, in a lot of NLP tasks (POS tagging, named entity recognition, document classification), the inputs are sequences of words. In most recent works, variants of deep neural networks (feed-forward, recurrent and convolutional) are now being applied with very impressive results. However, these libraries supporting these neural network algorithms receive their inputs as vectors of real numbers. Therefore, to take advantage of the effectiveness of these algorithms, it makes sense to use distributed representation of words (and word sequences) which the word embedding models offer. In fact, the use of embedding models is so commonplace in NLP that currently, the actual research efforts is focused on training the embedding model that best models the language i.e. one that capture better the structures, the meanings and the relationships of the words of the language.

Although the use of embedding models did not perform as well as other methods on our task, it is important to note that we did not focus on exploring all possible options for training and evaluating specific models for Igbo diacritic restoration. We used simple approaches that trained embedding models with the default parameters in the *Gensim* [84] library using small Igbo data and also a simple projection method that transfer pre-trained English embedding, (with no Igbo language information) to Igbo.

Therefore, our future work with embedding models on this task will be more explorative. We will focus on optimising the training parameters on the benchmark diacritic restoration data. Also, a hybrid approach using machine learning (or even deep neural models) could be used, with the embedding models as input, for the task.

### 8.6.4 Model evaluation

Using the full restoration pipeline described in §8.2, the restoration system applied the best configurations from the three methods used in this work. These were evaluated using the expanded ambiguous *all data* as well as a small evaluation dataset, *eval set*, constructed from the text from the *jw.org* website. The results (see 8.4) show a substantial improvement in metric scores on the full data compared to the results on the experimental datasets consisting of only ambiguous words. This is not surprising given that the test set includes mostly unambiguous and heavily-skewed ambiguous sets.

However, our best result 98.71% now compares better with those reported for other languages. For example Yarowsky [112] achieved up to 99% on Spanish and French; Mihalcea [64] got 99% on Romanian; Nguyen & Ock [70] achieved 94.7% on Vietnamese; Scannell's [91] best scores on some low-resource languages range from 88.8% to 99.5%; Cocks & Keegan [16] also achieved 99% on Māori.

An experiment to test the impact of diacritics (or lack of them) on machine translation systems was performed. GTranslate was used for the evaluation of the text from *Jehova Witness*'s website jw.org. An improvement of approximately 13% was observed when compared the score on a stripped text and its correctly marked version. Although the output of our system may not have achieved the same level of improvement as the human marked and manually checked diacritic text, it has been established that diacritics does enhance performance of machine translation systems.

In the next Chapter, a summary of the entire thesis will be presented and the future research directions will be discussed.

# Chapter 9

# Summary and Conclusion

In this chapter, we present a summary of all the activities we embarked on in the course of this PhD study. The structure of our presentation will cover the entire thesis. This will include the original motivation for this work and the project core objectives as well as a highlight of the main content of each of the chapters. The rest of the sections will contain a brief summary of the experimental results, a note on our published works, and then the conclusion and future work.

## 9.1 Review of the Research Questions

In Chapter 1, we established that African languages, such as Igbo, are generally low-resourced with regards to natural language processing research. We also identified the problem of diacritic restoration of Igbo which, though not a common NLP task, is a very essential pre-processing task for other downstream tasks e.g. machine translation, automatic speech recognition, text-to-speech systems.

We highlighted the role of diacritics marks in meaning disambiguation and correct pronunciation of Igbo words and then set out to address the following research questions:

1. *Can we construct a standard dataset for the Igbo diacritic restoration task?*:
2. *Can we build a robust automatic diacritic restoration system for the Igbo language?*
3. *Can we take advantage of existing high resource language models in diacritic restoration?*

A brief discussion of the extent to which the research questions were addressed is presented below:

**Reasearch Question 1 (RQ1):** At the beginning of this research work, we identified the need to develop a standard dataset against which we, and other researchers, could benchmark our models since none existed. We set out to define a simple generic framework (§4.2.3) for building the dataset for diacritic restoration experiments from diacritised text.

At the core of this framework is the identification of the location and the context of all the possible variants of a stripped word, and extracting and labelling appropriately for training. It could easily be applied to other Latin-based languages with diacritics and automatically generates the dataset on a one-off basis for future use.

By developing this generic framework and creating a standard dataset for Igbo diacritic restoration task in this process, RQ1 has been addressed.

**Reasearch Question 2 (RQ2):** In addressing the RQ2 for this work, we designed and conducted experiments for diacritic restoration systems built with three different methods: *n-gram* (Chapter 5), *machine learning* (Chapter 6), *embedding* (Chapter 7). For the n-gram models, we ran experiments with n-grams from unigram to 5-grams, beyond which there is no more improvements on the performance of the restoration task. With the machine learning approach, we trained and evaluated classification models with 12 different learning algorithms.

The models were evaluated on both the performance on the task and the efficiency in training. The embedding models were trained on the Igbo data as well as projected from other well-resourced languages. The performance of the best systems from these methods were analysed and compared in §8.4.1.

We have therefore addressed RQ2 by building a high performance diacritic restoration model for Igbo which could achieve an accuracy of 98.71% and compares well with similar systems for other languages.

**Reasearch Question 3 (RQ3):** Rather than re-inventing the wheel, a key approach to tackling the low-resource NLP challenges is adapting existing resources from well-resourced languages. These resources are then fine-tuned and applied to tasks that address the needs of a particular low-resource language. In addressing RQ3, we made an attempt to extend the resources for Igbo by creating embedding models with Igbo data. We also applied some transfer learning techniques to adapt the embeddings trained with a large collection of English text. For diacritic restoration, we devised a method for applying embedding models to the task.

We also built datasets for intrinsic evaluations of these models using the tasks of *odd-word*, *word-similarity* and *analogy.* These tasks are not directly related to our task but do share some similarity with regards to the concepts of the relatedness of words. For example, given that without diacritics *akwa* could be **ákwá**(cry) or **ákwá**(cloth), a context embedding closer to either of these – (*weeping, weep, cry, howl, tears*) or (*cloth, cloths, put, upon, headdress, garment*) may help us decide which variant we need.

Therefore, by applying transfer learning techniques in the construction of embedding models for Igbo from English data and developing methods for applying them to diacritic restoration and other tasks, we have answered RQ3.

## 9.2 Dataset and Experiments

The experiments and analysis done in this project were reported in chapter 4, chapter 5, chapter 6 and chapter 7. Also, a key part of these experiments was the report on the development of a framework for creating a standard dataset for the diacritic restoration task as well as the core evaluation methods and metrics which were presented in chapter 4.

The dataset created in this work for Igbo diacritic restoration provides a standard training set for our future work, the research community and those who may wish to extend our work. But equally important is the development of a robust framework for building the dataset. This makes it easier to test out method on other languages with a similar orthographic structure to Igbo.

### 9.2.1 Dataset: Key Considerations

As presented in §4.2.3, some key considerations were made while designing the framework for generating the dataset and they are summarised below:

- For a given wordkey, each of its variants should, at least, be up to a defined minimum (as explained in §4.2.3). This condition is given to avoid the noise often introduced by a few minor mistakes that tend to create ambiguity where there is none. For example, looking at *mmadụ* with 3474 instances and *mmadu* with 5 instances, our framework will consider the word unambiguous and will restore every occurrence of *mmadu* with *mmadụ*. This is controlled by variable *varntRep* and has a default value of 5% i.e. every variant should have a minimum of 5% representation in the wordkey.

- Besides the representation of the variants, the framework also imposes a minimum representation score for each wordkey. We refer to this as *wdkeyRep* (i.e. wordkey representation) and the default setting insists on the 0.0001 or 1 wordkey/1000 words to build, at least, $\approx 100$ training instances for the wordkey. Wordkeys that do not meet this condition are excluded from the dataset and their restoration will be based on their most common variants.

- Another key condition from the framework is the *variant distribution* (or *varnt-Distrb).* This considers the distribution of variants of a given wordkey with the aim of ensuring that no variant heavily dominates the others. The default value is set to 75% which basically removes any ambiguous set that achieves 75% accuracy by choosing the most common variant. We introduced this to ensure that robust models are built by removing the heavily skewed ambiguous set during the training stage. We will recommend that when the production model is being trained, it should be set to 100% thereby allowing all balanced and non-balanced sets.

### 9.2.2 Dataset: Limitations

A key limitation of the dataset we used in this work is that it is built mostly ($> 94\%$) from the Bible text. So clearly it does not reflect the contemporary written and spoken Igbo. Given that we focused on the word-based approach to the restoration task, it is obvious that many words (and wordkeys) used today may not be found in our dataset. This is likely to affect the performance of the system when deployed to more modern Igbo text. A possible solution to this problem which we have included in our future work is to increase the amount of data we have with a more contemporary text. However, these texts are likely to come from online sources (e.g. social media) and might be written with proper diacritic marks. Therefore, these texts should be corrected manually or using a semi-automatic approach to ensure that the quality of the text is good enough.

## 9.3 Future Work

As we earlier indicated, we embarked on an explorative research for igboNLP and therefore touched on a number of key concepts and stages in a typical NLP pipeline. Given that we had not much to start with, our aim was to see just how much we could achieve with *only* the data and almost no human annotation – hence corpus-based.

Also, because the data was not large enough for unsupervised language model training such as the embedding models used in this work, we also explored some transfer learning concepts. With these done, our next areas of concentration will include but will not be limited to:

**increasing the corpus size:** The methods we used in our experiments are corpus based and depend mostly on automatic techniques for pre-processing and dataset building. At approximately 1m tokens, our data size is still comparatively lower than those reported in literature which often range from hundreds of millions to even billions. It has been shown that, in practice, statistical NLP and machine learning models generally perform better as the size of data increases [43]. We will therefore, among other things, focus on increasing the size of the Igbo data used in training our models to measure the improvement (or otherwise) on the performance diacritic restoration.

**machine learning parameters:** In this work, we trained 12 classification models using the most common machine learning algorithms. However, because of time constraints, we adopted the *vanilla* versions the machine learning models used i.e. only their default configurations on Scikit-learn were applied. In future work, we will probe the best performing models further by optimising their parameters on the task and observe the impact on the overall performance of our model.

**character-based and sub-word embeddings:** Our work focused on the representation of diacritic variants as words. Some authors [105, 64] presented character-based approaches in their work on other low-resource languages. Also, character or sub-word (i.e. character ngrams) representations are recently being promoted as more robust models than just word embedding [52] in both machine learning and deep neural models. Our future experiments will explore the training and application of such models for the diacritic restoration task.

**multi-lingual word embeddings:** The transfer learning method applied in our experiments involved training embedding models only on monolingual data [42]. The trained model is then projected unto the embedding space of the new language using some alignment-based mapping. The downside of this approach is that it only learns the patterns in one language and attempts to "force" it on the process. However, it is possible to build models that learn simultaneously from more than one language [46] and our future work will be exploring this method.

**integrating other systems:** In this work, we applied a corpus-based approach in which we basically designed an experimental pipeline that accepts data as the only input. The dataset is automatically generated from the data and presented for the experiments. Depending on the restoration method, the data will be slightly re-processed to fit the input specification of the experiment. However, no other pre-built systems were included in the pipeline. In future work, we will explore the integration of other systems like an Igbo part-of-speech tagger and morphological segmentation tool to see how the performance of our system will be improved.

**extending the system:** With regards to diacritic restoration, there are many other similar (i.e. written with latin scripts and containing diacritics) languages that could benefit from this work in Africa or even in Nigeria. For example, the Hausa and Yoruba languages have similar diacritic challenges to Igbo. In our future work, we will be testing the generalizability of techniques on the other languages with similar orthography and diacritic content.

## 9.4 Relevance of the diacritic restoration

According to Yarowsky [111], for languages with diacritics, the restoration of diacritics on texts has ‘immediate and practical’ applications. The restoration tool can be packaged as a standalone tool or integrated to the front-end component of a language processing system.

A number of research works have demonstrated the capacity of the restoration tool to improve other, more mainstream, language and speech systems. Some of the common examples of tools that can directly benefit from the diacritic restoration tool either at a pre-processing or post-processing stage, or even as a key tool for the task, are discussed below:

**Spelling and grammar checking** Diacritic restoration is actually a special case of spelling and grammar correction. By replacing the wordkeys with their correct variants in context, the correct spellings of words, using the right characters, are achieved. It also supports grammar checking by identifying the correct word for a given role in the sentence. For instance, given the wordkey *egbe* in the context of a word, say *shooting*, one expects the most probable variant to be **égbè** i.e. *gun* rather than **égbé**, *kite*.

**Auto-completion** A major reason for not writing texts with proper diacritics is the lack of adequate keyboard support for languages with diacritics. This is made worse by the increase in the use of smart and hand-held devices in generating texts. An auto-completion system supported by a diacritic restoration can improve the production of texts with better marked diacritics [110]. This is necessary for enhancing the correctness of what is typed as well as ensuring that the right pages or documents are returned from a search query. For example a search with *resume* may also return pages with for *résumé*.

**Text-to-speech synthesis** A good text-to-speech synthesis (TTS) tool depends a lot on the correctness of the written text [103] to create good speech synthesis from text. The machine is often not capable of resolving the ambiguities occasioned by lack of proper diacritics. So there is a need to integrate a diacritic restoration tool to improve the correctness of the written text.

**Automatic speech recogniser** Similar to the TTS is the automatic speech recognition (ASR) system that supports the recognition and translation of human speech to text. It has been reported that diacritic restoration can improve the performance of the ASR – also referred to as *speech-to-text* or *computer speech recognition* – by up to 12% in Romanian [80].

**Machine translation** There are references to the similarities between machine translation and diacritic restoration in literature. For example, both Schlippe *et al.* [92] and Do *et al.* [27] presented the restoration process as a machine translation problem emphasizing the commonalities in both tasks. However, as we have demonstrated in §4.1.4, it is possible to improve the performance of an existing machine translation system by passing a better marked text.

## 9.5 Main Contributions

In the course of this research work, we have reviewed a substantial number of previous works on diacritic restoration which informed our choice of approach to the task for Igbo. For the actual implementations, we have built the datasets for the task and also developed the baseline and enhanced methods. Embedding models were also introduced as a novel approach to solving the task. These contributions have been presented in different sections in this thesis but a brief description of each of them is given below:

**Review of previous works:** In this work presents an extensive report on the review of approaches to diacritic restorations as well as the works done on restoring Igbo diacritics. This will be a handy reference point for anyone who wants to embark of similar research effort.

**Building diacritic restoration datasets:** Since none existed, we thought it appropriate to create a standard dataset for the key diacritically ambiguous words found in our experimental data. This will enable other researchers to test their methods on our dataset, compare performances with ours and even improve on the dataset.

**Developing the baseline models:** For experimental purposes, we also had to develop basic standard baseline $n$-gram models for the task. Although we reported better performing methods in this work, we consider the baseline models useful for future explorations on the task that might require trading off any part of the pipeline for our enhanced approaches.

**Building classification models:** In the course of our work, we had modelled the problem as a classification task at some point. Although we believe now that the approach is not as efficient as the ones we discovered later, the details of the implementation and accompanying resources are also available for interested researchers.

**Using embedding models:** This is a key novel approach that we introduced to this task. To our knowledge, embedding models have not been previously applied to diacritic restoration. However, given that the task shares some similarities with the sense disambiguation task, we considered adapting some of the basic embedding based approaches to sense disambiguation and we achieved good results.

**Multiple embedding projection:** A major challenge to using embedding models for any NLP task in Igbo is the availability of large training data for the embedding models. So we considered approaches to taking advantage of the rich embedding models trained with a very large amount of English data by projecting trained English embedding models to an Igbo embedding space. There are several approaches to building multi-lingual and cross-lingual embedding models but we adopted a method similar to the one described in [42], which uses an Igbo-English alignment dictionary built from as small parallel corpora.

**Intrinsic evaluation for embeddings:** Embedding models are mostly generalizable and often not tailored to one task. For the purposes of extending the training and application of Igbo embedding models for other tasks, there will be a need to independently and intrinsically evaluate the models. Embedding models are often evaluated on a number of intrinsic tasks such as *word-similarity* and *analogy* tasks. Unfortunately, Igbo has no dataset for such tasks. We therefore defined a strategy and applied this to adapting the existing datasets for English to Igbo which gave rise to Igbo word-similarity and analogy datasets.

## 9.6 Conclusion

In this project, we set out with the intention to develop some resources, tools and techniques to support the NLP research for Igbo language and, by extension, low resource languages. We focused on diacritic restoration, which is a very essential task for languages like Igbo. This is because, beyond being a useful pre-processing task for creating good quality Igbo text, it also supports other NLP applications e.g. machine translation, speech recognition, text generation systems.

Focusing on diacritic restoration, we were able to delve into the different aspects of the major project e.g. building corpora, creating datasets, training and projecting embedding models as well as designing validation experiments for intrinsic evaluation tasks. Different methods and processes were deployed to build models that were applied to the task with varying degrees of success.

In conclusion, this work has delivered on most of the key objectives outlined at the beginning which include: review of literature, building datasets, defining the baseline model, developing diacritic restoration systems. We have also presented and published parts of our work at different stages at key NLP conferences like TSD, EACL, NAACL and COLING.

# Appendix A: Results of IDR with N-Grams

# Appendix A: Results of IDR with N-Grams

| Wordkey | Counts | No of Variants | 1-gram | 2-gram | 3-gram | 4-gram | 5-gram | BestScore (BS) | Best Model | Wdkey Improvement | Baseline Improvement | Error Reduction |
|---|---|---|---|---|---|---|---|---|---|---|---|---|
| agbago | 99 | 2 | 0.2677 | 0.2677 | 0.2677 | 0.2677 | 0.2677 | 0.2677 | 1-gram | 0.00% | -5.53% | 0.00% |
| buuru | 180 | 2 | 0.3333 | 0.4844 | 0.5153 | 0.5153 | 0.5153 | 0.5153 | 3-gram | 18.20% | 19.23% | 27.30% |
| juru | 306 | 2 | 0.2680 | 0.6027 | 0.6126 | 0.6126 | 0.6126 | 0.6126 | 3-gram | 34.46% | 28.96% | 47.08% |
| inu | 156 | 2 | 0.2885 | 0.7030 | 0.6895 | 0.6895 | 0.6895 | 0.7030 | 2-gram | 41.45% | 38.00% | 58.26% |
| o | 31446 | 2 | 0.3677 | 0.6242 | 0.6605 | 0.6925 | 0.7081 | 0.7081 | 5-gram | 34.04% | 38.51% | 53.84% |
| i | 5347 | 2 | 0.3369 | 0.6971 | 0.7015 | 0.7288 | 0.7307 | 0.7307 | 5-gram | 39.38% | 40.77% | 59.39% |
| ruru | 488 | 2 | 0.2520 | 0.7397 | 0.7466 | 0.7484 | 0.7484 | 0.7484 | 4-gram | 49.64% | 42.54% | 66.36% |
| ikpo | 133 | 2 | 0.3083 | 0.7051 | 0.7547 | 0.7547 | 0.7547 | 0.7547 | 3-gram | 44.64% | 43.17% | 64.54% |
| ama | 1353 | 3 | 0.1404 | 0.7376 | 0.7559 | 0.7575 | 0.7575 | 0.7575 | 4-gram | 61.71% | 43.45% | 71.79% |
| iru | 333 | 2 | 0.2658 | 0.7320 | 0.7586 | 0.7586 | 0.7586 | 0.7586 | 3-gram | 49.28% | 43.56% | 67.12% |
| ibu | 682 | 2 | 0.3211 | 0.7525 | 0.7634 | 0.7646 | 0.7737 | 0.7737 | 5-gram | 45.26% | 45.07% | 66.67% |
| onya | 160 | 3 | 0.1771 | 0.4274 | 0.7853 | 0.7853 | 0.7853 | 0.7853 | 3-gram | 60.82% | 46.23% | 73.91% |
| doo | 120 | 2 | 0.3500 | 0.7857 | 0.7222 | 0.7059 | 0.7059 | 0.7857 | 2-gram | 43.57% | 46.27% | 67.03% |
| ukwu | 1432 | 3 | 0.1741 | 0.7478 | 0.7908 | 0.7964 | 0.7964 | 0.7964 | 4-gram | 62.23% | 47.34% | 75.35% |
| iso | 201 | 2 | 0.2512 | 0.7957 | 0.7988 | 0.7988 | 0.7988 | 0.7988 | 3-gram | 54.76% | 47.58% | 73.13% |
| aku | 384 | 3 | 0.1632 | 0.8064 | 0.8054 | 0.8092 | 0.8092 | 0.8092 | 4-gram | 64.60% | 48.62% | 77.20% |
| oke | 2267 | 3 | 0.2607 | 0.7818 | 0.7969 | 0.8114 | 0.8076 | 0.8114 | 4-gram | 55.07% | 48.84% | 74.49% |
| otu | 5947 | 2 | 0.3332 | 0.7966 | 0.8076 | 0.8150 | 0.8168 | 0.8168 | 5-gram | 48.36% | 49.38% | 72.53% |
| igwe | 1392 | 4 | 0.1652 | 0.7787 | 0.8040 | 0.8252 | 0.8220 | 0.8252 | 4-gram | 66.00% | 50.22% | 79.06% |
| ju | 97 | 2 | 0.3711 | 0.8397 | 0.8284 | 0.8284 | 0.8284 | 0.8397 | 2-gram | 46.86% | 51.67% | 74.51% |
| okpukpu | 211 | 2 | 0.3555 | 0.8275 | 0.8410 | 0.8410 | 0.8410 | 0.8410 | 3-gram | 48.55% | 51.80% | 75.33% |
| too | 125 | 2 | 0.3560 | 0.8124 | 0.8447 | 0.8447 | 0.8447 | 0.8447 | 3-gram | 48.87% | 52.17% | 75.89% |
| akwa | 1191 | 3 | 0.1411 | 0.8340 | 0.8472 | 0.8551 | 0.8551 | 0.8551 | 4-gram | 71.40% | 53.21% | 83.13% |
| bu | 16999 | 2 | 0.3220 | 0.8386 | 0.8575 | 0.8630 | 0.8640 | 0.8640 | 5-gram | 54.20% | 54.10% | 79.94% |
| si | 9039 | 2 | 0.2866 | 0.8702 | 0.8753 | 0.8723 | 0.8745 | 0.8753 | 3-gram | 58.87% | 55.23% | 82.52% |
| nku | 285 | 2 | 0.3070 | 0.8889 | 0.8924 | 0.8924 | 0.8924 | 0.8924 | 3-gram | 58.54% | 56.94% | 84.47% |
| doro | 205 | 2 | 0.3195 | 0.8961 | 0.8961 | 0.8961 | 0.8961 | 0.8961 | 2-gram | 57.66% | 57.31% | 84.73% |
| odo | 154 | 2 | 0.3636 | 0.9185 | 0.9375 | 0.9375 | 0.9375 | 0.9375 | 3-gram | 57.39% | 61.45% | 90.18% |
| wuru | 112 | 2 | 0.3348 | 1.0000 | 1.0000 | 1.0000 | 1.0000 | 1.0000 | 2-gram | 66.52% | 67.70% | 100.00% |
| **Baseline = 1gram; %Error = 67.70%** | | | **32.30%** | **73.52%** | **75.82%** | **77.48%** | **78.15%** | **78.19%** | **Performance Analysis** | Model | Improvement | Error Reduction |
| **Best Model Counts** | | | **1** | **5** | **11** | **7** | **5** | | *Best Score* | ***5-gram*** | **45.86%** | **67.73%** |
| **Model Error Reduction:** | | | 0.00% | 60.88% | 64.28% | 66.74% | 67.73% | 67.79% | *Best Counts* | ***3-gram*** | **43.52%** | **64.28%** |

Table 1 **N-Gram Precision:** Table showing the full raw precision scores. [*Color code indicates the worst(red)-to-best(green) scores on metric performance results for the wordkey and global baselines.*]

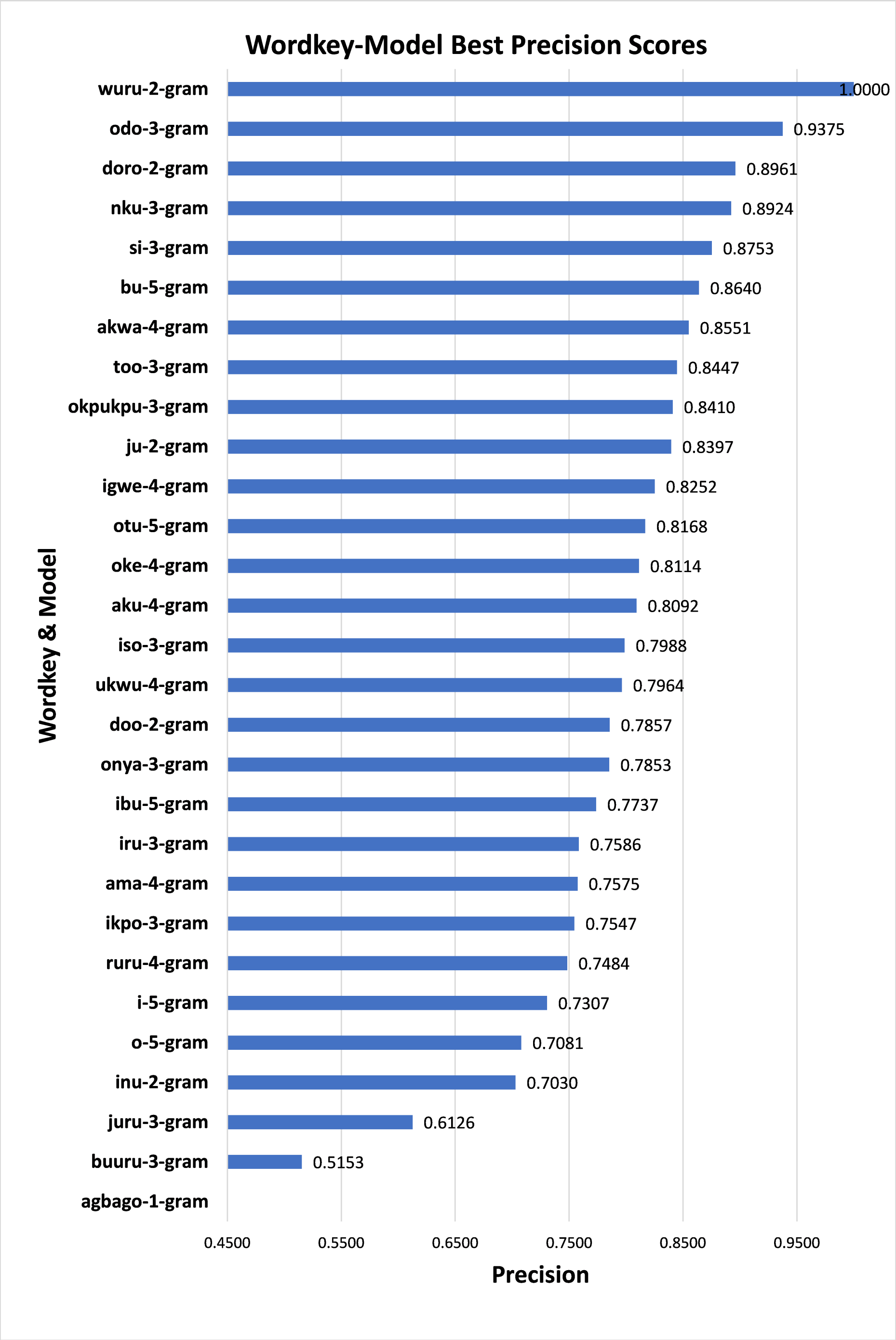

Fig. 1 **N-Gram Precision:** Graph showing the best performing model on for each of the wordkeys sorted in the ascending order of precision scores

# Appendix A: Results of IDR with N-Grams

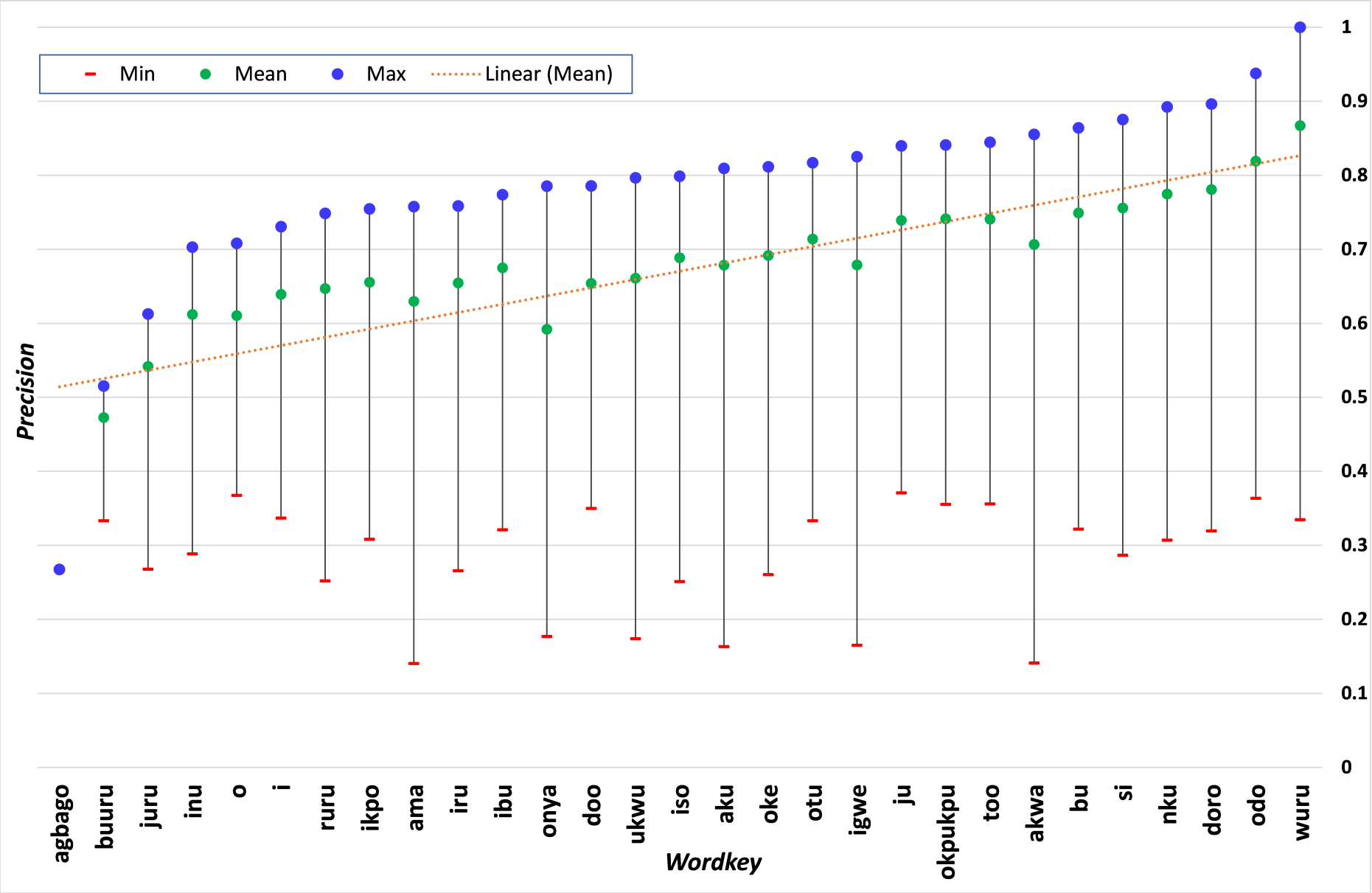

Fig. 2 **N-Gram Precision:** Graph showing minimum, mean and maximum precision scores and the linear trend line on the mean scores.

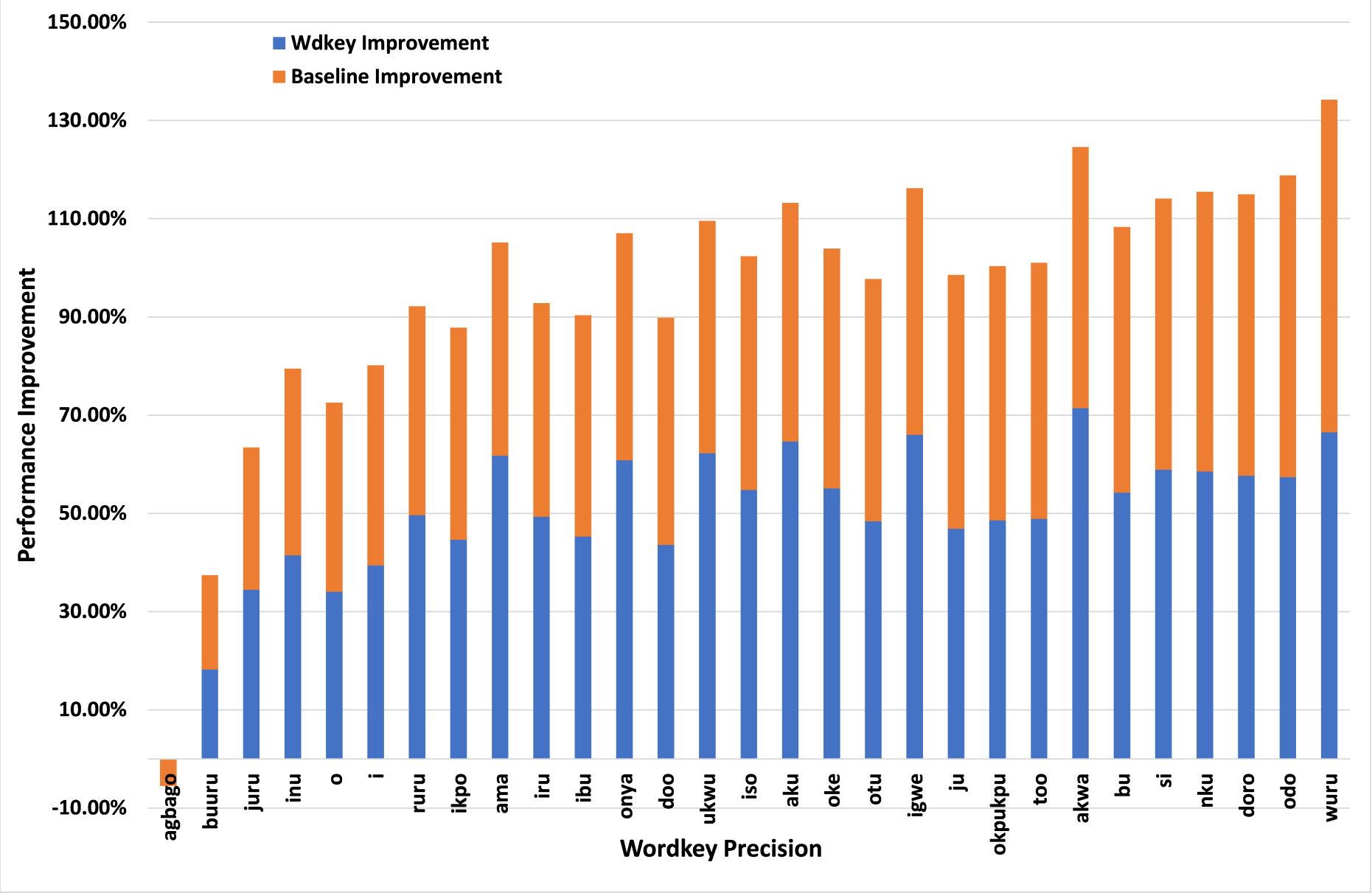

Fig. 3 **N-Gram Precision:** Graph showing the stacked column-chart of the improved precision scores on both the wordkey and the global baselines.

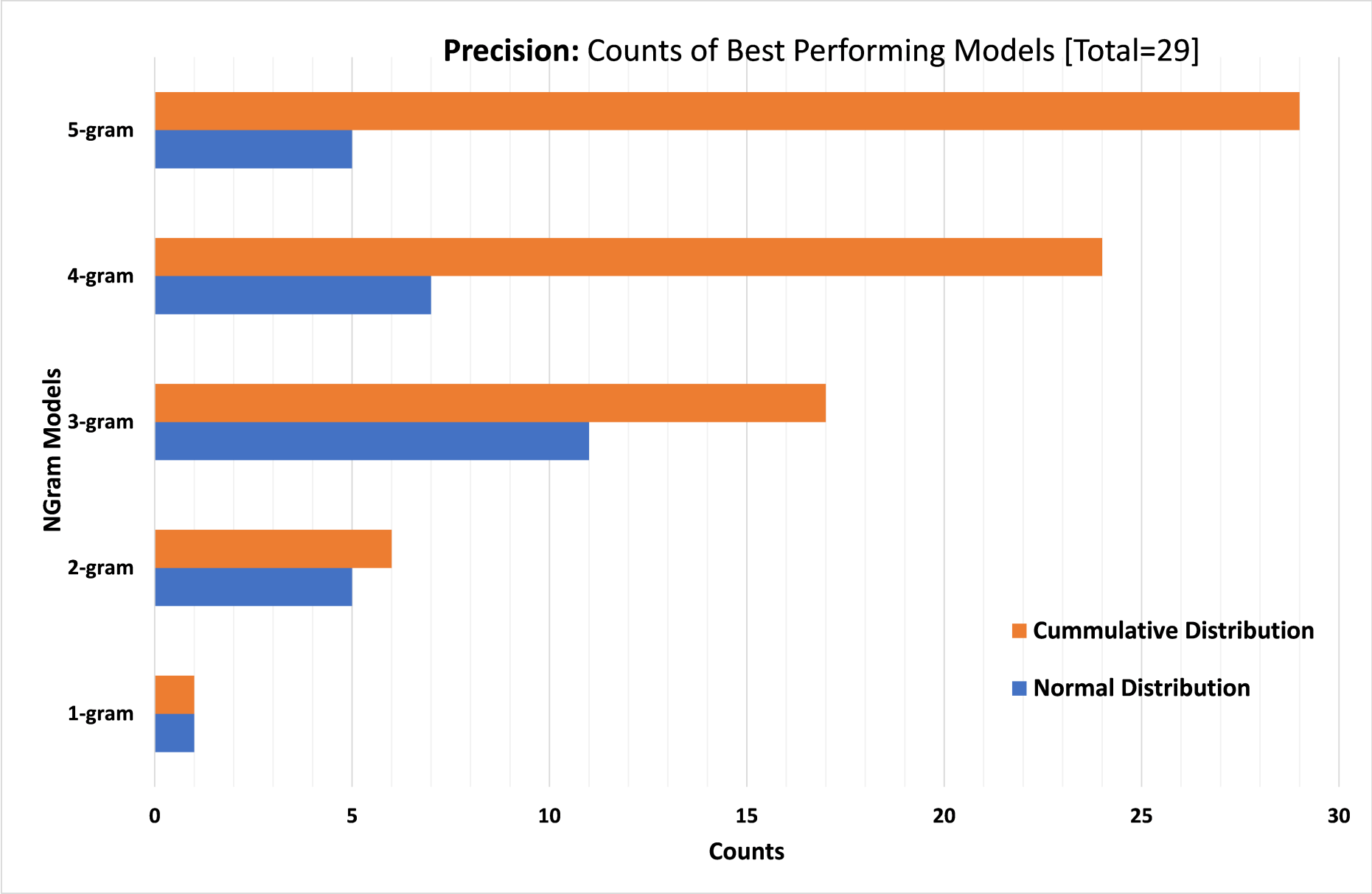

Fig. 4 **N-Gram Precision:** Graph showing the distribution of best precision scores across all ngram models.

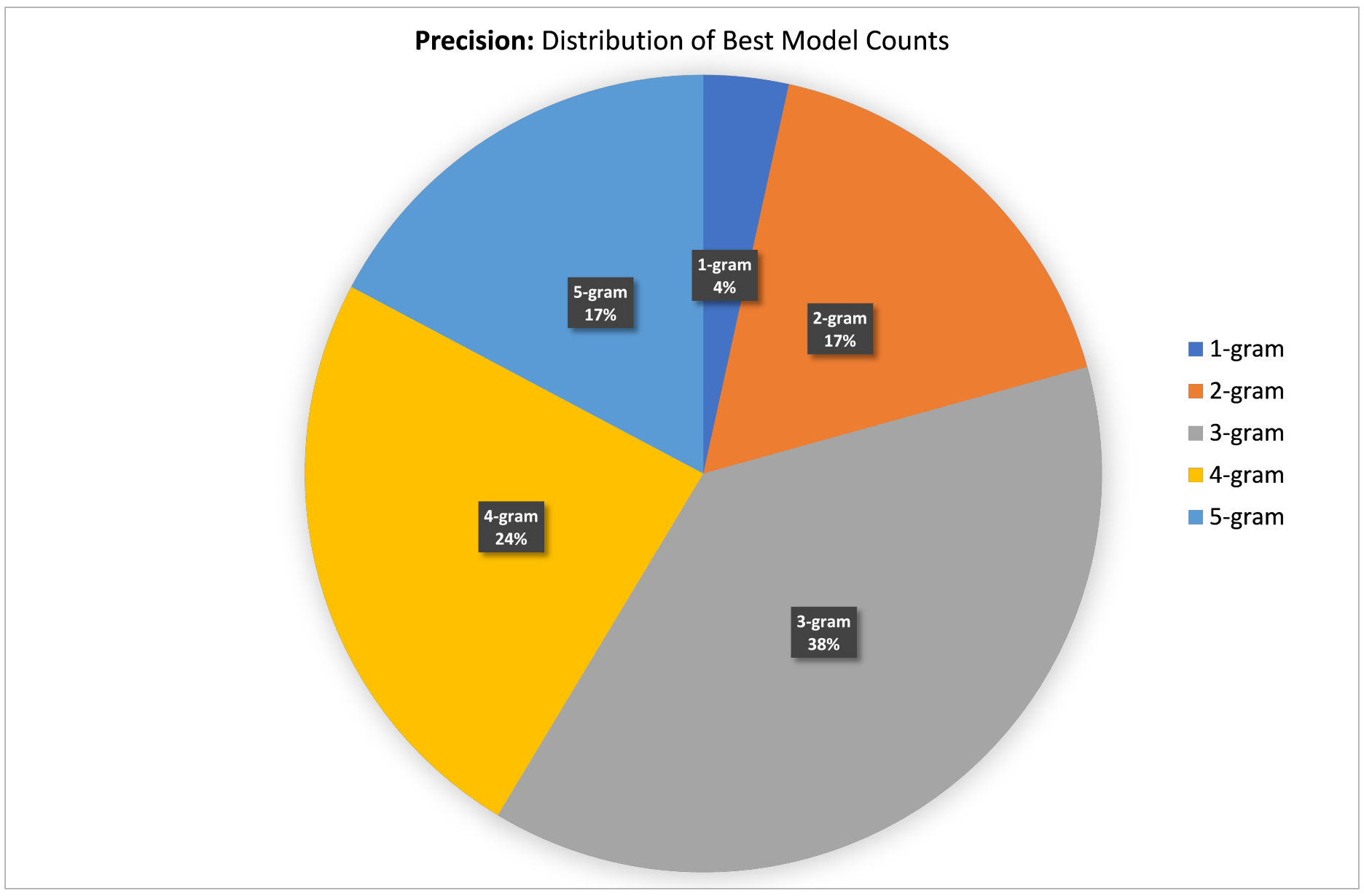

Fig. 5 **N-Gram Precision:** Piechart showing the percentage best distribution of the precision scores across all ngram models.

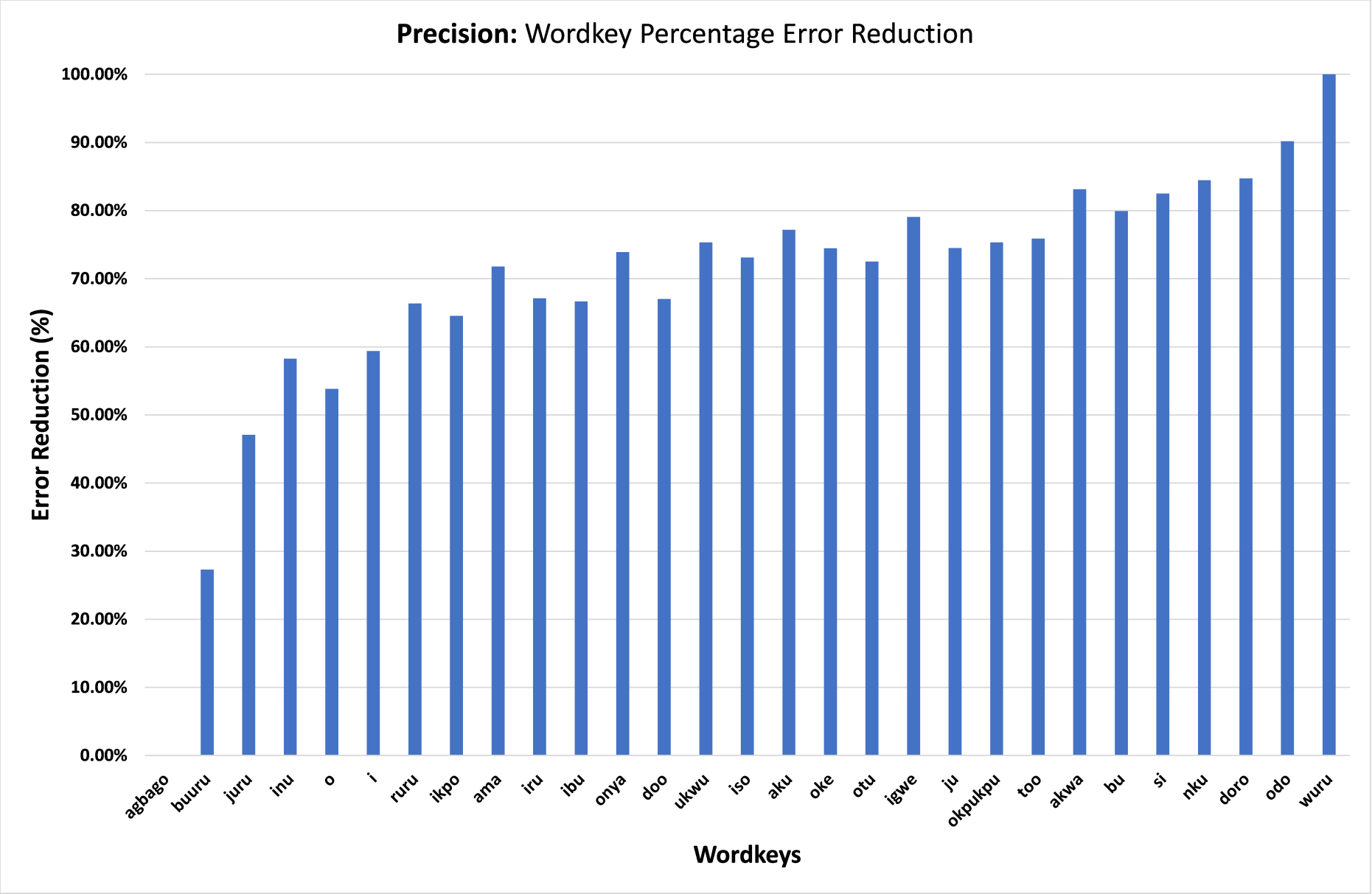

Fig. 6 **N-Gram Precision:** Graph showing the percentage precision error reduction on each wordkey by its best performing model.

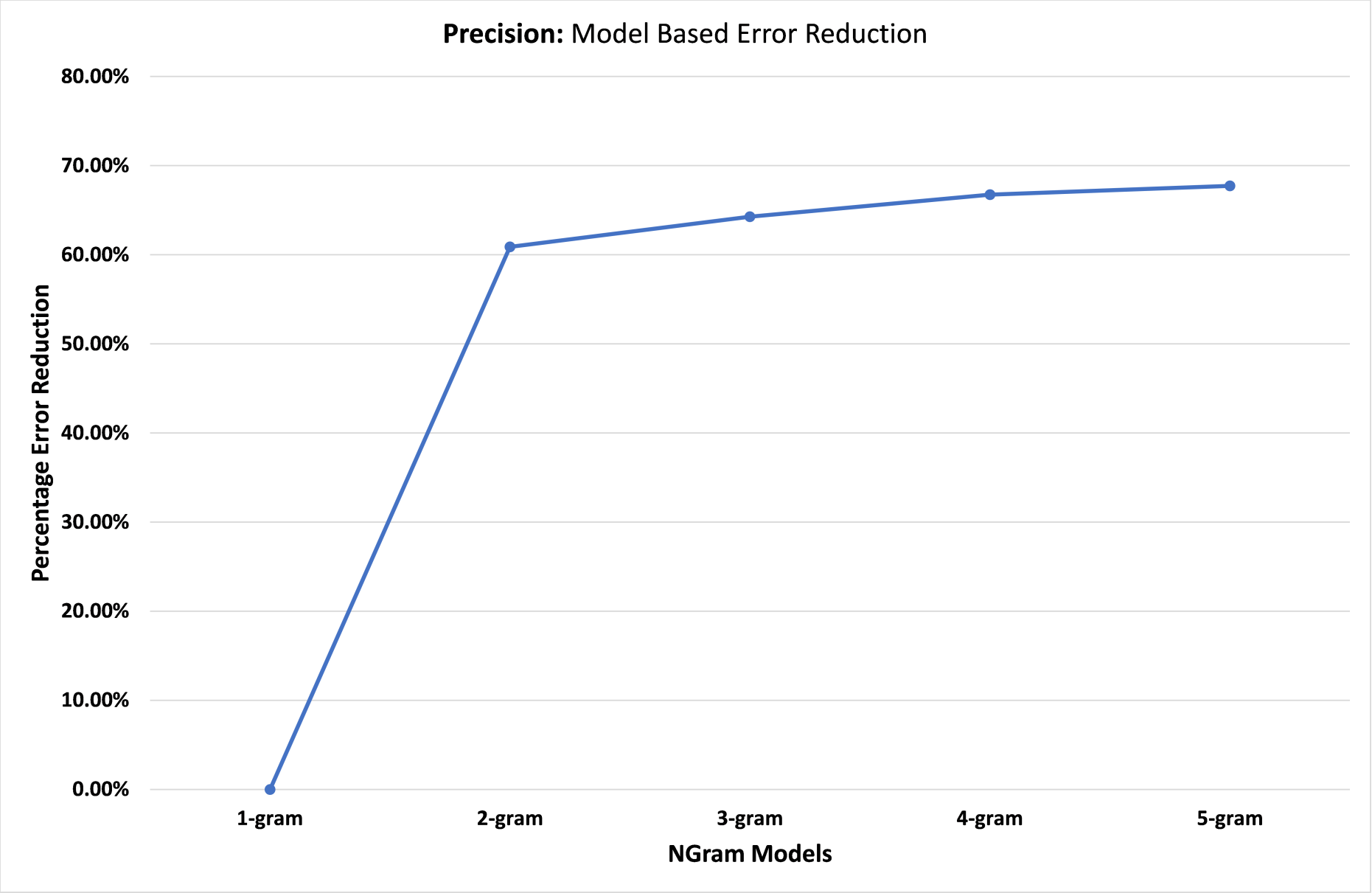

Fig. 7 **N-Gram Precision:** Column chart showing the percentage precision error reduction from the global baseline error by each model.

| Wordkey | Counts | No of Variants | 1-gram | 2-gram | 3-gram | 4-gram | 5-gram | BestScore (BS) | Best Model | Wdkey Improvement | Baseline Improvement | Error Reduction |
|---|---|---|---|---|---|---|---|---|---|---|---|---|
| onya | 160 | 3 | 0.3333 | 0.4446 | 0.4973 | 0.4973 | 0.4973 | 0.4973 | 3-gram | 16.40% | 1.56% | 24.60% |
| agbago | 99 | 2 | 0.5000 | 0.5000 | 0.5000 | 0.5000 | 0.5000 | 0.5000 | 1-gram | 0.00% | 1.83% | 0.00% |
| igwe | 1392 | 4 | 0.2500 | 0.4669 | 0.5045 | 0.5154 | 0.5155 | 0.5155 | 5-gram | 26.55% | 3.38% | 35.40% |
| buuru | 180 | 2 | 0.5000 | 0.4833 | 0.5167 | 0.5167 | 0.5167 | 0.5167 | 3-gram | 1.67% | 3.50% | 3.34% |
| juru | 306 | 2 | 0.5000 | 0.5771 | 0.5881 | 0.5881 | 0.5881 | 0.5881 | 3-gram | 8.81% | 10.64% | 17.62% |
| ama | 1353 | 3 | 0.3333 | 0.6255 | 0.6320 | 0.6326 | 0.6326 | 0.6326 | 4-gram | 29.93% | 15.09% | 44.89% |
| o | 31446 | 2 | 0.5000 | 0.5463 | 0.5909 | 0.6214 | 0.6360 | 0.6360 | 5-gram | 13.60% | 15.43% | 27.20% |
| ibu | 682 | 2 | 0.5000 | 0.6329 | 0.6422 | 0.6393 | 0.6416 | 0.6422 | 3-gram | 14.22% | 16.05% | 28.44% |
| oke | 2267 | 3 | 0.3333 | 0.5178 | 0.6311 | 0.6394 | 0.6428 | 0.6428 | 5-gram | 30.95% | 16.11% | 46.42% |
| doo | 120 | 2 | 0.5000 | 0.6488 | 0.6528 | 0.6468 | 0.6468 | 0.6528 | 3-gram | 15.28% | 17.11% | 30.56% |
| inu | 156 | 2 | 0.5000 | 0.6586 | 0.6551 | 0.6551 | 0.6551 | 0.6586 | 2-gram | 15.86% | 17.69% | 31.72% |
| i | 5347 | 2 | 0.5000 | 0.5805 | 0.6381 | 0.6611 | 0.6661 | 0.6661 | 5-gram | 16.61% | 18.44% | 33.22% |
| ikpo | 133 | 2 | 0.5000 | 0.6682 | 0.6926 | 0.6926 | 0.6926 | 0.6926 | 3-gram | 19.26% | 21.09% | 38.52% |
| iru | 333 | 2 | 0.5000 | 0.6863 | 0.7005 | 0.7005 | 0.7005 | 0.7005 | 3-gram | 20.05% | 21.88% | 40.10% |
| ju | 97 | 2 | 0.5000 | 0.7061 | 0.6861 | 0.6861 | 0.6861 | 0.7061 | 2-gram | 20.61% | 22.44% | 41.22% |
| okpukpu | 211 | 2 | 0.5000 | 0.6834 | 0.7080 | 0.7080 | 0.7080 | 0.7080 | 3-gram | 20.80% | 22.63% | 41.60% |
| aku | 384 | 3 | 0.3333 | 0.7054 | 0.7184 | 0.7202 | 0.7202 | 0.7202 | 4-gram | 38.69% | 23.85% | 58.03% |
| ruru | 488 | 2 | 0.5000 | 0.7308 | 0.7412 | 0.7433 | 0.7433 | 0.7433 | 4-gram | 24.33% | 26.16% | 48.66% |
| iso | 201 | 2 | 0.5000 | 0.7452 | 0.7502 | 0.7502 | 0.7502 | 0.7502 | 3-gram | 25.02% | 26.85% | 50.04% |
| otu | 5947 | 2 | 0.5000 | 0.7280 | 0.7528 | 0.7593 | 0.7600 | 0.7600 | 5-gram | 26.00% | 27.83% | 52.00% |
| ukwu | 1432 | 3 | 0.3333 | 0.7607 | 0.7734 | 0.7811 | 0.7811 | 0.7811 | 4-gram | 44.78% | 29.94% | 67.17% |
| nku | 285 | 2 | 0.5000 | 0.7727 | 0.7818 | 0.7818 | 0.7818 | 0.7818 | 3-gram | 28.18% | 30.01% | 56.36% |
| akwa | 1191 | 3 | 0.3333 | 0.7755 | 0.7873 | 0.7950 | 0.7950 | 0.7950 | 4-gram | 46.17% | 31.33% | 69.25% |
| odo | 154 | 2 | 0.5000 | 0.8051 | 0.8095 | 0.8095 | 0.8095 | 0.8095 | 3-gram | 30.95% | 32.78% | 61.90% |
| too | 125 | 2 | 0.5000 | 0.7580 | 0.8135 | 0.8135 | 0.8135 | 0.8135 | 3-gram | 31.35% | 33.18% | 62.70% |
| si | 9039 | 2 | 0.5000 | 0.8093 | 0.8176 | 0.8211 | 0.8227 | 0.8227 | 5-gram | 32.27% | 34.10% | 64.54% |
| doro | 205 | 2 | 0.5000 | 0.8331 | 0.8331 | 0.8331 | 0.8331 | 0.8331 | 2-gram | 33.31% | 35.14% | 66.62% |
| bu | 16999 | 2 | 0.5000 | 0.8390 | 0.8568 | 0.8625 | 0.8634 | 0.8634 | 5-gram | 36.34% | 38.17% | 72.68% |
| wuru | 112 | 2 | 0.5000 | 1.0000 | 1.0000 | 1.0000 | 1.0000 | 1.0000 | 2-gram | 50.00% | 51.83% | 100.00% |
| **Baseline = 1gram; %Error = 51.83%** | | | **48.17%** | **66.67%** | **69.94%** | **71.55%** | **72.20%** | **72.21%** | **Performance Analysis** | Model | Improvement | Error Reduction |
| **Best Model Counts** | | | **1** | **4** | **12** | **5** | **7** | | *Best Score* | ***5-gram*** | **24.03%** | **46.37%** |
| **Model Error Reduction:** | | | 0.00% | 35.69% | 42.00% | 45.11% | 46.37% | 46.38% | *Best Counts* | ***3-gram*** | **21.77%** | **42.00%** |

Table 2 **N-Gram Recall:** Table showing the full raw recall scores. [*Color code indicates the worst(red)-to-best(green) scores on metric performance results for the wordkey and global baselines.*]

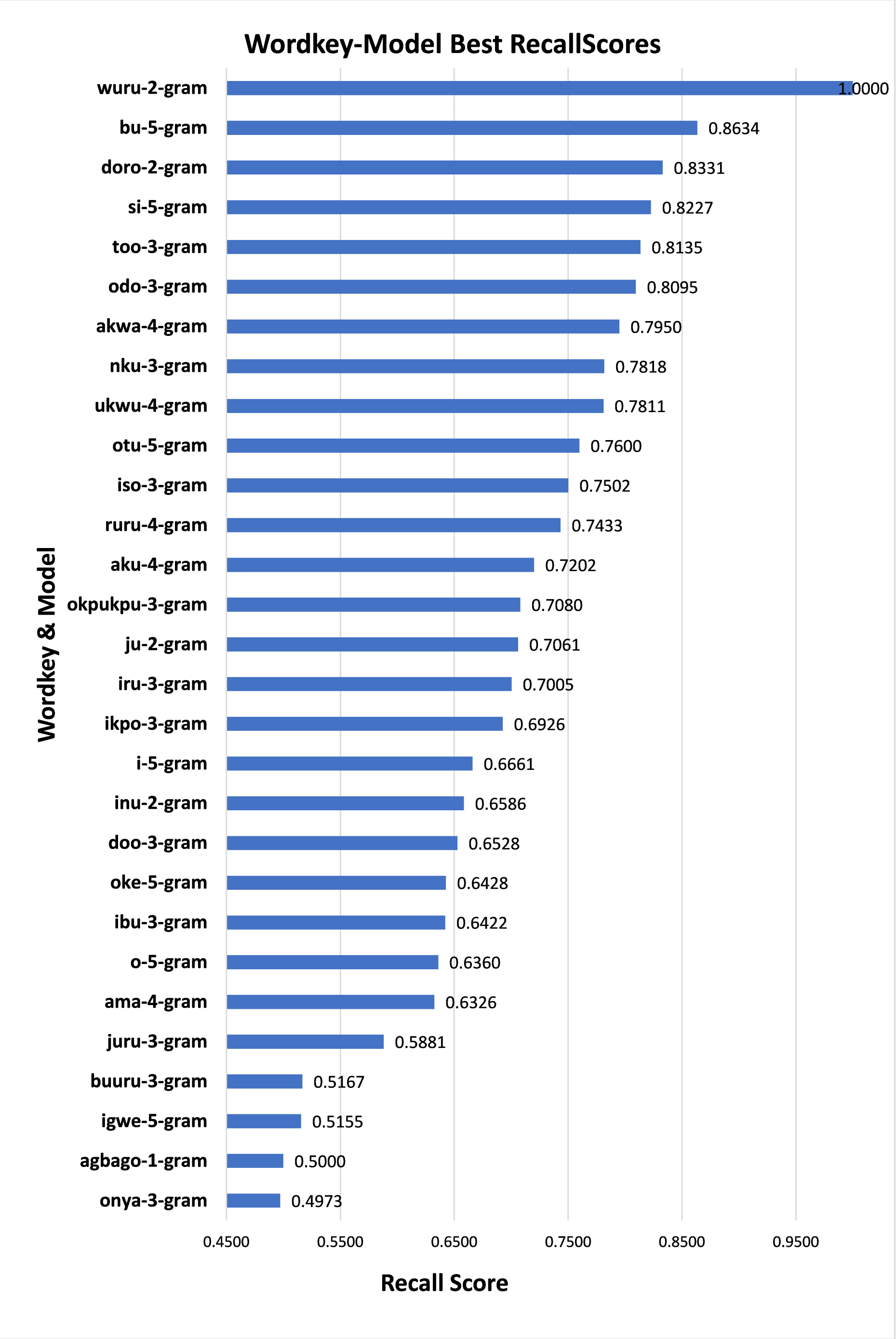

Fig. 8 **N-Gram Recall:** Figure showing the best performing model on for each of the wordkeys sorted in the ascending order of recall scores

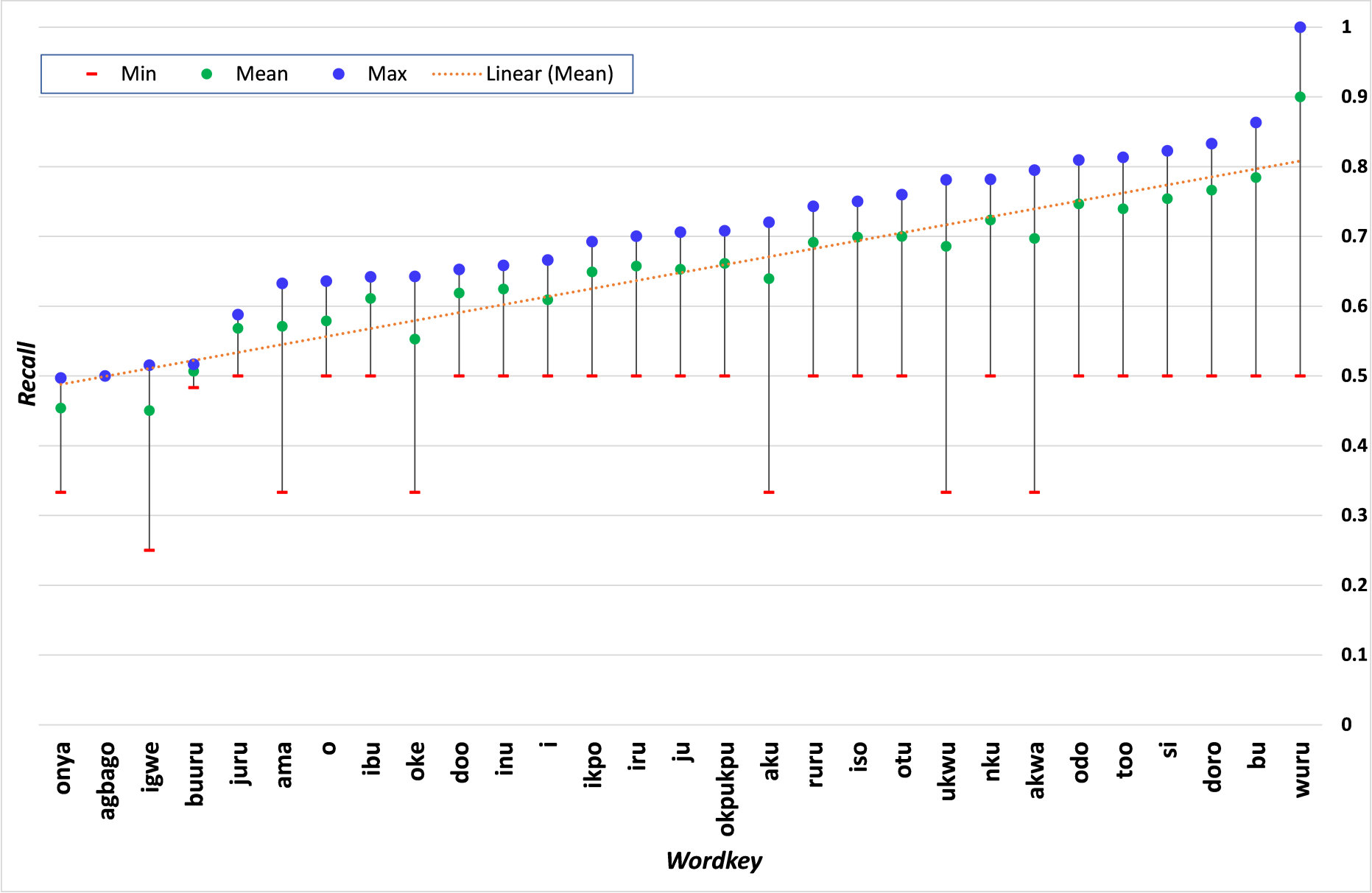

Fig. 9 **N-Gram Recall:** Graph showing minimum, mean and maximum recall scores and the linear trend line on the mean scores.

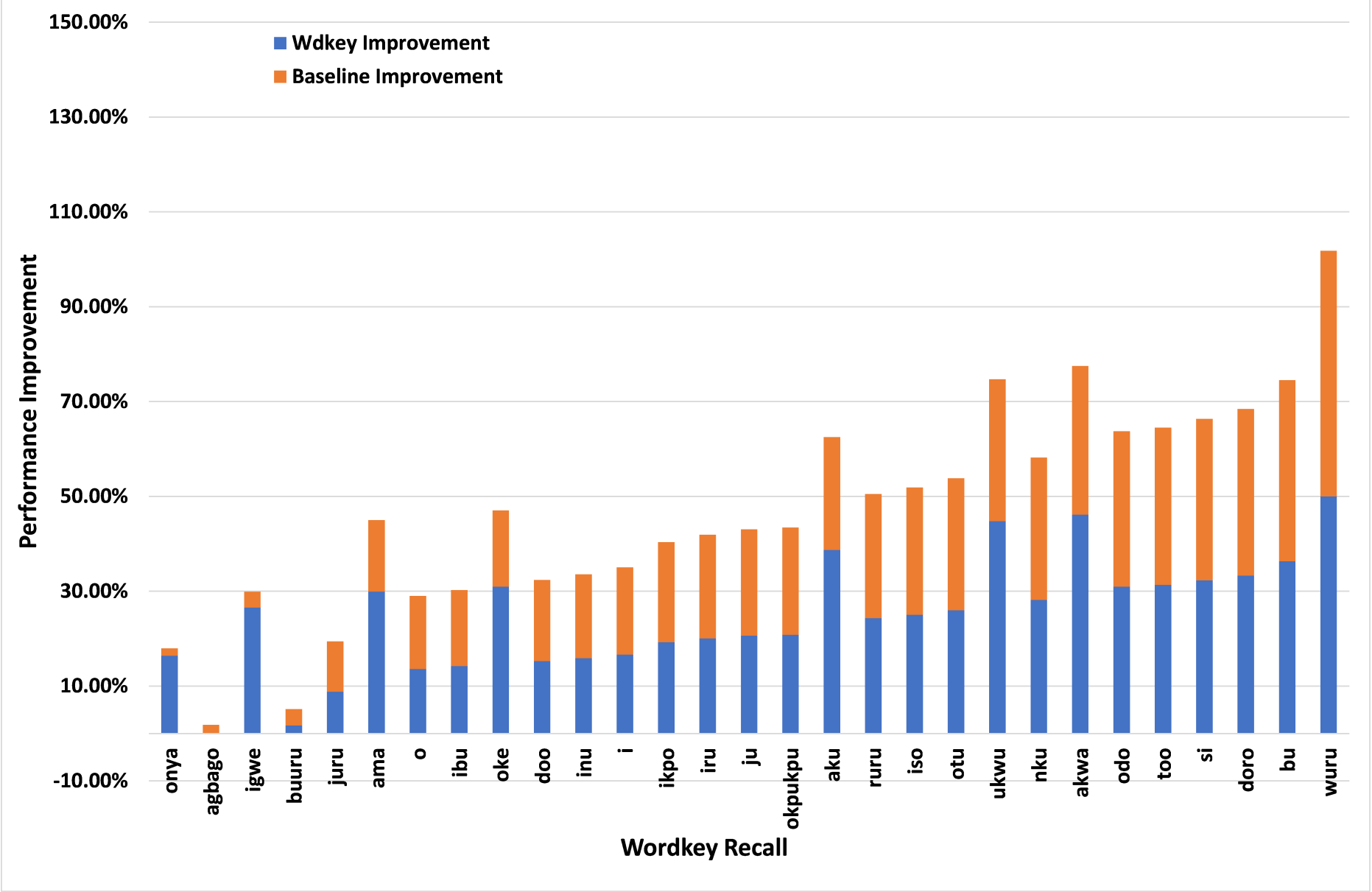

Fig. 10 **N-Gram Recall:** Graph showing the stacked column-chart of the improved recall scores on both the wordkey and the global baselines.

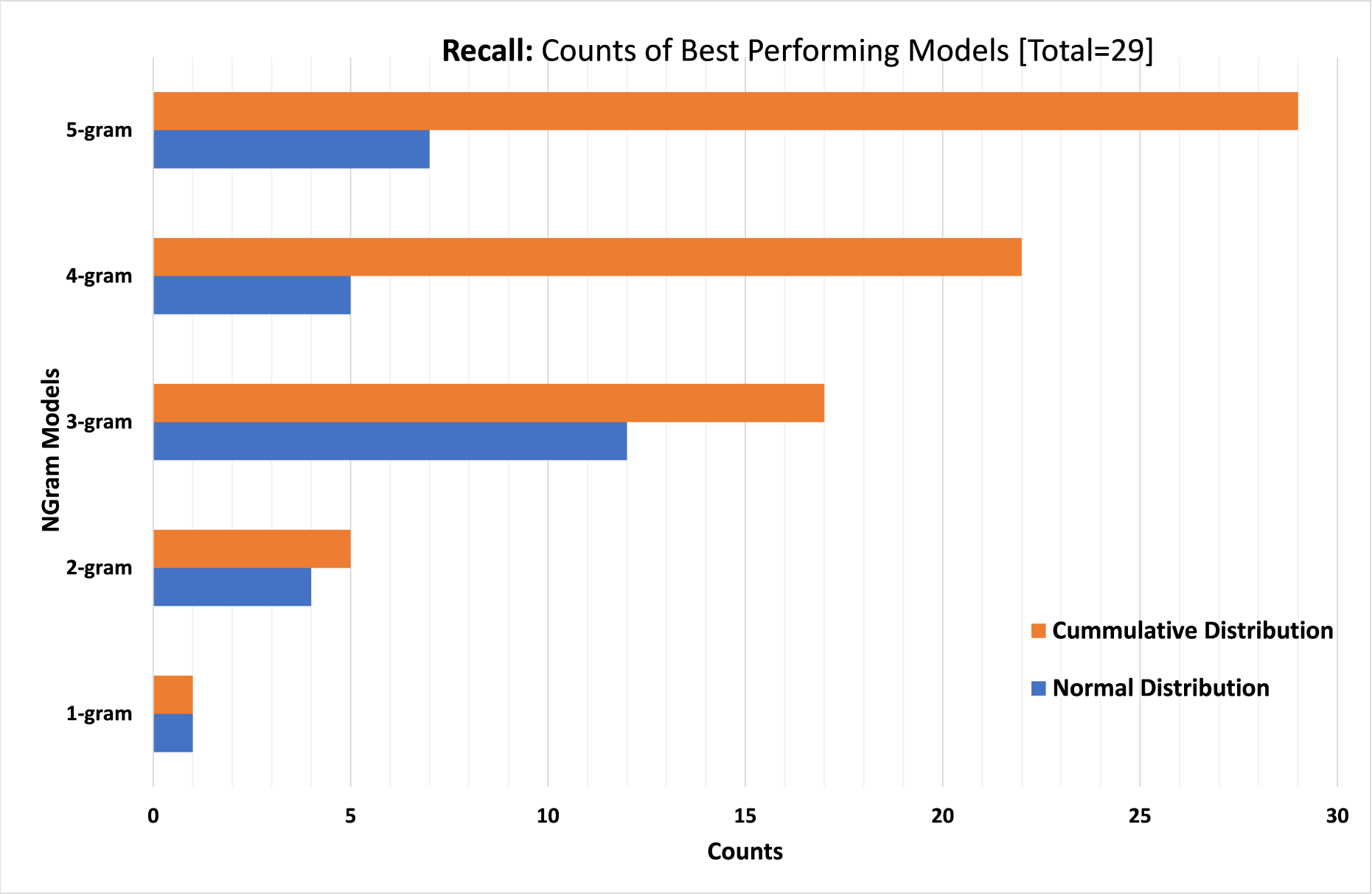

Fig. 11 **N-Gram Recall:** Graph showing the distribution of best recall scores across all ngram models.

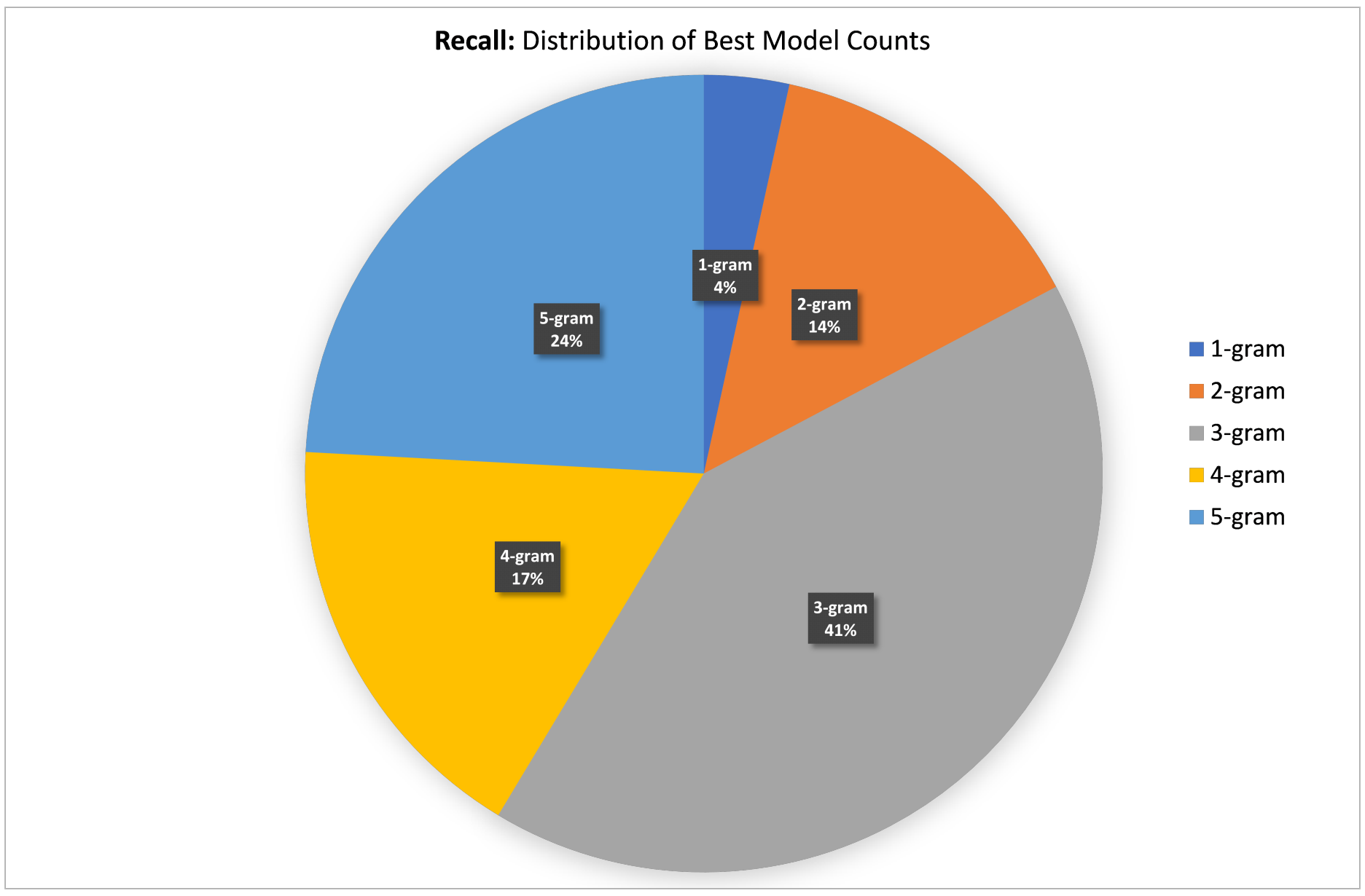

Fig. 12 **N-Gram Recall:** Piechart showing the percentage best distribution of the recall scores across all ngram models.

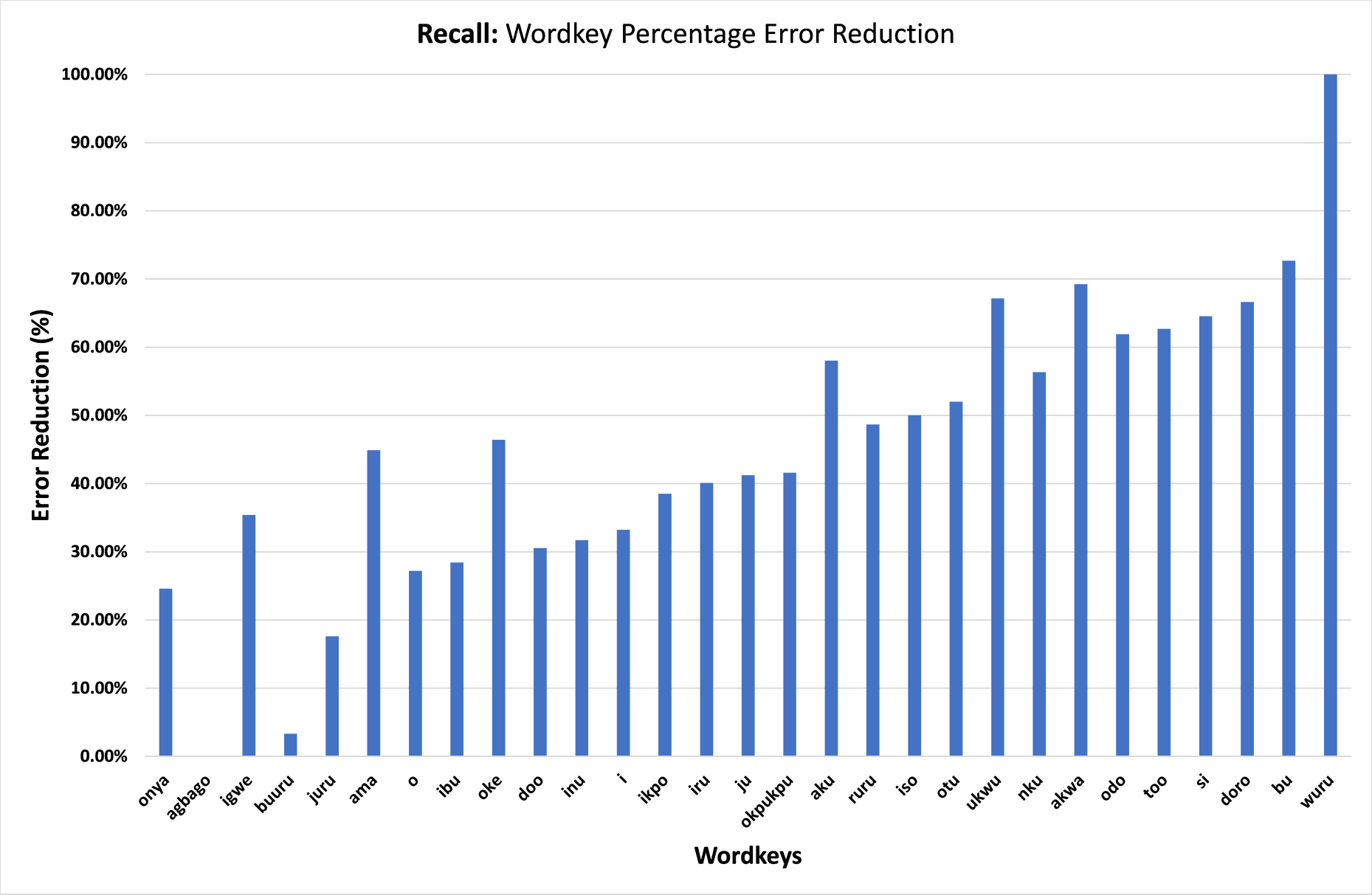

Fig. 13 **N-Gram Recall:** Graph showing the percentage recall error reduction on each wordkey by its best performing model.

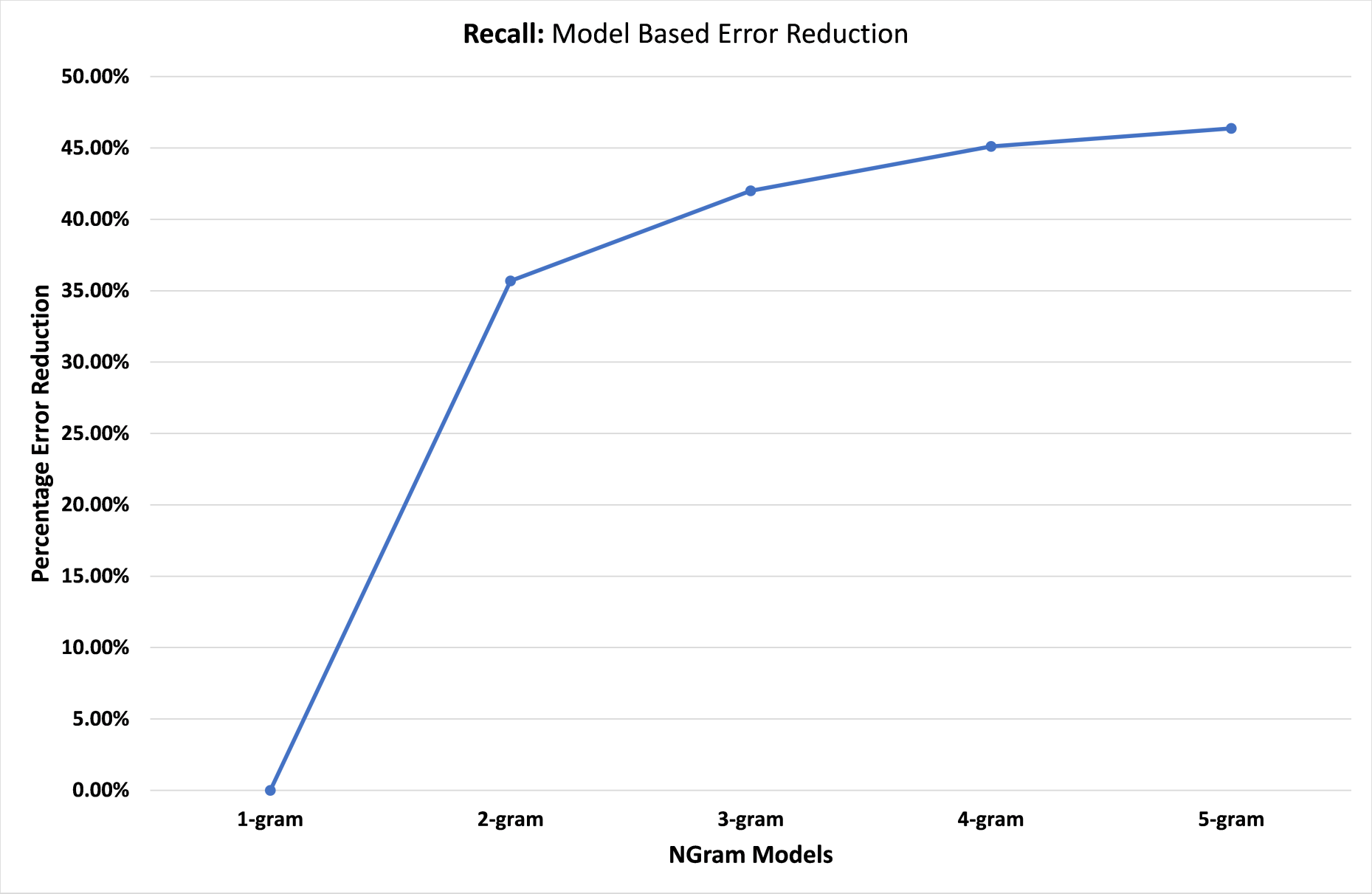

Fig. 14 **N-Gram Recall:** Column chart showing the percentage recall error reduction from the global baseline error by each model.

| Wordkey | Counts | No of Variants | 1-gram | 2-gram | 3-gram | 4-gram | 5-gram | BestScore (BS) | Best Model | Wdkey Improvement | Baseline Improvement | Error Reduction |
|---|---|---|---|---|---|---|---|---|---|---|---|---|
| agbago | 99 | 2 | 0.3487 | 0.3487 | 0.3487 | 0.3487 | 0.3487 | 0.3487 | 1-gram | 0.00% | -3.68% | 0.00% |
| buuru | 180 | 2 | 0.4000 | 0.4823 | 0.5126 | 0.5126 | 0.5126 | 0.5126 | 3-gram | 11.26% | 12.71% | 18.77% |
| onya | 160 | 3 | 0.2313 | 0.4327 | 0.5166 | 0.5166 | 0.5166 | 0.5166 | 3-gram | 28.53% | 13.11% | 37.11% |
| juru | 306 | 2 | 0.3489 | 0.5582 | 0.5732 | 0.5732 | 0.5732 | 0.5732 | 3-gram | 22.43% | 18.77% | 34.45% |
| igwe | 1392 | 4 | 0.1990 | 0.5335 | 0.5768 | 0.5905 | 0.5904 | 0.5905 | 4-gram | 39.15% | 20.50% | 48.88% |
| ama | 1353 | 3 | 0.1976 | 0.6100 | 0.6151 | 0.6159 | 0.6159 | 0.6159 | 4-gram | 41.83% | 23.04% | 52.13% |
| ibu | 682 | 2 | 0.3911 | 0.6317 | 0.6436 | 0.6396 | 0.6421 | 0.6436 | 3-gram | 25.25% | 25.81% | 41.47% |
| o | 31446 | 2 | 0.4237 | 0.5330 | 0.5972 | 0.6342 | 0.6511 | 0.6511 | 5-gram | 22.74% | 26.56% | 39.46% |
| inu | 156 | 2 | 0.3659 | 0.6571 | 0.6545 | 0.6545 | 0.6545 | 0.6571 | 2-gram | 29.12% | 27.16% | 45.92% |
| doo | 120 | 2 | 0.4118 | 0.6639 | 0.6664 | 0.6590 | 0.6590 | 0.6664 | 3-gram | 25.46% | 28.09% | 43.28% |
| i | 5347 | 2 | 0.4026 | 0.5673 | 0.6465 | 0.6728 | 0.6783 | 0.6783 | 5-gram | 27.57% | 29.28% | 46.15% |
| oke | 2267 | 3 | 0.2926 | 0.5753 | 0.6776 | 0.6895 | 0.6921 | 0.6921 | 5-gram | 39.95% | 30.66% | 56.47% |
| iru | 333 | 2 | 0.3471 | 0.6788 | 0.6922 | 0.6922 | 0.6922 | 0.6922 | 3-gram | 34.51% | 30.67% | 52.86% |
| ikpo | 133 | 2 | 0.3814 | 0.6735 | 0.7003 | 0.7003 | 0.7003 | 0.7003 | 3-gram | 31.89% | 31.48% | 51.55% |
| okpukpu | 211 | 2 | 0.4155 | 0.7081 | 0.7359 | 0.7359 | 0.7359 | 0.7359 | 3-gram | 32.04% | 35.04% | 54.82% |
| ju | 97 | 2 | 0.4260 | 0.7382 | 0.7161 | 0.7161 | 0.7161 | 0.7382 | 2-gram | 31.22% | 35.27% | 54.39% |
| iso | 201 | 2 | 0.3344 | 0.7347 | 0.7404 | 0.7404 | 0.7404 | 0.7404 | 3-gram | 40.60% | 35.49% | 61.00% |
| ruru | 488 | 2 | 0.3351 | 0.7288 | 0.7402 | 0.7424 | 0.7424 | 0.7424 | 4-gram | 40.73% | 35.69% | 61.26% |
| aku | 384 | 3 | 0.2191 | 0.7292 | 0.7437 | 0.7457 | 0.7457 | 0.7457 | 4-gram | 52.66% | 36.02% | 67.44% |
| ukwu | 1432 | 3 | 0.2287 | 0.7367 | 0.7704 | 0.7770 | 0.7770 | 0.7770 | 4-gram | 54.83% | 39.15% | 71.09% |
| otu | 5947 | 2 | 0.3999 | 0.7453 | 0.7693 | 0.7762 | 0.7772 | 0.7772 | 5-gram | 37.73% | 39.17% | 62.87% |
| nku | 285 | 2 | 0.3804 | 0.7904 | 0.8002 | 0.8002 | 0.8002 | 0.8002 | 3-gram | 41.98% | 41.47% | 67.75% |
| akwa | 1191 | 3 | 0.1982 | 0.7931 | 0.8050 | 0.8129 | 0.8129 | 0.8129 | 4-gram | 61.47% | 42.74% | 76.67% |
| too | 125 | 2 | 0.4159 | 0.7772 | 0.8267 | 0.8267 | 0.8267 | 0.8267 | 3-gram | 41.08% | 44.12% | 70.33% |
| si | 9039 | 2 | 0.3644 | 0.8190 | 0.8275 | 0.8307 | 0.8325 | 0.8325 | 5-gram | 46.81% | 44.70% | 73.65% |
| odo | 154 | 2 | 0.4211 | 0.8412 | 0.8490 | 0.8490 | 0.8490 | 0.8490 | 3-gram | 42.79% | 46.35% | 73.92% |
| doro | 205 | 2 | 0.3899 | 0.8523 | 0.8523 | 0.8523 | 0.8523 | 0.8523 | 2-gram | 46.24% | 46.68% | 75.79% |
| bu | 16999 | 2 | 0.3918 | 0.8347 | 0.8533 | 0.8584 | 0.8593 | 0.8593 | 5-gram | 46.75% | 47.38% | 76.87% |
| wuru | 112 | 2 | 0.4011 | 1.0000 | 1.0000 | 1.0000 | 1.0000 | 1.0000 | 2-gram | 59.89% | 61.45% | 100.00% |
| **Baseline = 1gram; %Error = 61.45%** | | | **38.55%** | **66.45%** | **70.66%** | **72.55%** | **73.30%** | **73.31%** | **Performance Analysis** | Model | Improvement | Error Reduction |
| **Best Model Counts** | | | **1** | **4** | **12** | **6** | **6** | | *Best Score* | ***5-gram*** | **34.75%** | **56.55%** |
| **Model Error Reduction:** | | | 0.00% | 45.41% | 52.26% | 55.33% | 56.55% | 56.56% | *Best Counts* | ***3-gram*** | **32.12%** | **52.26%** |

Table 3 **N-Gram F1-Score:** Table showing the full raw f1 scores. [*Color code indicates the worst(red)-to-best(green) scores on metric performance results for the wordkey and global baselines.*]

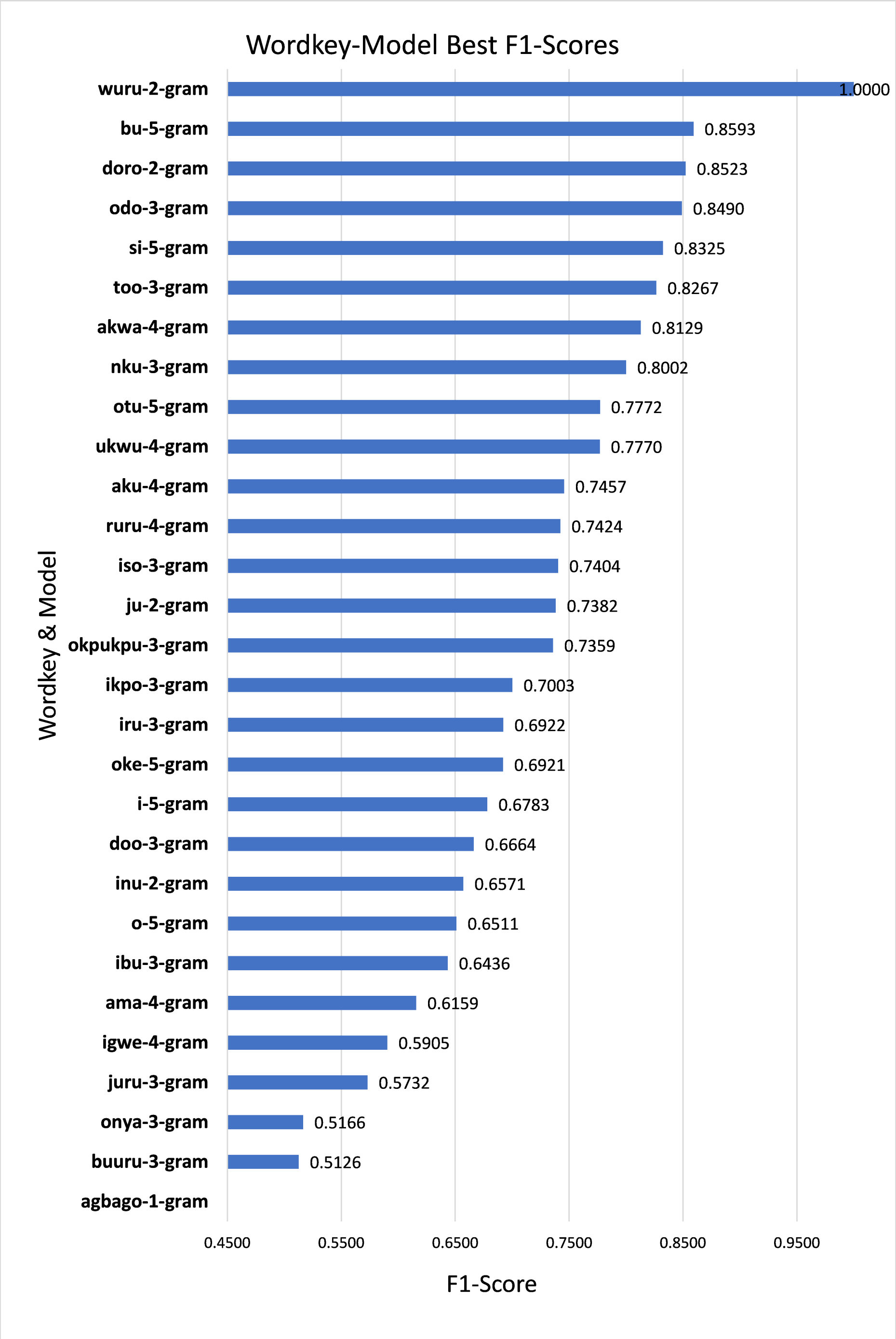

Fig. 15 **N-Gram F1-Score:** Figure showing the best performing model on for each of the wordkeys sorted in the ascending order of f1 scores

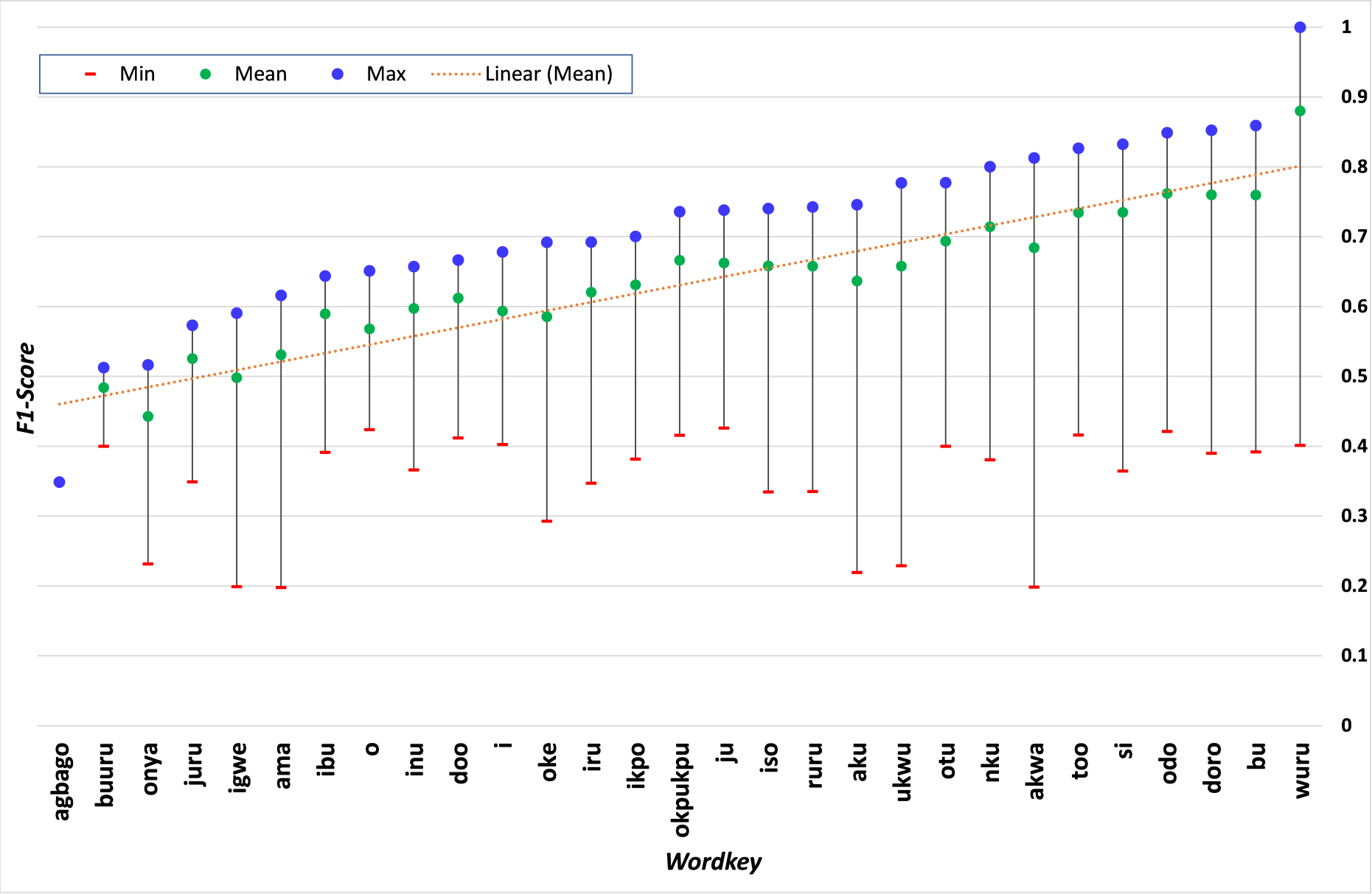

Fig. 16 **N-Gram F1-Score:** Graph showing minimum, mean and maximum F1 scores and the linear trend line on the mean scores.

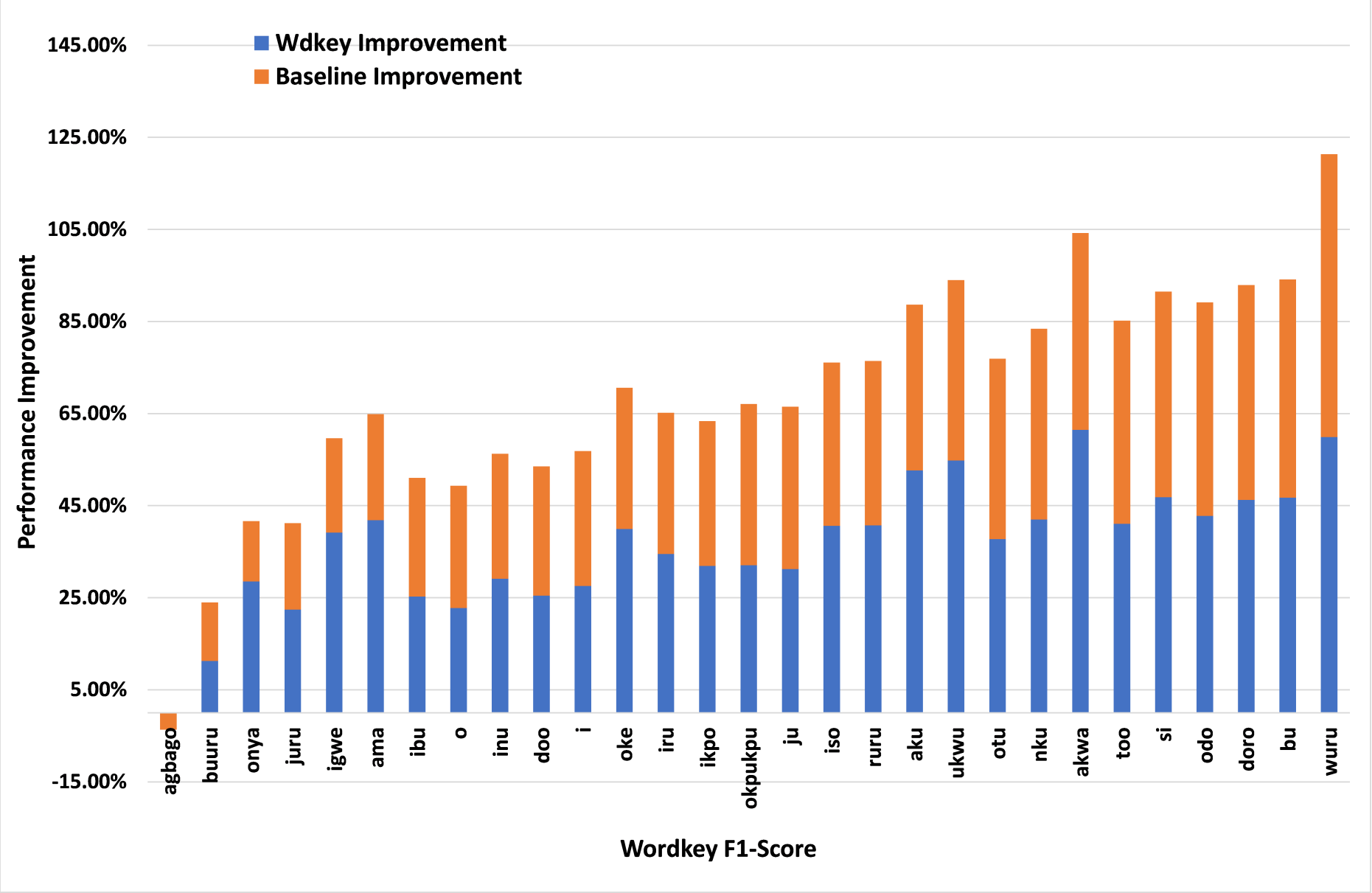

Fig. 17 **N-Gram F1-Score:** Graph showing the stacked column-chart of the improved F1 scores on both the wordkey and the global baselines.

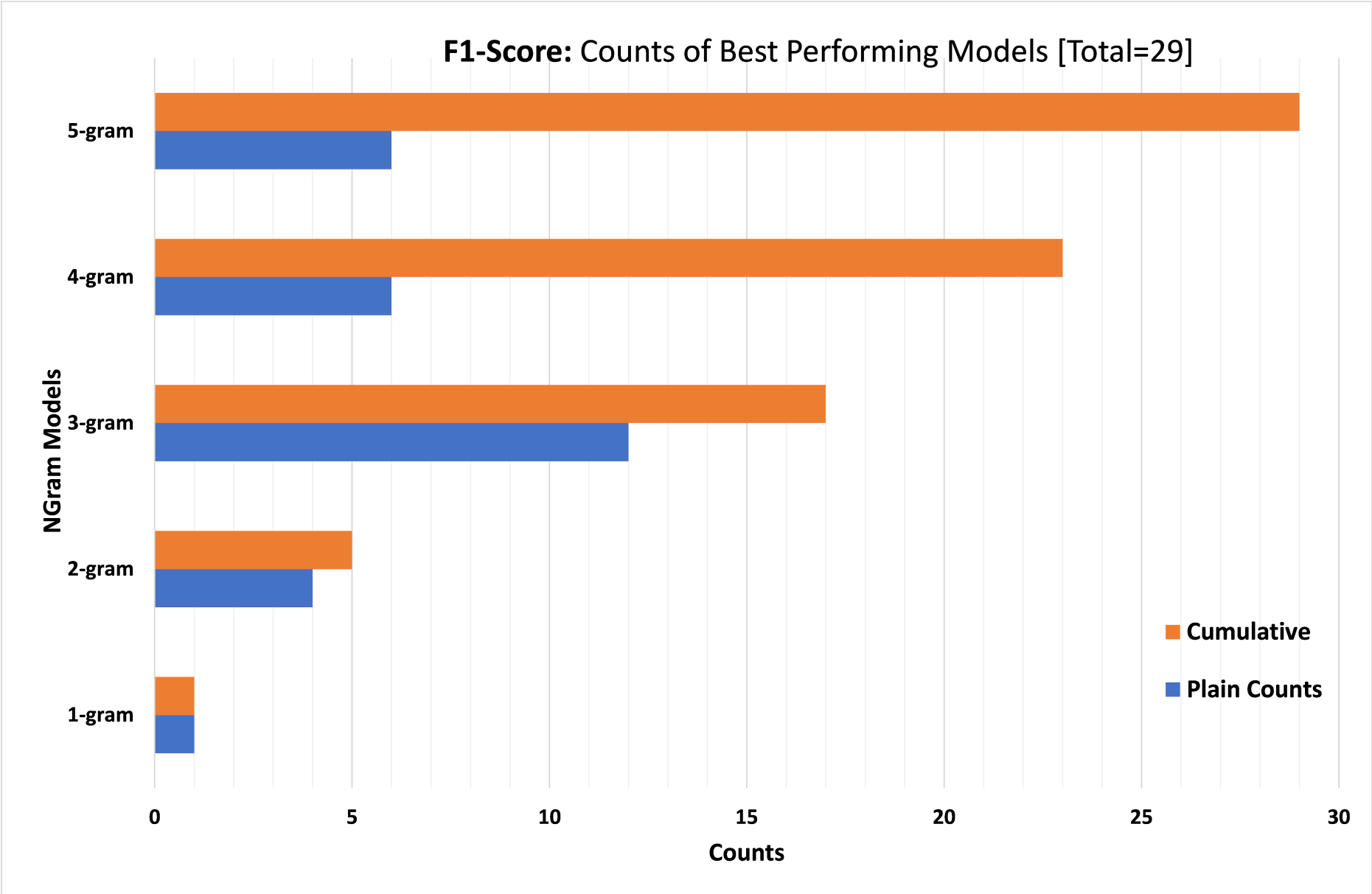

Fig. 18 **N-Gram F1-Score:** Graph showing the distribution of best f1 scores across all ngram models.

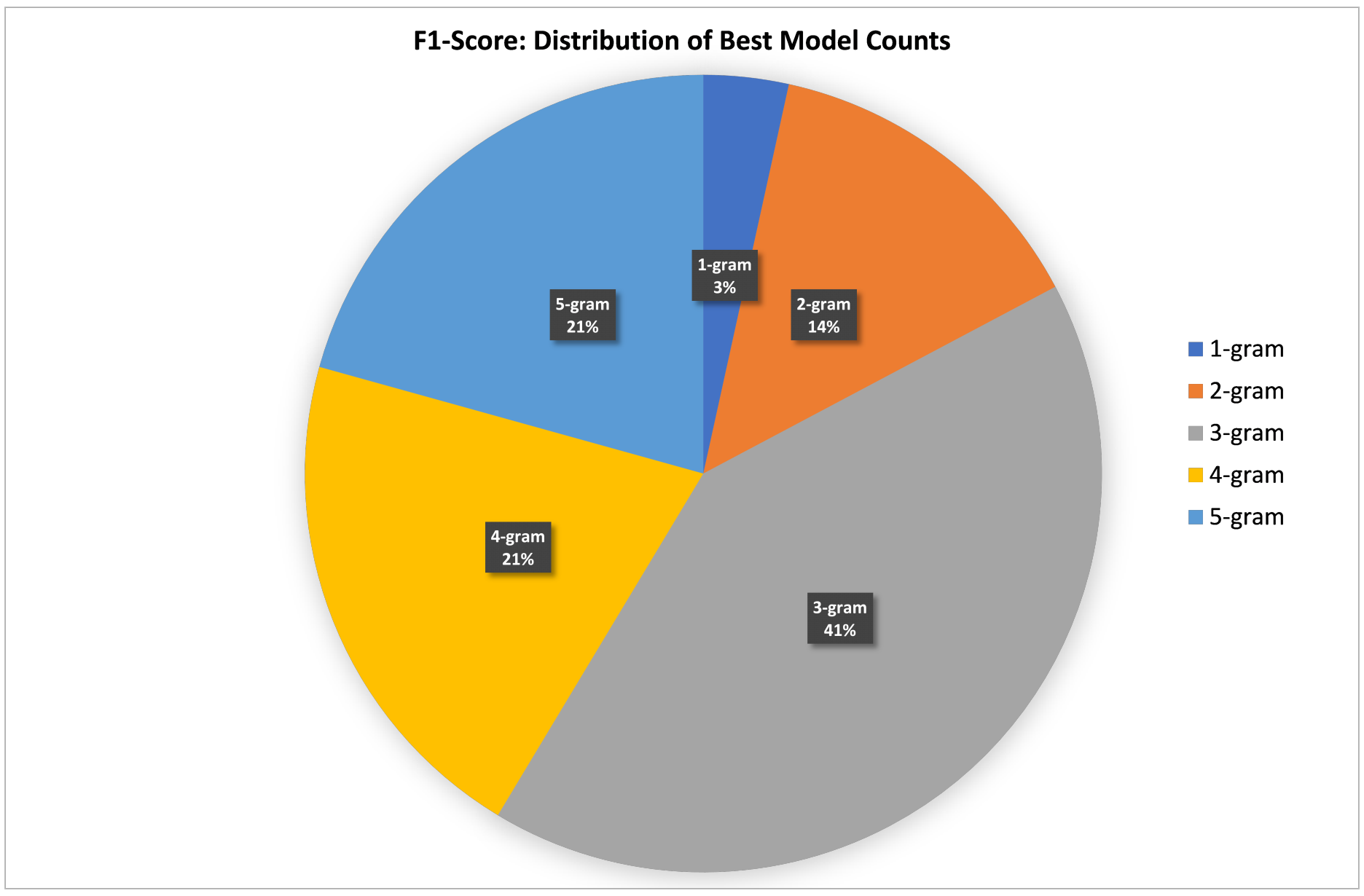

Fig. 19 **N-Gram F1-Score:** Piechart showing the percentage best distribution of the f1 scores across all ngram models.

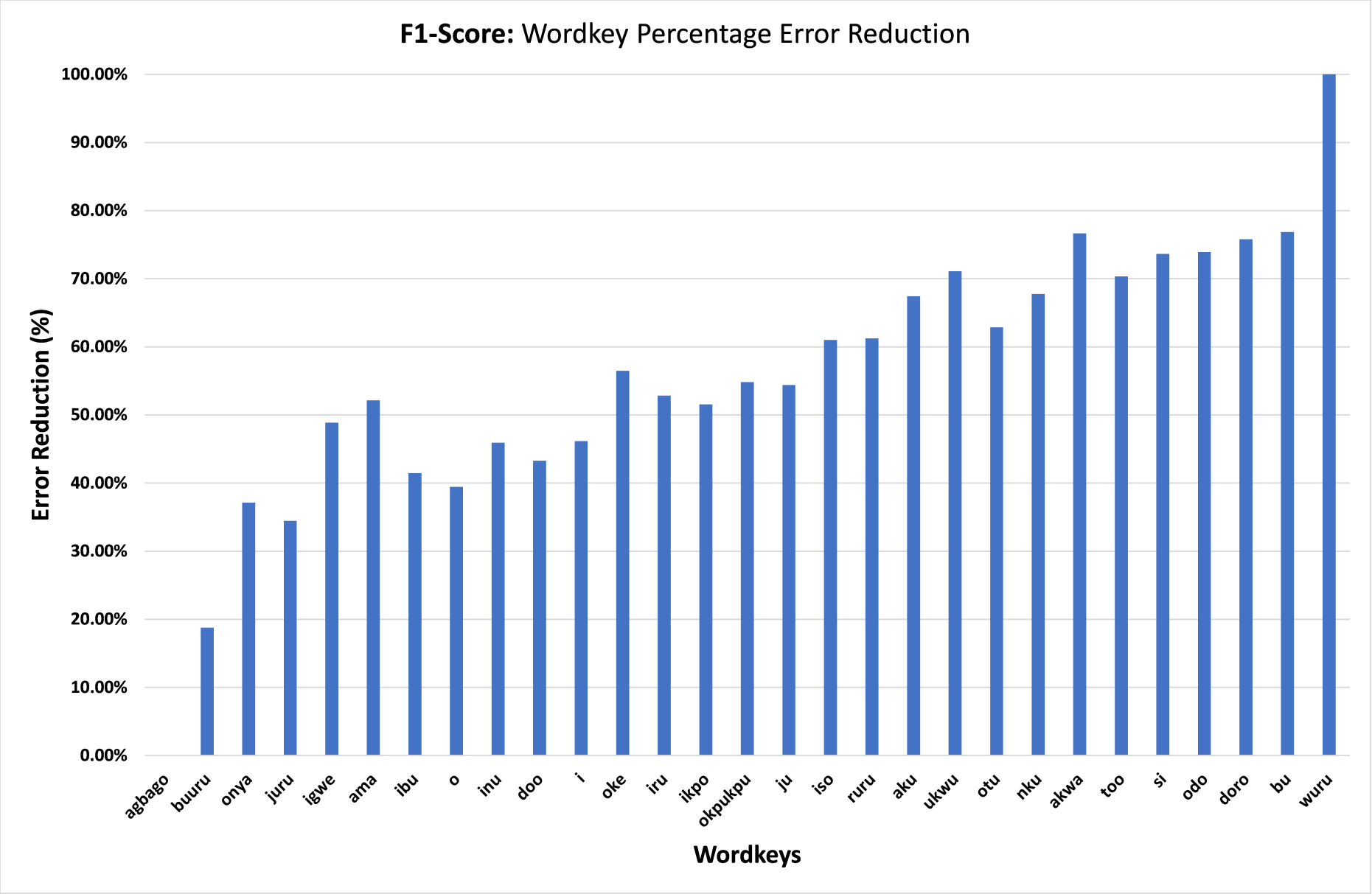

Fig. 20 **N-Gram F1-Score:** Graph showing the percentage F1 error reduction on each wordkey by its best performing model.

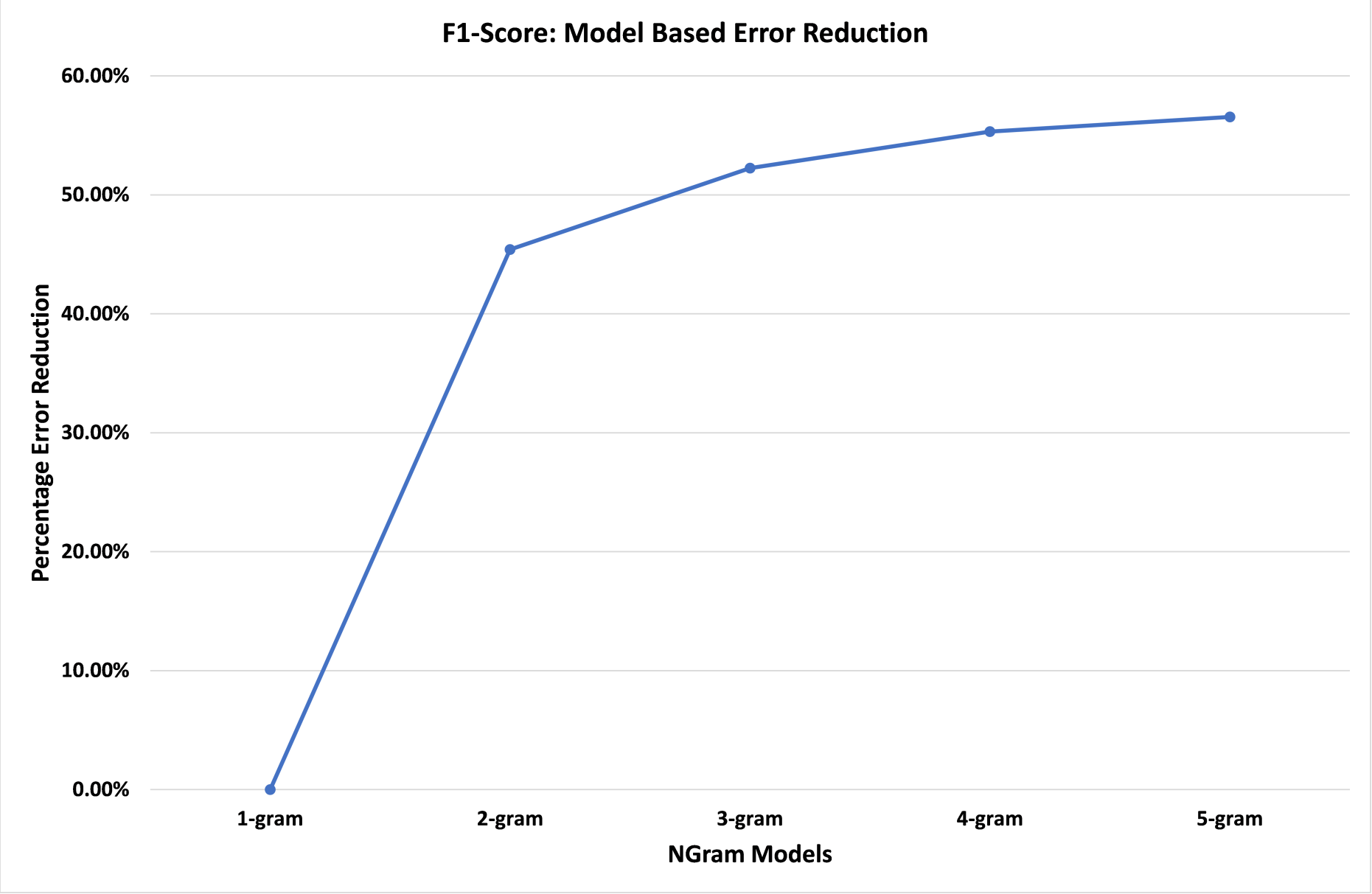

Fig. 21 **N-Gram F1-Score:** Column chart showing the percentage F1 error reduction from the global baseline error by each model.

# Appendix B: Results of IDR with Classification

| Wordkey | Cnt | nVar | 1Gram | SVEC | KNNC | PCPT | MNNB | DCTC | RFCL | BNNB | ADAB | BAGG | LSVC | SGDC | LRCV | BestScore (BS) | Best Model | Wdkey Imprvt | Baseline Imprvt | Error Reduction |
|---|---|---|---|---|---|---|---|---|---|---|---|---|---|---|---|---|---|---|---|---|
| buuru | 180 | 2 | 0.6667 | 0.6667 | 0.6722 | 0.7167 | 0.6722 | 0.7000 | 0.7056 | 0.6667 | 0.7222 | 0.7000 | 0.7000 | 0.6889 | 0.6889 | 0.7222 | ADAB | 5.55% | 5.47% | 16.65% |
| bu | 16999 | 2 | 0.6441 | 0.6441 | 0.6456 | 0.6240 | 0.6684 | 0.6946 | 0.7147 | 0.7006 | 0.7244 | 0.7234 | 0.7015 | 0.7212 | 0.7300 | 0.7300 | LRCV | 8.59% | 6.25% | 24.14% |
| onya | 160 | 3 | 0.5312 | 0.5312 | 0.6937 | 0.7312 | 0.7000 | 0.6000 | 0.6500 | 0.7188 | 0.6000 | 0.6562 | 0.7125 | 0.6813 | 0.6687 | 0.7312 | PCPT | 20.00% | 6.37% | 42.66% |
| ukwu | 1432 | 3 | 0.5223 | 0.5223 | 0.6634 | 0.6816 | 0.7137 | 0.6669 | 0.6927 | 0.7214 | 0.6627 | 0.7011 | 0.7486 | 0.7165 | 0.7416 | 0.7486 | LSVC | 22.63% | 8.11% | 47.37% |
| aku | 384 | 3 | 0.4896 | 0.4896 | 0.6719 | 0.7370 | 0.6823 | 0.6250 | 0.6562 | 0.6927 | 0.6276 | 0.6484 | 0.7630 | 0.7500 | 0.7604 | 0.7630 | LSVC | 27.34% | 9.55% | 53.57% |
| akwa | 1191 | 3 | 0.4232 | 0.4232 | 0.7170 | 0.6910 | 0.7590 | 0.6692 | 0.7053 | 0.7683 | 0.6944 | 0.7053 | 0.7498 | 0.7254 | 0.7322 | 0.7683 | BNNB | 34.51% | 10.08% | 59.83% |
| i | 5347 | 2 | 0.6738 | 0.6738 | 0.7193 | 0.7191 | 0.7339 | 0.7361 | 0.7629 | 0.7831 | 0.7883 | 0.7681 | 0.7945 | 0.8001 | 0.7986 | 0.8001 | SGDC | 12.63% | 13.26% | 38.72% |
| iso | 201 | 2 | 0.5025 | 0.5124 | 0.7512 | 0.7562 | 0.7861 | 0.7114 | 0.6766 | 0.8010 | 0.6617 | 0.6915 | 0.7910 | 0.7711 | 0.7910 | 0.8010 | BNNB | 29.85% | 13.35% | 59.19% |
| ibu | 682 | 2 | 0.6422 | 0.6422 | 0.7595 | 0.8050 | 0.7478 | 0.7375 | 0.7698 | 0.7903 | 0.7683 | 0.7419 | 0.8138 | 0.8079 | 0.8167 | 0.8167 | LRCV | 17.45% | 14.92% | 48.77% |
| ama | 1353 | 3 | 0.4213 | 0.4213 | 0.7081 | 0.7967 | 0.7524 | 0.7650 | 0.7679 | 0.7938 | 0.7775 | 0.7908 | 0.8219 | 0.8152 | 0.8211 | 0.8219 | LSVC | 40.06% | 15.44% | 69.22% |
| nku | 285 | 2 | 0.6140 | 0.6140 | 0.8281 | 0.7930 | 0.8035 | 0.7754 | 0.7895 | 0.8281 | 0.8035 | 0.7684 | 0.8281 | 0.8211 | 0.8211 | 0.8281 | KNNCBNNBLSVC | 21.41% | 16.06% | 55.47% |
| agbago | 99 | 2 | 0.5354 | 0.5354 | 0.7475 | 0.7576 | 0.7980 | 0.7778 | 0.7980 | 0.8283 | 0.7576 | 0.7879 | 0.7980 | 0.7677 | 0.8081 | 0.8283 | BNNB | 29.29% | 16.08% | 63.04% |
| si | 9039 | 2 | 0.5733 | 0.5733 | 0.7548 | 0.7624 | 0.7911 | 0.7640 | 0.7950 | 0.8140 | 0.7974 | 0.7935 | 0.8219 | 0.8289 | 0.8302 | 0.8302 | LRCV | 25.69% | 16.27% | 60.21% |
| ju | 97 | 2 | 0.7423 | 0.7423 | 0.7938 | 0.8144 | 0.7423 | 0.7629 | 0.7938 | 0.7526 | 0.8144 | 0.7320 | 0.8144 | 0.8351 | 0.8041 | 0.8351 | SGDC | 9.28% | 16.76% | 36.01% |
| otu | 5947 | 2 | 0.6664 | 0.6664 | 0.7553 | 0.7569 | 0.7711 | 0.8006 | 0.8017 | 0.8098 | 0.8176 | 0.8288 | 0.8261 | 0.8298 | 0.8371 | 0.8371 | LRCV | 17.07% | 16.96% | 51.17% |
| juru | 306 | 2 | 0.5359 | 0.5359 | 0.7549 | 0.8366 | 0.8137 | 0.7418 | 0.7418 | 0.8105 | 0.7549 | 0.7745 | 0.8399 | 0.8366 | 0.8431 | 0.8431 | LRCV | 30.72% | 17.56% | 66.19% |
| ruru | 488 | 2 | 0.5041 | 0.5041 | 0.7439 | 0.8320 | 0.8197 | 0.8012 | 0.8156 | 0.8299 | 0.7951 | 0.8053 | 0.8422 | 0.8422 | 0.8443 | 0.8443 | LRCV | 34.02% | 17.68% | 68.60% |
| inu | 156 | 2 | 0.5769 | 0.5769 | 0.7308 | 0.8013 | 0.7885 | 0.8205 | 0.7821 | 0.8141 | 0.7821 | 0.7949 | 0.8462 | 0.8141 | 0.8333 | 0.8462 | LSVC | 26.93% | 17.87% | 63.65% |
| o | 31446 | 2 | 0.7353 | 0.7353 | 0.7754 | 0.7952 | 0.7747 | 0.8056 | 0.8265 | 0.8279 | 0.8368 | 0.8330 | 0.8490 | 0.8454 | 0.8514 | 0.8514 | LRCV | 11.61% | 18.39% | 43.86% |
| too | 125 | 2 | 0.7120 | 0.7120 | 0.8480 | 0.8560 | 0.7680 | 0.7680 | 0.8400 | 0.8240 | 0.7520 | 0.8480 | 0.8560 | 0.8560 | 0.8480 | 0.8560 | PCPTLSVCSGDC | 14.40% | 18.85% | 50.00% |
| ikpo | 133 | 2 | 0.6165 | 0.6165 | 0.8271 | 0.8496 | 0.7669 | 0.7820 | 0.7744 | 0.7820 | 0.7744 | 0.7293 | 0.8571 | 0.8647 | 0.8346 | 0.8647 | SGDC | 24.82% | 19.72% | 64.72% |
| oke | 2267 | 3 | 0.7821 | 0.7821 | 0.8394 | 0.8478 | 0.7966 | 0.8015 | 0.8430 | 0.8306 | 0.8262 | 0.8416 | 0.8844 | 0.8778 | 0.8778 | 0.8844 | LSVC | 10.23% | 21.69% | 46.95% |
| igwe | 1392 | 4 | 0.6609 | 0.6609 | 0.8269 | 0.8779 | 0.7234 | 0.8463 | 0.8635 | 0.8312 | 0.8068 | 0.8685 | 0.8994 | 0.8879 | 0.8966 | 0.8994 | LSVC | 23.85% | 23.19% | 70.33% |
| okpukpu | 211 | 2 | 0.7109 | 0.7109 | 0.8341 | 0.9147 | 0.7678 | 0.9005 | 0.8389 | 0.8057 | 0.8768 | 0.9052 | 0.9052 | 0.9100 | 0.8957 | 0.9147 | PCPT | 20.38% | 24.72% | 70.49% |
| iru | 333 | 2 | 0.5315 | 0.5315 | 0.8468 | 0.8889 | 0.8949 | 0.9039 | 0.9099 | 0.9399 | 0.9009 | 0.9189 | 0.9189 | 0.9069 | 0.9279 | 0.9399 | BNNB | 40.84% | 27.24% | 87.17% |
| doo | 120 | 2 | 0.7000 | 0.7000 | 0.8417 | 0.9250 | 0.8167 | 0.9000 | 0.9167 | 0.9000 | 0.9000 | 0.8917 | 0.9417 | 0.9417 | 0.9417 | 0.9417 | LSVCSGDCLRCV | 24.17% | 27.42% | 80.57% |
| wuru | 112 | 2 | 0.6696 | 0.6696 | 0.9018 | 0.9107 | 0.8839 | 0.9643 | 0.9286 | 0.9464 | 0.9554 | 0.9464 | 0.9643 | 0.9643 | 0.9643 | 0.9643 | DCTCLSVCSGDCLRCV | 29.47% | 29.68% | 89.19% |
| doro | 205 | 2 | 0.6390 | 0.6390 | 0.9024 | 0.9463 | 0.8976 | 0.9317 | 0.9415 | 0.9512 | 0.9317 | 0.9463 | 0.9561 | 0.9659 | 0.9561 | 0.9659 | SGDC | 32.69% | 29.84% | 90.55% |
| odo | 154 | 2 | 0.7273 | 0.7273 | 0.9156 | 0.9545 | 0.9740 | 0.9416 | 0.9351 | 0.9286 | 0.9351 | 0.9416 | 0.9610 | 0.9675 | 0.9610 | 0.9740 | MNNB | 24.67% | 30.65% | 90.47% |
| **Baseline = 1gram; %Error = 33.25%** | | | **66.75%** | **66.76%** | **73.96%** | **74.87%** | **75.01%** | **76.68%** | **78.75%** | **79.14%** | **79.43%** | **79.45%** | **80.74%** | **80.98%** | **81.55%** | **81.71%** | **Performance Analysis** | **Model** | **Imprvnt** | **Error Reduction** |
| **Best Model Counts** | | | | **0** | **1** | **3** | **1** | **1** | **0** | **5** | **1** | **0** | **10** | **7** | **9** | | *Best Score* | ***LRCV*** | **14.80%** | **44.51%** |
| **Model Error Reduction:** | | | 0.00% | 0.01% | 21.67% | 24.43% | 24.85% | 29.87% | 36.09% | 37.26% | 38.14% | 38.19% | 42.06% | 42.78% | 44.51% | 44.98% | *Best Counts* | ***LSVC*** | **12.70%** | **38.19%** |

Table 4 **ML Accuracy:** Table showing the raw accuracy score of *all* machine learning models as compared with only the baseline (1-gram). The summary table for the 3 best performing models is presented in Table 6.3. [*NB:* The color code indicates the worst(red)-to-best(green) based on the accuracy performances and improvements on wordkey and global baselines].

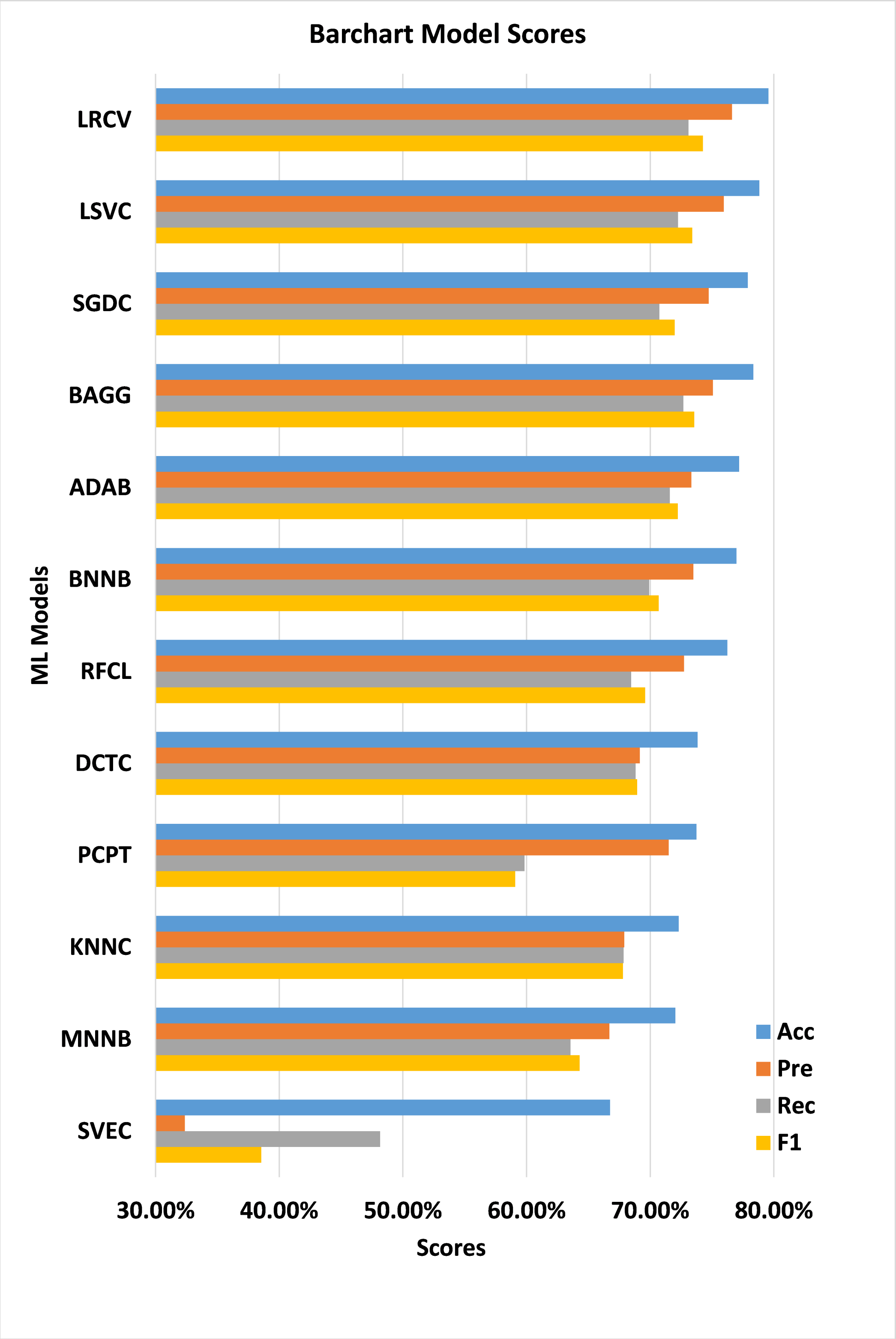

Fig. 22 **ML All Metrics:** Performance of all models across all metrics starting from the worst (SVEC) to the best (LRCV).

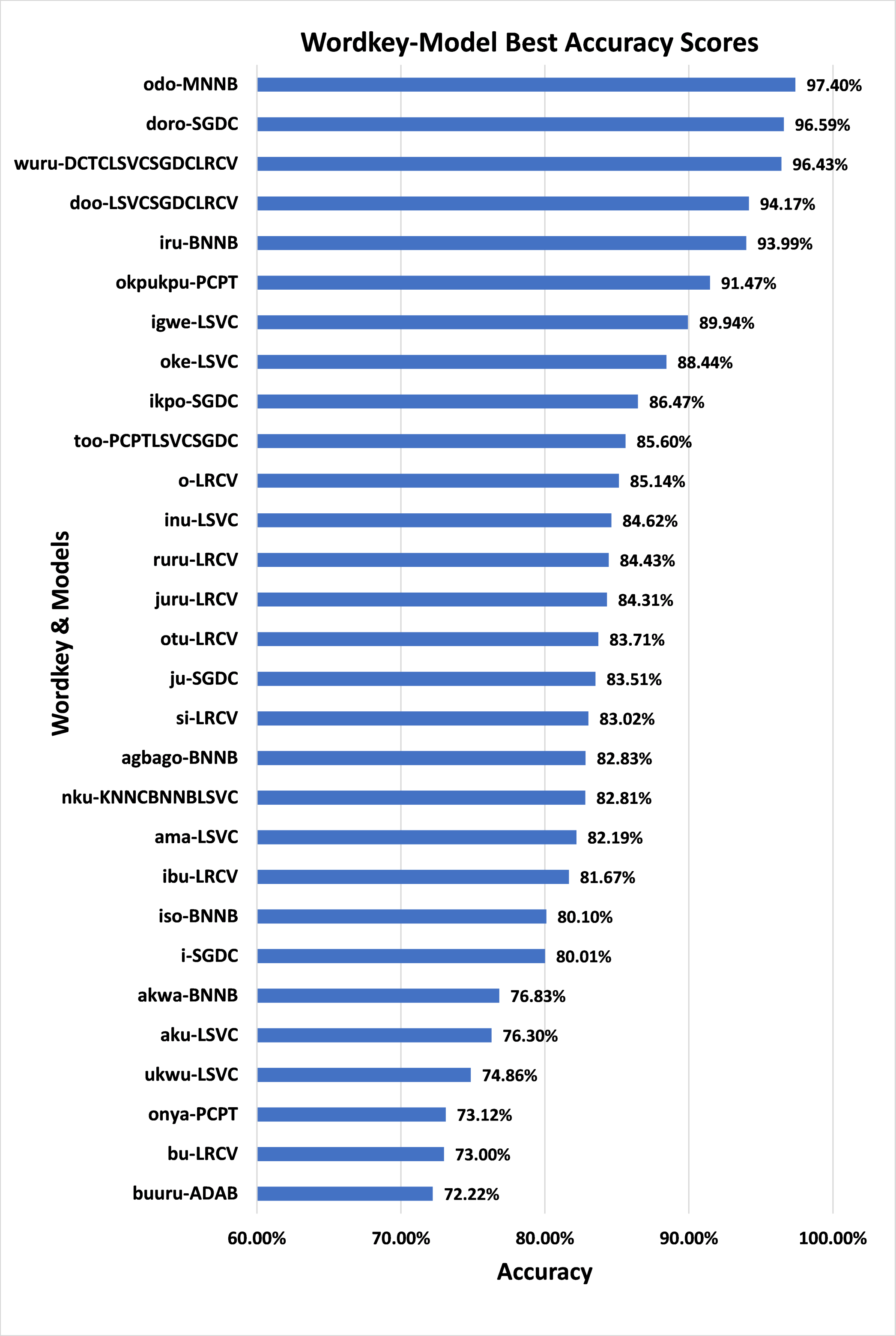

Fig. 23 **ML Accuracy:** Best accuracy scores obtained for all wordkeys and all machine learning models that achieved the score. Note that, in some cases, multiple models obtained the best score.

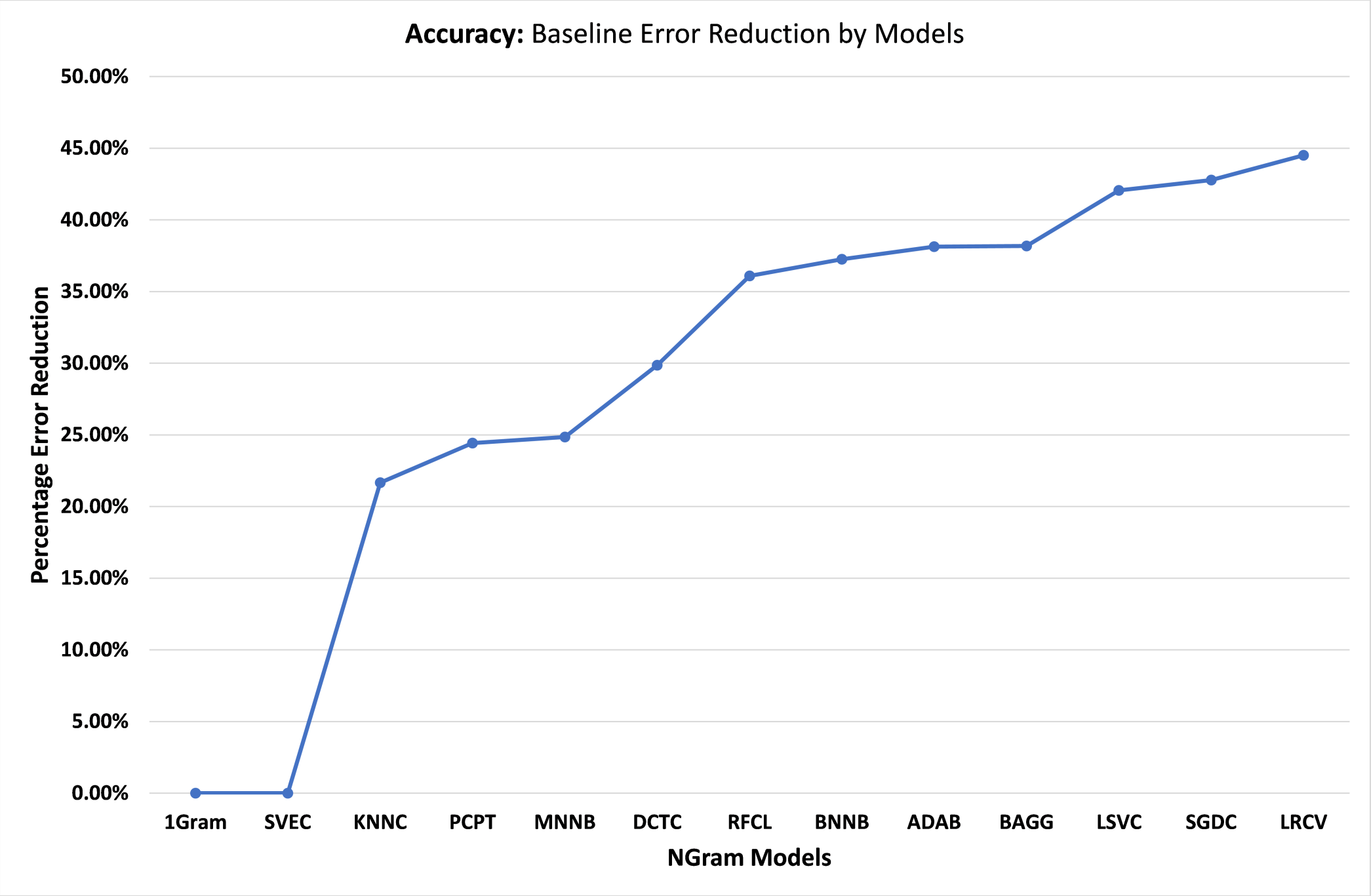

Fig. 24 **ML Accuracy:** Graph showing the percentage error reduction on accuracy achieved by each machine learning model with respect to the initial remaining error obtained by the baseline unigram model.

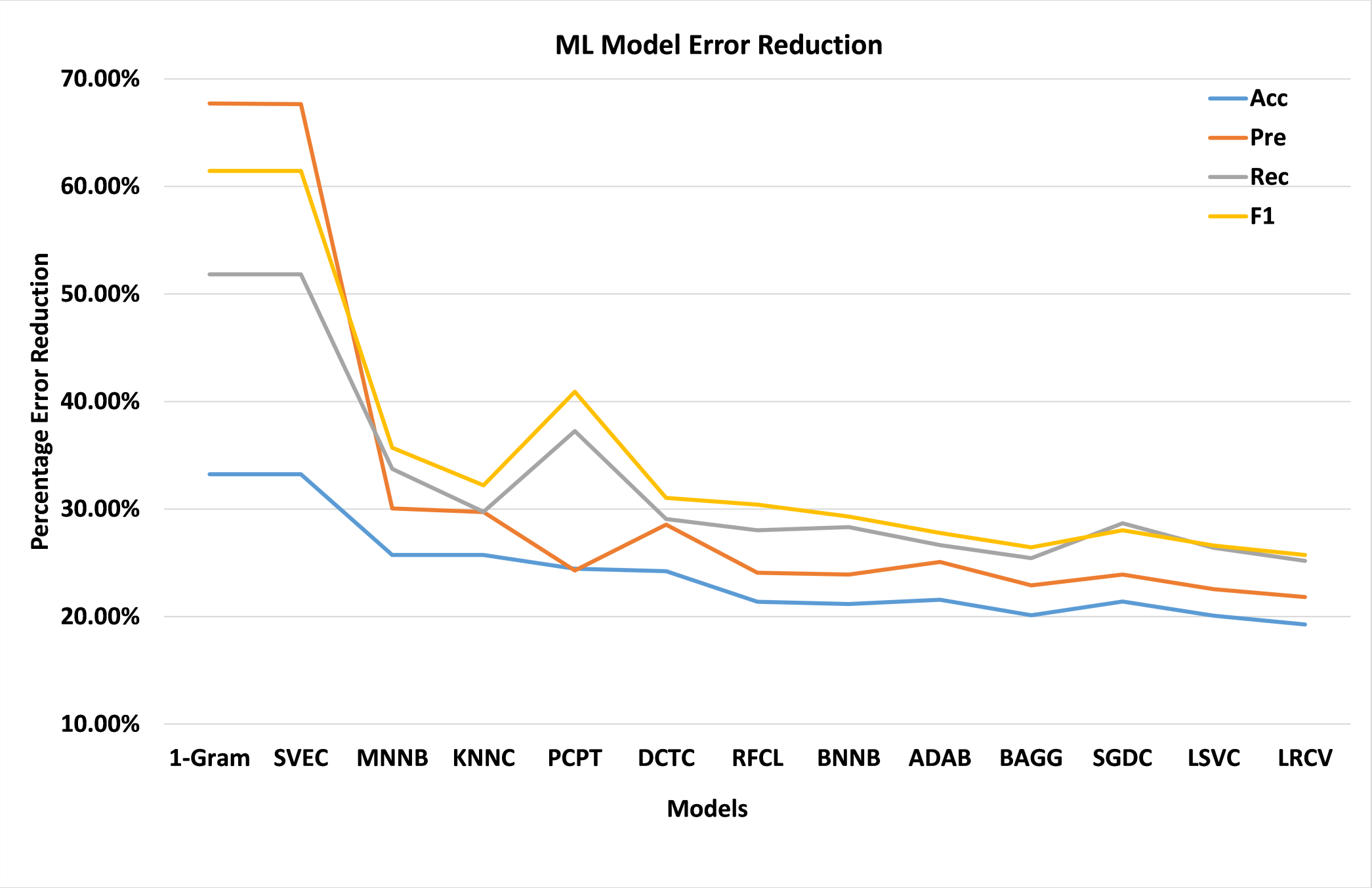

Fig. 25 **ML Accuracy:** Graph showing the plot of the remaining error on all metrics as obtained by all the models starting with the worst to the best. It is similar to Figure 24 but measures the percentage *remaining* errors and not the percentage *error reduction*.

# Appendix C: Results of IDR with Embedding Models

| Wordkey | Counts | No of Variants | 1-gram | igWikCrwl | igWikSbwd | igWikNews | igEnModel | igGglNews | igTnModel | BestScore (BS) | Best Model | Wdkey Improvement | Baseline Improvement | Error Reduction |
|---|---|---|---|---|---|---|---|---|---|---|---|---|---|---|
| onya | 160 | 3 | 0.3333 | 0.3799 | 0.3597 | 0.3687 | 0.4125 | 0.3620 | 0.4616 | 0.4616 | igTnModel | 12.83% | -2.01% | 19.24% |
| igwe | 1392 | 4 | 0.2500 | 0.4725 | 0.4049 | 0.4137 | 0.5106 | 0.4963 | 0.5412 | 0.5412 | igTnModel | 29.12% | 5.95% | 38.83% |
| aku | 384 | 3 | 0.3333 | 0.4754 | 0.3564 | 0.3956 | 0.4378 | 0.3640 | 0.5428 | 0.5428 | igTnModel | 20.95% | 6.11% | 31.42% |
| akwa | 1191 | 3 | 0.3333 | 0.5148 | 0.4139 | 0.4582 | 0.5600 | 0.4683 | 0.5871 | 0.5871 | igTnModel | 25.38% | 10.54% | 38.07% |
| ukwu | 1432 | 3 | 0.3333 | 0.3516 | 0.3527 | 0.3643 | 0.4859 | 0.4308 | 0.5974 | 0.5974 | igTnModel | 26.41% | 11.57% | 39.62% |
| ama | 1353 | 3 | 0.3333 | 0.3817 | 0.3835 | 0.3894 | 0.4785 | 0.3849 | 0.6048 | 0.6048 | igTnModel | 27.15% | 12.31% | 40.72% |
| too | 125 | 2 | 0.5000 | 0.5694 | 0.5443 | 0.5638 | 0.5748 | 0.6111 | 0.6138 | 0.6138 | igTnModel | 11.38% | 13.21% | 22.75% |
| bu | 16999 | 2 | 0.5000 | 0.5043 | 0.5234 | 0.4888 | 0.5438 | 0.5710 | 0.6151 | 0.6151 | igTnModel | 11.51% | 13.34% | 23.02% |
| ju | 97 | 2 | 0.5000 | 0.5000 | 0.5000 | 0.5000 | 0.5000 | 0.5000 | 0.6261 | 0.6261 | igTnModel | 12.61% | 14.44% | 25.22% |
| agbago | 99 | 2 | 0.5000 | 0.6122 | 0.4984 | 0.6013 | 0.5453 | 0.6097 | 0.6352 | 0.6352 | igTnModel | 13.52% | 15.34% | 27.03% |
| oke | 2267 | 3 | 0.3333 | 0.4925 | 0.4079 | 0.4739 | 0.5261 | 0.3779 | 0.6404 | 0.6404 | igTnModel | 30.71% | 15.87% | 46.06% |
| doo | 120 | 2 | 0.5000 | 0.5437 | 0.4564 | 0.5318 | 0.6191 | 0.5000 | 0.6429 | 0.6429 | igTnModel | 14.29% | 16.12% | 28.57% |
| juru | 306 | 2 | 0.5000 | 0.5827 | 0.5350 | 0.5491 | 0.5630 | 0.5977 | 0.6459 | 0.6459 | igTnModel | 14.59% | 16.42% | 29.17% |
| o | 31446 | 2 | 0.5000 | 0.5007 | 0.5114 | 0.5015 | 0.5965 | 0.4949 | 0.6509 | 0.6509 | igTnModel | 15.09% | 16.91% | 30.17% |
| ibu | 682 | 2 | 0.5000 | 0.6217 | 0.6202 | 0.6517 | 0.6769 | 0.6420 | 0.6720 | 0.6769 | igEnModel | 17.69% | 19.52% | 35.39% |
| nku | 285 | 2 | 0.5000 | 0.5534 | 0.5443 | 0.5335 | 0.6614 | 0.6231 | 0.6818 | 0.6818 | igTnModel | 18.18% | 20.01% | 36.36% |
| otu | 5947 | 2 | 0.5000 | 0.5267 | 0.5451 | 0.5143 | 0.6701 | 0.5115 | 0.6974 | 0.6974 | igTnModel | 19.74% | 21.57% | 39.49% |
| buuru | 180 | 2 | 0.5000 | 0.6792 | 0.6458 | 0.6417 | 0.7042 | 0.6208 | 0.6333 | 0.7042 | igEnModel | 20.42% | 22.25% | 40.83% |
| i | 5347 | 2 | 0.5000 | 0.5252 | 0.5093 | 0.5248 | 0.5702 | 0.5485 | 0.7061 | 0.7061 | igTnModel | 20.61% | 22.44% | 41.23% |
| okpukpu | 211 | 2 | 0.5000 | 0.5751 | 0.5754 | 0.5836 | 0.6981 | 0.5954 | 0.7087 | 0.7087 | igTnModel | 20.87% | 22.70% | 41.74% |
| si | 9039 | 2 | 0.5000 | 0.5782 | 0.6291 | 0.5402 | 0.6492 | 0.5483 | 0.7090 | 0.7090 | igTnModel | 20.90% | 22.73% | 41.81% |
| ikpo | 133 | 2 | 0.5000 | 0.7021 | 0.7127 | 0.7606 | 0.6942 | 0.6831 | 0.6868 | 0.7606 | igWikNews | 26.06% | 27.89% | 52.13% |
| iso | 201 | 2 | 0.5000 | 0.5852 | 0.5000 | 0.5000 | 0.7503 | 0.5000 | 0.7608 | 0.7608 | igTnModel | 26.08% | 27.91% | 52.17% |
| ruru | 488 | 2 | 0.5000 | 0.6037 | 0.6667 | 0.6725 | 0.7142 | 0.6265 | 0.7694 | 0.7694 | igTnModel | 26.94% | 28.77% | 53.87% |
| doro | 205 | 2 | 0.5000 | 0.6490 | 0.6579 | 0.6704 | 0.7724 | 0.6857 | 0.7929 | 0.7929 | igTnModel | 29.29% | 31.12% | 58.58% |
| inu | 156 | 2 | 0.5000 | 0.6783 | 0.6556 | 0.6576 | 0.7783 | 0.6631 | 0.8404 | 0.8404 | igTnModel | 34.04% | 35.87% | 68.08% |
| wuru | 112 | 2 | 0.5000 | 0.7065 | 0.6532 | 0.6463 | 0.7593 | 0.6332 | 0.8860 | 0.8860 | igTnModel | 38.60% | 40.42% | 77.19% |
| iru | 333 | 2 | 0.5000 | 0.8305 | 0.6667 | 0.7429 | 0.7759 | 0.8701 | 0.9065 | 0.9065 | igTnModel | 40.65% | 42.48% | 81.30% |
| odo | 154 | 2 | 0.5000 | 0.9107 | 0.7902 | 0.8125 | 0.9077 | 0.8214 | 0.8929 | 0.9107 | igWikCrwl | 41.07% | 42.90% | 82.14% |
| **Baseline = 1gram; %Error = 51.83%** | | | **48.17%** | **51.49%** | **52.21%** | **50.39%** | **59.06%** | **52.14%** | **65.51%** | **65.54%** | **Performance Analysis** | Model | Improvement | Error Reduction |
| **Best Model Counts** | | | **0** | **1** | **0** | **1** | **2** | **0** | **25** | | *Best Score* | ***igTnModel*** | **17.34%** | **33.45%** |
| **Model Error Reduction:** | | | 0.00% | 6.40% | 7.79% | 4.29% | 21.02% | 7.65% | 33.45% | 33.52% | *Best Counts* | ***igTnModel*** | **17.34%** | **33.45%** |

Table 5 **Emb Recall:** Table showing the full raw recall scores. [*Color code indicates the worst(red)-to-best(green) scores on metric performance results for the wordkey and global baselines.*]

| Wordkey | Counts | No of Variants | 1-gram | igWikCrwl | igWikSbwd | igWikNews | igEnModel | igGglNews | igTnModel | BestScore (BS) | Best Model | Wdkey Improvement | Baseline Improvement | Error Reduction |
|---|---|---|---|---|---|---|---|---|---|---|---|---|---|---|
| onya | 160 | 3 | 0.2313 | 0.3259 | 0.2901 | 0.3020 | 0.3954 | 0.3307 | 0.4312 | 0.4312 | igTnModel | 19.99% | 4.58% | 26.01% |
| igwe | 1392 | 4 | 0.1990 | 0.4857 | 0.4107 | 0.4372 | 0.4295 | 0.4780 | 0.4836 | 0.4857 | igWikCrwl | 28.67% | 10.02% | 35.79% |
| ukwu | 1432 | 3 | 0.2287 | 0.2283 | 0.2219 | 0.2434 | 0.3696 | 0.3617 | 0.5118 | 0.5118 | igTnModel | 28.31% | 12.63% | 36.70% |
| aku | 384 | 3 | 0.2191 | 0.4107 | 0.2872 | 0.3047 | 0.4123 | 0.2793 | 0.5404 | 0.5404 | igTnModel | 32.13% | 15.49% | 41.14% |
| akwa | 1191 | 3 | 0.1982 | 0.4734 | 0.3297 | 0.3966 | 0.5344 | 0.4189 | 0.5648 | 0.5648 | igTnModel | 36.66% | 17.93% | 45.72% |
| ama | 1353 | 3 | 0.1976 | 0.2730 | 0.2848 | 0.2860 | 0.4226 | 0.3306 | 0.5666 | 0.5666 | igTnModel | 36.90% | 18.12% | 45.99% |
| bu | 16999 | 2 | 0.3918 | 0.4224 | 0.4807 | 0.4888 | 0.5396 | 0.5679 | 0.6166 | 0.6166 | igTnModel | 22.48% | 23.11% | 36.95% |
| oke | 2267 | 3 | 0.2926 | 0.4536 | 0.3677 | 0.4394 | 0.4326 | 0.3690 | 0.6175 | 0.6175 | igTnModel | 32.49% | 23.20% | 45.92% |
| too | 125 | 2 | 0.4159 | 0.5478 | 0.5135 | 0.5421 | 0.5677 | 0.6139 | 0.6201 | 0.6201 | igTnModel | 20.42% | 23.46% | 34.95% |
| agbago | 99 | 2 | 0.3487 | 0.6029 | 0.4762 | 0.5900 | 0.5447 | 0.5865 | 0.6238 | 0.6238 | igTnModel | 27.51% | 23.83% | 42.24% |
| ibu | 682 | 2 | 0.3911 | 0.5689 | 0.5382 | 0.6181 | 0.6302 | 0.6169 | 0.5910 | 0.6302 | igEnModel | 23.91% | 24.47% | 39.26% |
| juru | 306 | 2 | 0.3489 | 0.5366 | 0.4482 | 0.4728 | 0.5134 | 0.5623 | 0.6427 | 0.6427 | igTnModel | 29.38% | 25.72% | 45.12% |
| ju | 97 | 2 | 0.4260 | 0.4260 | 0.4260 | 0.4260 | 0.4260 | 0.4260 | 0.6434 | 0.6434 | igTnModel | 21.74% | 25.79% | 37.87% |
| otu | 5947 | 2 | 0.3999 | 0.3440 | 0.3847 | 0.3284 | 0.6206 | 0.2830 | 0.6502 | 0.6502 | igTnModel | 25.03% | 26.47% | 41.71% |
| doo | 120 | 2 | 0.4118 | 0.5433 | 0.3833 | 0.5238 | 0.6273 | 0.2308 | 0.6563 | 0.6563 | igTnModel | 24.45% | 27.08% | 41.57% |
| nku | 285 | 2 | 0.3804 | 0.4928 | 0.4765 | 0.4530 | 0.6656 | 0.5486 | 0.6613 | 0.6656 | igEnModel | 28.52% | 28.02% | 46.04% |
| o | 31446 | 2 | 0.4237 | 0.4417 | 0.4689 | 0.4454 | 0.5878 | 0.4307 | 0.6671 | 0.6671 | igTnModel | 24.34% | 28.16% | 42.23% |
| i | 5347 | 2 | 0.4026 | 0.5130 | 0.4960 | 0.5118 | 0.4859 | 0.5402 | 0.6734 | 0.6734 | igTnModel | 27.08% | 28.79% | 45.33% |
| si | 9039 | 2 | 0.3644 | 0.5572 | 0.6296 | 0.4892 | 0.6506 | 0.5077 | 0.7098 | 0.7098 | igTnModel | 34.54% | 32.43% | 54.34% |
| buuru | 180 | 2 | 0.4000 | 0.6672 | 0.6063 | 0.6164 | 0.7120 | 0.6256 | 0.6390 | 0.7120 | igEnModel | 31.20% | 32.65% | 52.00% |
| okpukpu | 211 | 2 | 0.4155 | 0.3860 | 0.4066 | 0.4107 | 0.6871 | 0.4394 | 0.7147 | 0.7147 | igTnModel | 29.92% | 32.92% | 51.18% |
| ikpo | 133 | 2 | 0.3814 | 0.6690 | 0.7055 | 0.7354 | 0.6955 | 0.6874 | 0.6903 | 0.7354 | igWikNews | 35.40% | 34.99% | 57.22% |
| iso | 201 | 2 | 0.3344 | 0.5139 | 0.3344 | 0.3344 | 0.7404 | 0.3344 | 0.7599 | 0.7599 | igTnModel | 42.55% | 37.44% | 63.92% |
| ruru | 488 | 2 | 0.3351 | 0.5522 | 0.6584 | 0.6596 | 0.7106 | 0.5851 | 0.7657 | 0.7657 | igTnModel | 43.06% | 38.02% | 64.76% |
| doro | 205 | 2 | 0.3899 | 0.6522 | 0.5828 | 0.6787 | 0.7666 | 0.6846 | 0.8052 | 0.8052 | igTnModel | 41.53% | 41.98% | 68.08% |
| inu | 156 | 2 | 0.3659 | 0.6706 | 0.6408 | 0.6455 | 0.7809 | 0.6509 | 0.8418 | 0.8418 | igTnModel | 47.59% | 45.63% | 75.04% |
| wuru | 112 | 2 | 0.4011 | 0.6158 | 0.5402 | 0.5417 | 0.7098 | 0.5102 | 0.8729 | 0.8729 | igTnModel | 47.18% | 48.74% | 78.78% |
| iru | 333 | 2 | 0.3471 | 0.8173 | 0.6128 | 0.7141 | 0.7772 | 0.8611 | 0.9088 | 0.9088 | igTnModel | 56.17% | 52.33% | 86.03% |
| odo | 154 | 2 | 0.4211 | 0.8548 | 0.6878 | 0.7180 | 0.9234 | 0.7300 | 0.9207 | 0.9234 | igEnModel | 50.23% | 53.79% | 86.77% |
| **Baseline = 1gram; %Error = 61.45%** | | | **38.55%** | **44.99%** | **47.39%** | **45.35%** | **56.90%** | **46.74%** | **65.13%** | **65.19%** | **Performance Analysis** | Model | Improvement | Error Reduction |
| **Best Model Counts** | | | **0** | **1** | **0** | **1** | **4** | **0** | **23** | | *Best Score* | ***igTnModel*** | **26.58%** | **43.26%** |
| **Model Error Reduction:** | | | 0.00% | 10.48% | 14.39% | 11.07% | 29.87% | 13.33% | 43.26% | 43.36% | *Best Counts* | ***igTnModel*** | **26.58%** | **43.26%** |

Table 6 **Emb F1-Score:** Table showing the full raw f1 scores. [*Color code indicates the worst(red)-to-best(green) scores on metric performance results for the wordkey and global baselines.*]

| *Wordkey* | *Counts* | *No of Variants* | 1-gram | igWikCrwl | igWikSbwd | igWikNews | igEnModel | igGglNews | igTnModel | *BestScore (BS)* | *Best Model* | *Wdkey Improvement* | *Baseline Improvement* | *Error Reduction* |
|---|---|---|---|---|---|---|---|---|---|---|---|---|---|---|
| **onya** | 160 | 3 | 0.1771 | 0.4910 | 0.4687 | 0.5185 | 0.4188 | 0.3629 | 0.4511 | 0.5185 | igWikNews | 34.14% | 19.56% | 41.49% |
| **igwe** | 1392 | 4 | 0.1652 | 0.5059 | 0.5509 | 0.5508 | 0.4576 | 0.5459 | 0.5088 | 0.5509 | igWikSbwd | 38.57% | 22.79% | 46.20% |
| **ama** | 1353 | 3 | 0.1404 | 0.4891 | 0.4896 | 0.5343 | 0.4834 | 0.5495 | 0.5785 | 0.5785 | igTnModel | 43.81% | 25.56% | 50.97% |
| **aku** | 384 | 3 | 0.1632 | 0.5000 | 0.4452 | 0.4625 | 0.4298 | 0.5051 | 0.6009 | 0.6009 | igTnModel | 43.77% | 27.79% | 52.30% |
| **ukwu** | 1432 | 3 | 0.1741 | 0.4771 | 0.6140 | 0.6072 | 0.4504 | 0.4102 | 0.5383 | 0.6140 | igWikSbwd | 43.99% | 29.10% | 53.26% |
| **bu** | 16999 | 2 | 0.3220 | 0.5053 | 0.5234 | 0.4888 | 0.5556 | 0.5954 | 0.6187 | 0.6187 | igTnModel | 29.67% | 29.58% | 43.77% |
| **agbago** | 99 | 2 | 0.2677 | 0.6347 | 0.4979 | 0.6243 | 0.5451 | 0.6269 | 0.6422 | 0.6422 | igTnModel | 37.45% | 31.92% | 51.13% |
| **oke** | 2267 | 3 | 0.2607 | 0.5135 | 0.3638 | 0.4832 | 0.5093 | 0.6523 | 0.6068 | 0.6523 | igGglNews | 39.16% | 32.93% | 52.96% |
| **akwa** | 1191 | 3 | 0.1411 | 0.6085 | 0.5925 | 0.5876 | 0.6603 | 0.5790 | 0.6712 | 0.6712 | igTnModel | 53.01% | 34.83% | 61.72% |
| **juru** | 306 | 2 | 0.2680 | 0.6670 | 0.6252 | 0.6517 | 0.6272 | 0.6722 | 0.6625 | 0.6722 | igGglNews | 40.42% | 34.93% | 55.22% |
| **otu** | 5947 | 2 | 0.3332 | 0.5775 | 0.5959 | 0.5442 | 0.6557 | 0.6096 | 0.6786 | 0.6786 | igTnModel | 34.54% | 35.57% | 51.81% |
| **i** | 5347 | 2 | 0.3369 | 0.5395 | 0.5082 | 0.5397 | 0.5759 | 0.5769 | 0.6813 | 0.6813 | igTnModel | 34.44% | 35.83% | 51.93% |
| **ibu** | 682 | 2 | 0.3211 | 0.6231 | 0.6432 | 0.6420 | 0.6714 | 0.6310 | 0.6986 | 0.6986 | igTnModel | 37.75% | 37.56% | 55.60% |
| **si** | 9039 | 2 | 0.2866 | 0.6236 | 0.6414 | 0.5950 | 0.6609 | 0.5979 | 0.7109 | 0.7109 | igTnModel | 42.43% | 38.79% | 59.47% |
| **o** | 31446 | 2 | 0.3677 | 0.5060 | 0.5562 | 0.5112 | 0.5853 | 0.4500 | 0.7162 | 0.7162 | igTnModel | 34.85% | 39.33% | 55.12% |
| **okpukpu** | 211 | 2 | 0.3555 | 0.6429 | 0.6175 | 0.6336 | 0.6812 | 0.6301 | 0.7226 | 0.7226 | igTnModel | 36.71% | 39.96% | 56.95% |
| **buuru** | 180 | 2 | 0.3333 | 0.6645 | 0.6313 | 0.6259 | 0.7262 | 0.6411 | 0.7256 | 0.7262 | igEnModel | 39.29% | 40.32% | 58.92% |
| **ikpo** | 133 | 2 | 0.3083 | 0.6981 | 0.7037 | 0.7482 | 0.6973 | 0.6978 | 0.6972 | 0.7482 | igWikNews | 43.99% | 42.52% | 63.59% |
| **doo** | 120 | 2 | 0.3500 | 0.5540 | 0.4564 | 0.5278 | 0.6800 | 0.1500 | 0.7596 | 0.7596 | igTnModel | 40.96% | 43.67% | 63.02% |
| **nku** | 285 | 2 | 0.3070 | 0.7537 | 0.7411 | 0.7597 | 0.6816 | 0.6590 | 0.6735 | 0.7597 | igWikNews | 45.27% | 43.67% | 65.32% |
| **ju** | 97 | 2 | 0.3711 | 0.3711 | 0.3711 | 0.3711 | 0.3711 | 0.3711 | 0.7866 | 0.7866 | igTnModel | 41.55% | 46.37% | 66.07% |
| **ruru** | 488 | 2 | 0.2520 | 0.6972 | 0.6868 | 0.7065 | 0.7275 | 0.7158 | 0.7920 | 0.7920 | igTnModel | 54.00% | 46.91% | 72.20% |
| **iso** | 201 | 2 | 0.2512 | 0.7101 | 0.2512 | 0.2512 | 0.7988 | 0.2512 | 0.7664 | 0.7988 | igEnModel | 54.76% | 47.58% | 73.13% |
| **doro** | 205 | 2 | 0.3195 | 0.7544 | 0.6796 | 0.7332 | 0.7629 | 0.6837 | 0.8296 | 0.8296 | igTnModel | 51.01% | 50.67% | 74.97% |
| **inu** | 156 | 2 | 0.2885 | 0.8041 | 0.7917 | 0.7769 | 0.7853 | 0.7959 | 0.8434 | 0.8434 | igTnModel | 55.49% | 52.05% | 77.99% |
| **wuru** | 112 | 2 | 0.3348 | 0.7161 | 0.6901 | 0.6726 | 0.7313 | 0.6805 | 0.8643 | 0.8643 | igTnModel | 52.95% | 54.13% | 79.60% |
| **too** | 125 | 2 | 0.3560 | 0.8708 | 0.6989 | 0.7864 | 0.6910 | 0.8803 | 0.7907 | 0.8803 | igGglNews | 52.43% | 55.74% | 81.42% |
| **iru** | 333 | 2 | 0.2658 | 0.8611 | 0.7847 | 0.8158 | 0.7850 | 0.8861 | 0.9154 | 0.9154 | igTnModel | 64.96% | 59.25% | 88.48% |
| **odo** | 154 | 2 | 0.3636 | 0.8387 | 0.7360 | 0.7500 | 0.9431 | 0.7561 | 0.9628 | 0.9628 | igTnModel | 59.92% | 63.99% | 94.16% |
| **Baseline = 1gram; %Error = 67.70%** | | | **32.30%** | **53.67%** | **55.71%** | **53.05%** | **58.96%** | **53.73%** | **67.85%** | **68.26%** | **Performance Analysis** | Model | Improvement | Error Reduction |
| **Best Model Counts** | | | **0** | **0** | **2** | **3** | **2** | **3** | **19** | | *Best Score* | ***igTnModel*** | **35.56%** | **52.52%** |
| **Model Error Reduction:** | | | 0.00% | 31.57% | 34.58% | 30.65% | 39.38% | 31.66% | 52.52% | 53.12% | *Best Counts* | ***igTnModel*** | **35.56%** | **52.52%** |

Fig. 26 **Emb Precision:** Graph showing the best performing model on for each of the wordkeys sorted in the ascending order of precision scores

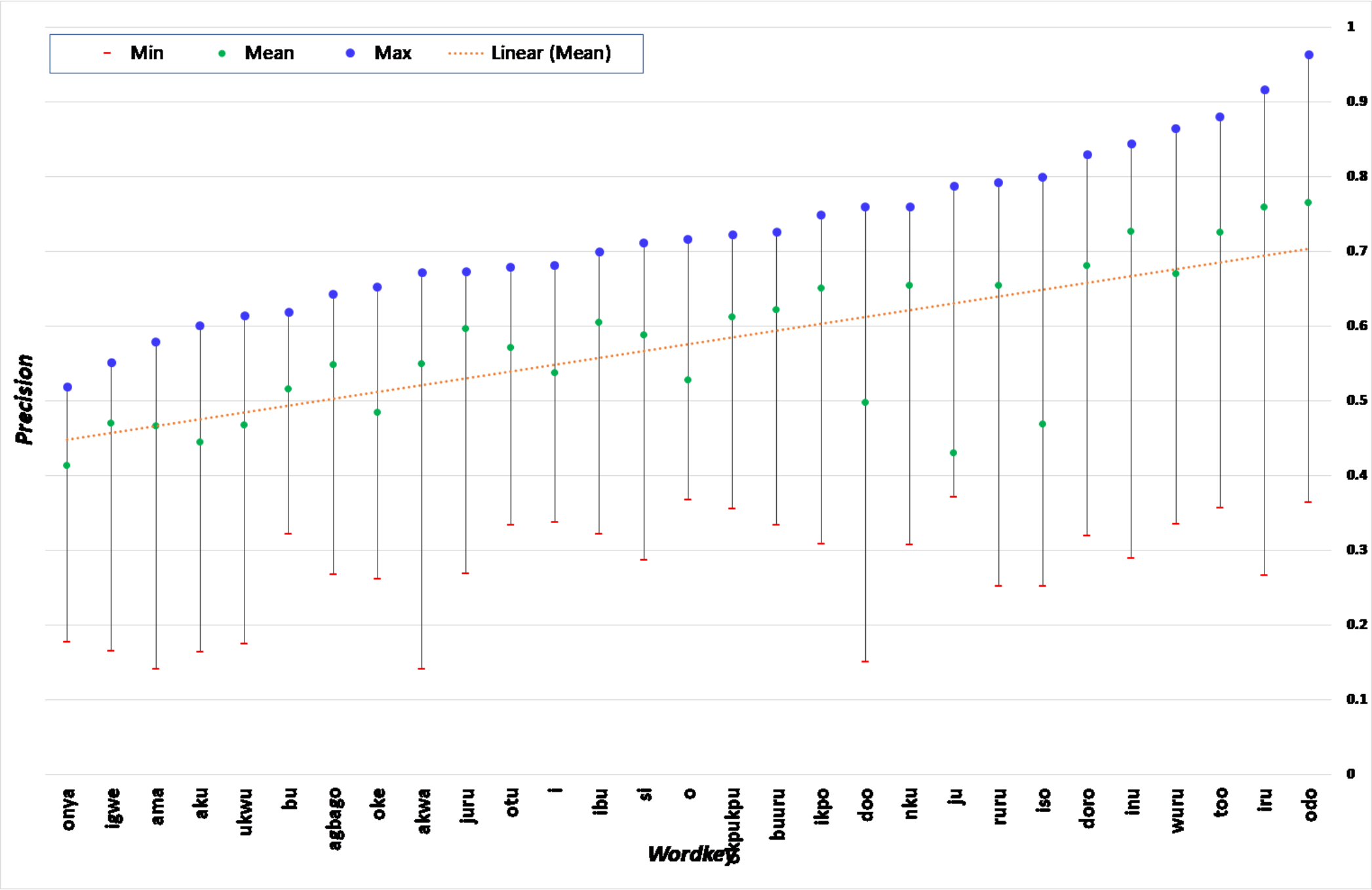

Fig. 27 **Emb Precision:** Graph showing minimum, mean and maximum precision scores and the linear trend line on the mean scores.

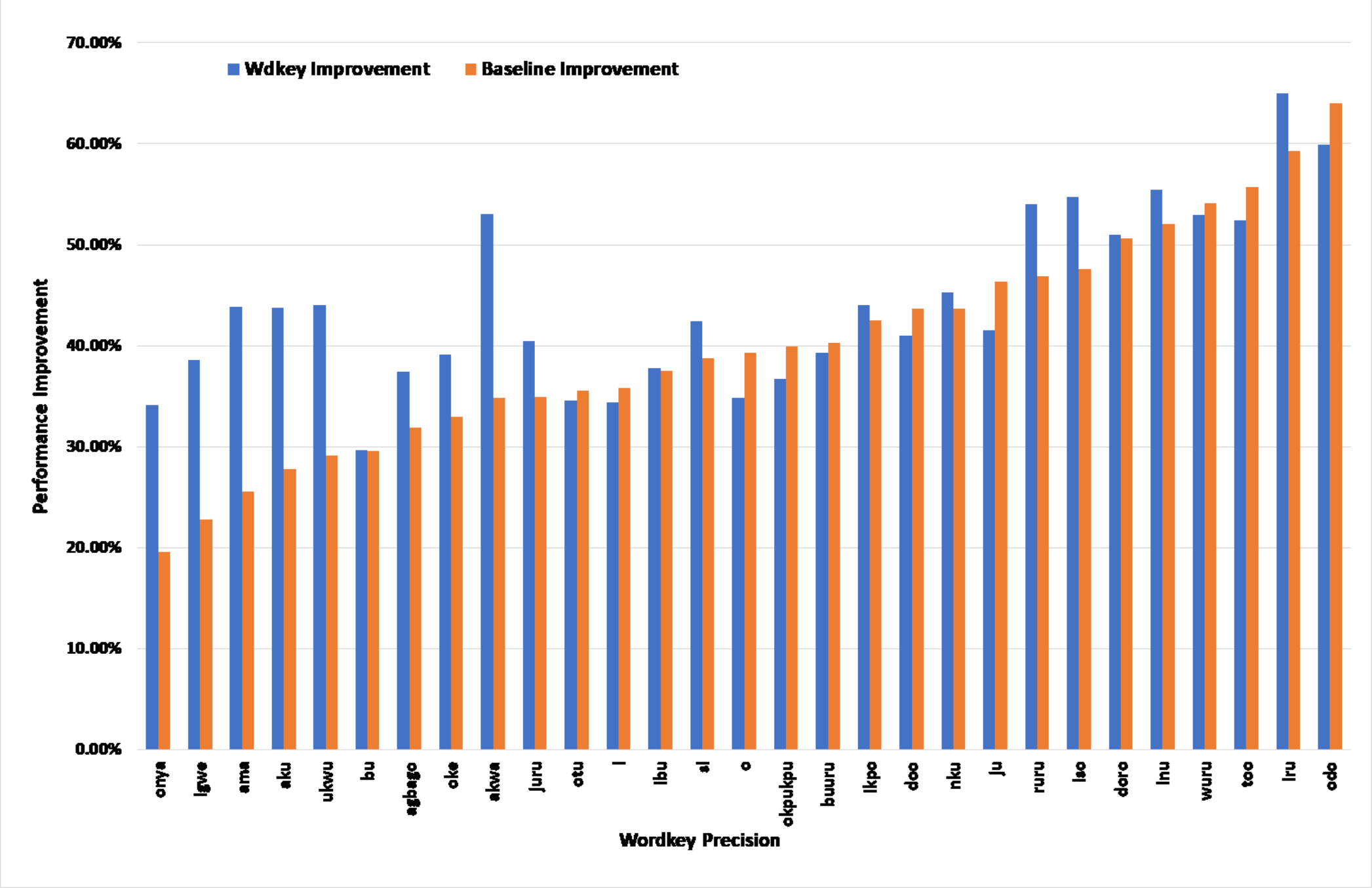

Fig. 28 **Emb Precision:** Graph showing the stacked column-chart of the improved precision scores on both the wordkey and the global baselines.

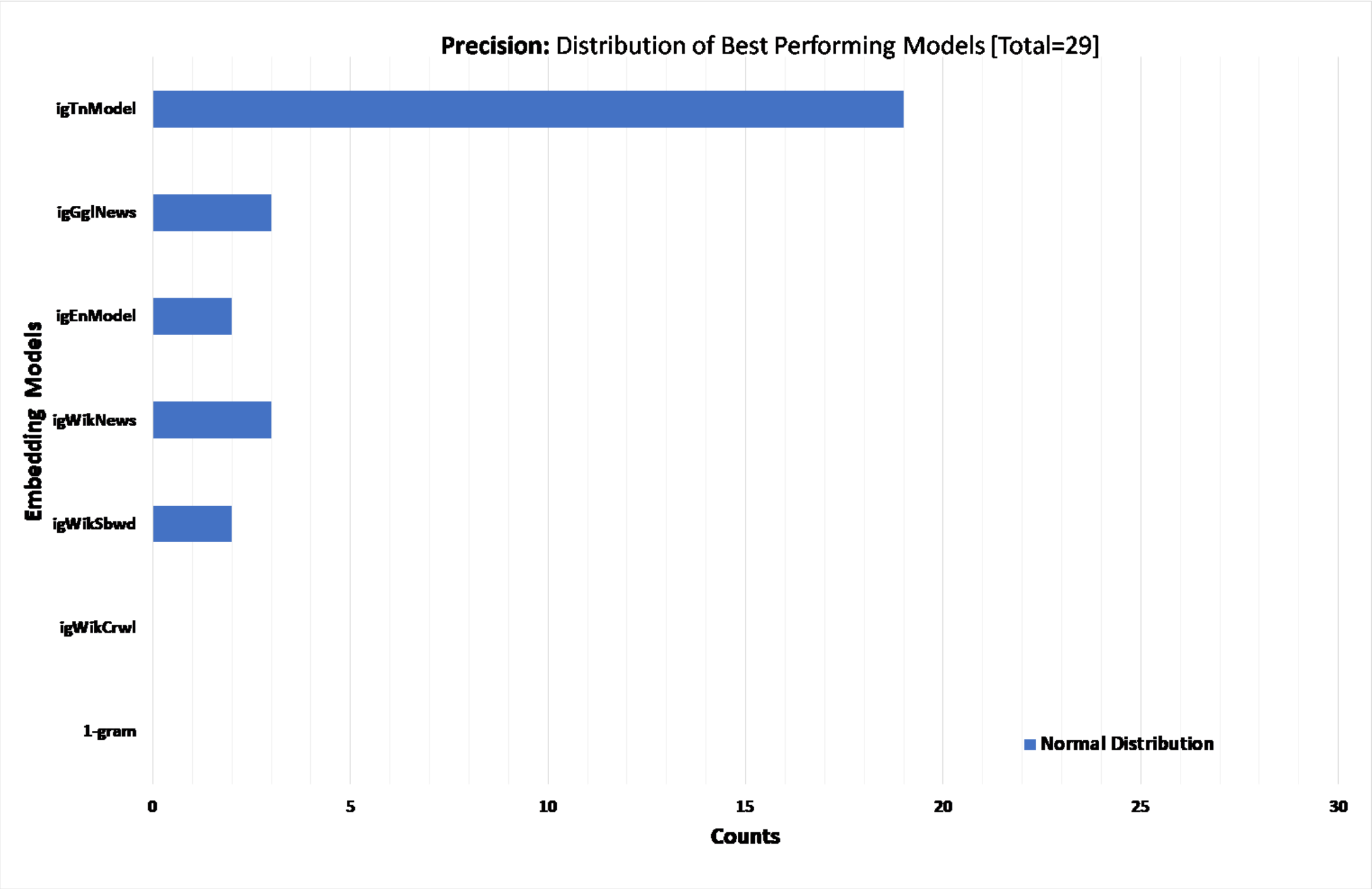

Fig. 29 **Emb Precision:** Graph showing the distribution of best precision scores across all models.

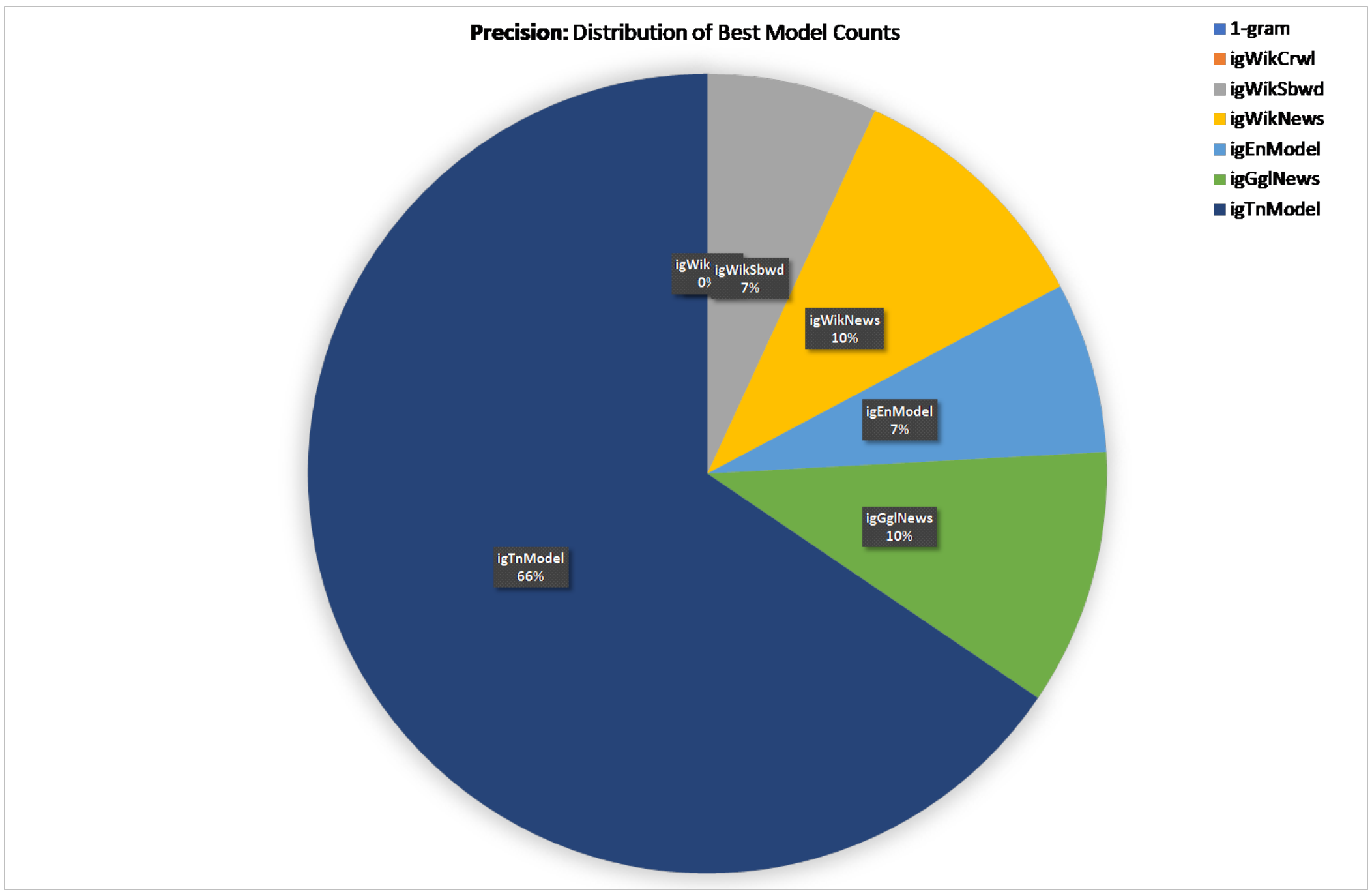

Fig. 30 **Emb Precision:** Piechart showing the percentage best distribution of the precision scores across all emb models.

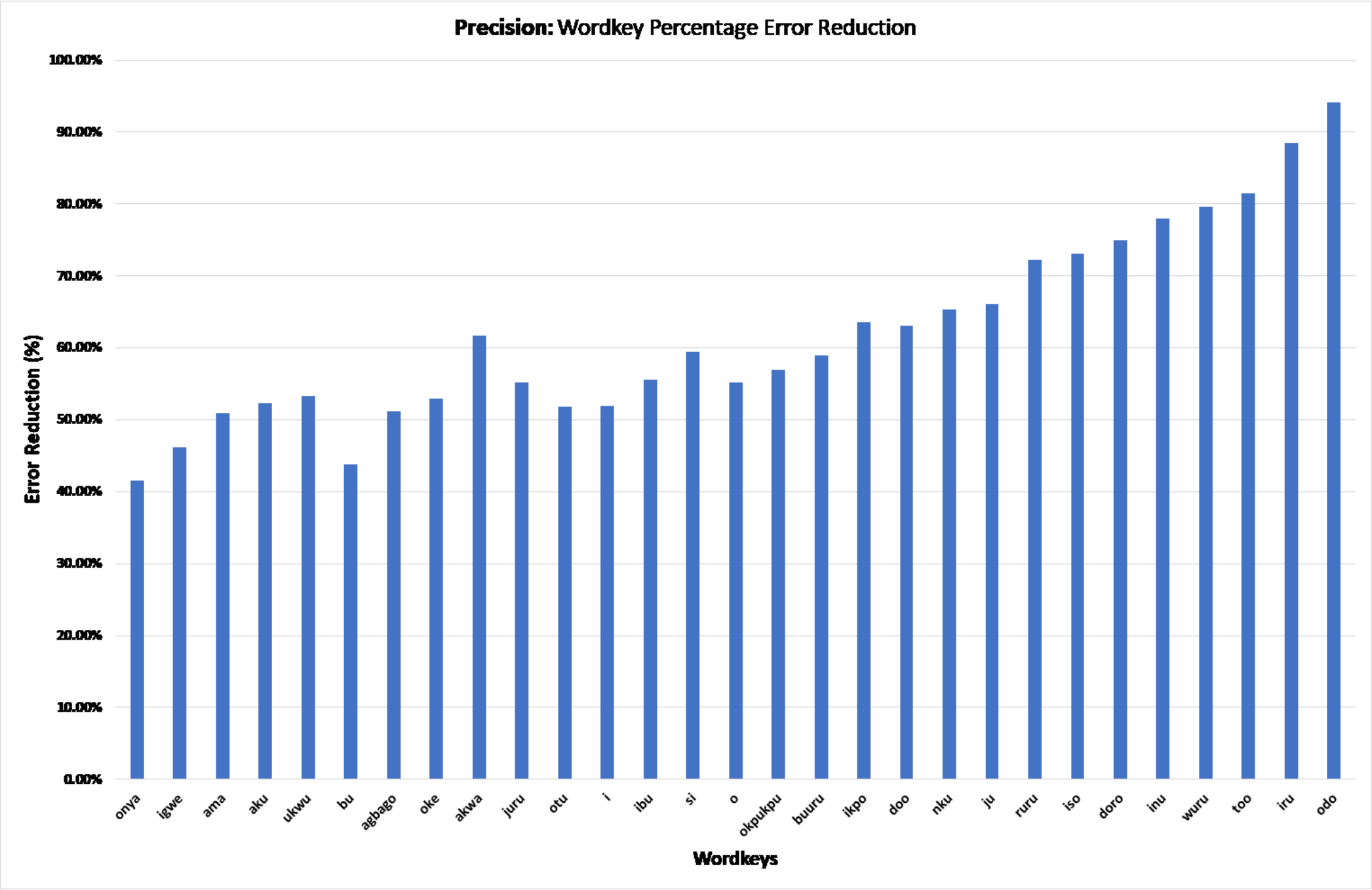

Fig. 31 **Emb Precision:** Graph showing the percentage precision error reduction on each wordkey by its best performing model.

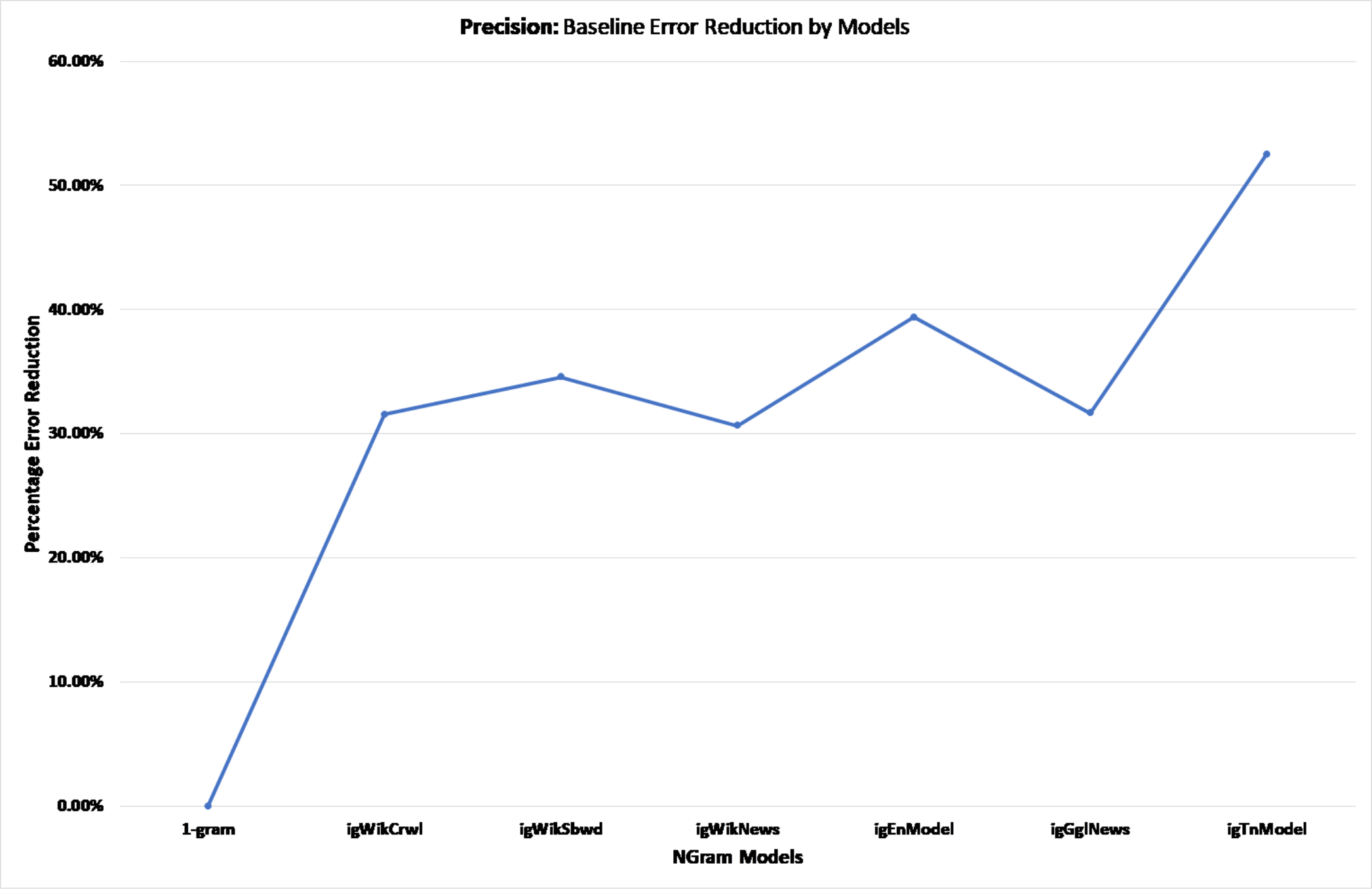

Fig. 32 **Emb Precision:** Column chart showing the percentage precision error reduction from the global baseline error by each model.

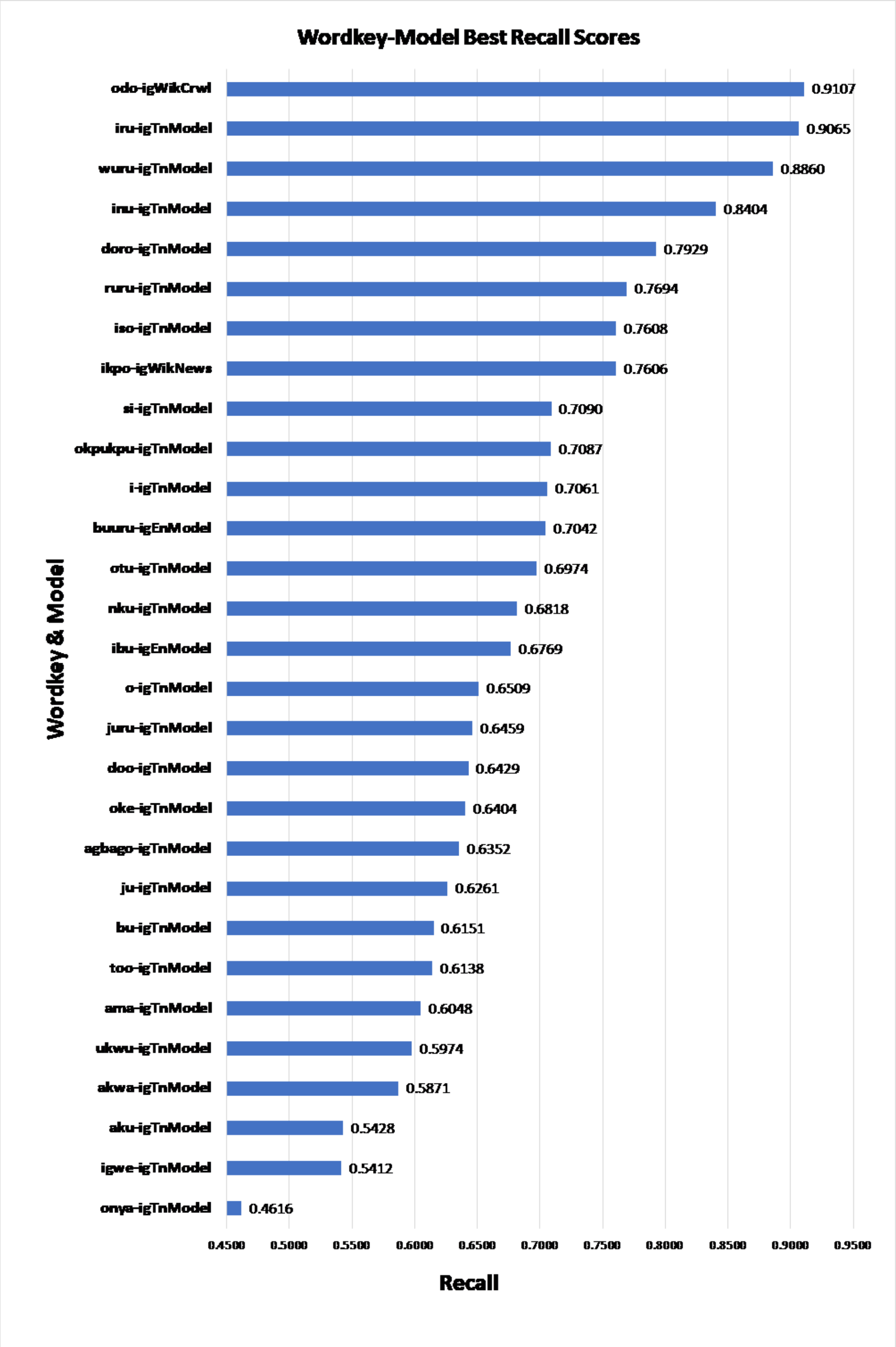

Fig. 33 **Emb Recall:** Figure showing the best performing model on for each of the wordkeys sorted in the ascending order of recall scores

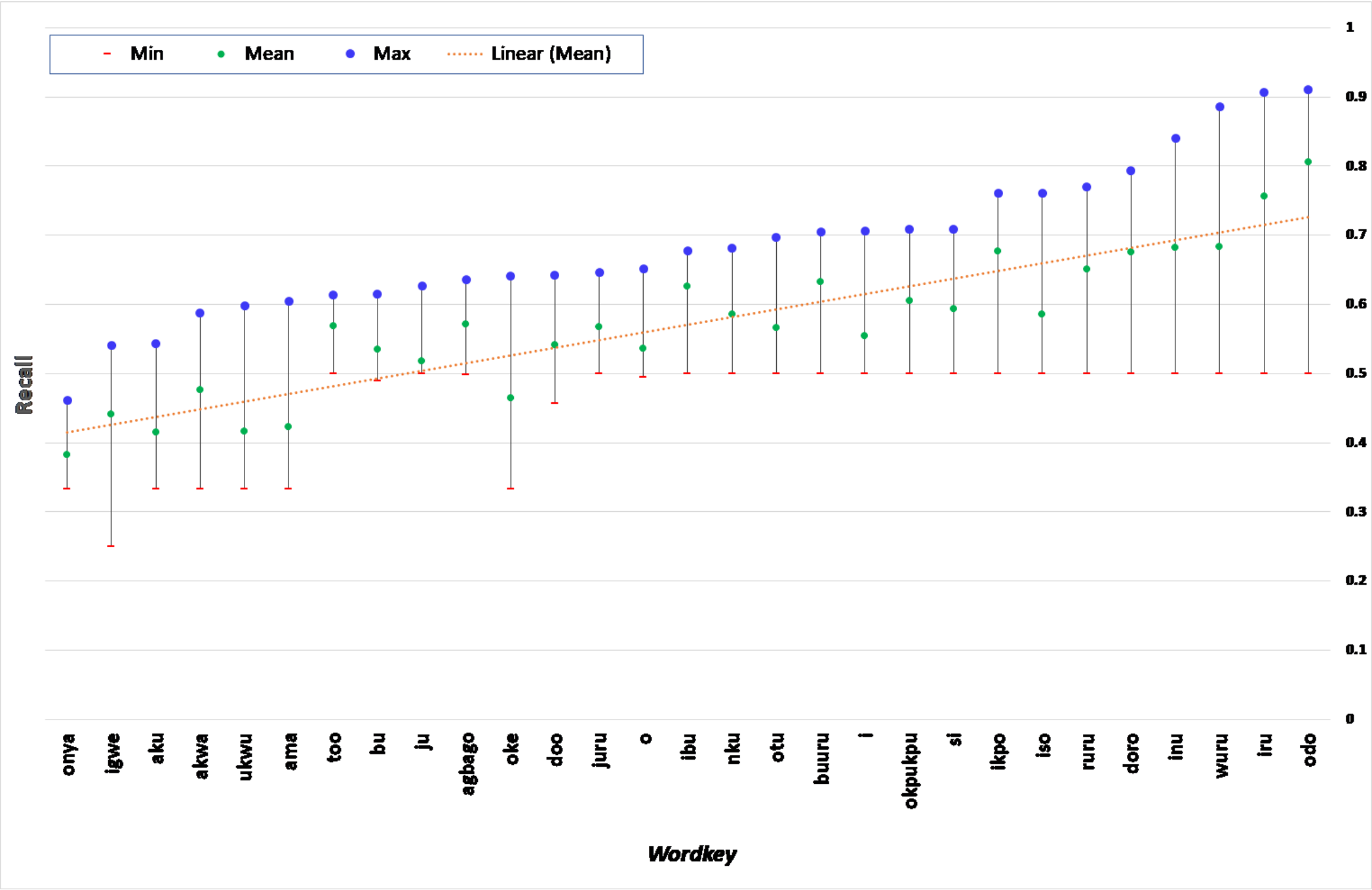

Fig. 34 **Emb Recall:** Graph showing minimum, mean and maximum recall scores and the linear trend line on the mean scores.

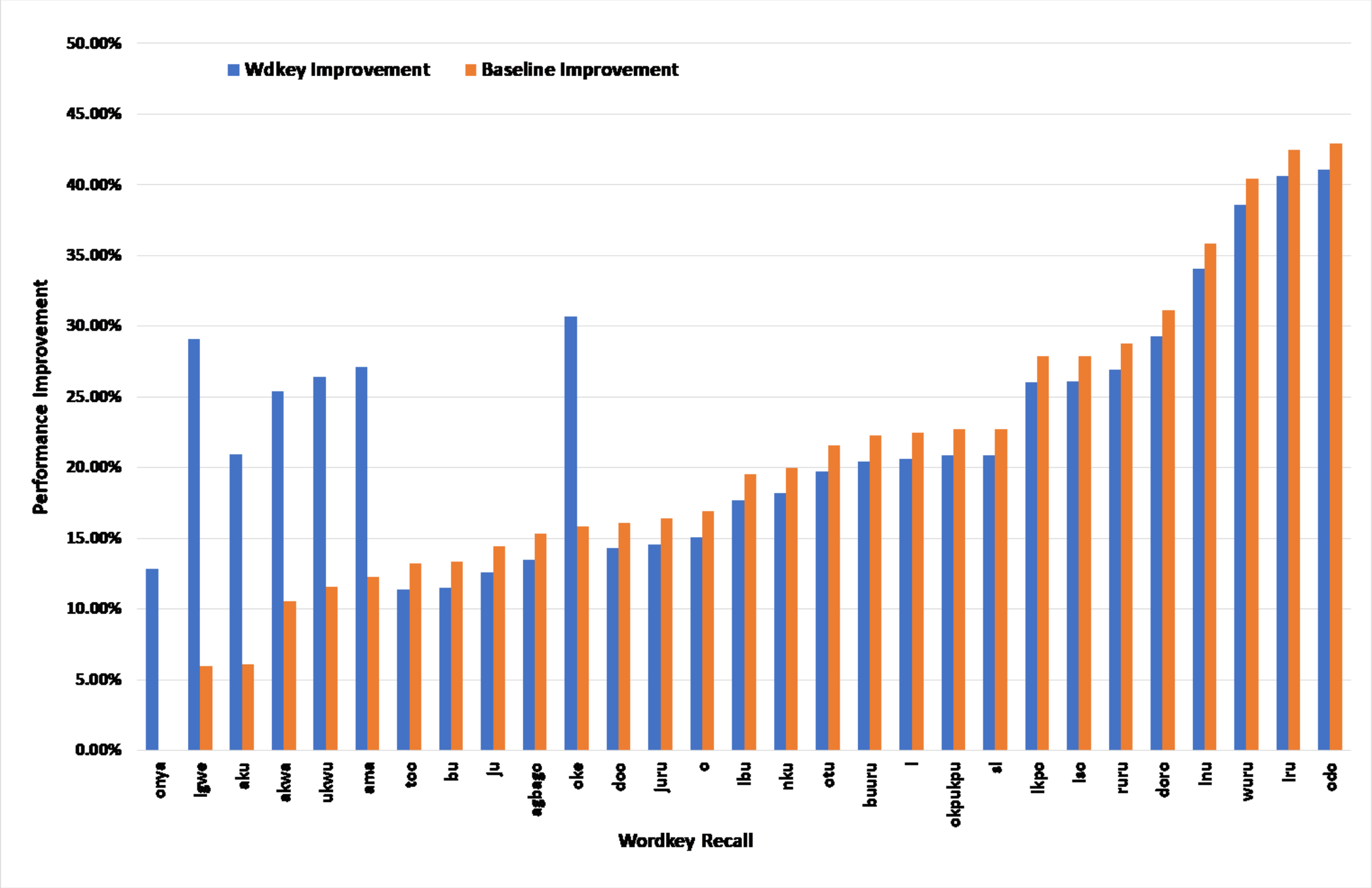

Fig. 35 **Emb Recall:** Graph showing the stacked column-chart of the improved recall scores on both the wordkey and the global baselines.

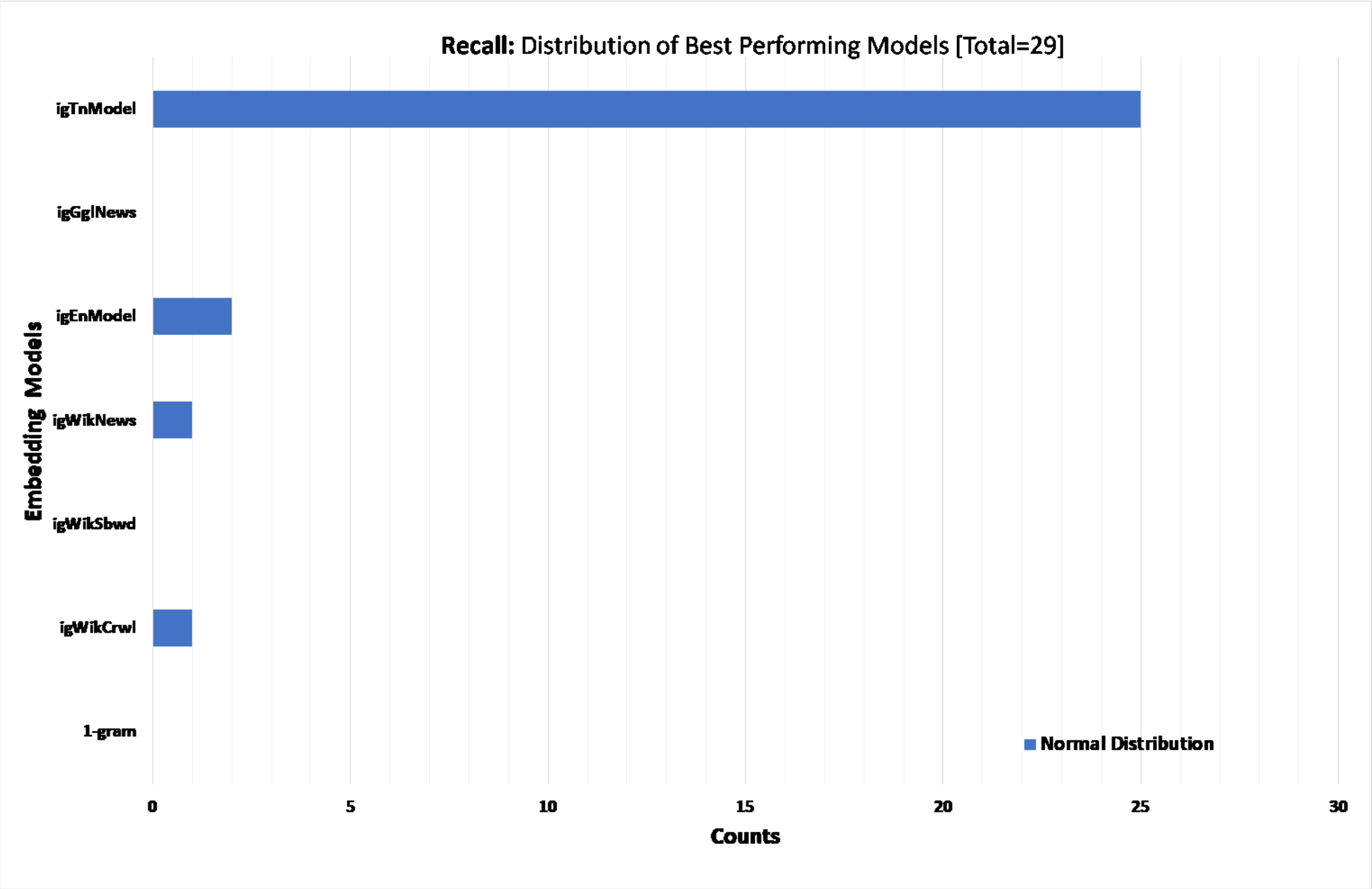

Fig. 36 **Emb Recall:** Graph showing the distribution of best recall scores across all ngram models.

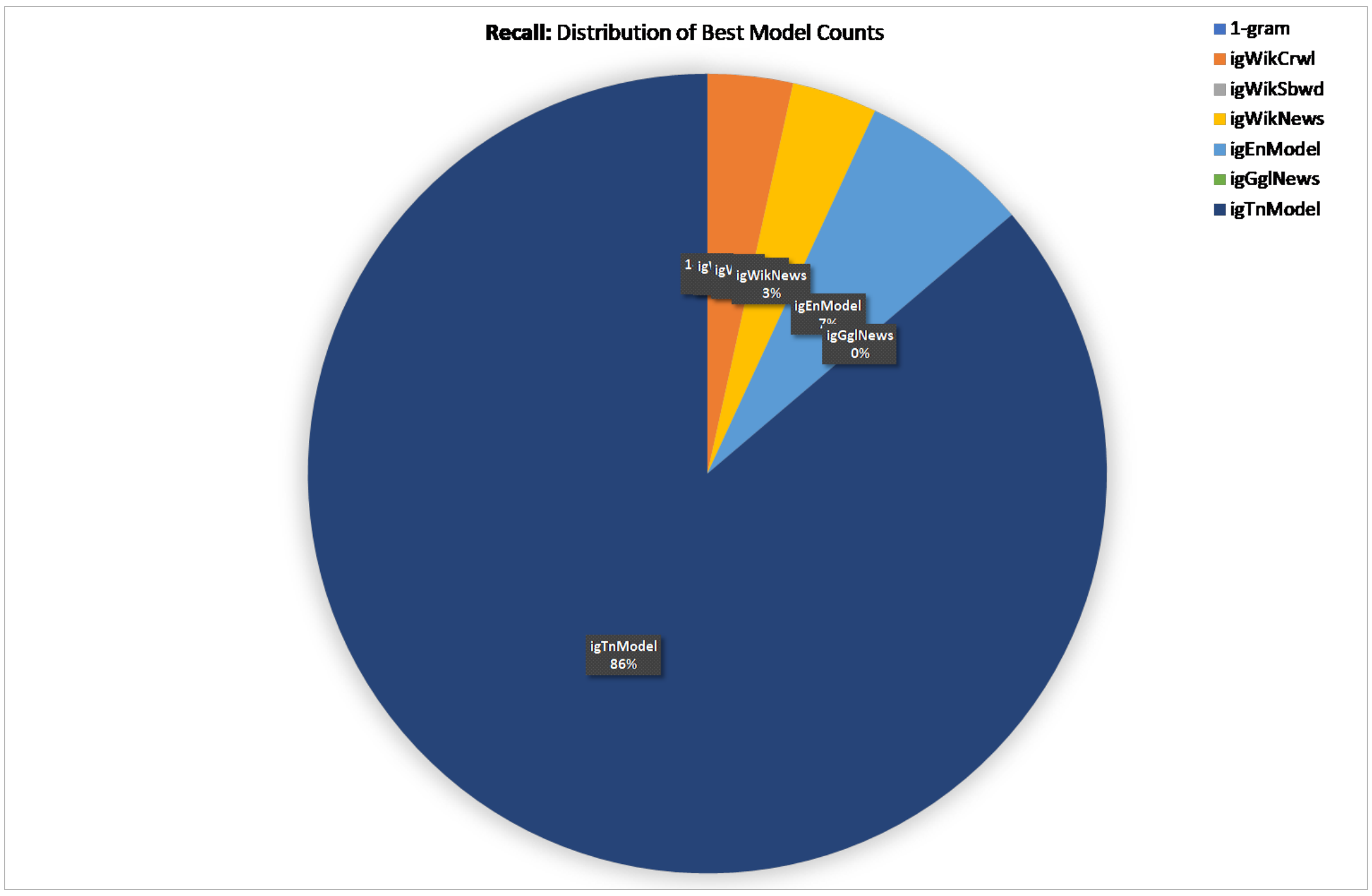

Fig. 37 **Emb Recall:** Piechart showing the percentage best distribution of the recall scores across all ngram models.

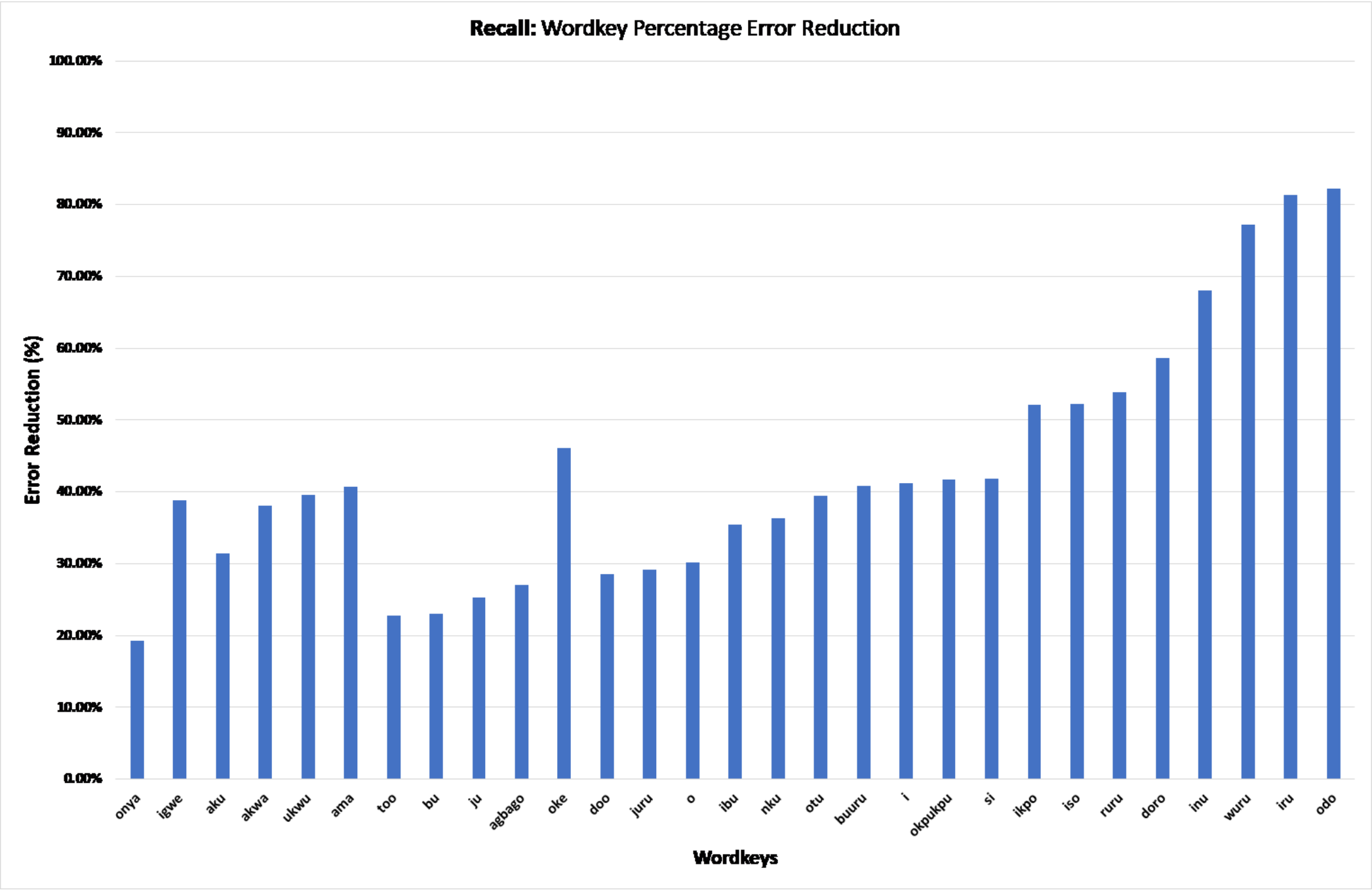

Fig. 38 **Emb Recall:** Graph showing the percentage recall error reduction on each wordkey by its best performing model.

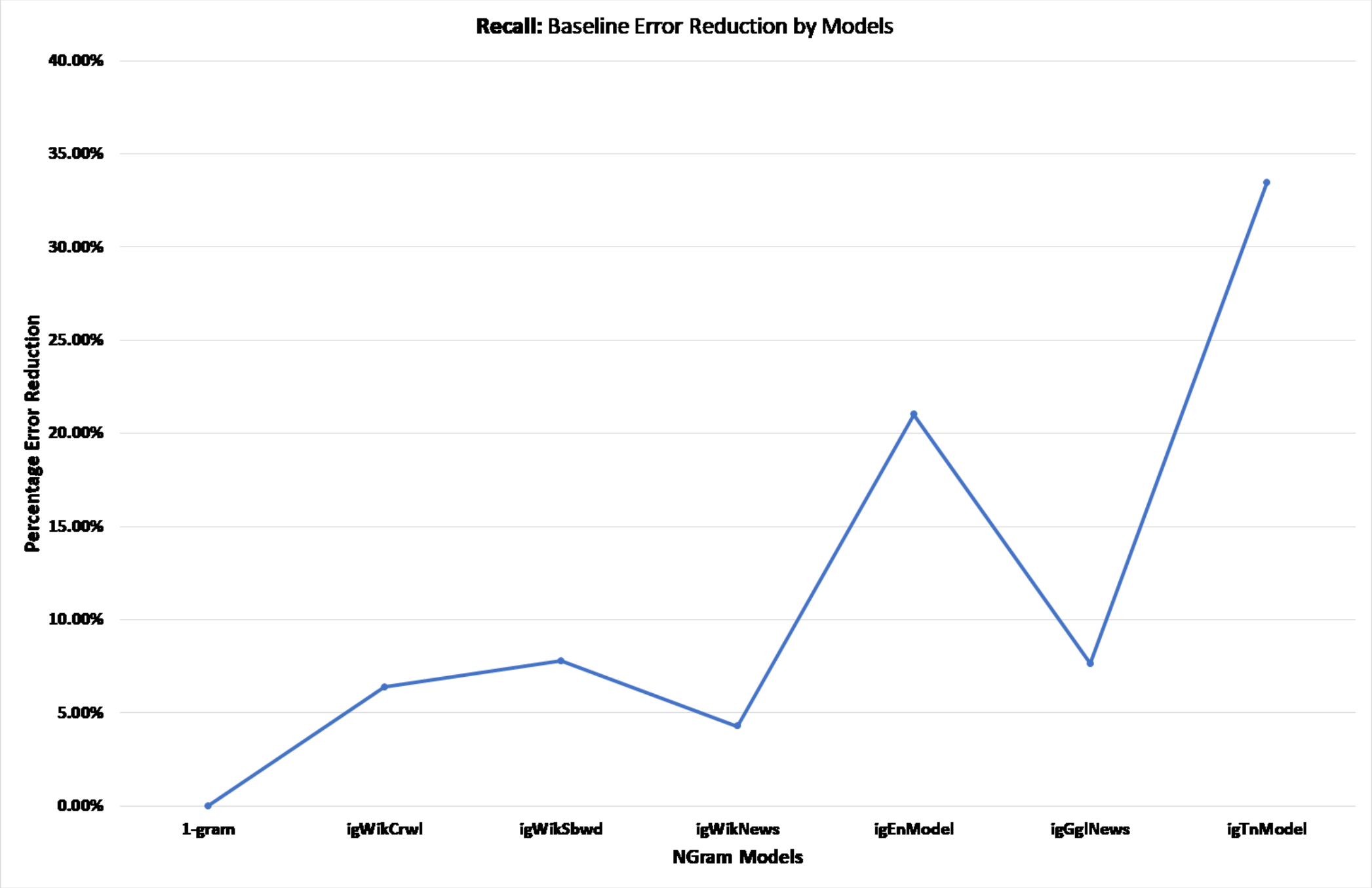

Fig. 39 **Emb Recall:** Column chart showing the percentage recall error reduction from the global baseline error by each model.

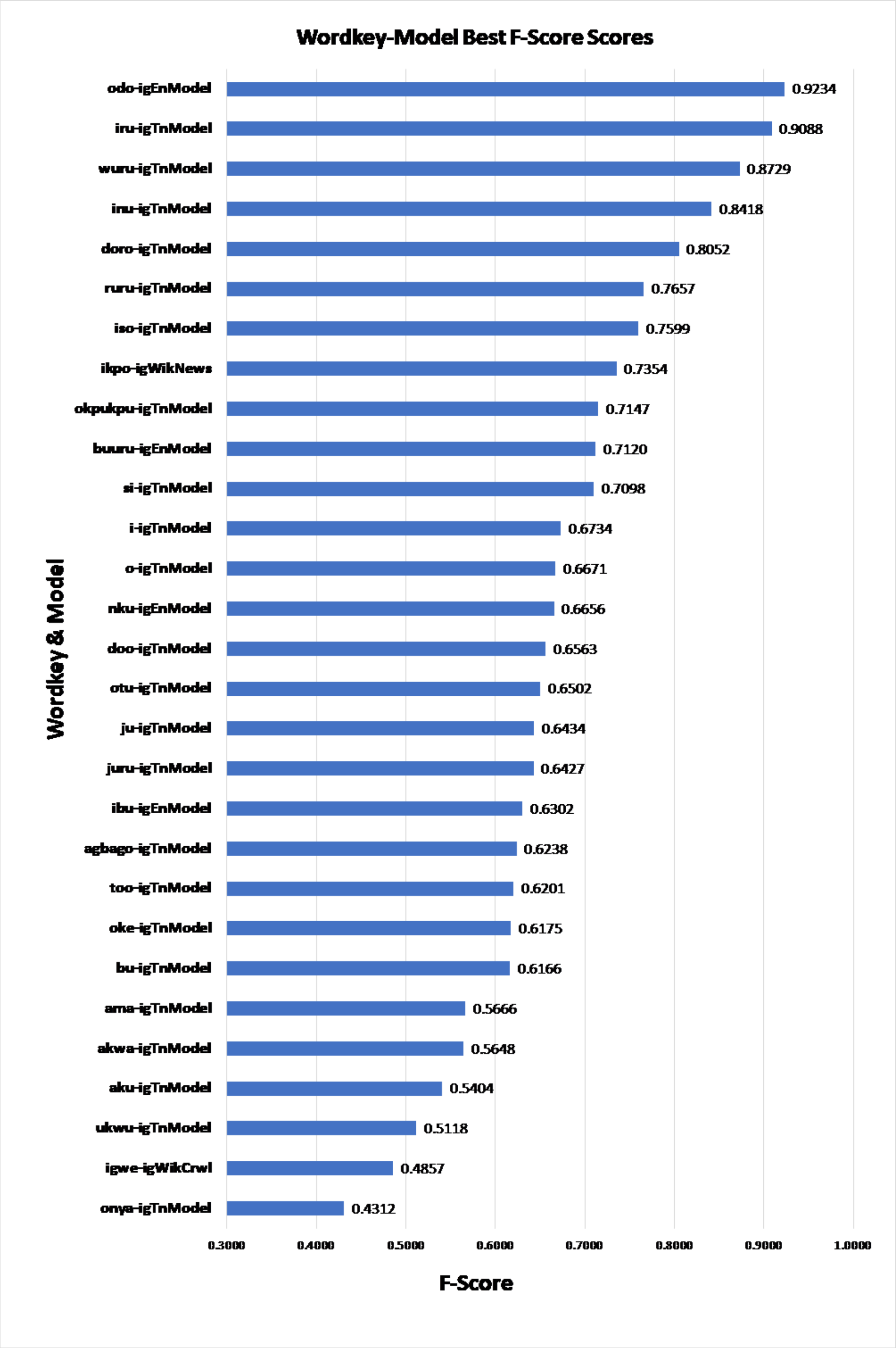

Fig. 40 **Emb F1-Score:** Figure showing the best performing model on for each of the wordkeys sorted in the ascending order of f1 scores

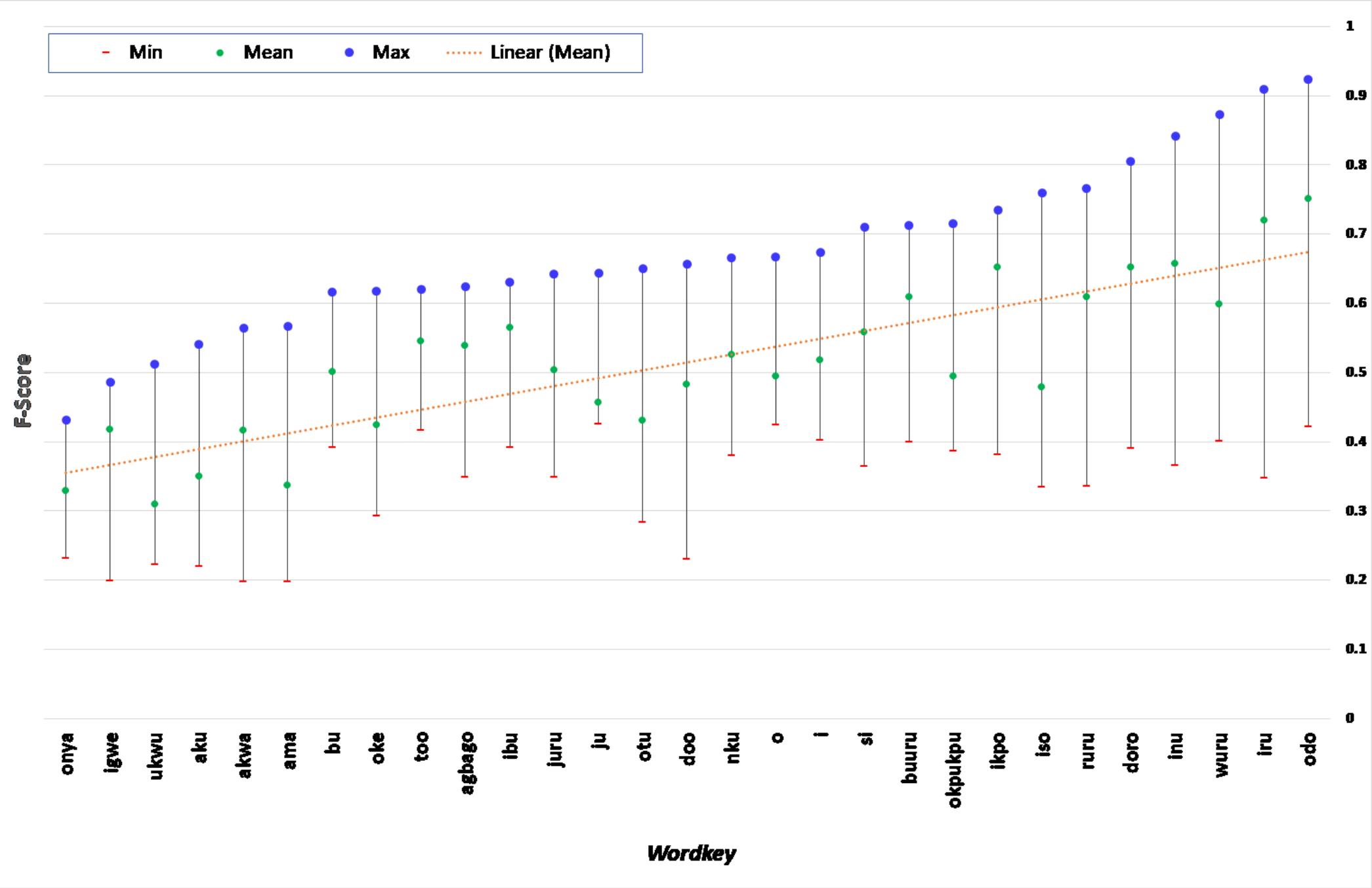

Fig. 41 **Emb F1-Score:** Graph showing minimum, mean and maximum F1 scores and the linear trend line on the mean scores.

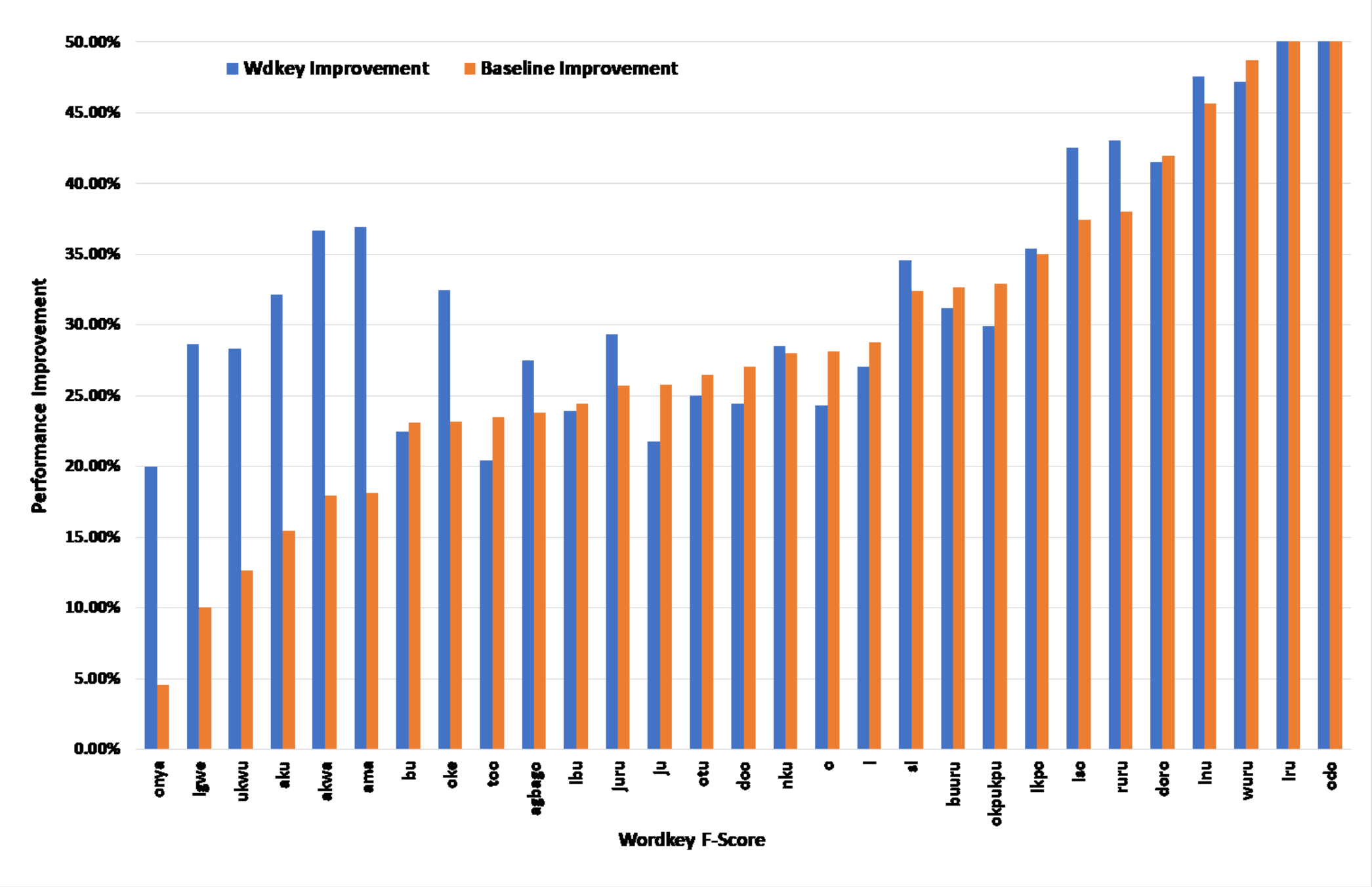

Fig. 42 **Emb F1-Score:** Graph showing the stacked column-chart of the improved F1 scores on both the wordkey and the global baselines.

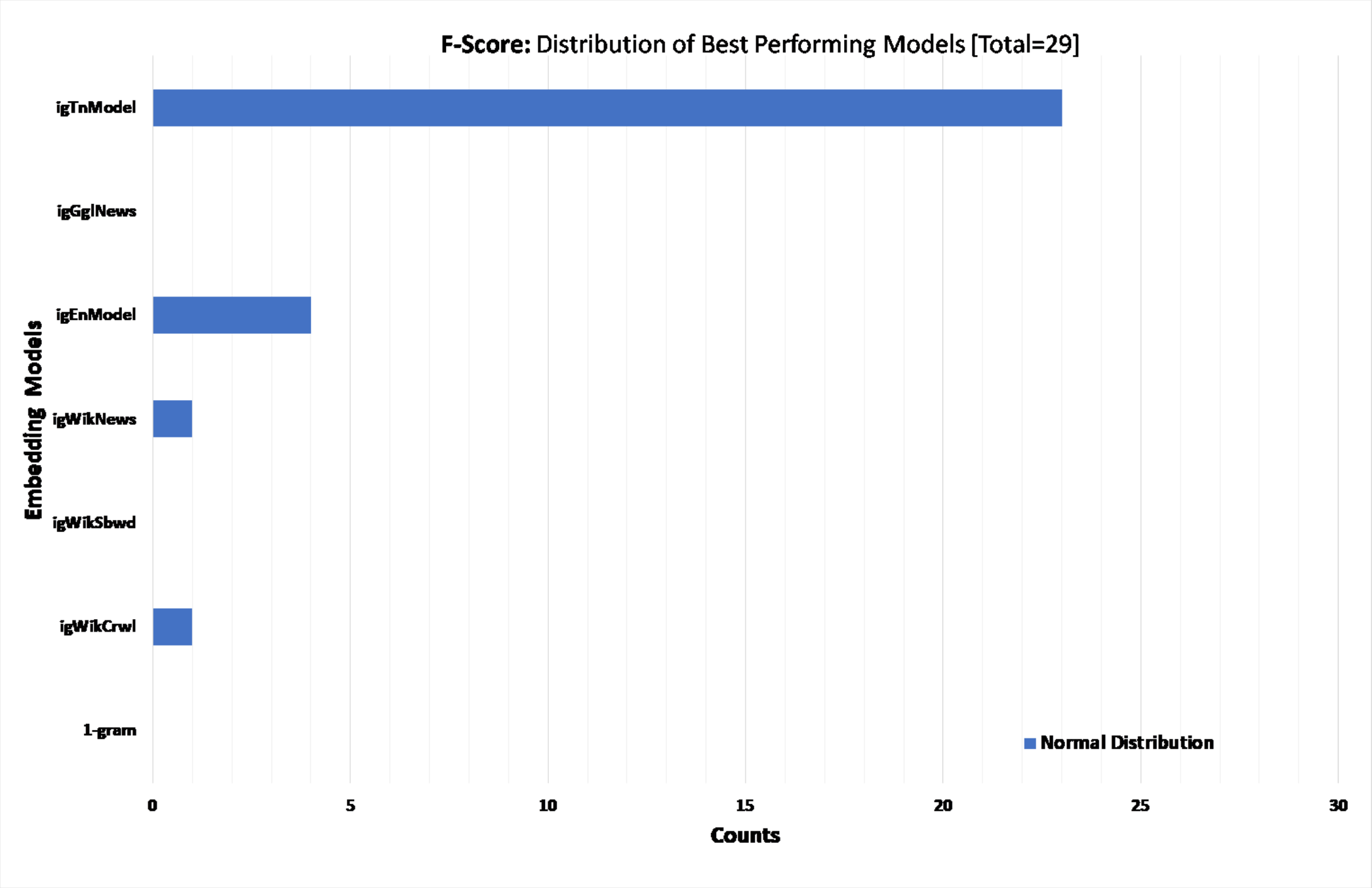

Fig. 43 **Emb F1-Score:** Graph showing the distribution of best f1 scores across all ngram models.

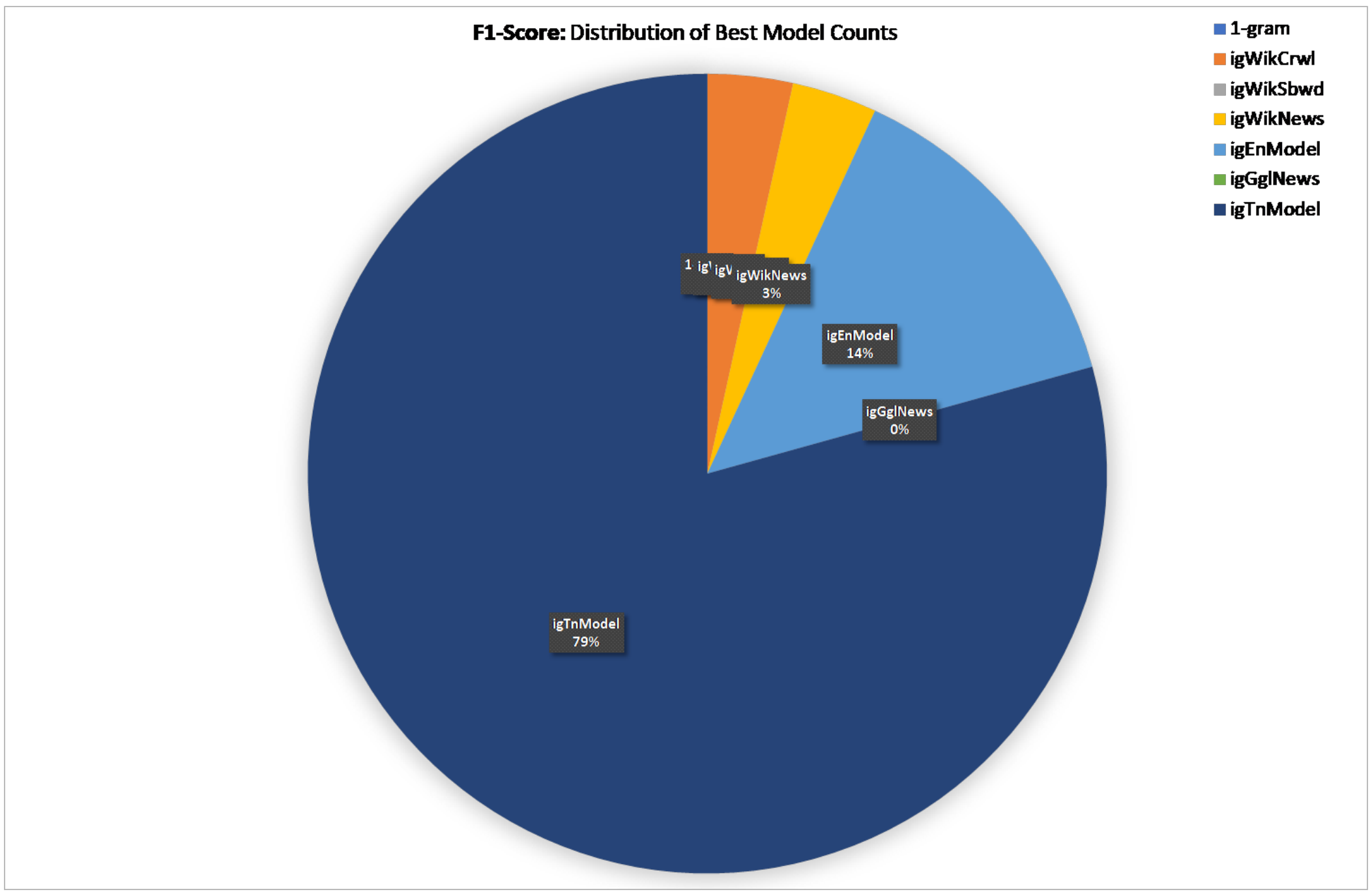

Fig. 44 **Emb F1-Score:** Piechart showing the percentage best distribution of the f1 scores across all ngram models.

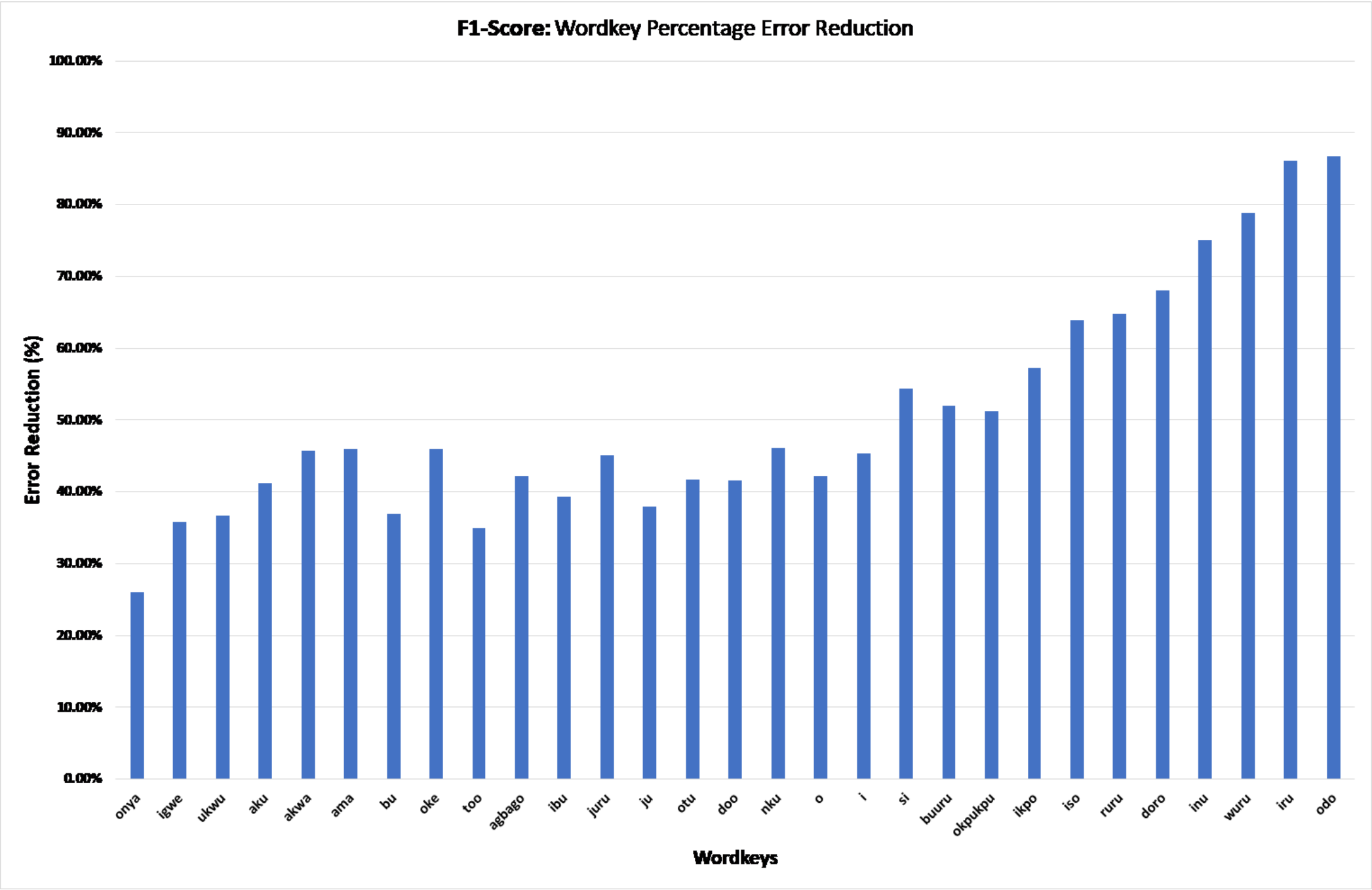

Fig. 45 **Emb F1-Score:** Graph showing the percentage F1 error reduction on each wordkey by its best performing model.

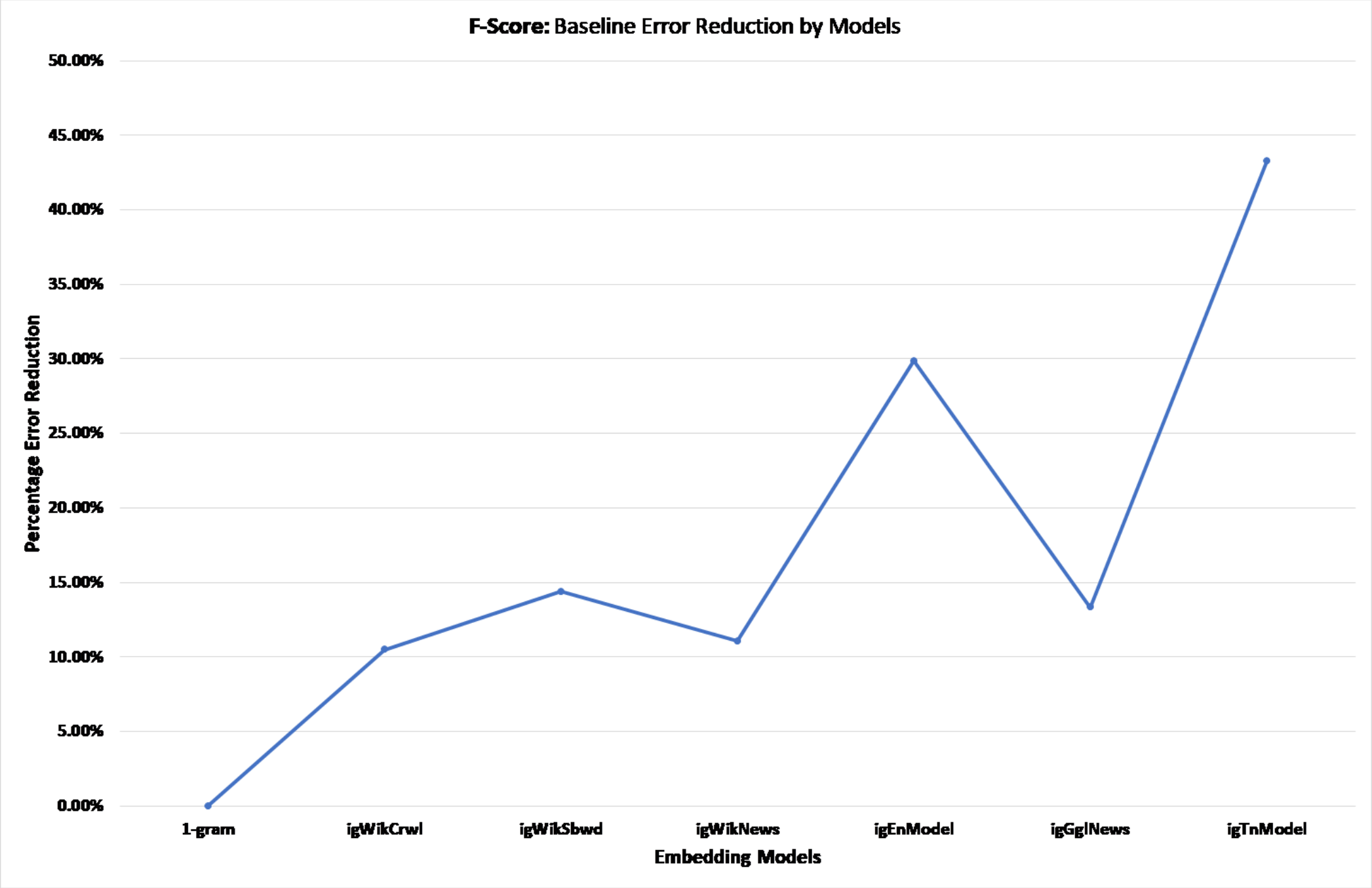

Fig. 46 **Emb F1-Score:** Column chart showing the percentage F1 error reduction from the global baseline error by each model.